\documentclass{article}

\usepackage{microtype}
\usepackage{graphicx}
\usepackage{subfigure}
\usepackage{booktabs} 

\usepackage{hyperref}

\usepackage{amsmath,amsthm,amssymb}
\usepackage{algorithm} 
\usepackage{tabularx}
\usepackage{dsfont}
\newtheorem{theorem}{Theorem}

\DeclareMathOperator{\sign}{sign}
\DeclareMathOperator*{\argmin}{\arg\min}
\DeclareMathOperator*{\argmax}{\arg\max}

\newcommand{\ST}{\mathrm{s.t.}}

\newcommand{\bR}{\mathbb{R}}

\newcommand{\cF}{\mathcal{F}}
\newcommand{\cX}{\mathcal{X}}
\newcommand{\cY}{\mathcal{Y}}
\newcommand{\bx}{{x}}
\newcommand{\bxprime}{{x}'}
\newcommand{\bxtidle}{\tilde{{x}}}

\newcommand{\epsball}{\mathcal{B}_\epsilon}
\newcommand{\xadv}{\tilde{{x}}}

\usepackage[accepted]{icml2020}

\icmltitlerunning{Friendly Adversarial Training}

\begin{document}

\twocolumn[
\icmltitle{Attacks Which Do Not Kill Training Make Adversarial Learning Stronger}




\icmlsetsymbol{equal}{*}
\icmlsetsymbol{intern}{$\dagger$}

\begin{icmlauthorlist}
\icmlauthor{Jingfeng Zhang}{equal,intern,nus}
\icmlauthor{Xilie Xu}{equal,sdu_taishan}
\icmlauthor{Bo Han}{buhk,riken}
\icmlauthor{Gang Niu}{riken}
\icmlauthor{Lizhen Cui}{sdu_software}\\
\icmlauthor{Masashi Sugiyama}{riken,tokyo_u}
\icmlauthor{Mohan Kankanhalli}{nus}
\end{icmlauthorlist}

\icmlaffiliation{nus}{School of Computing, National University of Singapore, Singapore} 
\icmlaffiliation{sdu_taishan}{Taishan College, Shandong University, Jinan, China}
\icmlaffiliation{buhk}{Department of Computer Science, Hong Kong Baptist University, Hong Kong, China}
\icmlaffiliation{sdu_software}{School of Software $\&$ Joint SDU-NTU Centre for Artificial Intelligence Research (C-FAIR), Shandong University, Jinan, China}
\icmlaffiliation{riken}{RIKEN Center for Advanced Intelligence Project (AIP), Tokyo, Japan}
\icmlaffiliation{tokyo_u}{Graduate School of Frontier Sciences, The University of Tokyo, Tokyo, Japan}

\icmlcorrespondingauthor{Jingfeng Zhang}{j-zhang@comp.nus.edu.sg}

\icmlkeywords{Machine Learning, ICML}

\vskip 0.3in
]

\newcommand{\icmlIntern}{\textsuperscript{$\dagger$}Preliminary work was done during an internship at RIKEN AIP. }
\printAffiliationsAndNotice{\icmlEqualContribution\icmlIntern} 

\begin{abstract}
Adversarial training based on the \emph{minimax} formulation is necessary 
for obtaining \emph{adversarial robustness} of trained models.
However, it is conservative or even pessimistic so that it sometimes hurts the \emph{natural generalization}. In this paper, we raise a fundamental question---do we have to trade off natural generalization for adversarial robustness? We argue that adversarial training is to employ confident adversarial data for updating the current model. We propose a novel formulation of \emph{friendly adversarial training} (FAT): rather than employing most adversarial data maximizing the loss, we search for least adversarial data (i.e., \textit{friendly adversarial data}) minimizing the loss, among the adversarial data that are confidently misclassified. Our novel formulation is easy to implement by just stopping the most adversarial data searching algorithms such as PGD (projected gradient descent) early, which we call \emph{early-stopped PGD}. Theoretically, FAT is justified by an upper bound of the adversarial risk. Empirically, early-stopped PGD allows us to answer the earlier question negatively---adversarial robustness can indeed be achieved without compromising the natural generalization.
\end{abstract}
\section{Introduction}
\label{introduction}
Safety-critical nature of some areas such as medicine~\cite{buch2018artificial_medicine} and automatic driving~\cite{litman2017autonomous}, necessitates the need for deep neural networks (DNNs) to be adversarially robust that generalize well. 
Recent research focuses on improving their robustness mainly by two defense approaches, i.e., certified defense and empirical defense. Certified defense tries to learn provably robust DNNs against norm-bounded (e.g., $\ell_2$ and $\ell_\infty$) perturbations~\cite{Jeremy_cohen_certified_robustness_random_smoothing,Eric_Wong_provable_defence_convex_polytope,Tsuzuku_Lipschitz_margin_training_scalable_certification,Lecuyer_certified_robustness_with_DP,Weng_Evaluating_robustness_extreme_value_approach,balunovic2020adversarial_certified_robustness,zhang2020towards_certifiable}. 
Empirical defense incorporates adversarial data into the training process~\cite{Goodfellow14_Adversarial_examples,Madry_adversarial_training,Cai_CAT,Zhang_trades,Wang_Xingjun_MA_FOSC_DAT,wang2020improving_MART}.
For instance, empirical defense has been used to train Wide ResNet~\cite{zagoruyko2016WRN} with natural data and its adversarial variants to make the trained network robust against strong adaptive attacks~\cite{Athalye_ICML_18_Obfuscated_Gradients,Carlini017_CW}. This paper belongs to the empirical defense category.

Existing empirical defense methods formulate the adversarial training as a minimax optimization problem (Section~\ref{section:learning_obj_madry})~\cite{Madry_adversarial_training}.
To conduct this minimax optimization, projected gradient descent (PGD) is a common method to generate the most adversarial data that maximizes the loss, updating the current model.
PGD perturbs the natural data for a fixed number of steps with small step size. After each step of perturbation, PGD projects the adversarial data back onto the $\epsilon$-norm ball of the natural data.

\begin{figure}[tp!]
    \centering
    \includegraphics[scale=0.22]{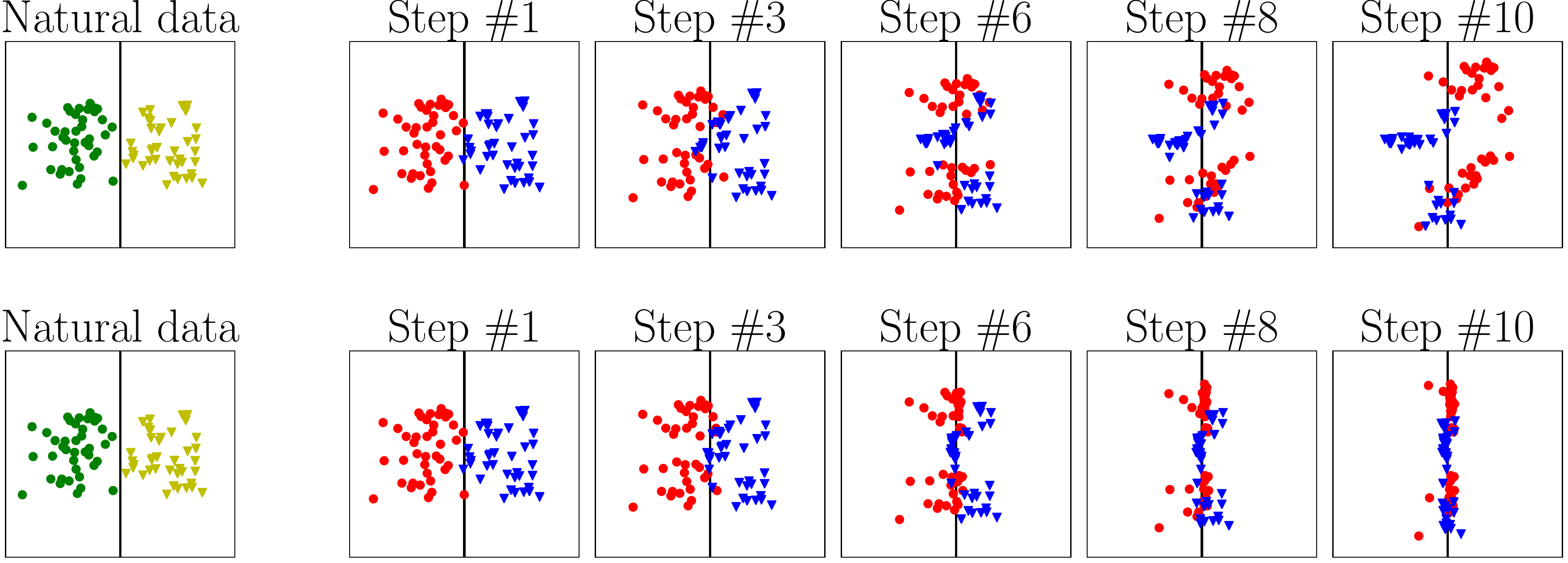}
    \caption{Green circles and yellow triangles are natural data for binary classification. Red circles and blue triangles are adversarial variants of green circles and yellow triangles, respectively. Black solid line is the decision boundary representing the current classifier. 
    Top: the adversarial data generated by PGD. Bottom: the adversarial data generated by early-stopped PGD.}
    \label{fig:motivation_fig}
    \vspace{-4mm}
\end{figure}

However, this minimax formulation is conservative (or even pessimistic), such that it sometimes hurts the natural generalization~\cite{Tsipras19_robustness_at_odd}. 
For example, the top panels in Figure~\ref{fig:motivation_fig} show that at step $\#6$ to $\#10$ in PGD, the adversarial variants of the natural data significantly cross over the decision boundary and are located at their peer's (natural data) area. Since adversarial training aims to fit natural data and its adversarial variants simultaneously, such the cross-over mixture makes adversarial training extremely difficult. Therefore, the most adversarial data generated by the \mbox{PGD-10} (i.e., step $\#$10 in top panel of Figure~\ref{fig:motivation_fig}) directly ``kill'' the training, thus rendering the training unsuccessful.

Inspired by philosopher Friedrich Nietzsche's quote ``\emph{that which does not kill us makes us stronger}," we propose \textit{friendly adversarial training} (FAT): rather than employing the most adversarial data for updating the current model, we search for the \textit{friendly adversarial data} minimizing the loss. The friendly adversarial data are confidently misclassified by the current model.
We design the learning objective of FAT and theoretically justify it by deriving an upper bound of the adversarial risk (Section~\ref{Section:FAT_obj_theory}).
Essentially, FAT updates the current model using friendly adversarial data. FAT trains a DNN using the wrongly-predicted adversarial data minimizing the loss and the correctly-predicted adversarial data maximizing the loss. 
%

FAT is a reasonable strategy due to two reasons: It removes the existing inconsistency between attack and defense, and it adheres to the spirit of curriculum learning.  
First, the ways of generating adversarial data by adversarial attackers and  adversarial defense methods are inconsistent. Adversarial attacks~\cite{szegedy,Carlini017_CW,Athalye_ICML_18_Obfuscated_Gradients} 
aim to find the adversarial data (not maximizing the loss) to confidently fool the model. On the other hand, existing adversarial defense methods generate the most adversarial data maximizing the loss regardless of the model's predictions. These two should be harmonized.
Second, the curriculum learning strategy has been shown to be effective~\cite{bengio2009curriculum}. Fitting most adversarial data initially makes the learning extremely difficult, sometimes even killing the training. Instead, FAT learns initially from the least adversarial data and progressively utilizes increasingly adversarial data.

FAT is easy to implement by just early stopping most-adversarial-data searching algorithms such as PGD, which we call early-stopped PGD (Section~\ref{section:PGD-K-t}). 
Once adversarial data is misclassified by the current model, we stop the PGD iterations early.
Early-stopped PGD has the benefit of alleviating the cross-over mixture problem. For example, as shown in the bottom panels of Figure~\ref{fig:motivation_fig}, adversarial data generated by early-stopped PGD will not be located at their peer areas (extensive details in Section~\ref{Section:Mixture_alleviation}). Thus, it will not hurt the generalization ability much. 
In addition, FAT based on early-stopped PGD progressively employs stronger and stronger adversarial data (with more PGD steps), engendering increasingly enhanced robustness of the model over the training progression (Section~\ref{section:curriculum_learning}). This implies that attacks that do not kill the training indeed make the adversarial learning stronger.

A brief overview of our contributions is as follows. We propose a novel formulation for adversarial learning (Section~\ref{Section:learning_obj_FAT}) and theoretically justify it by an upper bound of the adversarial risk (Section~\ref{Section:upper_bound_on_adversarial_risk}). Our FAT approximately realizes this formulation by just stopping PGD early.
FAT has the following benefits.
\begin{itemize}
    \item Conventional adversarial training methods, e.g., standard adversarial training~\cite{Madry_adversarial_training}, TRADES~\cite{Zhang_trades} and MART~\cite{wang2020improving_MART}, can be easily modified to become friendly adversarial training counterparts, i.e., FAT, FAT for TRADES, and FAT for MART (Section~\ref{Sec:Friendly-AdvTraining}).
    \item Compared with conventional adversarial training, FAT has a better standard accuracy for natural data, while keeping a competitively robust accuracy for adversarial data (Sections~\ref{Section:slippery_step_tau} and~\ref{section:SOTA_results}). 
    \item FAT is computationally efficient because the early stopped PGD saves a large number of backward propagations for searching adversarial data (Section~\ref{Section:Computaional_efficient}).
    \item FAT can enable larger values of the perturbation bound, i.e., $\epsilon_{train}$ (Section~\ref{section:fat_enable_lager_epsilon}), due to that FAT can alleviate the cross-over mixture problem (Section~\ref{Section:Mixture_alleviation}). 
\end{itemize}
With these benefits, FAT allows us to answer that adversarial robustness can indeed be achieved without compromising the natural generalization.

\section{Standard Adversarial Training}
\label{Section:background_and_notations}
Let $(\cX,d_\infty)$ be the input feature space $\cX$ with the infinity distance metric $d_{\inf}(\bx,\bxprime)=\|\bx-\bxprime\|_\infty$, and
\begin{align}
\label{eqn:perturbation_ball}
\epsball[\bx] = \{\bxprime \in \cX \mid d_{\inf}(\bx,\bx')\le\epsilon\}
\end{align}
be the closed ball of radius $\epsilon>0$ centered at $\bx$ in $\cX$.

\subsection{Learning Objective}
\label{section:learning_obj_madry}
\begin{figure*}[h!]
    \centering
    \includegraphics[scale=0.4]{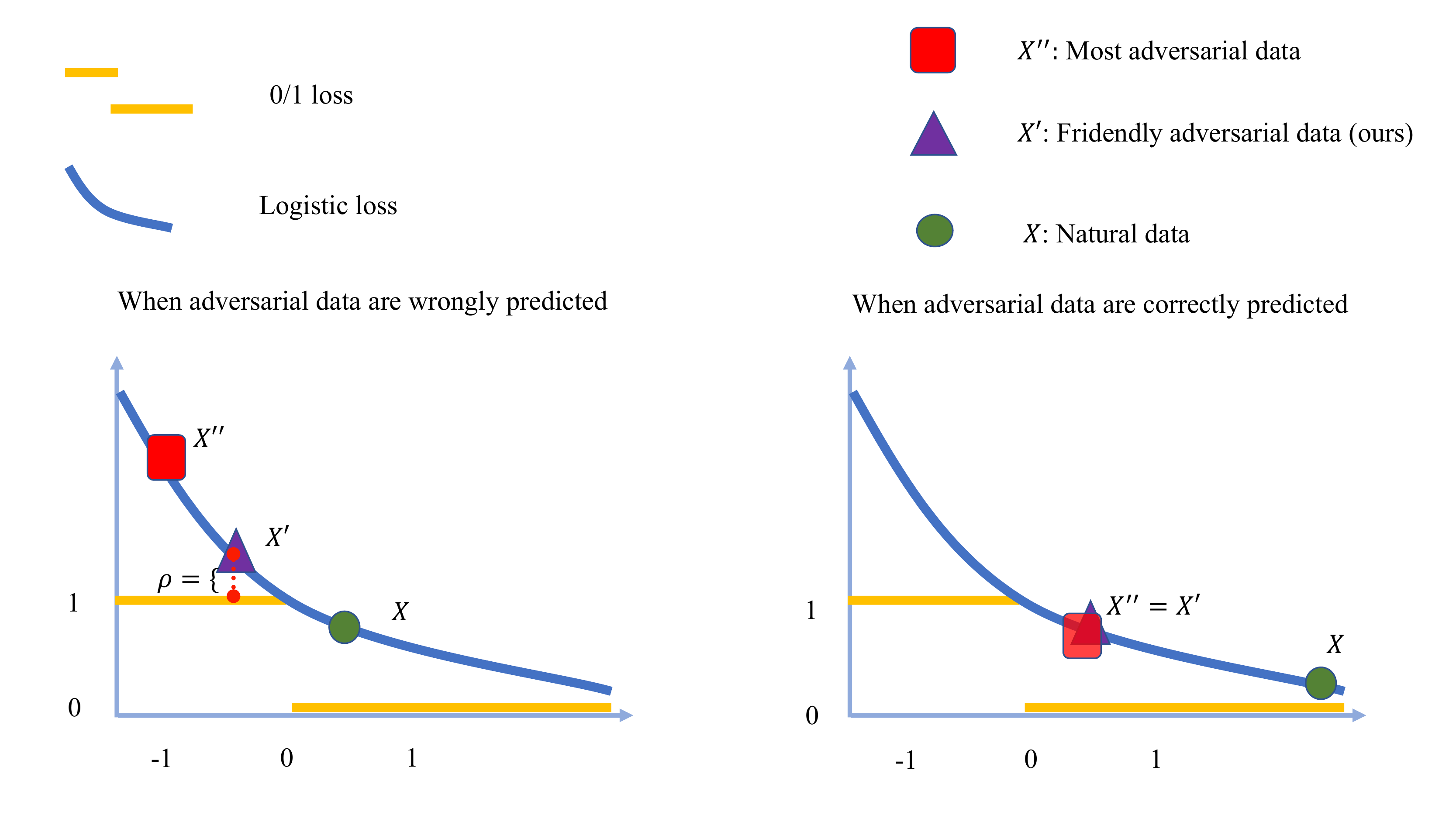}
    \caption{In adversarial training, the adversarial data is generated within the perturbation ball $\epsball[X]$ of natural data $X$ given the current model $f$. The existing adversarial training generates most adversarial data $X^{''}$ that maximizes the inner loss regardless of their predictions. FAT takes their predictions into account. When the model makes the wrong predictions of adversarial data, our friendly adversarial data $X'$ minimizes the inner loss by a violation of a small constant $\rho$.}
    \label{fig:lemma_show}
\end{figure*}
Given a dataset $S = \{ ({x}_i, y_i)\}^n_{i=1}$, where ${x}_i \in \cX$ and $y_i \in \cY =  \{0, 1, ..., C-1\}$, 
the objective function of standard adversarial training~\cite{Madry_adversarial_training} is 
\begin{equation}
\label{madry_adversarial_training}
\min_{f\in\cF} \frac{1}{n}\sum_{i=1}^n \left\{\max_{\bxtidle \in \epsball[\bx_i]} \ell(f(\bxtidle),y_i)\right\},
\end{equation}
where $\bxtidle$ is the adversarial data within the $\epsilon$-ball centered at ${x}$, $f(\cdot):\cX\to\bR^C$ is a score function, and the loss function $\ell:\bR^C\times\cY\to\bR$ is a composition of a base loss $\ell_\textrm{B}:\Delta^{C-1}\times\cY\to\bR$ (e.g., the cross-entropy loss) and an inverse link function $\ell_\textrm{L}:\bR^C\to\Delta^{C-1}$ (e.g., the soft-max activation), in which $\Delta^{C-1}$ is the corresponding probability simplex---in other words, $\ell(f(\cdot),y)=\ell_\textrm{B}(\ell_\textrm{L}(f(\cdot)),y)$.

For the sake of conceptual consistency, the objective function (i.e., Eq.~\eqref{madry_adversarial_training}) can also be re-written as
\begin{align}
\label{eq:adv-obj}
\min_{f\in\cF} \frac{1}{n}\sum_{i=1}^n \ell(f(\xadv_i),y_i),
\end{align}
where
\begin{align}
\label{Eq:madry_inner_maximization}
\xadv_i = \argmax\nolimits_{\xadv\in\epsball[\bx_i]} \ell(f(\xadv),y_i).
\end{align}
It implies the optimization of adversariallly robust network, with one step maximizing loss to find adversarial data and one step minimizing loss on the adversarial data w.r.t. the network parameters $\theta$. 

\subsection{Projected Gradient Descent (PGD)}
\label{section_PGD}
To generate adversarial data, standard adversarial training uses PGD to approximately solve the inner maximization of Eq.~\eqref{Eq:madry_inner_maximization}~\cite{Madry_adversarial_training}.

PGD formulates the problem of finding adversarial data as a constrained optimization problem. Namely, given a starting point ${x}^{(0)} \in \cX$ and step size $\alpha > 0$, PGD works as follows:
\begin{equation}
    \label{PGD-k}
    {x}^{(t+1)} = \Pi_{\mathcal{B}[{x}^{(0)}]} \big( {x}^{(t)} +\alpha \sign (\nabla_{{x}^{(t)}} \ell(f_{\theta}({x}^{(t)}), y )  )  \big ) , \forall { t \geq 0}
\end{equation}
until a certain stopping criterion is satisfied. For example, the criterion can be a fixed number of iterations $K$, namely the PGD-$K$ algorithm~\cite{Madry_adversarial_training,wang2020improving_MART}. In Eq.~\eqref{PGD-k}, $\ell$ is the loss function in Eq.~\eqref{Eq:madry_inner_maximization}; ${x}^{(0)}$ refers to natural data or natural data corrupted by a small Gaussian or uniform random noise; $y$ is the corresponding label for natural data; ${x}^{(t)}$ is adversarial data at step $t$; and $\Pi_{\epsball[{x}_0]}(\cdot)$ is the projection function that projects the adversarial data back into the $\epsilon$-ball centered at ${x}^{(0)}$ if necessary. 

There are also other ways to generate adversarial data, e.g., the fast gradient signed method~\cite{szegedy,Goodfellow14_Adversarial_examples}, the CW attack~\cite{Carlini017_CW}, deformation attack~\cite{Alaifari19_iclr_deformation,Chaowei_iclr18_deformation}, and Hamming distance method~\cite{shamir2019simple}.
\paragraph{PGD adversarial training.} Besides the standard adversarial training, several improvements to PGD adversarial training have also been proposed, such as Lipschitz regularization~\cite{cisse2017parseval,hein2017formal,yan2018deep,farnia2018generalizable}, curriculum adversarial training~\cite{Cai_CAT,Wang_Xingjun_MA_FOSC_DAT}, computationally efficient adversarial learning~\cite{Ali_NIPS19_adversarial_training_for_free,Lu_yiping_NIPS19_yopo,wong2020fast_zico_kolter}, ensemble adversarial training~\cite{tramer2018ensemble_iclr,Pang_ICML_19_AT_Ensemble}, and adversarial training by utilizing unlabeled data~\cite{carmon2019unlabeled,najafi2019robustness,DeepMind_useto}.
In addition, TRADES~\cite{Zhang_trades} and MART~\cite{wang2020improving_MART} are effective adversarial training methods, which trains on both natural data ${x}$ and adversarial data $\Tilde{{x}}$ (the learning objectives are reviewed in Appendices~\ref{Appendix:Learning_obj_for_trades} and~\ref{Appendix:Learning_obj_for_mart} respectively).
Moreover, there are interesting analyses of PGD adversarial training, such as showing overfitting in PGD-adversarial training~\cite{rice2020overfitting}, disentangling robust and non-robust features through PGD-adversarially trained network~\cite{robust_features_nips2019_madry}, showing different feature representations by robust model and non-robust model~\cite{Tsipras19_robustness_at_odd}, and providing a new explanation for the tradeoff between robustness and accuracy of PGD-adversarial training~\cite{raghunathan2020understanding}.

\section{Friendly Adversarial Training}
\label{Section:FAT_obj_theory}
In this section, we develop a novel learning objective for friendly adversarial training (FAT). Theoretically, we justify FAT by deriving a tight upper bound of the adversarial risk. 
\subsection{Learning Objective}
\label{Section:learning_obj_FAT}
Let $\rho>0$ be a margin such that our adversarial data would be misclassified with a certain amount of confidence. 

The outer minimization still follows Eq.~\eqref{eq:adv-obj}. However, instead of generating $\xadv_i$ via inner maximization, we generate $\xadv_i$ as follows:
\begin{align*}
\label{Eq:FAT_obj_function}
\xadv_i = \argmin_{\xadv\in\epsball[x_i]} &\; \ell(f(\xadv),y_i) \nonumber \\
\ST &\; \ell(f(\xadv),y_i) - \min\nolimits_{y\in\cY}\ell(f(\xadv),y) \ge \rho.
\end{align*}
Note that the operator $\argmax$ in Eq.~\eqref{Eq:madry_inner_maximization} is replaced with $\argmin$ here, and there is a constraint on the margin of loss values (i.e., the misclassification confidence).

The constraint firstly ensures $y_i \neq \argmin\nolimits_{y\in\cY} \ell(f(\xadv),y)$ or $\xadv$ is misclassified, and secondly ensures for $\xadv$ the wrong prediction is better than the desired prediction $y_i$ by at least $\rho$ in terms of the loss value. Among all such $\xadv$ satisfying the constraint, we select the one minimizing $\ell(f(\xadv),y_i)$. Namely, we minimize the adversarial loss given that some confident adversarial data has been found. This $\xadv_i$ could be regarded as a ``friend'' among the adversaries, which is termed \emph{friendly adversarial data}. Figure~\ref{fig:lemma_show} illustrates the differences between our learning objective and the conventional minimax formulation. 

\subsection{Upper Bound on Adversarial Risk}
\label{Section:upper_bound_on_adversarial_risk}
In this subsection, we derive a tight upper bound on the adversarial risk, and provide our analysis for adversarial risk minimization. Let $X$ and $Y$ represent random variables. We employ the definition of the adversarial risk given by~\citet{Zhang_trades}, i.e., $\mathcal{R}_{\mathrm{rob}}(f) := \mathbb{E}_{(X,Y) \sim \mathcal{D}} \mathds{1} \{ \exists X' \in \epsball[X]: f(X') \neq Y\} $.
\begin{theorem} \upshape
\label{lemma}
For any classifier $f$, any non-negative surrogate loss function $\ell$ which upper bounds the $0/1$ loss, and any probability distribution $\mathcal{D}$, we have
\begin{align*}
   \mathcal{R}_{\mathrm{rob}}(f)  &\leq  \underbrace{ \mathbb{E}_{(X,Y) \sim \mathcal{D}} \ell (f(X), Y) }_{\text{For standard test accuracy}} \\ 
   &+ \underbrace{ \mathbb{E}_{(X,Y) \sim \mathcal{D}, X' \in \epsball[X, \epsilon]} \ell^{*} (f(X'), Y) }_{\text{For robust test accuracy}},
\end{align*}
where \[
    \ell^{*} = 
\begin{cases}
    \min \ell(f(X'), Y) + \rho,& \text{if } f(X') \neq Y, \\
    \max \ell(f(X'), Y),& \text{if } f(X') = Y.
\end{cases}
\]
\end{theorem}
Note that $\rho$ is the small margin such that friendly adversarial data would be misclassified with a certain amount of confidence. The proof is in Appendix~\ref{APPENDIX:Proof}. From Theorem~\ref{lemma}, our upper bound on the adversarial risk is tighter than that of conventional adversarial training, e.g.,TRADES~\cite{Zhang_trades}, where they maximize the loss regardless of model prediction, i.e., $\ell^{*} = \max \ell(f(X'), Y)$. By contrast, our bound takes the model prediction into consideration. When the model makes correct prediction on adversarial data $X'$ (i.e., $f(X') = Y$), we still maximize the loss; while the model makes wrong prediction on adversarial data $X'$ (i.e., $f(X') \neq Y$), we minimize the inner loss by violation of a small constant $\rho$. To better understand the nature of adversarial training and Theorem~\ref{lemma}, we provide supporting schematics in Figure~\ref{fig:lemma_show} and Figure~\ref{fig:theorem_show} (in Appendix~\ref{APPENDIX:Proof}).

Minimizing the adversarial risk based on our upper bound aids in fine-tuning the decision boundary using friendly adversarial data. On one side, the wrongly-predicted adversarial data have a small distance $\rho$ (in term of the loss value) from the decision boundary (e.g., ``Step $\#$10 panel'' at the bottom series in Figure~\ref{fig:motivation_fig}) so that it will not cause the severe issue of cross-over mixture but fine-tunes the decision boundary. On the other side, correctly-predicted adversarial data maintain the largest distance (in term of maximizing the loss value) from their natural data so that the decision boundary is kept far away.
\section{Key Component of FAT}
\label{section:key_component_of_FAT}
To search friendly adversarial data, 
we design an efficient early-stopped PGD algorithm called PGD-$K$-$\tau$ (Section~\ref{section:PGD-K-t}), which
could alleviate the cross-over mixture problem (Section~\ref{Section:Mixture_alleviation}) and therefore helps adversarial training. 
Note that besides PGD-$K$-$\tau$, there are other ways to search for friendly adversarial data. We show one example in Appendix~\ref{appendix:pgd_no_projection}.
\subsection{PGD-$K$-$\tau$ Algorithm}
\label{section:PGD-K-t}
\newcommand{\algorithmicbreak}{\textbf{break}}
\newcommand{\BREAK}{\STATE \algorithmicbreak}
\begin{algorithm}[tp!]
   \caption{PGD-$K$-$\tau$}
   \label{alg:PGD-k-t}
\begin{algorithmic}
   \STATE {\bfseries Input:} data ${x}\in \cX$, label $y \in \cY$, model $f$, loss function $\ell$, maximum PGD step $K$, step $\tau$, perturbation bound $\epsilon$, step size $\alpha$
   \STATE {\bfseries Output:} $\Tilde{{x}}$  
   
   \STATE $\Tilde{{x}} \gets {x}$
    \WHILE{$K > 0 $}
    \IF{$\arg\max_{i} f (\Tilde{ {x} }) \neq y$ and $\tau = 0$}
    \BREAK
    \ELSIF{$\arg\max_{i} f (\Tilde{ {x} }) \neq y$}
     \STATE $\tau \gets \tau - 1$
   \ENDIF
   \STATE $\Tilde{{x}} \gets \Pi_{\mathcal{B}[{x},\epsilon]}\big( \alpha\sign(\nabla_{\Tilde{{x}}} \ell(f(\Tilde{{x}}), y))  +  \Tilde{{x}} \big) $ 
   \STATE $K \gets K-1$
    \ENDWHILE
\end{algorithmic}
\end{algorithm}
In Algorithm~\ref{alg:PGD-k-t}, $\Pi_{\mathcal{B}[{x},\epsilon]}$ is the projection operator that projects adversarial data $\Tilde{{x}}$ onto the $\epsilon$-norm ball centered at ${x}$, and $\arg\max_{i} f (\Tilde{ {x} })$ returns the predicted label of adversarial data $\Tilde{{x}}$, where $f (\Tilde{ {x} }) = \big (f^{i}(\Tilde{{x}}) \big )^{\top}_{i=0,\dots,C-1}$ measures the probabilistic predictions over $C$ classes. Unlike the conventional PGD-$K$ generating adversarial data by maximizing the loss function $\ell$ regardless of model prediction, our PGD-$K$-$\tau$ generates the adversarial data which takes model prediction into consideration.

Algorithm~\ref{alg:PGD-k-t} returns the misclassified adversarial data with small loss values or correctly classified adversarial data with large loss values. Step $\tau$ controls the extent of loss minimization when misclassified adversarial data are found. When $\tau$ is larger, the misclassified adversarial data with slightly larger loss values are returned, and vice versa. $\tau \times \alpha$ is an approximation to $\rho$ in our learning objective. Note that when $\tau = K$, the conventional PGD-$K$ is the special case of our PGD-$K$-$\tau$. As $\tau$ is an important hyper-parameter of PGD-$K$-$\tau$ for FAT (Section~\ref{Sec:Friendly-AdvTraining}), we discuss how to select $\tau$ in Sections~\ref{Section:slippery_step_tau} and \ref{Section:Computaional_efficient} in detail.

\subsection{PGD-$K$-$\tau$ Alleviates Cross-over Mixture}
\label{Section:Mixture_alleviation}
In deep neural networks, the cross-over mixture problem may not trivially appear in the original input space, but occur in the output of the intermediate layer. Our proposed PGD-$K$-$\tau$ is an effective solution to overcome this problem, which leads to successful adversarial training.

\begin{figure}[tp!]
    \centering
    \includegraphics[scale=0.18]{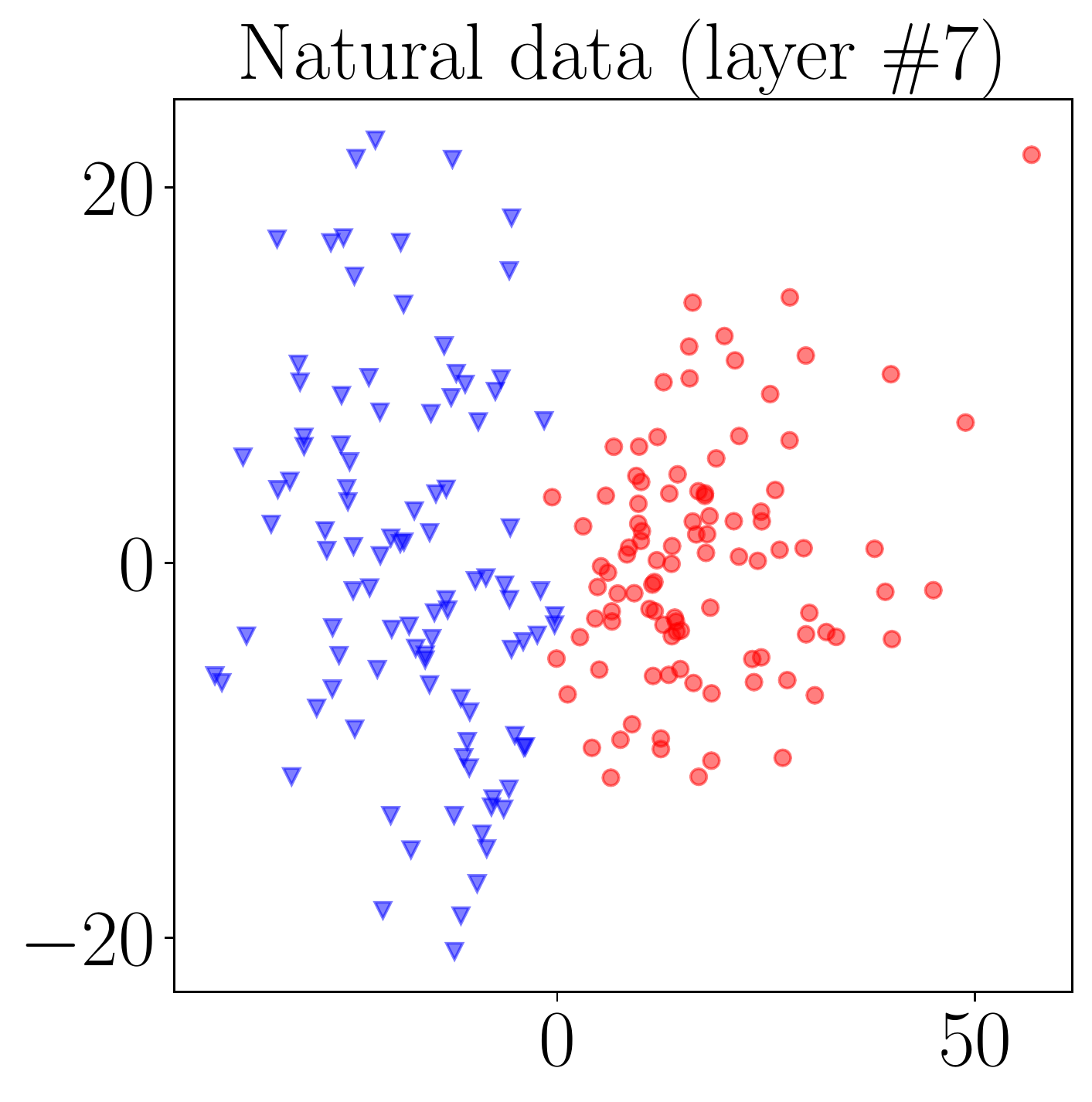}
    \includegraphics[scale=0.18]{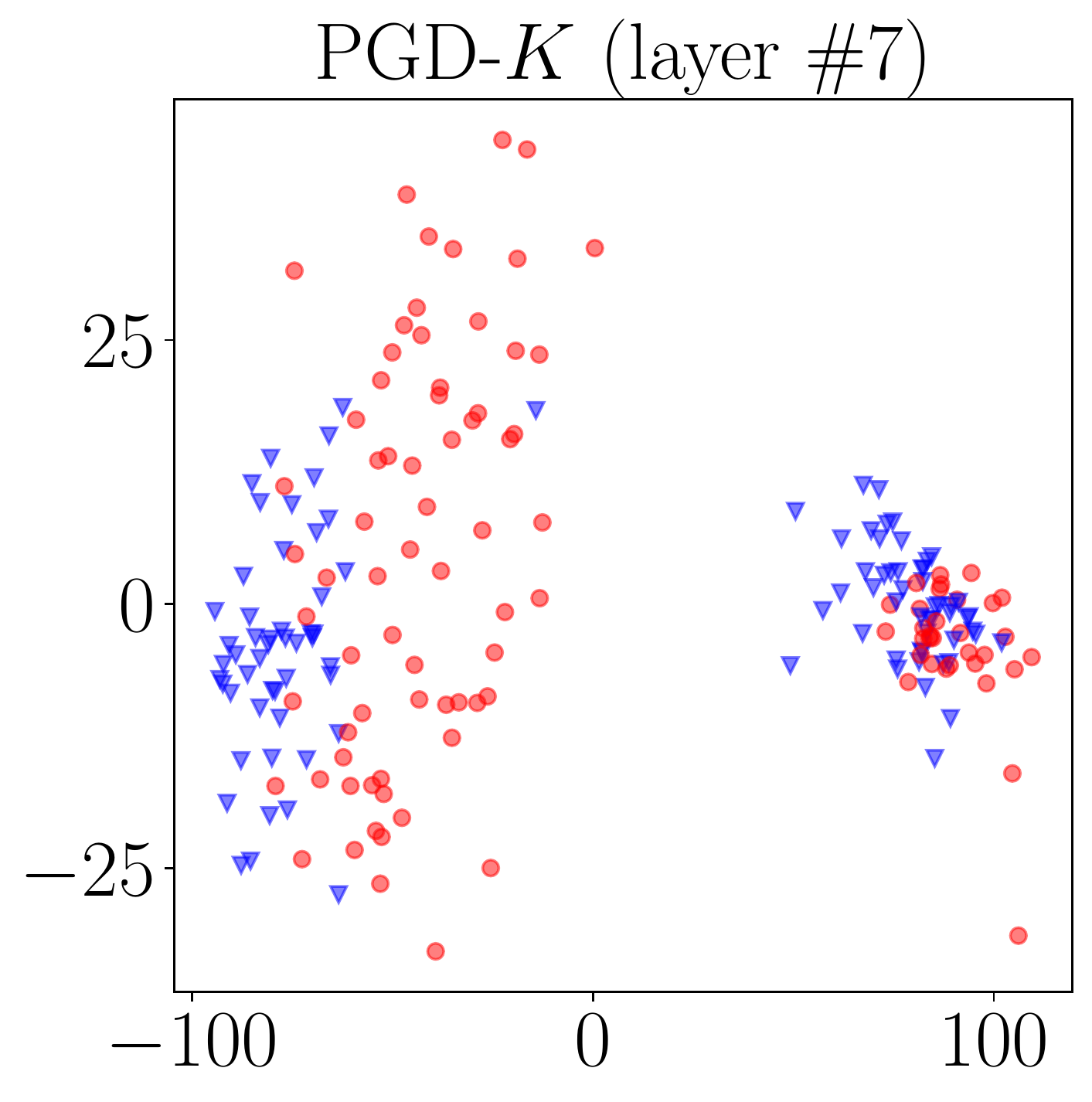}
    \includegraphics[scale=0.18]{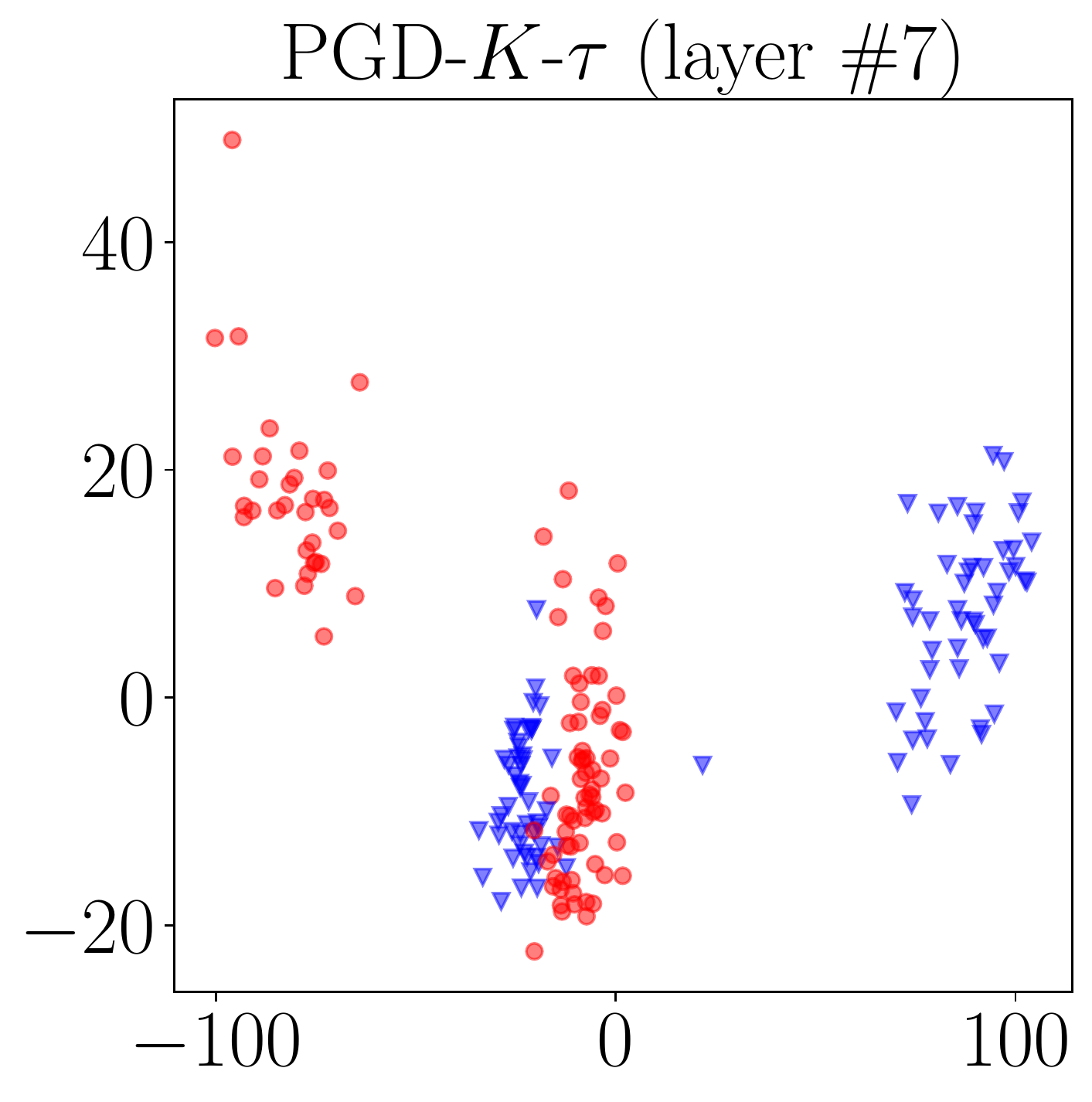}
    \caption{Left: layer $\#7$'s output distribution on natural data (not mixed). Middle: layer $\#7$'s output distribution on adversarial data generated by PGD-$20$ (significantly mixed). Right: layer $\#7$'s output distribution on friendly adversarial data generated by PGD-$20$-$0$ (not significantly mixed).}
    \label{fig:overshoot_cifar}
    \vspace{-3mm}
\end{figure}

In Figure~\ref{fig:overshoot_cifar}, we trained an 8-layer convolutional neural network (6 convolutional layers and 2 fully-connected layers, namely, Small CNN) on images of two selected classes in CIFAR-10. We conducted a warm-up training using natural training data, then included their adversarial variants generated by PGD-$20$ (middle panel) and PGD-$20$-$0$ (right panel), where $\tau = 0$ means PGD iterations stop immediately once adversarial data are wrongly predicted by the current network. 

Figure~\ref{fig:overshoot_cifar} shows the output distribution of layer $\#7$ by principal component analysis (PCA, \cite{abdi2010_PCA}), which projects high-dimensional output into a two-dimensional subspace. As shown in Figure~\ref{fig:overshoot_cifar}, the output distribution on natural data (left panel) of the intermediate layer is clearly not mixed. By contrast, the conventional PGD-$K$ (middle panel) leads to severe mixing between outputs of adversarial data with different classes. It is more difficult to fit these mixed adversarial data, which leads to inaccurate classifiers. By comparing with PGD-$K$, our PGD-$K$-$\tau$ (right panel) could greatly overcome the mixture issue of adversarial data. Thus, it helps the training algorithm to return an accurate classifier while ensuring adversarial robustness.

To further justify the above fact, we plot output distributions of other layers, e.g., layer $\#6$ and layer $\#8$. Moreover, we train a Wide ResNet (WRN-40-4)~\cite{zagoruyko2016WRN} on 10 classes and randomly select 3 classes for illustrating their output distributions of its intermediate layers. Instead of PCA, we also use a non-linear technique for dimensionality reduction, i.e., t-distributed stochastic neighbor embedding (t-SNE)~\cite{maaten2008visualizing_tsne} to visualize output distributions of different classes. All these can be found in Appendix~\ref{appendix:mixture_alleviation}.

\section{Realization of FAT}\label{Sec:Friendly-AdvTraining}
\begin{algorithm}[h!]
   \caption{Friendly Adversarial Training (FAT)}
   \label{alg:FAT}
\begin{algorithmic}
   \STATE {\bfseries Input:} network $f_{\mathbf{\theta}}$, training dataset $S = \{(\bx_i, y_i) \}^{n}_{i=1}$, learning rate $\eta$, number of epochs $T$, batch size $m$, number of batches $M$
   \STATE {\bfseries Output:} adversarially robust network $f_{\mathbf{\theta}}$  
  \FOR{epoch $= 1$, $\dots$, $T$}
    \FOR {mini-batch $=1$, $\dots$, $M$ }
    \STATE Sample a mini-batch $\{(\bx_i, y_i) \}^{m}_{i=1}$ from $S$
        \FOR{$i = 1$, $\dots$, $m$ (in parallel) }
         \STATE Obtain adversarial data $\bxtidle_i$of $\bx_i$ by Algorithm~\ref{alg:PGD-k-t}
         \ENDFOR
    \STATE $\mathbf{\theta} \gets \mathbf{\theta} - \eta \frac{1}{m} \sum^{m}_{i - 1} \nabla_{\mathbf{\theta}}\ell(f_{\mathbf{\theta}}(\bxtidle_i), y_i)   $
  \ENDFOR
 \ENDFOR
\end{algorithmic}
\end{algorithm}

Based on the proposed PGD-$K$-$\tau$, we have a new algorithm termed FAT (Algorithm~\ref{alg:FAT}). FAT treats the standard adversarial training~\cite{Madry_adversarial_training} as a special case when we set $\tau = K$ in Algorithm~\ref{alg:PGD-k-t}. Besides, we also design FAT for TRADES (Appendix~\ref{Appendix:fat_for_trades}) and FAT for MART (Appendix~\ref{Appendix:fat_for_mart_realization}), making two effective adversarial training methods (TRADES~\cite{Zhang_trades} and MART~\cite{wang2020improving_MART}) special cases when $\tau = K$. 
Since the essential component of FAT is PGD-$K$-$\tau$, we should discuss the effects of step $\tau$ w.r.t. standard accuracy and adversarial robustness (Section~\ref{Section:slippery_step_tau}) and computational efficiency (Section~\ref{Section:Computaional_efficient}).  Besides, we should discuss the relation between FAT and curriculum learning (Section~\ref{section:curriculum_learning}), since FAT is a progressive training strategy. 
It is worth noting that \citet{Chawin_ATES} independently propose adversarial training with early stopping (ATES). Along with our FAT, ATES corroborates the new formulation (Section~\ref{Section:learning_obj_FAT}) for adversarial training. 

\subsection{Selection of Step $\tau$}
\label{Section:slippery_step_tau}
As shown in Section~\ref{section_PGD}, the conventional PGD-$K$ is a special case, when step $\tau = K$ in PGD-$K$-$\tau$. Thus, standard adversarial training is a special case of FAT. Here, we investigate how step $\tau$ affects the performance of FAT empirically, and summarize that larger $\tau$ may not increase adversarial robustness but hurt the standard test accuracy. Detailed experimental setups of Figure~\ref{fig:tau_effect} are in Appendix~\ref{APPENDIX:selection_of_slippery_step}.

Figure~\ref{fig:tau_effect} shows that, with the increase of $\tau$, the standard test accuracy for natural data decreases significantly; while the robust test accuracy for adversarial data increases at smaller values of $\tau$ but reaches its plateau at larger values of $\tau$. For example, when $\tau$ is bigger than 2, the standard test accuracy continues to decrease with larger $\tau$. However, the robust test accuracy begins to maintain a plateau for both Small CNN and ResNet-18. Such observation manifests that a larger step $\tau$ may not be necessary for adversarial training. Namely, it may not increase adversarial robustness but hurt the standard accuracy. This reflects a trade-off between the standard accuracy and adversarially robust accuracy \cite{Tsipras19_robustness_at_odd} and suggests that our step $\tau$ helps manage this trade-off. 

$\tau$ can be treated as a hyper-parameter. Based on the observations in Figure~\ref{fig:tau_effect}, it is enough to select $\tau$ from the set $\{0,1,2,3\}$. Note that the size of the set is also influenced by step size $\alpha$ and maximum PGD step $K$. In Section~\ref{section:SOTA_results}, we use $\tau$ to fine-tune the performance of FAT.

\subsection{Smaller $\tau$ is Computationally Efficient}
\label{Section:Computaional_efficient}
Adversarial training is time-consuming since it needs multiple backward propagations (BPs) to produce adversarial data. The time-consuming factor depends on the number of BPs used for generating adversarial data~\cite{Ali_NIPS19_adversarial_training_for_free,Lu_yiping_NIPS19_yopo,wong2020fast_zico_kolter,babu_2020_CVPR}.

\begin{figure}[tp!]
    \centering
    \includegraphics[scale=0.185]{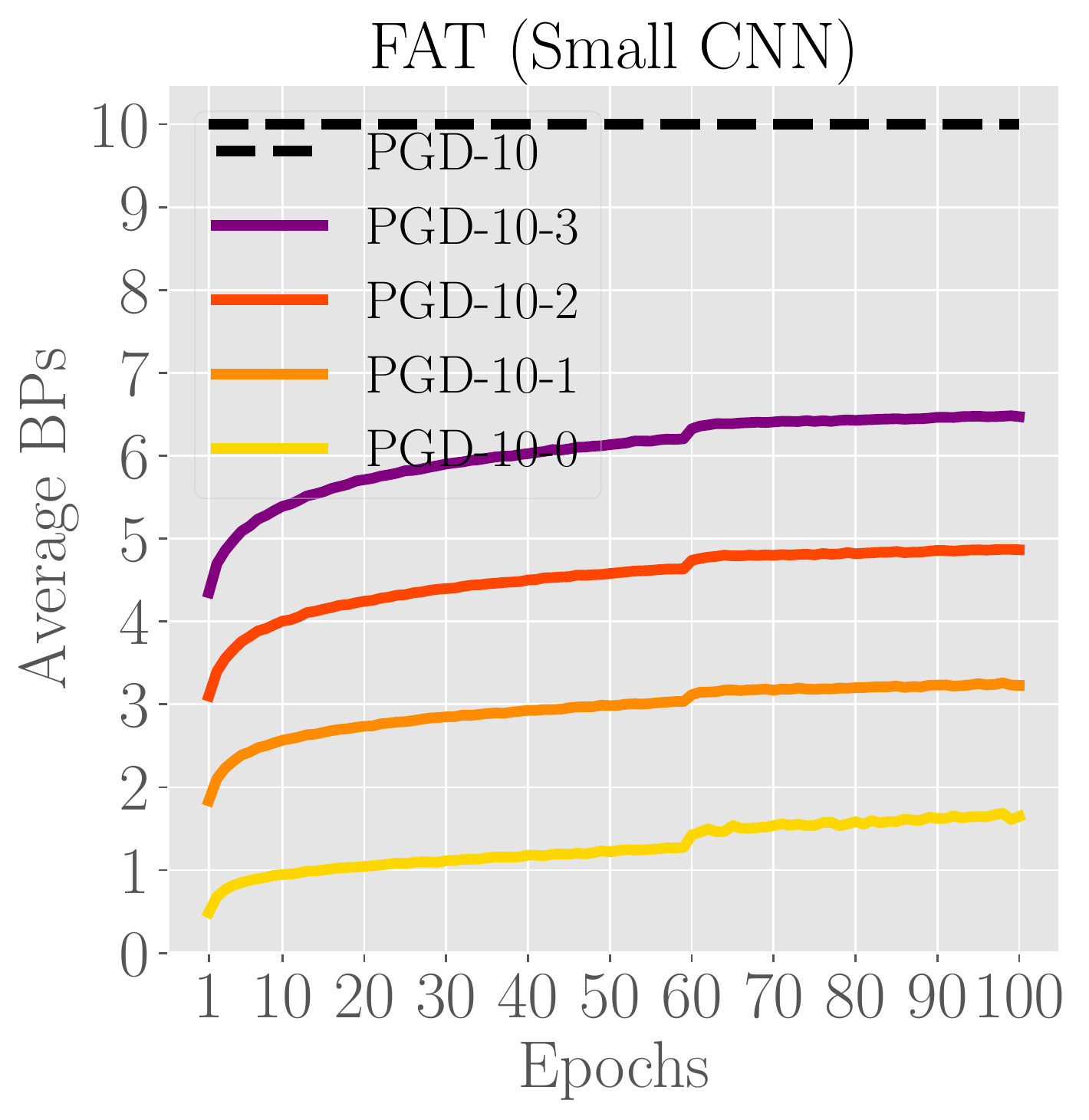}
    \includegraphics[scale=0.185]{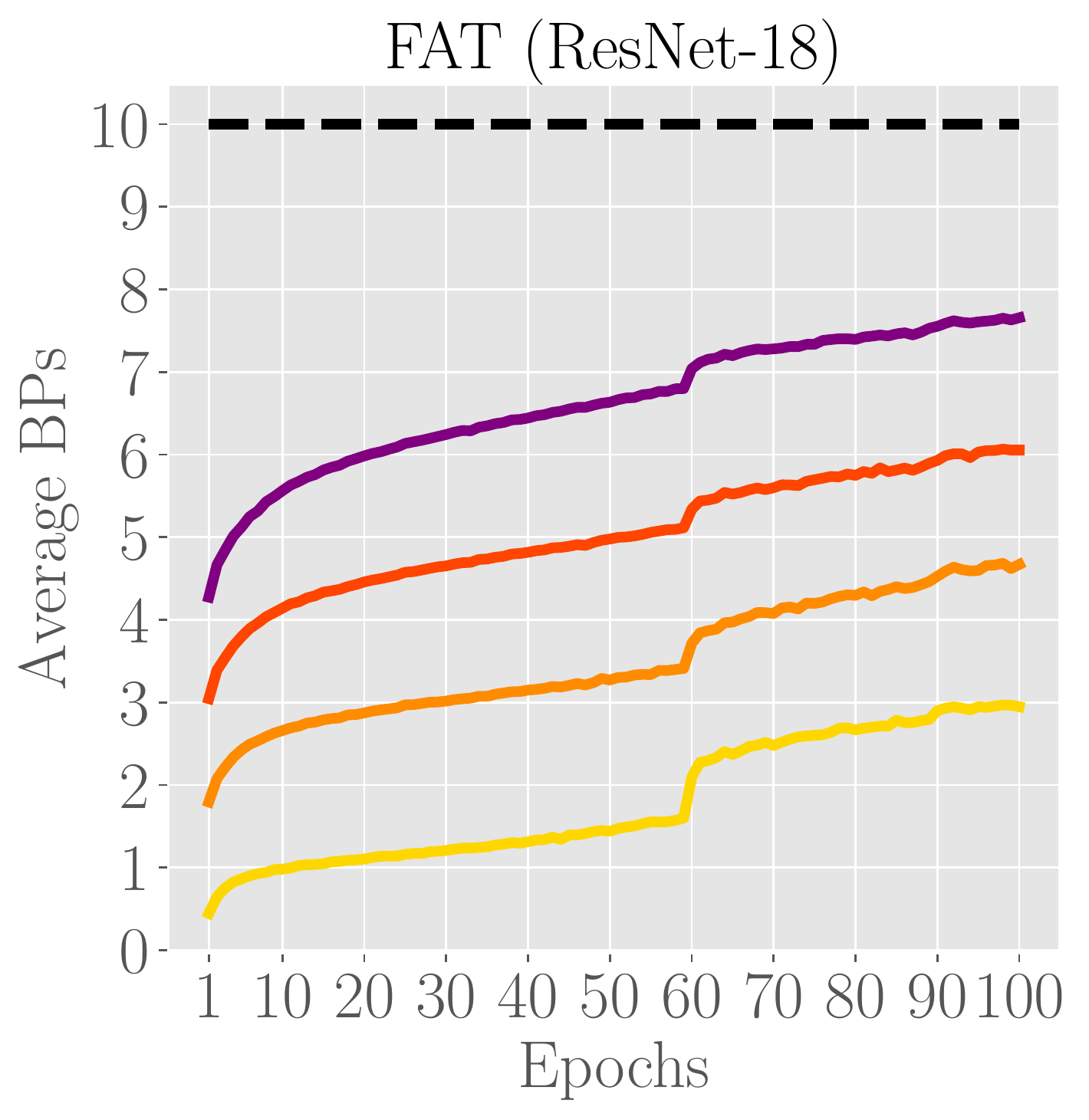} 
    \includegraphics[scale=0.185]{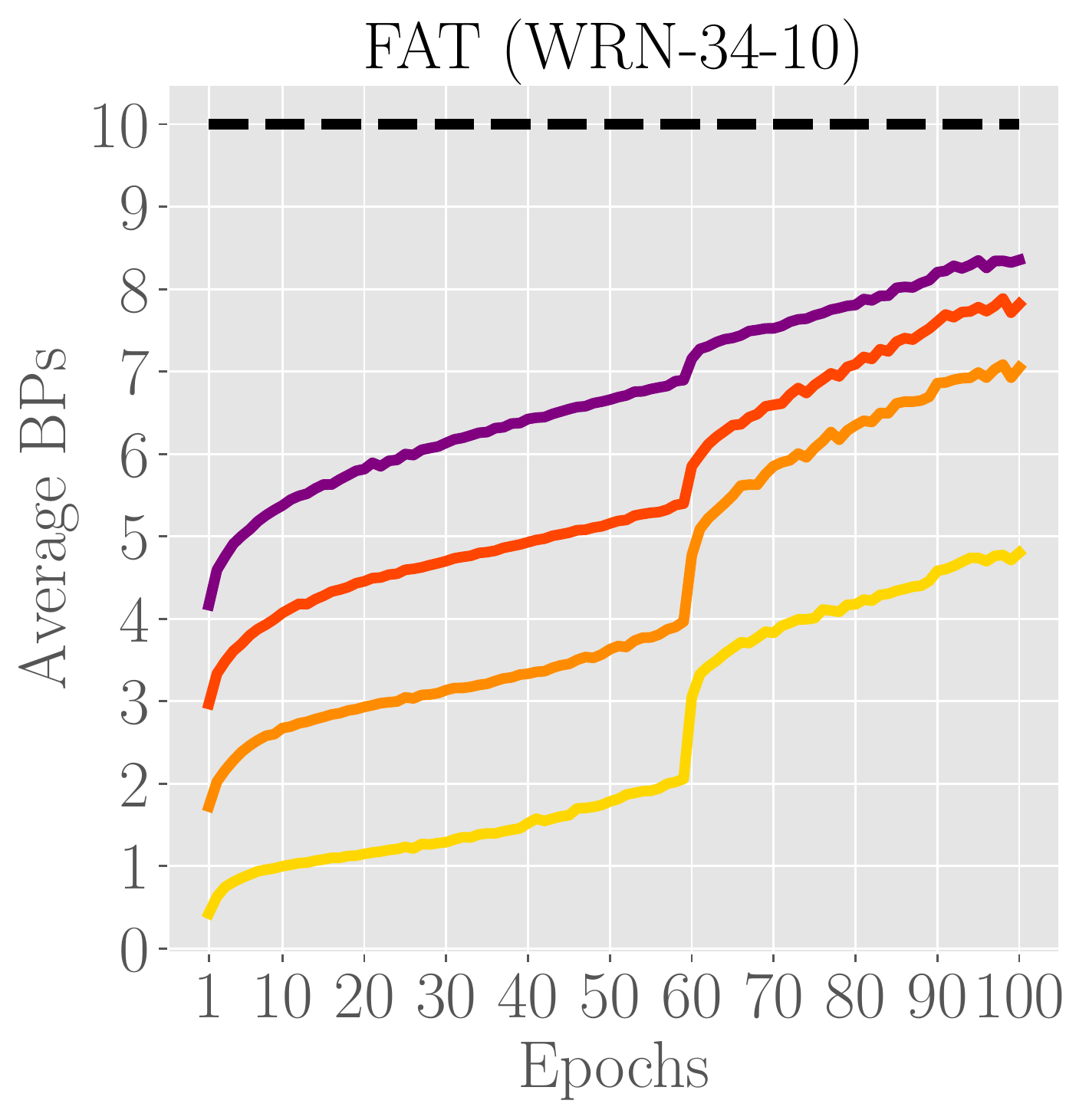}\\
    \includegraphics[scale=0.185]{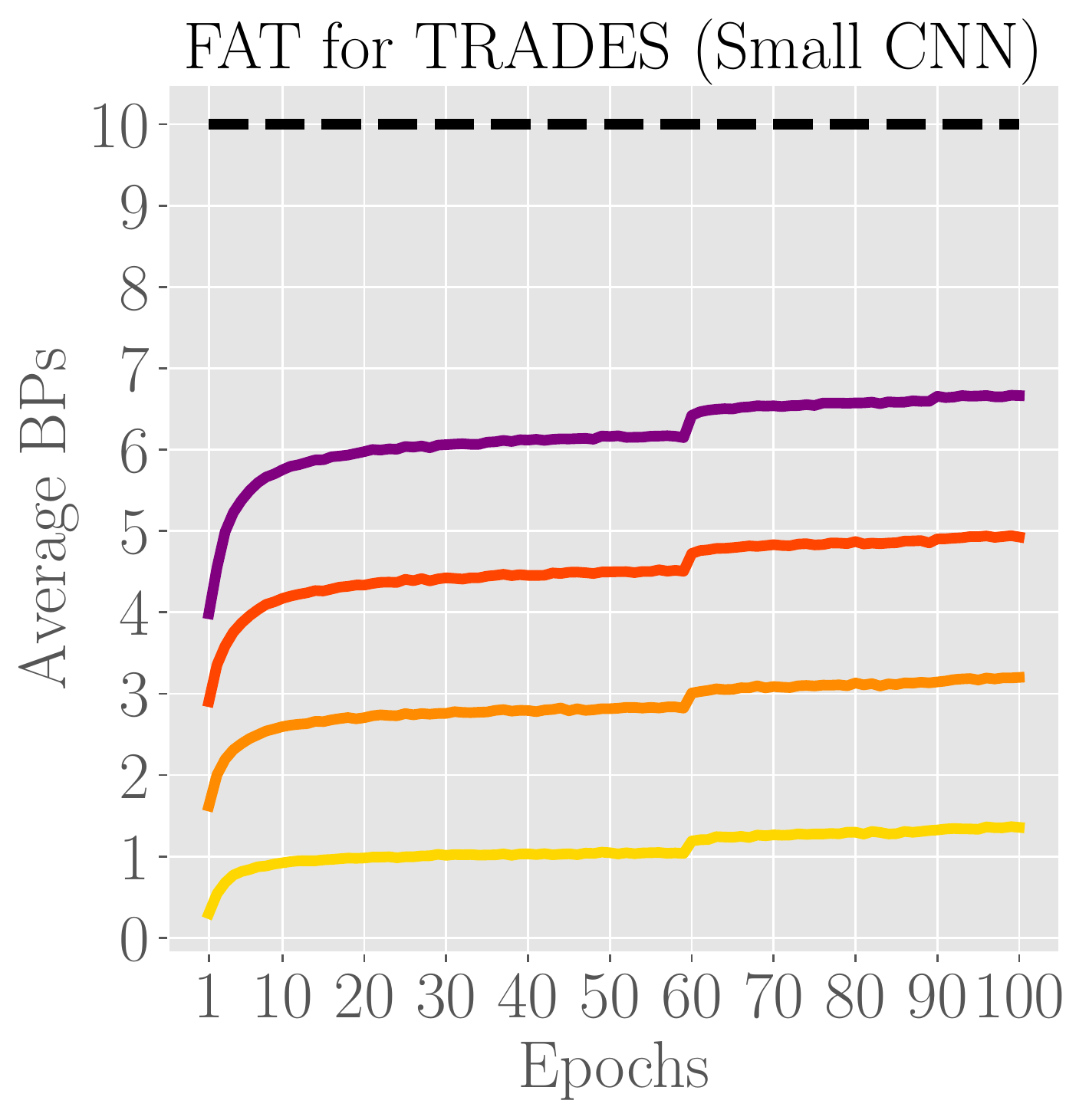}
    \includegraphics[scale=0.185]{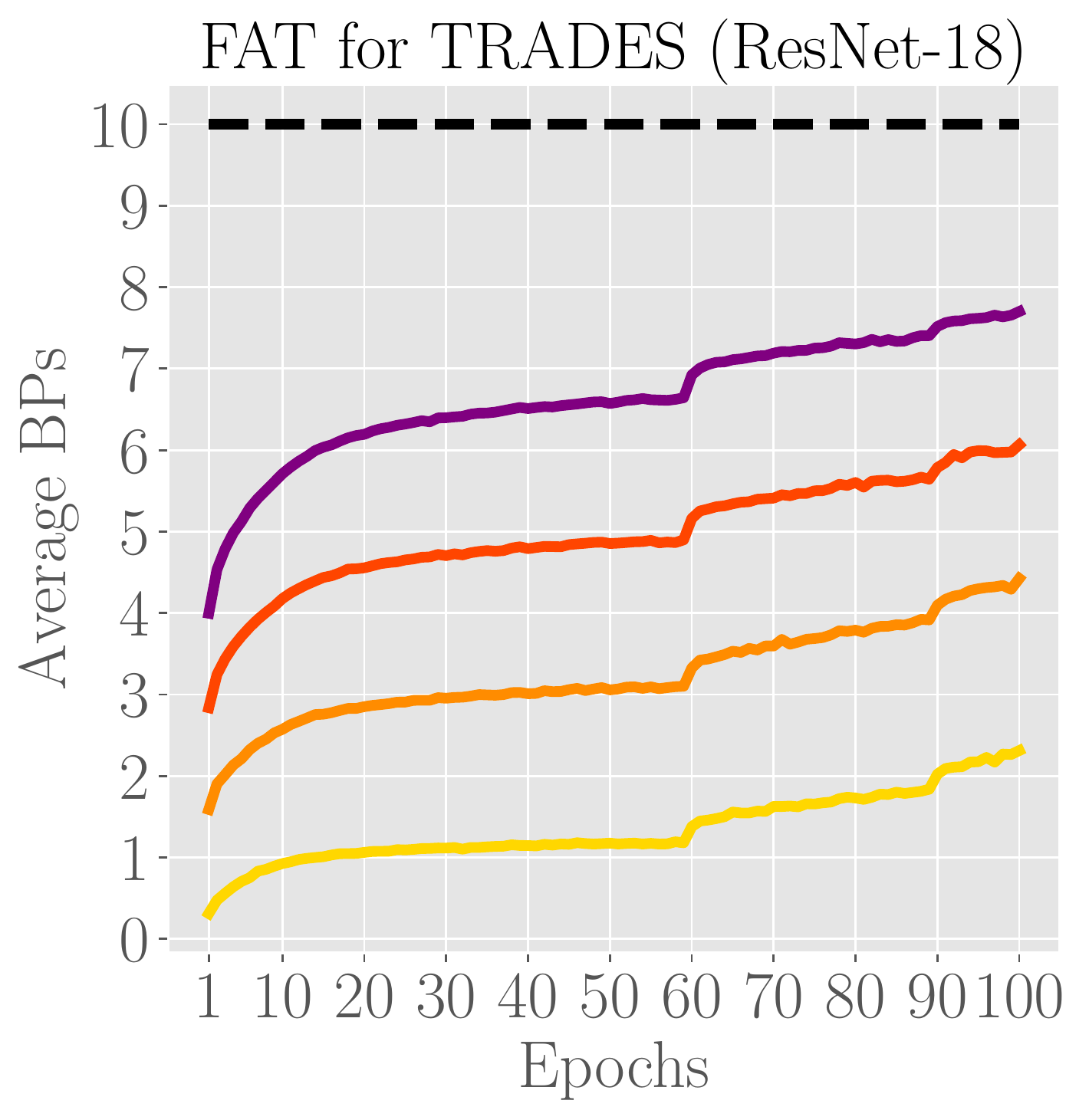}
    \includegraphics[scale=0.185]{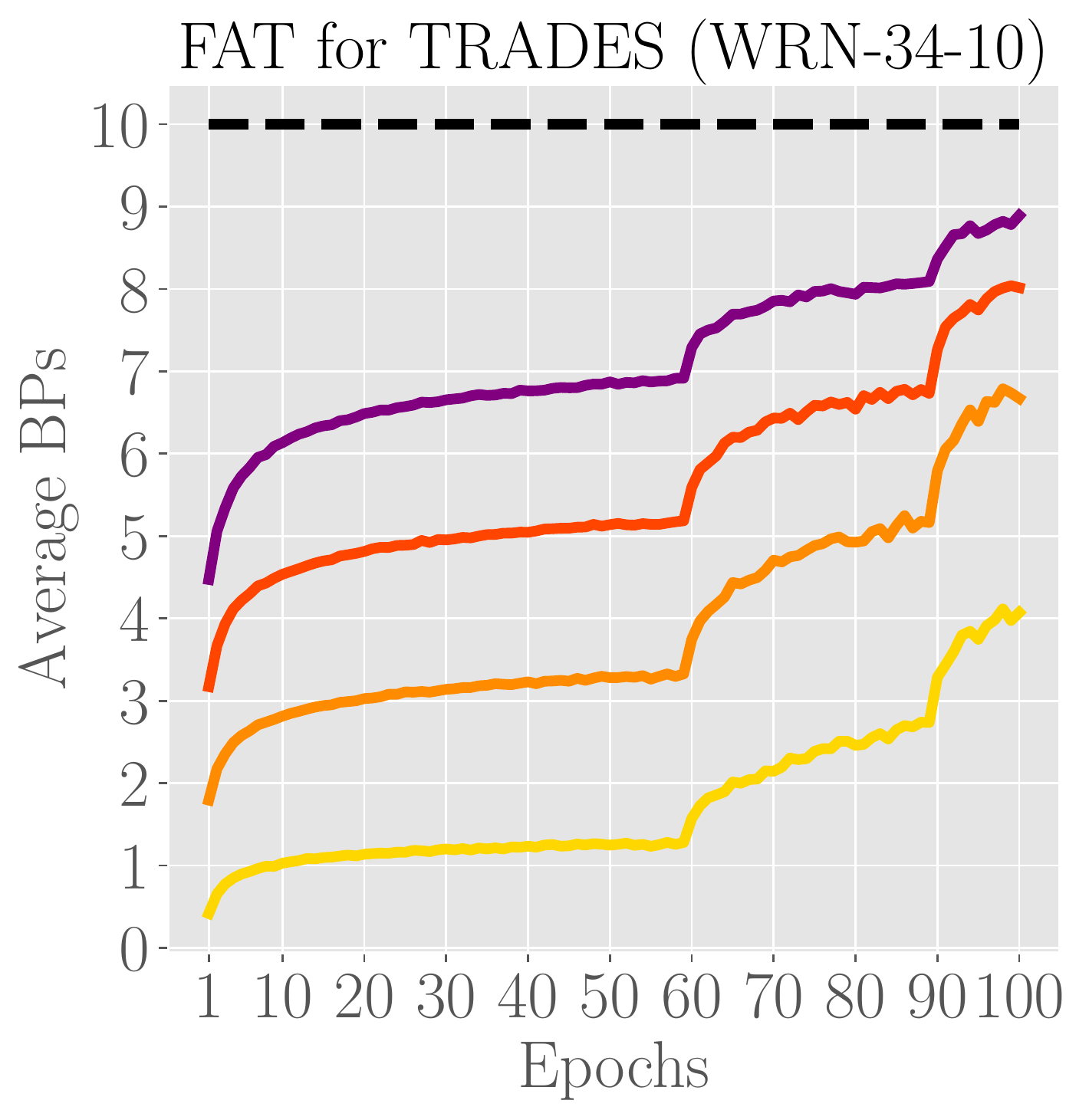}
    \caption{Training statistics on the average number of BPs needed for generating adversarial data. 
    Top three panels are adversarial training FAT on three networks, Small CNN (top left), ResNet-18 (top middle) and WRN-34-10 (top right).
    Bottom three panels are adversarial training FAT for TRADES on three networks, Small CNN (bottom left), ResNet-18 (bottom middle) and WRN-34-10 (bottom right).}
    \label{fig:computational_efficiency}
    \vspace{-3mm}
\end{figure}

Our FAT uses PGD-$K$-$\tau$ to generate adversarial data, and PGD-$K$-$\tau$ is early-stopped. This implies that FAT is computationally efficient, since FAT does not need to compute maximum $K$ BPs on each mini-batch. To illustrate this, we count the number of BPs for generating adversarial data during training. The training setup is the same as the one in Section~\ref{Section:slippery_step_tau}, but we only choose $\tau$ from $\{0, 1, 2, 3\}$ and train for 100 epochs with learning rate divided by 10 at epochs 60 and 90. In Figure~\ref{fig:computational_efficiency}, we compare the standard adversarial training~\cite{Madry_adversarial_training} (dashed line) with our FAT (solid line) and adversarial training TRADES~\cite{Zhang_trades} (dashed line) with our FAT for TRADES (solid line, detailed in Algorithm~\ref{alg:FAT_for_TRADES} in Appendix~\ref{Appendix:fat_for_trades}). For each epoch, we compute average BPs over all training data for generating the adversarial counterpart. 

\begin{figure}[tp!]
	\centering
	\includegraphics[scale=0.27]{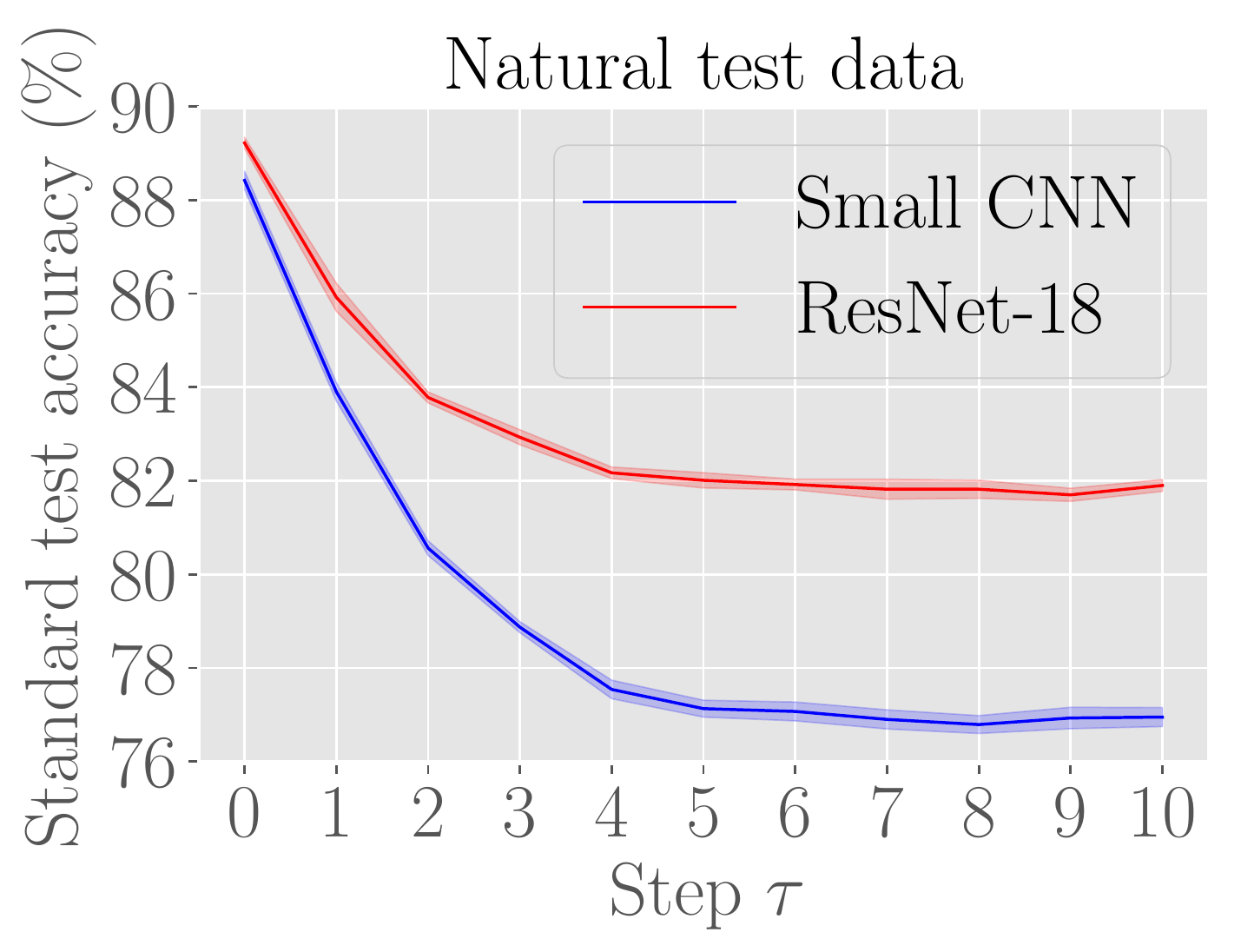}
	\includegraphics[scale=0.27]{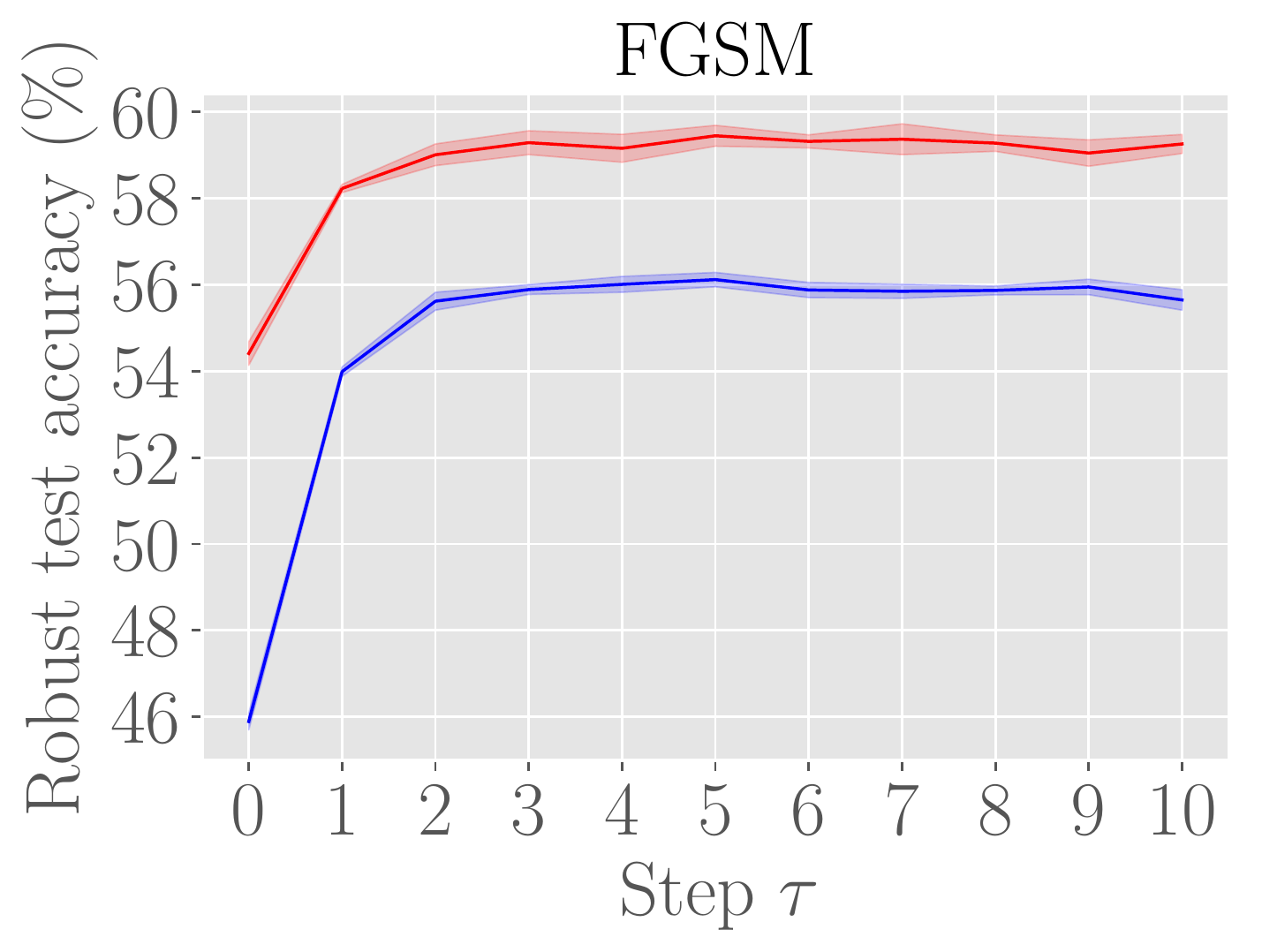} \\
	\includegraphics[scale=0.27]{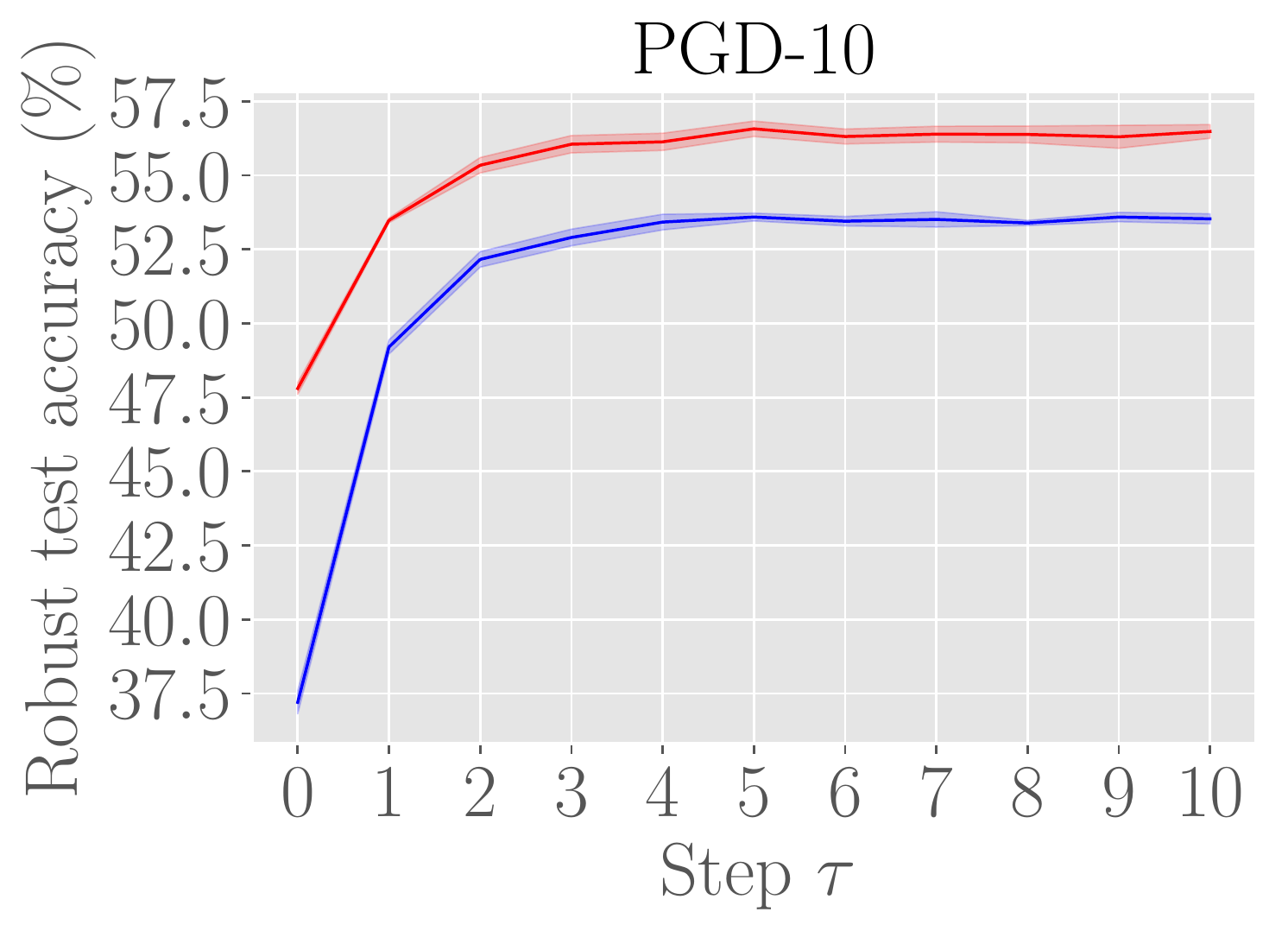}
	\includegraphics[scale=0.27]{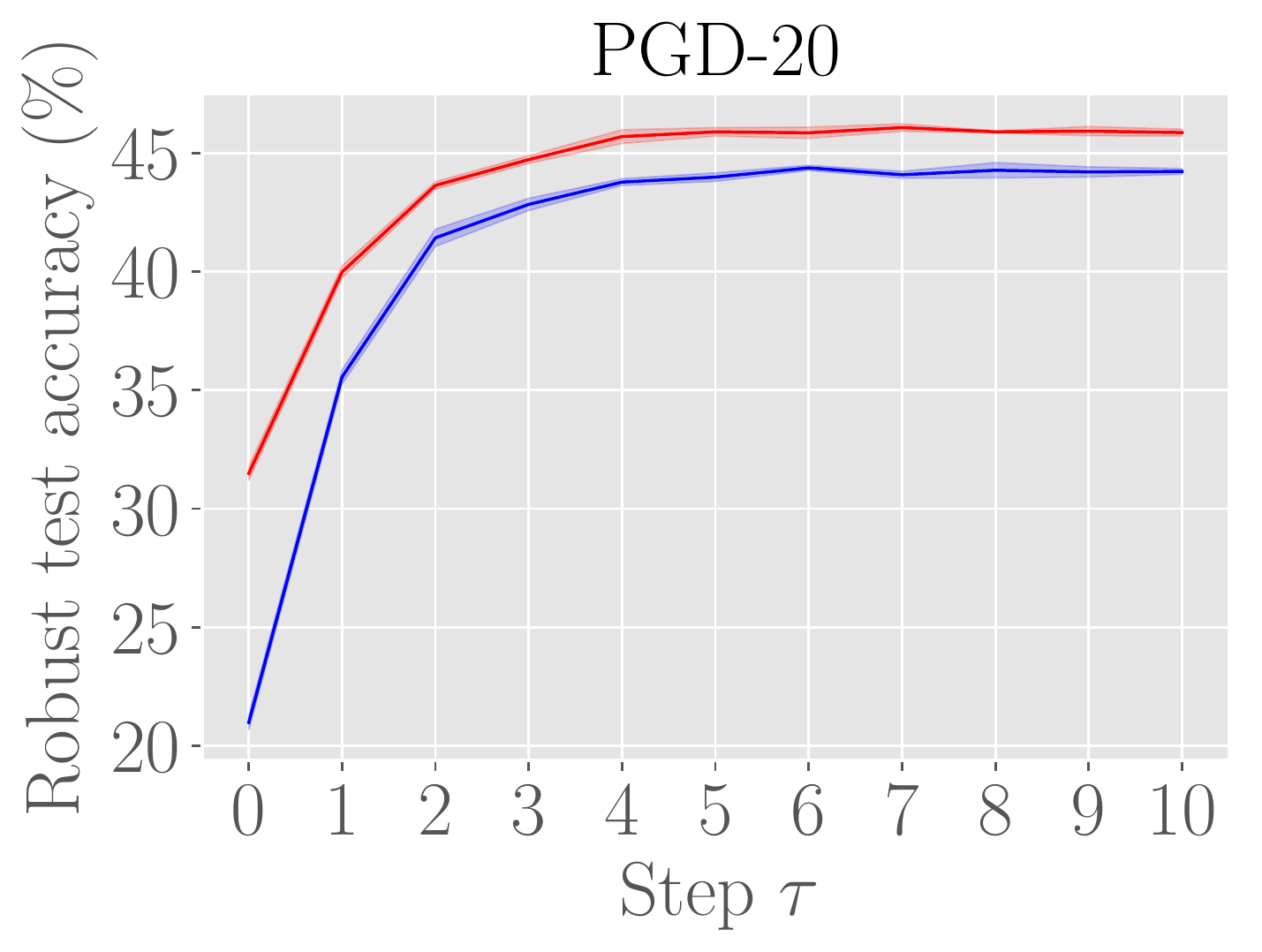}\\
	\includegraphics[scale=0.27]{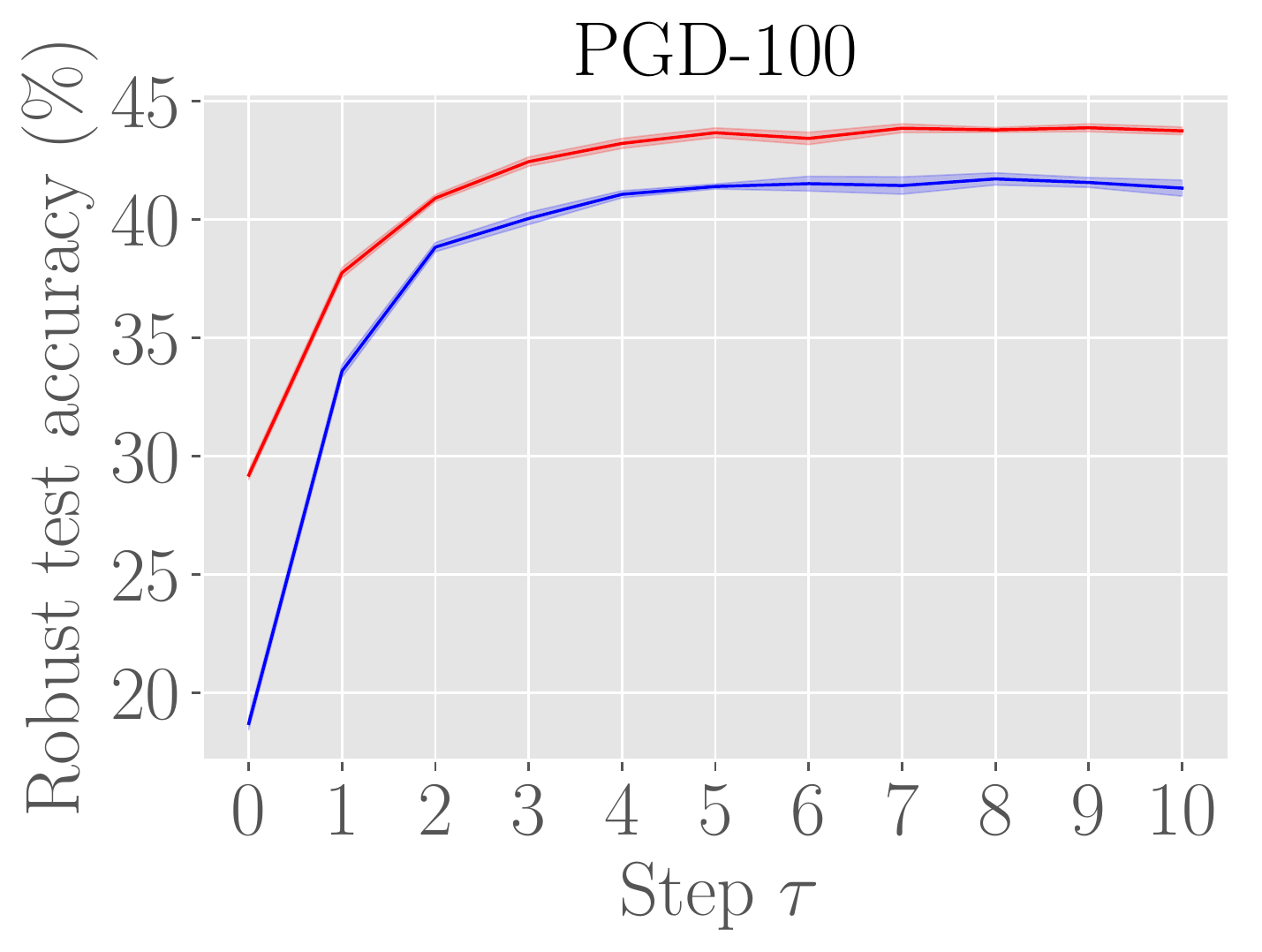}
	\includegraphics[scale=0.27]{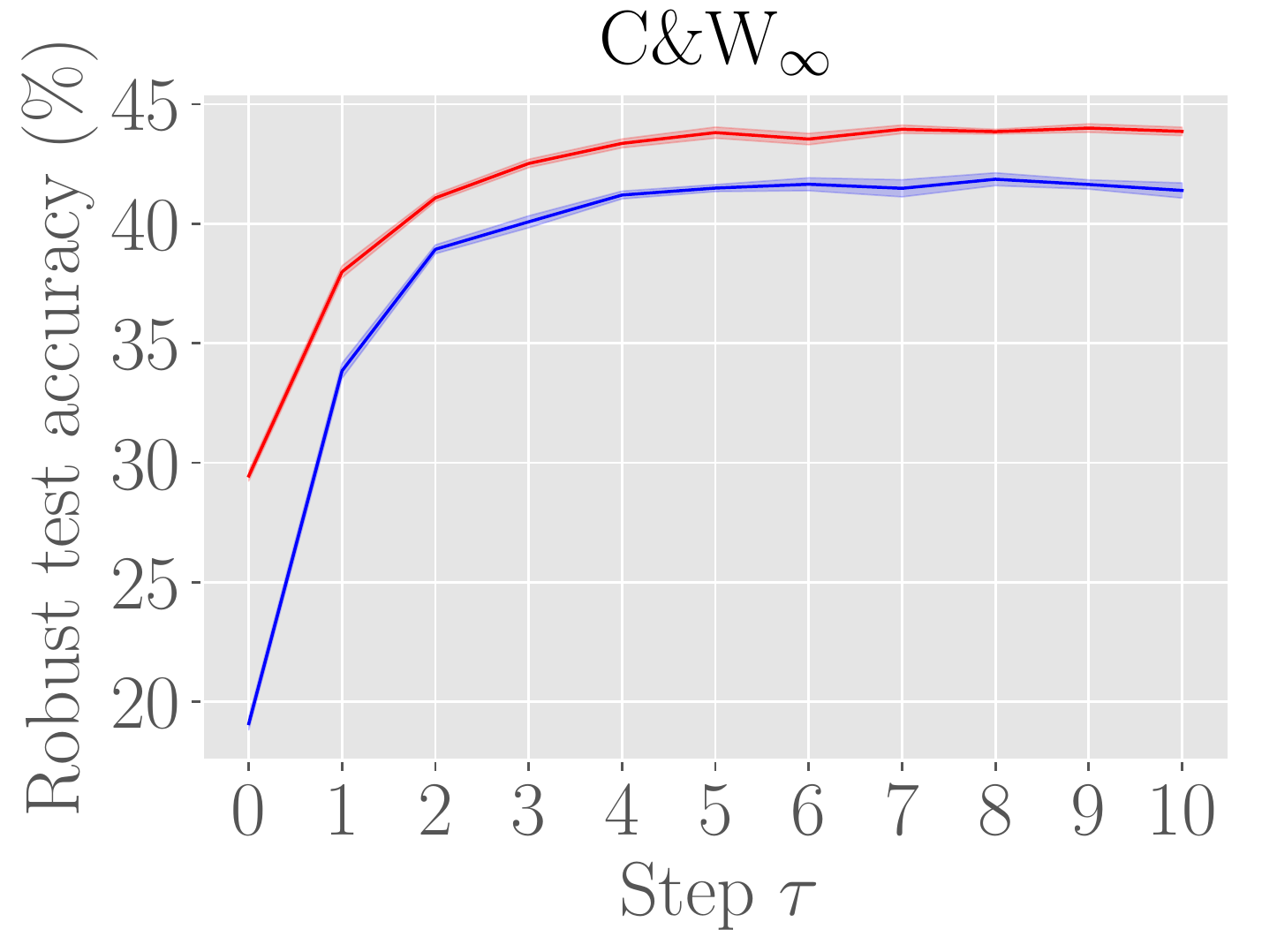}
	\caption{We conduct our adversarial training FAT with various values of step $\tau$ on two networks Small CNN (blue line) and ResNet-18 (red line). We evaluate the adversarial training performance according to networks'
		standard test accuracy for natural test data and robust test accuracy for adversarial test data generated by FGSM, PGD-10, PGD-20, PGD-100 and C$\&$W attack. We report the median test accuracy and its standard deviation as the shaded color over 5 repeated trials of adversarial training.}
	\label{fig:tau_effect}
	\vspace{-3mm}
\end{figure}

Figure~\ref{fig:computational_efficiency} shows that conventional adversarial training uses PGD-$K$ which takes $K$ BPs for generating adversarial data in each mini-batch. 
By contrast, our adversarial training FAT that uses PGD-$K$-$\tau$ significantly reduces the number of required BPs. 
In addition, with the smaller $\tau$, FAT needs less BPs on average for generating adversarial data. The magnitude of $\tau$ controls the number of extra BPs, once misclassified adversarial data is found.

Moreover, there are some interesting phenomena observed in our FAT (or FAT for TRADES). As the training progresses, the number of BPs gradually increases. This shows that more steps are needed by PGD to find misclassified adversarial data. 
This signifies that it is increasingly difficult to find adversarial data that misclassifies the model. Thus, DNNs become more and more adversarially robust over training epochs. In addition, there is a slight surge in average BPs at epochs 60 and 90, where we divided the learning rate by 10. This means that the robustness of the model gets substantially improved at epochs 60 and 90. It is a common trick to decrease the learning rate over training DNNs for good standard accuracy~\cite{he2016deep}. Figure~\ref{fig:computational_efficiency} confirms that it is similarly meaningful to decrease the learning rate during adversarial training. 

\subsection{Relation to Curriculum Learning}
\label{section:curriculum_learning}
Curriculum learning~\cite{bengio2009curriculum} is a machine learning strategy that gradually makes the learning task more difficult. 
Curriculum learning is shown effective in improving standard generalization and providing faster convergence~\cite{bengio2009curriculum}.

In adversarial training, curriculum learning can also be used to improve adversarial robustness. Namely, DNNs learn from milder adversarial data first, and gradually adapt to stronger adversarial data. There are different ways to determine the hardness of adversarial data. For example, curriculum adversarial training (CAT) uses the perturbation step $K$ of PGD as the hardness measure~\cite{Cai_CAT}. Dynamic adversarial training (DAT) uses their proposed criterion, the first-order stationary condition (FOSC), as the hardness measure~\cite{Wang_Xingjun_MA_FOSC_DAT}. However, both methods do not have a principled way to decide when the hardness should be increased during training. To increase the hardness at the right time, both methods need domain knowledge to fine-tune the curriculum training sequence. For example, CAT needs to decide when to increase step $K$ in PGD over training epochs; while DAT needs to decide FOSC for generating adversarial data at different training stages.

Our FAT can also be regarded as a type of curriculum training. As shown in Figure~\ref{fig:computational_efficiency}, as the training progresses, more and more backward propagations (BPs) are needed to generate adversarial data to fool the classifier. 
Thus, more and more PGD steps are needed to generate adversarial data. 
Meanwhile, the network gradually and automatically learns from stronger and stronger adversarial data (adversarial data generated by more and more PGD steps). Differently from CAT and DAT, FAT could automatically increase the hardness of friendly adversarial data based on the model's predictions. As a result, Table~\ref{table:sota_result_madry} in Section~\ref{section:SOTA_results} shows that empirical results of FAT can outperform the best results in CAT~\cite{Cai_CAT}, DAT~\cite{Wang_Xingjun_MA_FOSC_DAT} and standard Madry's adversarial training (not a curriculum learning)~\cite{Madry_adversarial_training}. 
Thus, attacks which do not kill training indeed make adversarial learning stronger.

\section{Experiments}
\label{Section:experiments}
To evaluate the efficacy of FAT, we firstly use CIFAR-10~\cite{krizhevsky2009learning_cifar10} and SVHN~\cite{netzer2011reading_SVHN} datasets to verify that FAT can help achieve a larger perturbation bound $\epsilon_{train}$. Then, we train Wide ResNet~\cite{zagoruyko2016WRN} on the CIFAR-10 dataset to achieve state-of-the-art results.
\subsection{FAT can Enable Larger Perturbation Bound $\epsilon_{train}$}
\label{section:fat_enable_lager_epsilon}
\begin{figure}[tp!]
    \centering
    \includegraphics[scale=0.25]{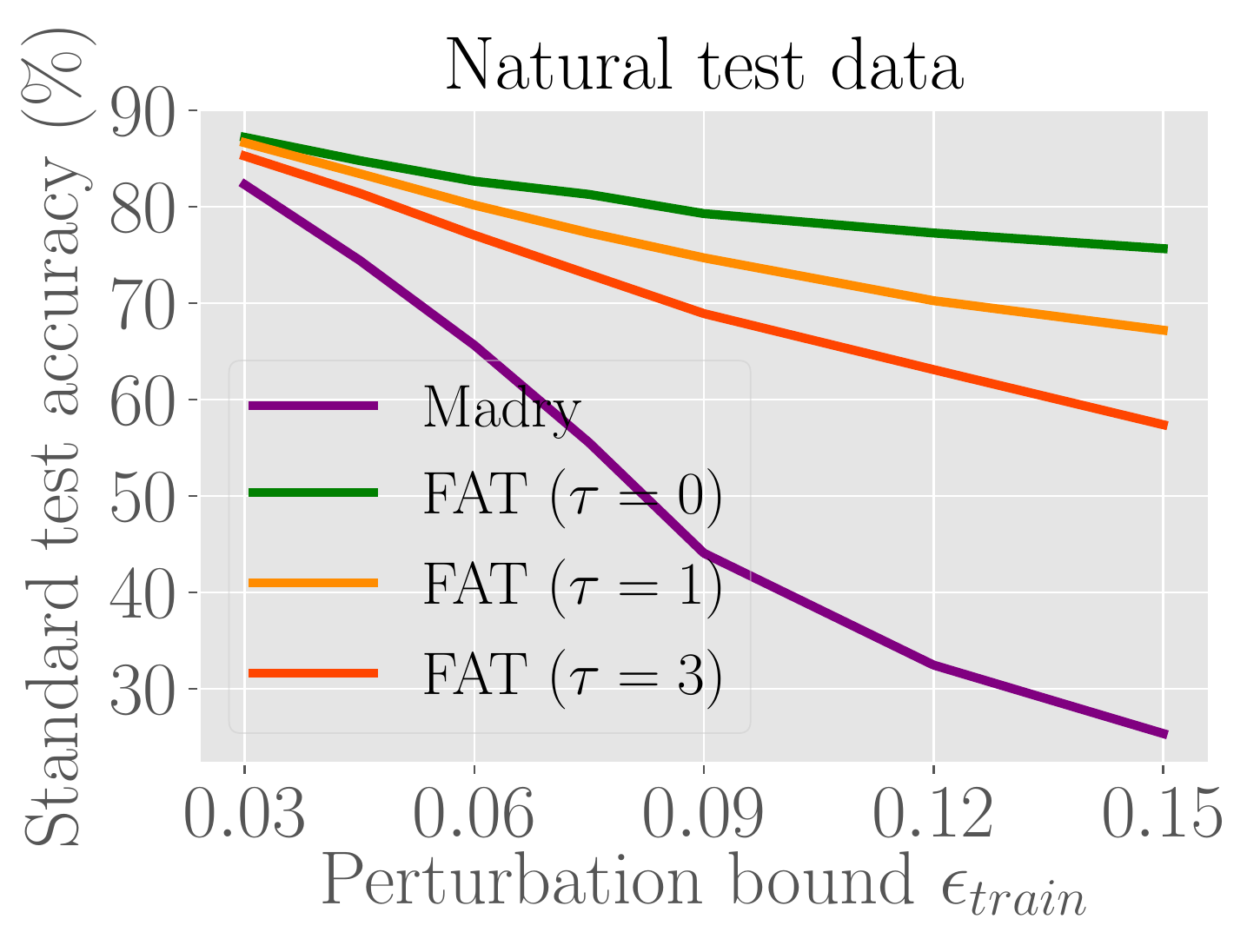}
    \includegraphics[scale=0.25]{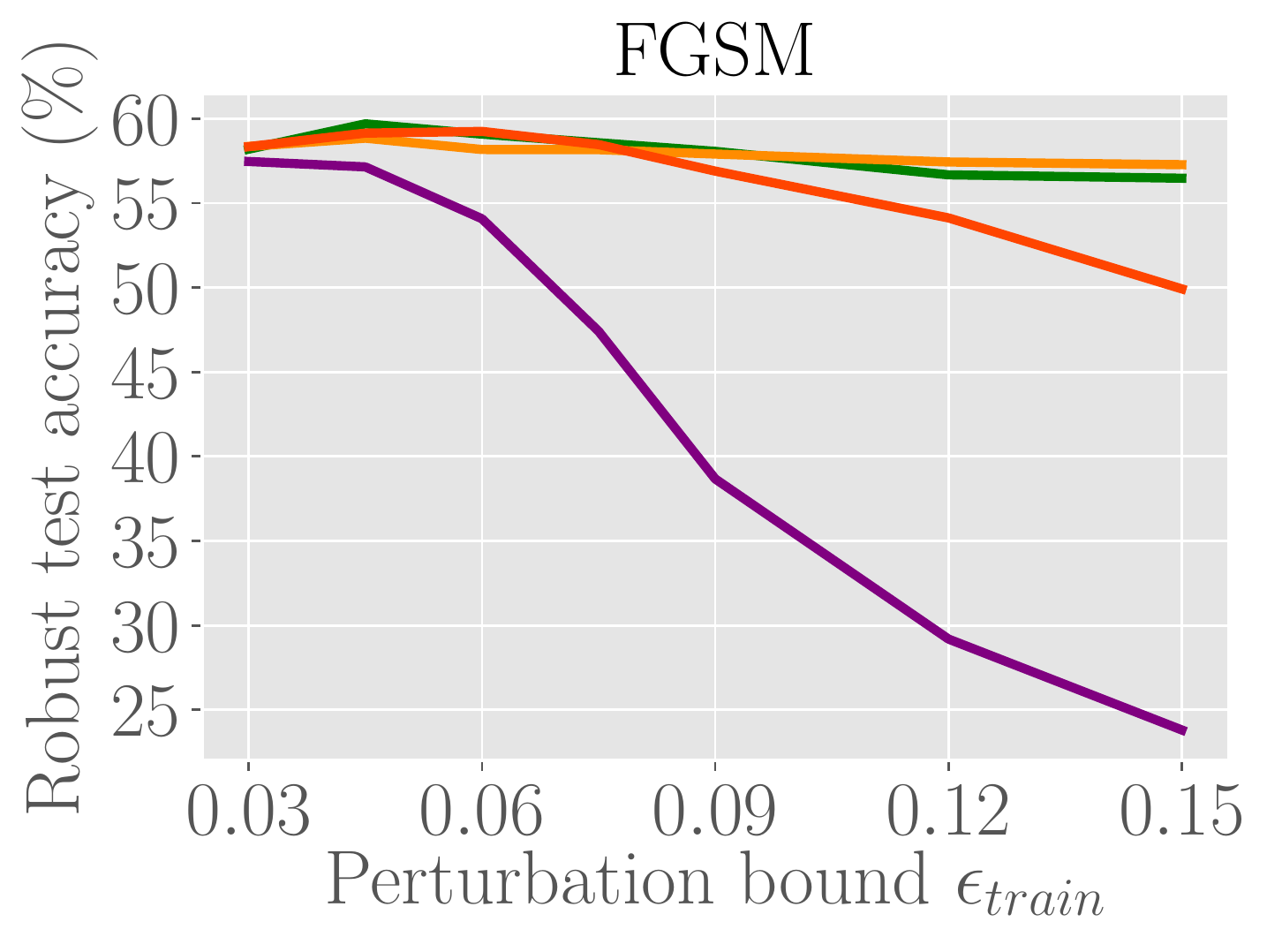} \\
    \includegraphics[scale=0.25]{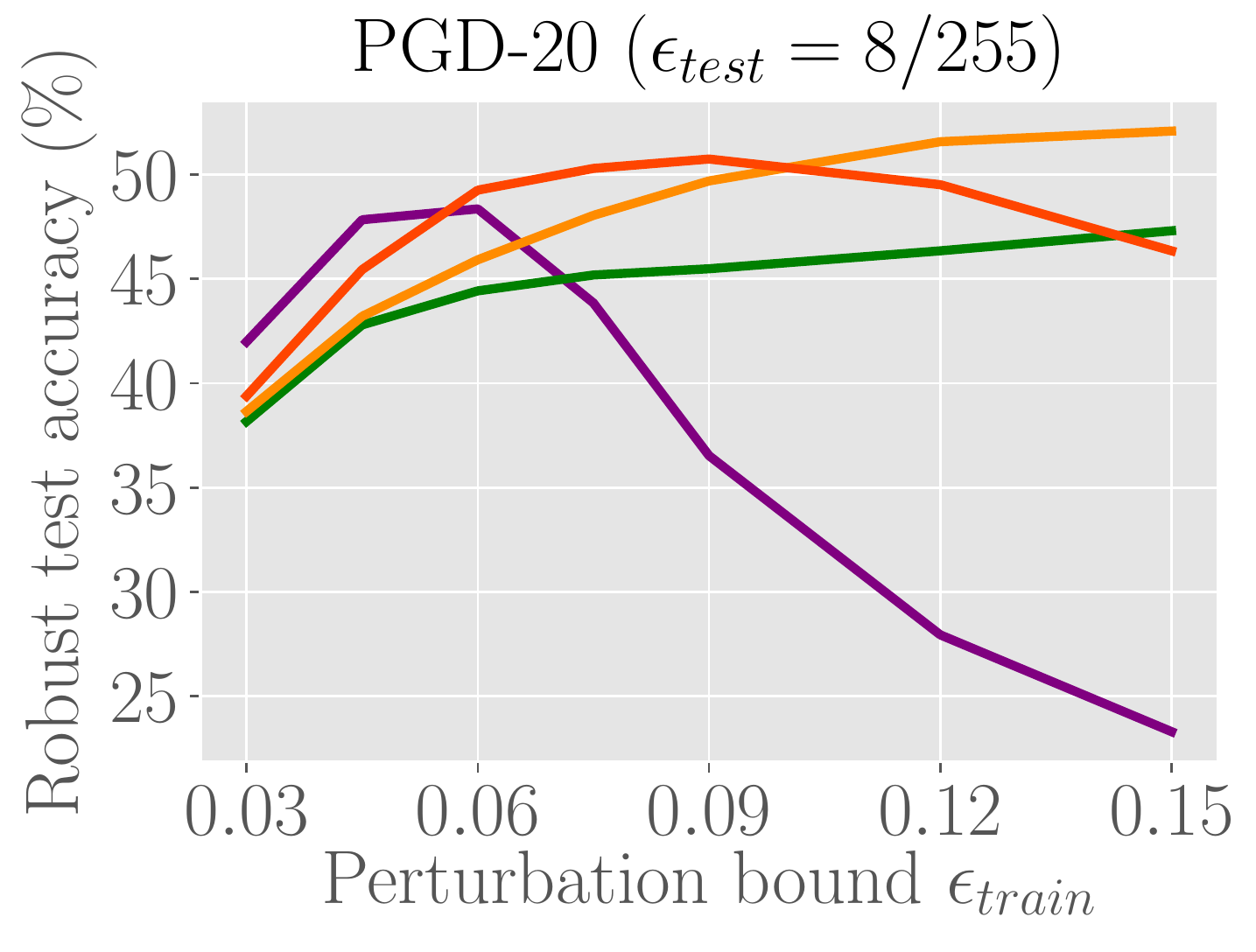}
    \includegraphics[scale=0.25]{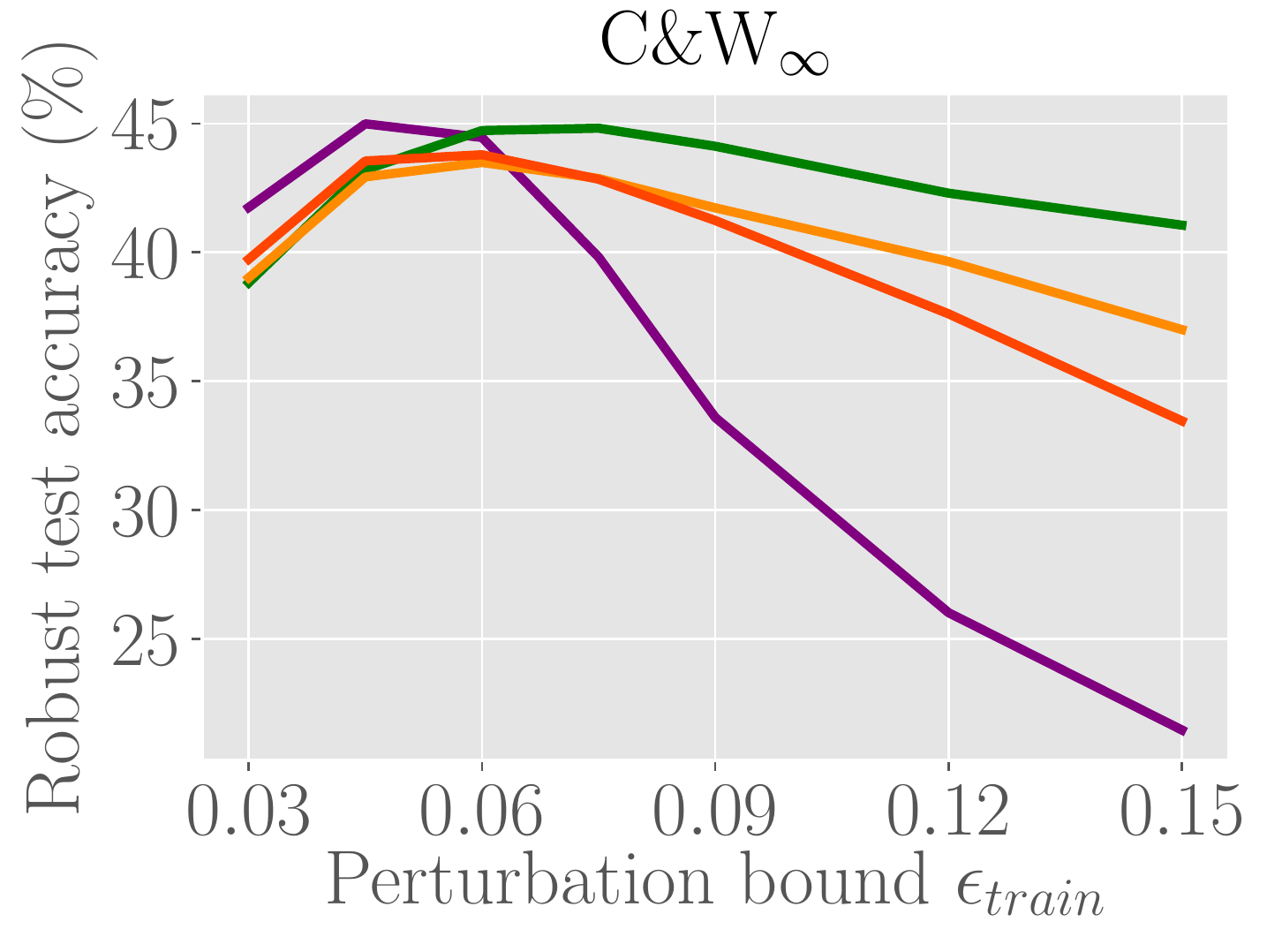}
    \\
    \includegraphics[scale=0.25]{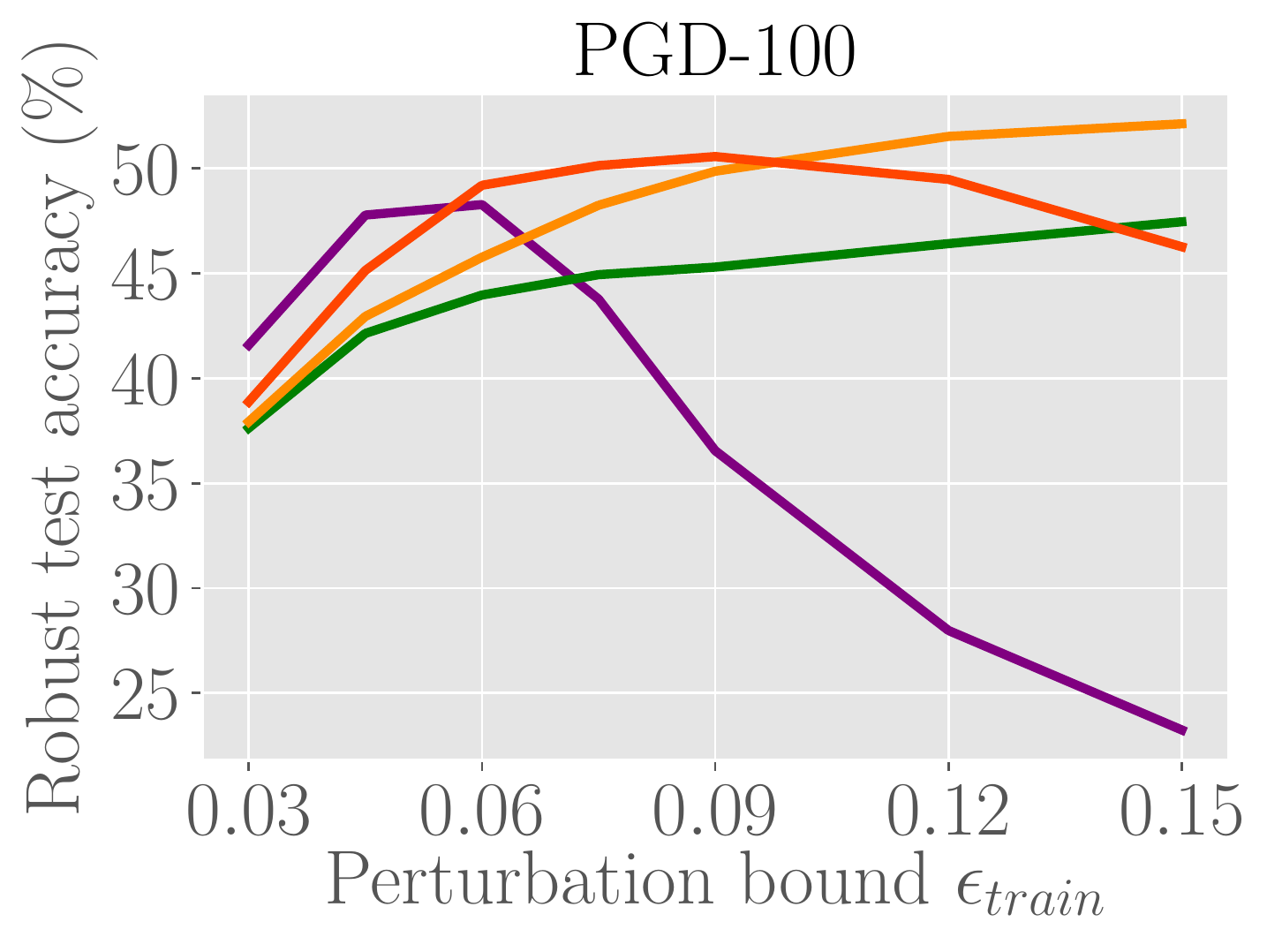}
    \includegraphics[scale=0.25]{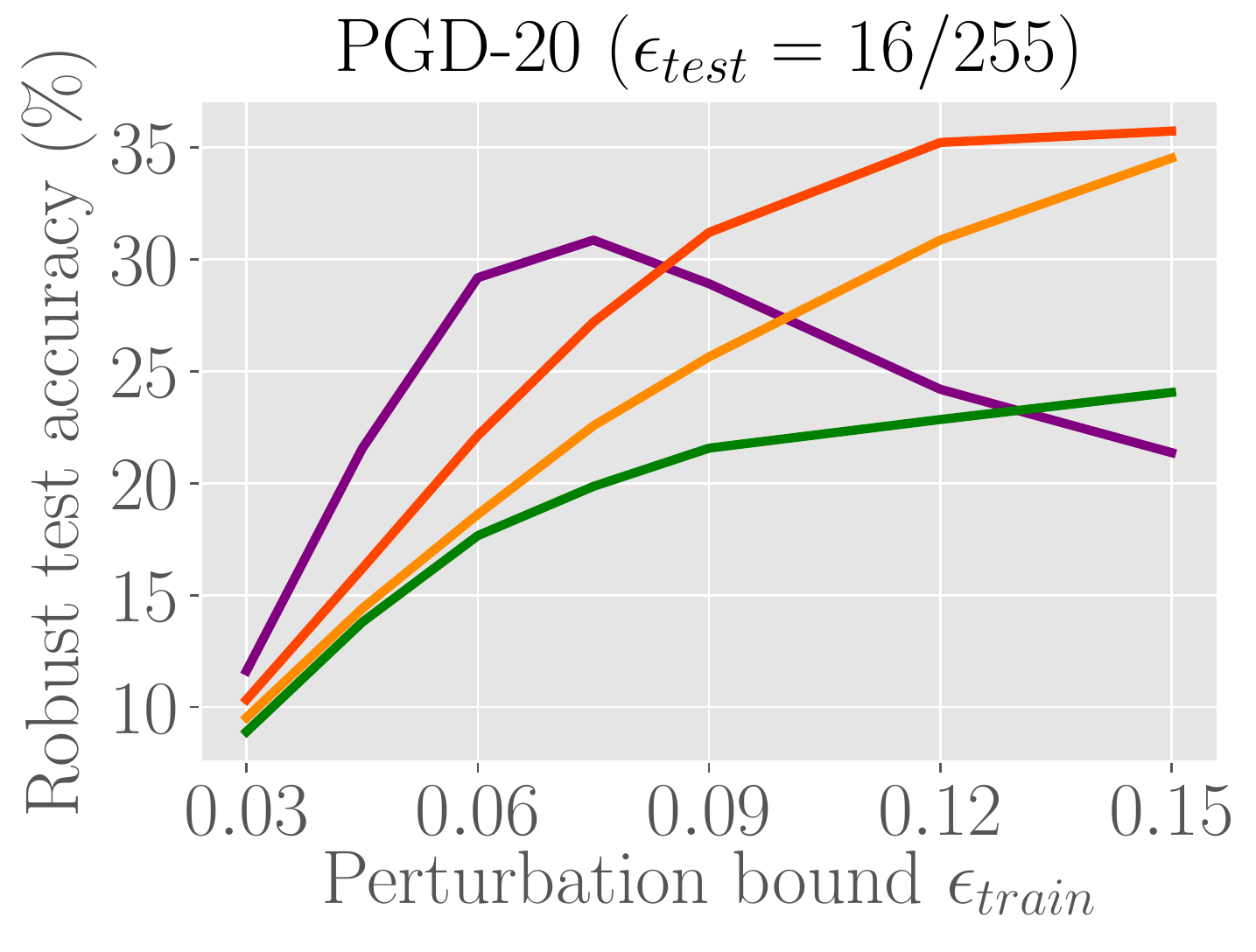}
    \caption{Comparisons on standard test accuracy and robust test accuracy of the deep model (ResNet-18) trained by standard adversarial training (Madry) and our friendly adversarial training ($\tau = 0, 1, 3$) respectively on CIFAR-10 dataset.}
    \label{fig:resnet18_cifar10_dynamic_epsball}
    \vspace{-3mm}
\end{figure}

\begin{figure}[tp!]
    \centering
    \includegraphics[scale=0.25]{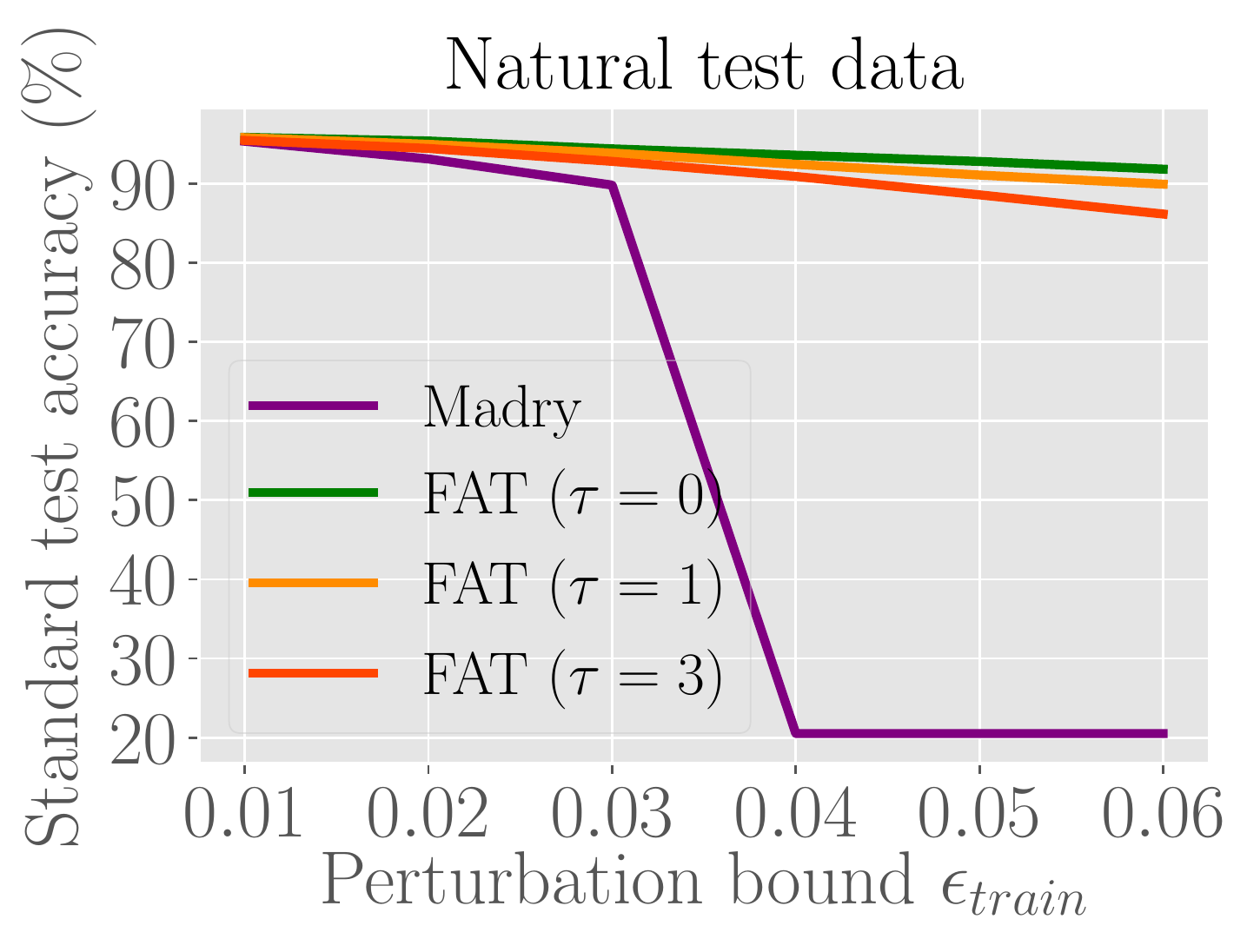}
    \includegraphics[scale=0.25]{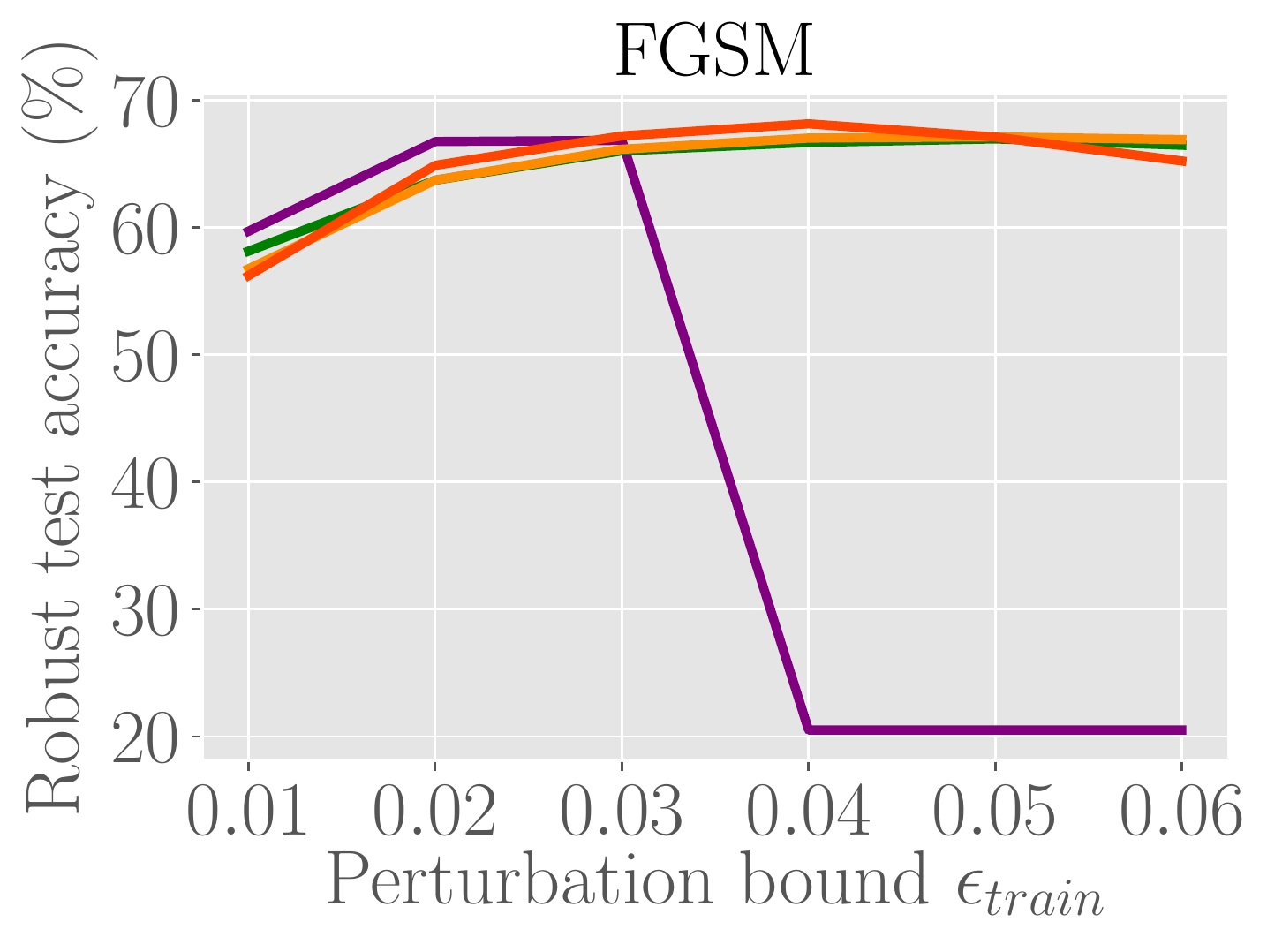} \\
    \includegraphics[scale=0.25]{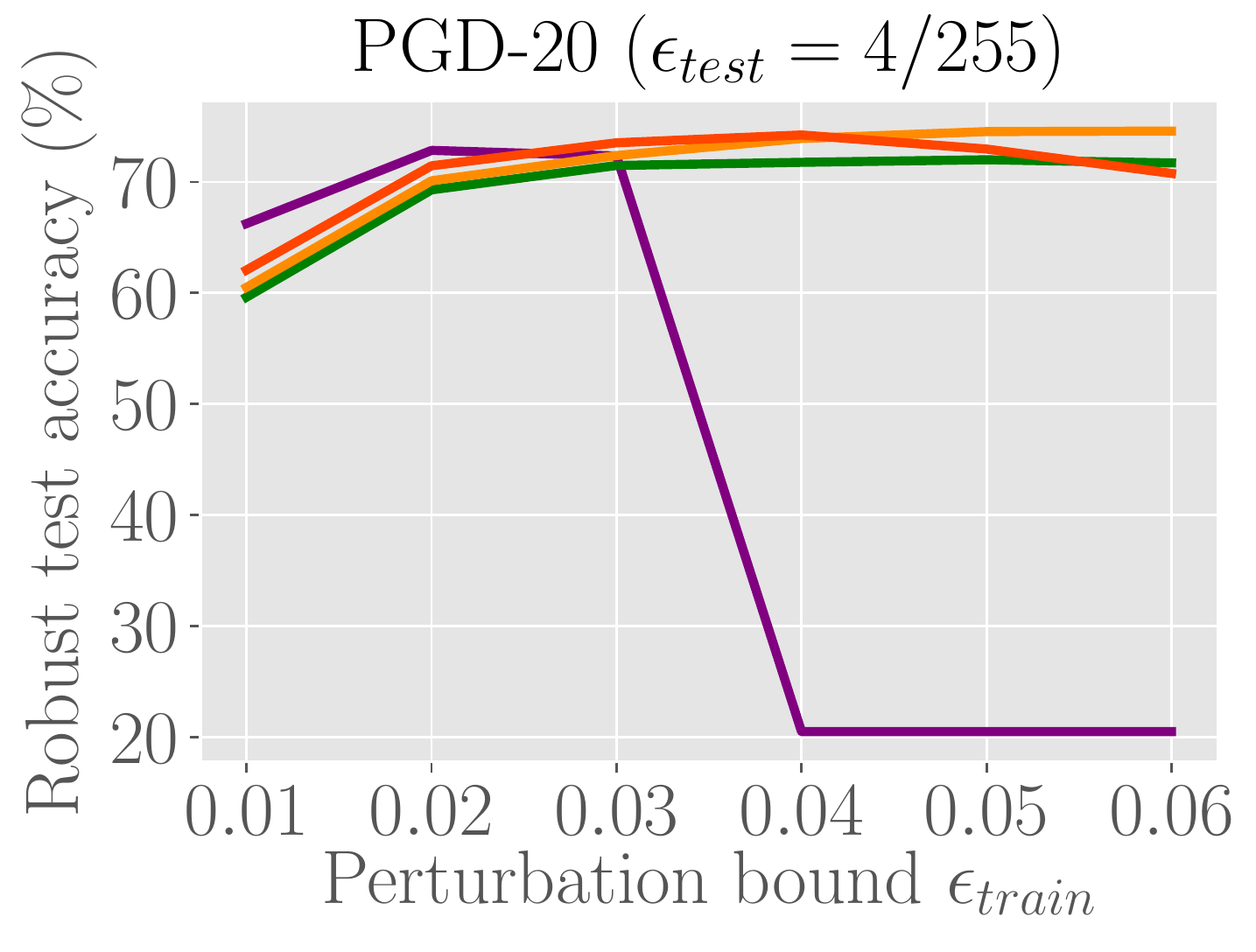}
    \includegraphics[scale=0.25]{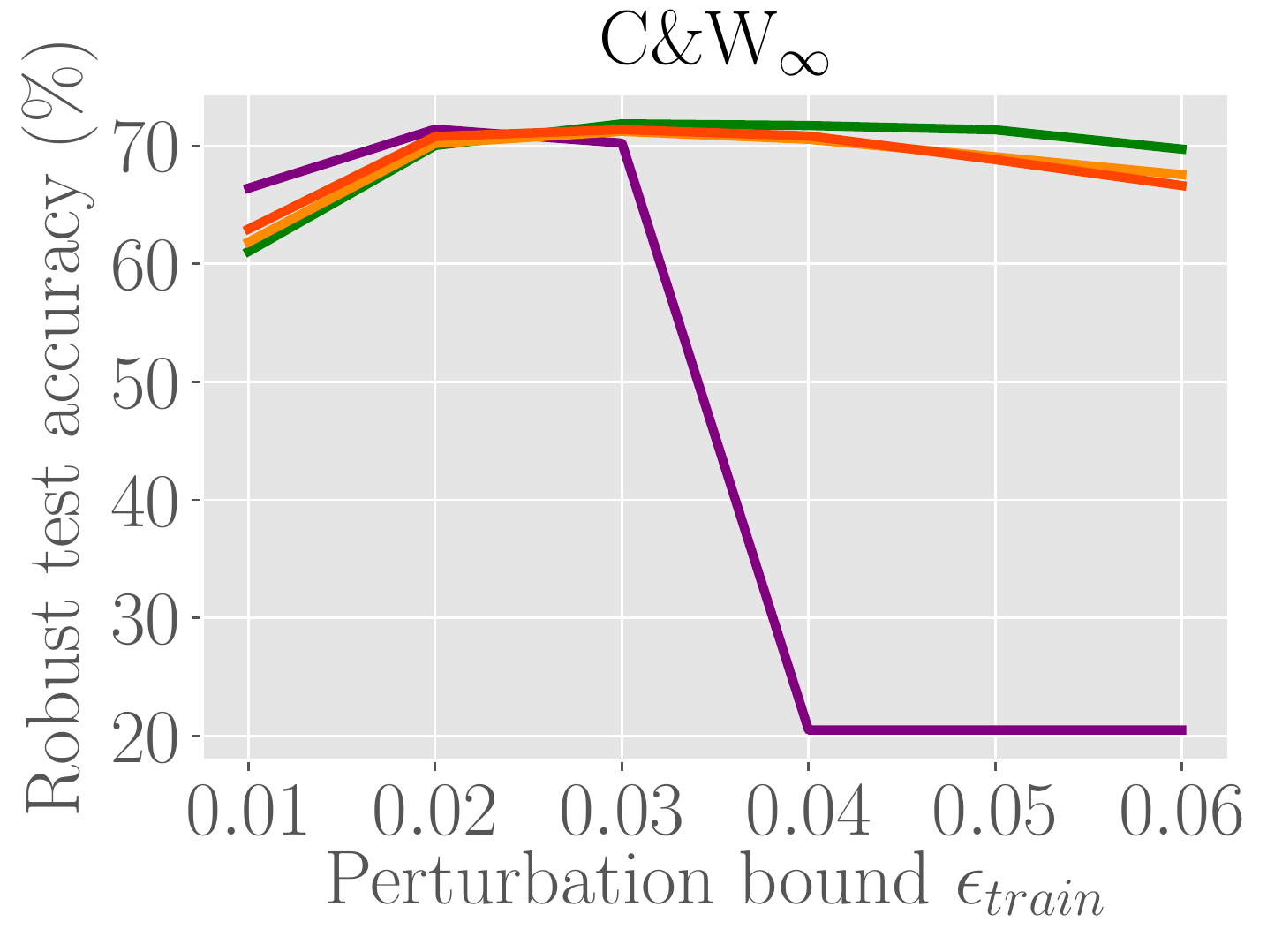}\\
    \includegraphics[scale=0.25]{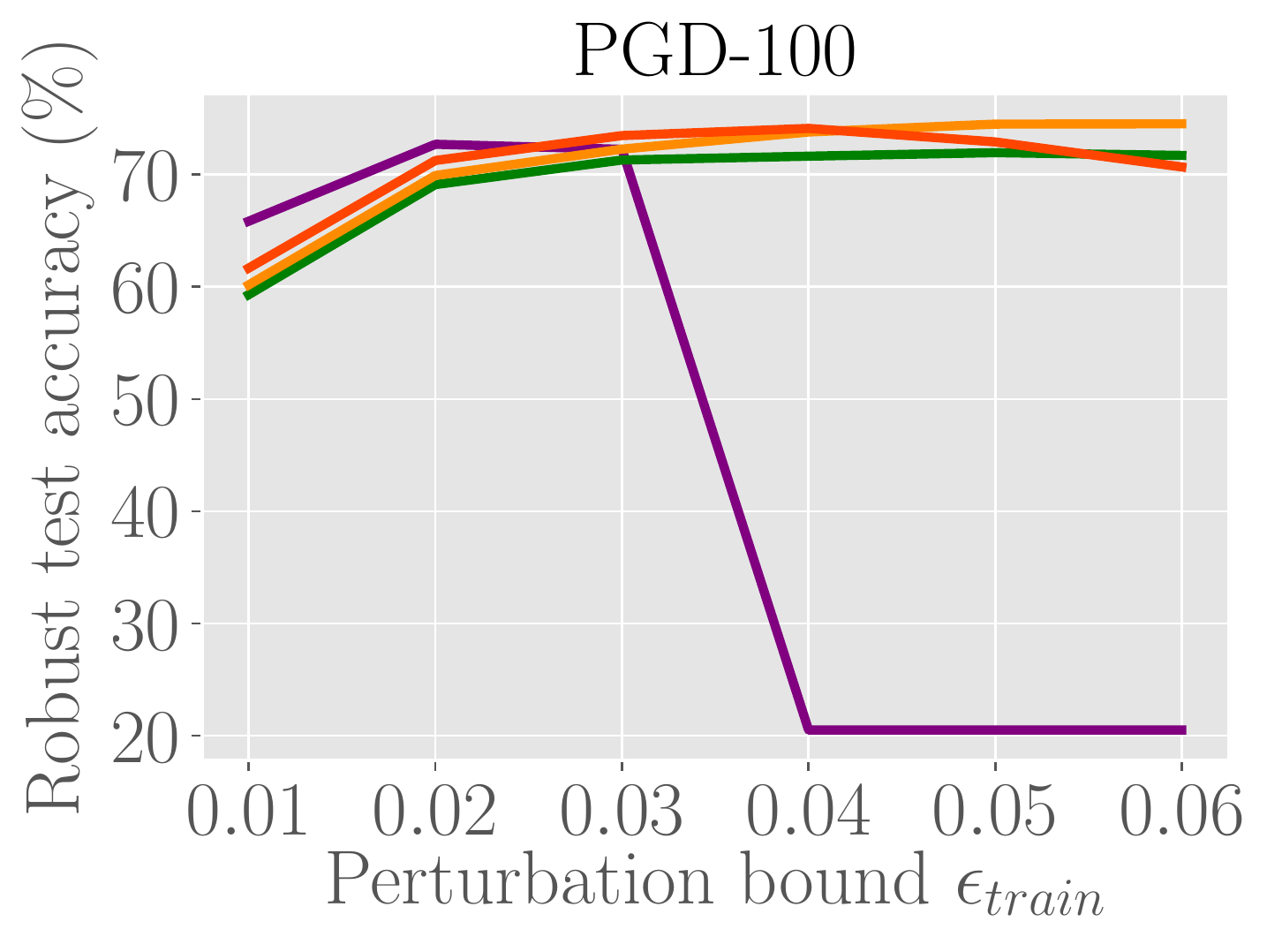}
    \includegraphics[scale=0.25]{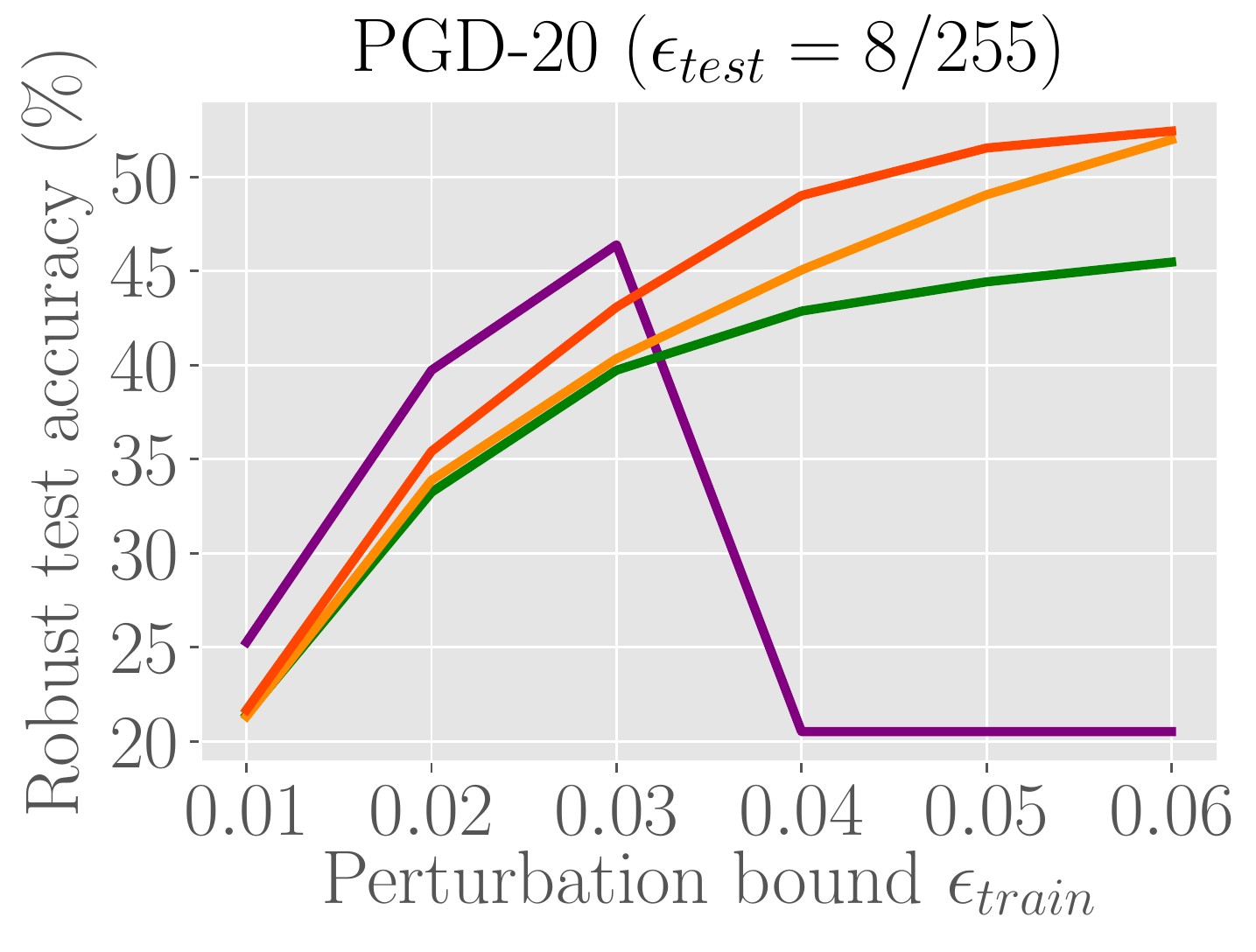}
    \caption{
    Comparisons on standard test accuracy and robust test accuracy of deep model (ResNet-18) trained by standard adversarial training (Madry) and our friendly adversarial training ($\tau = 0, 1, 3$) respectively on SVHN dataset.}
    \label{fig:resnet18_svhn_dynamic_epsball}
    \vspace{-3mm}
\end{figure}

All images of CIFAR-10 and SVHN are normalized into $[0,1]$. We compare our FAT ($\tau = 0, 1, 3$) and standard adversarial training (Madry) on ResNet-18 with different perturbation bounds $\epsilon_{train}$, i.e., $\epsilon_{train} \in [0.03, 0.15]$ for CIFAR-10 in Figure~\ref{fig:resnet18_cifar10_dynamic_epsball} and  $\epsilon_{train} \in [0.01, 0.06]$ for SVHN in Figure~\ref{fig:resnet18_svhn_dynamic_epsball}. 
The maximum PGD step $K = 10$, step size $\alpha = \epsilon / 10$. 
DNNs were trained using SGD with 0.9 momentum for 80 epochs with the initial learning rate of 0.01 divided by 10 at epoch 60. 

In addition, we have experiments of training a different DNN, e.g., Small CNN. We also set maximum PGD step $K = 20$. Besides, we compare our FAT for TRADES and TRADES~\cite{Zhang_trades} with different values of $\epsilon_{train}$.
These extensive results are presented in Appendix~\ref{APPENDIX:exp_larger_eps_ball}.

Figures~\ref{fig:resnet18_cifar10_dynamic_epsball} and~\ref{fig:resnet18_svhn_dynamic_epsball} show the performance of FAT ($\tau =0, 1, 3$) and standard adversarial training w.r.t. standard test accuracy and adversarially robust test accuracy of the DNNs.
We obtain standard test accuracy for natural test data and robust test accuracy for adversarial test data. The adversarial test data are bounded by $L_{\infty}$ perturbations with $\epsilon_{test} = 8/255$ for CIFAR-10 and $\epsilon_{test} = 4/255$ for SVHN, which are generated by FGSM, PGD-20, PGD-100 and C$\&$W$_{\infty}$ ($L_{\infty}$ version of C$\&$W optimized by PGD-30~\cite{Carlini017_CW}). 
Moreover, we also evaluate the robust DNNs using stronger adversarial test data generated by PGD-20 with a larger perturbation bound $\epsilon_{test} = 16/255$ for CIFAR-10 and $\epsilon_{test} = 8/255$ for SVHN (bottom right panels in Figures~\ref{fig:resnet18_cifar10_dynamic_epsball} and~\ref{fig:resnet18_svhn_dynamic_epsball}). 
All PGD attacks have random start, i.e, the uniformly random perturbation of $[-\epsilon_{test},\epsilon_{test}]$ added to the natural test data before PGD perturbations. Step size $\alpha$ of PGD is fixed to 2/255. 

From top-left panels of Figures~\ref{fig:resnet18_cifar10_dynamic_epsball} and \ref{fig:resnet18_svhn_dynamic_epsball}, DNNs trained by FAT ($\tau=0,1,3$) have higher standard test accuracy compared with those trained by standard adversarial training (i.e., Madry). This gap significantly widens as perturbation bound $\epsilon_{train}$ increases. 
Larger $\epsilon_{train}$ will allow the generated adversarial data deviate more from natural data. In standard adversarial training, natural generalization is significantly hurt with larger $\epsilon_{train}$ due to the cross-over mixture issue.
In contrast, DNNs trained by FAT could have better standard generalization, which is less affected by an increasing perturbation bound $\epsilon_{train}$.

\begin{table*}[tp!]
\small \centering
\caption{Evaluations (test accuracy) of deep models (WRN-32-10) on CIFAR-10 dataset}
\label{table:sota_result_madry}
\begin{tabular*}{\hsize}{@{}@{\extracolsep{\fill}}clccccc@{}}
\hline
\multicolumn{2}{c|}{Defense} & \multicolumn{1}{c}{Natural} & \multicolumn{1}{c}{FGSM} & \multicolumn{1}{c}{PGD-20} & \multicolumn{1}{c}{C$\&$W$_{\infty}$} & \multicolumn{1}{c}{PGD-100} \\ 
\hline
\multicolumn{2}{c|}{Madry} & 87.30  & 56.10 & 45.80 & 46.80 & - \\
\multicolumn{2}{c|}{CAT} & 77.43 & 57.17 & 46.06 & 42.28 & - \\
\multicolumn{2}{c|}{DAT} & 85.03 & 63.53 & 48.70 & 47.27 & -\\
\multicolumn{2}{c|}{FAT ($\epsilon_{train}={8}/{255}$)} & \textbf{89.34} $\pm$ 0.221  & 65.52 $\pm$ 0.355 & 46.13 $\pm$ 0.409 & 46.82 $\pm$ 0.517 & 45.31 $\pm$ 0.531\\
\multicolumn{2}{c|}{FAT ($\epsilon_{train} = {16}/{255}$)} & 87.00 $\pm$ 0.203 & \textbf{65.94} $\pm$ 0.244& \textbf{49.86} $\pm$ 0.328 & \textbf{48.65} $\pm$ 0.176& \textbf{49.56} $\pm$ 0.255 \\
\hline
\end{tabular*}
\vskip1ex%
{\footnotesize%
Results of Madry, CAT and DAT are reported in~\cite{Wang_Xingjun_MA_FOSC_DAT}. FAT has the same evaluations.
}
\vskip -0.1in
\end{table*}

\begin{table*}[tp!]
\small \centering
\caption{Evaluations (test accuracy) of deep models (WRN-34-10) on CIFAR-10 dataset}
\label{table:sota_result_trades}
\begin{tabular*}{\hsize}{@{}@{\extracolsep{\fill}}clccccc@{}}
\hline
\multicolumn{2}{c|}{Defense} & \multicolumn{1}{c}{Natural} & \multicolumn{1}{c}{FGSM} & \multicolumn{1}{c}{PGD-20} & \multicolumn{1}{c}{C$\&$W$_{\infty}$} & \multicolumn{1}{c}{PGD-100}\\
\hline
\multicolumn{2}{c|}{TRADES ($\beta = 1.0$)} & 88.64 & 56.38 & 49.14 & - & -\\ 
\multicolumn{2}{c|}{FAT for TRADES ($\epsilon_{train} = 8/255$)} & \textbf{89.94} $\pm$ 0.303 & \textbf{61.00} $\pm$ 0.418 & \textbf{49.70} $\pm$ 0.653 & 49.35 $\pm$ 0.363 & 48.35 $\pm$ 0.240\\ 
\hline
\multicolumn{2}{c|}{TRADES ($\beta = 6.0$)} & 84.92 & 61.06 & 56.61 &\textbf{54.47} & 55.47\\   
\multicolumn{2}{c|}{FAT for TRADES ($\epsilon_{train} = 8/255$)} & \textbf{86.60} $\pm$ 0.548 & \textbf{61.97} $\pm$ 0.570 & 55.98 $\pm$ 0.209 & 54.29 $\pm$ 0.173 & 55.34 $\pm$ 0.291 \\ 
\multicolumn{2}{c|}{FAT for TRADES ($\epsilon_{train} = 16/255$)} & 84.39 $\pm$ 0.030 & 61.73 $\pm$ 0.131 & \textbf{57.12} $\pm$ 0.233 & 54.36 $\pm$ 0.177 & \textbf{56.07} $\pm$ 0.155 \\ 
\hline
\end{tabular*}
\vskip1ex%
{\footnotesize%
Results of TRADES ($\beta = 1.0$ and $6.0$) are reported in~\cite{Zhang_trades}. FAT for TRADES has the same evaluations.
}
\vskip -0.1in
\end{table*}

In addition, with the increase of the perturbation bound $\epsilon_{train}$, robust test accuracy (e.g., PGD and C$\&$W) of DNNs trained by standard adversarial training (Madry) gets a slight increase first but is followed by a sharp drop. 
For a larger $\epsilon_{train}$ (e.g., $\epsilon_{train} > 0.06$ in CIFAR-10 and $\epsilon_{train} > 0.03$ in SVHN), standard adversarial training (Madry) basically fails, and thus its robust test accuracy drops sharply. 
Without early-stopped PGD, the generated adversarial data has a severe cross-over mixture problem, which makes the adversarial learning extremely difficult and sometimes even ``kills'' the learning.

However, it is still meaningful to enable a stronger defense over a weaker attack, i.e., $\epsilon_{train}$ in adversarial training should be larger than $\epsilon_{test}$ in adversarial attack. The right bottom panels in Figures~\ref{fig:resnet18_cifar10_dynamic_epsball} and~\ref{fig:resnet18_svhn_dynamic_epsball} show for the stronger attack ($\epsilon_{test} = 16/255$ for CIFAR-10 and $\epsilon_{test} = 8/255$ for SVHN), it is meaningful to have larger $\epsilon_{train}$ to attain a better robustness. 
Our FAT is able to achieve larger $\epsilon_{train}$. 
Figures~\ref{fig:resnet18_cifar10_dynamic_epsball} and~\ref{fig:resnet18_svhn_dynamic_epsball} show our FAT ($\tau = 0,1,3$) maintain higher robust accuracy with larger $\epsilon_{train}$.

Note that the performance of FAT with PGD-10-3 ($\tau=3$, red lines) in Figure~\ref{fig:resnet18_cifar10_dynamic_epsball} drops with $\epsilon_{train} > 0.09$. We believe that FAT with larger $\tau$ could also have the issue of cross-over mixture, which is detrimental to adversarial learning.
In addition, both standard adversarial training and friendly adversarial training do not perform well under C$\&$W attack with larger $\epsilon_{train}$ (e.g., $\epsilon_{train} > 0.075$ in Figure~\ref{fig:resnet18_cifar10_dynamic_epsball}), we believe it is due to the mismatch between PGD-adversarial training and C$\&$W attack. We discuss the reasons in detail in  Appendix~\ref{Appendix:reasons-cw-performance}. 

To sum up, deep models by FAT with $\tau = 0$ (green line) have higher standard test accuracy, but lower robust test accuracy. By increasing $\tau$ to 1, deep models have slightly reduced standard test accuracy but have the increased adversarial robustness. 
This sheds light on the importance of $\tau$, which handles the trade-off between robustness and standard accuracy.
In order not to ``kill'' the training at the initial stage, we could vary $\tau$ from a smaller value to a larger value over training epochs. 
In addition, due to benefits that our FAT could enable larger $\epsilon_{train}$, we could also make $\epsilon_{train}$ larger over training. Those tricks echo our paper's philosophy of ``attacks which do not kill training make adversarial training stronger''. 
In the next subsection, we unleash the full power of FAT (and FAT for TRADES) and show its superior performance over the state-of-the-art methods.

\subsection{Performance Evaluations on Wide ResNet}
\label{section:SOTA_results}
To manifest the full power of friendly adversarial training, we adversarially train Wide ResNet~\cite{zagoruyko2016WRN} to achieve the state-of-the-art performance on CIFAR-10.
Similar to~\cite{Wang_Xingjun_MA_FOSC_DAT,Zhang_trades}, we employ WRN-32-10 (Table~\ref{table:sota_result_madry}) and WRN-34-10 (Table~\ref{table:sota_result_trades}) as our deep models.


In Table~\ref{table:sota_result_madry}, we compare FAT with standard adversarial training (Madry)~\cite{Madry_adversarial_training}, CAT~\cite{Cai_CAT} and DAT~\cite{Wang_Xingjun_MA_FOSC_DAT} on WRN-32-10. Training and evaluation details are in Appendix~\ref{APPENDIX:training_details_sota_FAT}.
The performance evaluations are done exactly as in DAT~\cite{Wang_Xingjun_MA_FOSC_DAT}. 
In Table~\ref{table:sota_result_trades}, we compare FAT for TRADES with TRADES~\cite{Zhang_trades} on WRN-34-10.
Training and evaluation details are in Appendix~\ref{APPENDIX:training_details_sota_TRADES}. The performance evaluations are done exactly as in TRADES~\cite{Zhang_trades}.

Moreover, \citet{nakkiran2019adversarial} states that robust classification needs more complex classifiers (exponentially more complex). We employ FAT for TRADES on even larger WRN-58-10, the performance gets further improved over WRN-34-10 (Appendix~\ref{APPENDIX:training_details_sota_TRADES}). Moreover, we also apply the early-stopped PGD to MART, namely, FAT for MART~(in Appendix~\ref{Appendix:fat_for_mart_realization}). As a result, the performance gets improved (detailed in Appendix~\ref{APPENDIX:training_details_sota_MART}). 

Tables~\ref{table:sota_result_madry} and~\ref{table:sota_result_trades} and results in Appendix~\ref{appendix:sota-WRN} justify the efficacy of friendly adversarial training - adversarial robustness can indeed be achieved without compromising the natural generalization. In addition, we are even able to attain the state-of-the-art robustness. 

\section{Conclusion}
This paper has proposed a novel formulation for adversarial training. 
Friendly adversarial training (FAT) approximately realizes this formulation by stopping the PGD early. 
FAT is computationally efficient and adheres to the spirit of curriculum training. 
In addition, FAT helps to relieve the problem of cross-over mixture. 
As a result, FAT can train deep models with larger perturbation bounds $\epsilon_{train}$.
Finally, FAT can achieve competitive performance on the large capacity networks. 
Further research includes (a) how to choose optimal step $\tau$ in FAT algorithm,
(b) besides PGD-K-$\tau$, how to search for friendly adversarial data effectively, and (c) theoretically studying adversarially robust generalization~\cite{Dong_Yin_adv_gen}, e.g., through the lens of Rademacher complexity~\cite{bartlett2002rademacher}.

\clearpage
\subsection*{Acknowledgments}
This research is supported by the National Research Foundation, Singapore under its Strategic Capability Research Centres Funding Initiative (MK, JZ), JST AIP Acceleration Research Grant Number JPMJCR20U3, Japan (GN, MS), National Key R$\&$D Program No.2017YFB1400100, the NSFC No.91846205, the Shandong Key R$\&$D Program No.2018YFJH0506 (LC), the Early Career Scheme (ECS) through the Research Grants Council of Hong Kong under Grant No.22200720 (BH), HKBU Tier-1 Start-up Grant (BH) and HKBU CSD Start-up Grant (BH).
Any opinions, findings and conclusions or recommendations expressed in this material are those of the author(s) and do not reflect the views of National Research Foundation, Singapore.
%
%
%
\nocite{langley00}

\bibliography{references}
\bibliographystyle{icml2020}

\appendix
\onecolumn

\section{Friendly Adversarial Training}
For completeness, besides the learning objective by loss value, we also give the learning objective of FAT by class probability. 
Then, based on the learning objective by loss value, we give the proof of Theorem~\ref{lemma}, which theoretically justifies FAT. 
\subsection{Learning Objective}
\label{Appendix:learning_obj_fat}
\paragraph{Case 1 (by loss value, restated).} The outer minimization still follows Eq.~\eqref{eq:adv-obj}. However, instead of generating $\xadv_i$ via inner maximization, we generate $\xadv_i$ as follows:
\begin{align*}
\label{Eq:FAT_obj_function}
\xadv_i = \argmin_{\xadv\in\epsball[x_i]} &\; \ell(f(\xadv),y_i) \nonumber \\
\ST &\; \ell(f(\xadv),y_i) - \min\nolimits_{y\in\cY}\ell(f(\xadv),y) \ge \rho.
\end{align*}
Note that the operator $\argmax$ in Eq.~\eqref{Eq:madry_inner_maximization} is replaced with $\argmin$ here, and there is a constraint on the margin of loss values (i.e., the mis-classification confidence).

The constraint firstly ensures $y_i \neq \argmin\nolimits_{y\in\cY} \ell(f(\xadv),y)$ or $\xadv$ is mis-classified, and secondly ensures for $\xadv$ the wrong prediction is better than the desired prediction $y_i$ by at least $\rho$ in terms of the loss value. Among all such $\xadv$ satisfying the constraint, we select the one minimizing $\ell(f(\xadv),y_i)$. Namely, we minimize the adversarial loss given that a confident adversarial data has been found. This $\xadv_i$ could be regarded as a friend among the adversaries, which is termed as friendly adversarial data.

\paragraph{Case 2 (by class probability).}~
We redefine the above objective from the loss point of view (above) to the class probability point of view. 
The objective is still Eq.~\eqref{eq:adv-obj}, in which  $\ell(f(\bxtidle),y)=\ell_\textrm{B}(\ell_\textrm{L}(f(\bxtidle)),y)$. Hence, $\ell_\textrm{L}(f(\bxtidle))$ is an estimate of the class-posterior probability $p(y \mid \bxtidle)$, and for convenience, denote by $p_f(y \mid \bxtidle)$ the $y$-th element of the vector $\ell_\textrm{L}(f(\bxtidle))$.
\begin{align*}
\xadv_i = \argmax_{\xadv\in\epsball[x_i]} &\; p_f(y_i \mid \xadv)\\
\ST &\; \max\nolimits_{y\in\cY}p_f(y \mid \xadv) - p_f(y_i \mid \xadv) \ge \rho.
\end{align*}
The constraint ensures $\xadv$ is misclassified by at least $\rho$, but here the margin $\rho$ is applied to the class probability instead of the loss value. Hence, $\xadv_i$ should usually be different from the one according to the loss value.

\subsection{Proofs}
\label{APPENDIX:Proof}

We derive a tight upper bound on adversarially robust risk (adversarial risk), and provide our theoretical analysis for the adversarial risk minimization. $X$ and $Y$ represent random variables. Adversarial risk $\mathcal{R}_{\mathrm{rob}}(f) := \mathbb{E}_{(X,Y) \sim \mathcal{D}} \mathds{1} \{ \exists X' \in \epsball[X]: f(X') \neq Y\} $. 
$\mathcal{R}_{\mathrm{rob}}(f)$ can be decomposed, i.e., $ \mathcal{R}_{\mathrm{rob}}(f) = \mathcal{R}_{nat}(f) + \mathcal{R}_{\mathrm{bdy}}(f)$, where natural risk $\mathcal{R}_{\mathrm{nat}}(f) = \mathbb{E}_{(X,Y) \sim \mathcal{D}} \mathds{1} \{ f(X) \neq Y\}  $  and boundary risk $\mathcal{R}_{\mathrm{bdy}}(f) = \mathbb{E}_{(X,Y) \sim \mathcal{D}} \mathds{1} \{ X \in \epsball[DB(f)], f(X) = Y\} $. Note that $\epsball[DB(f)]$ is the set denoting the decision boundary of $f$, i.e., $\{\bx \in \cX:  \exists \bxprime \in \epsball[\bx] \quad \mathrm{s.t.} \quad f(\bx)\neq f(\bxprime)\}$.

\begin{figure}[h!]
    \centering
    \includegraphics[scale=0.5]{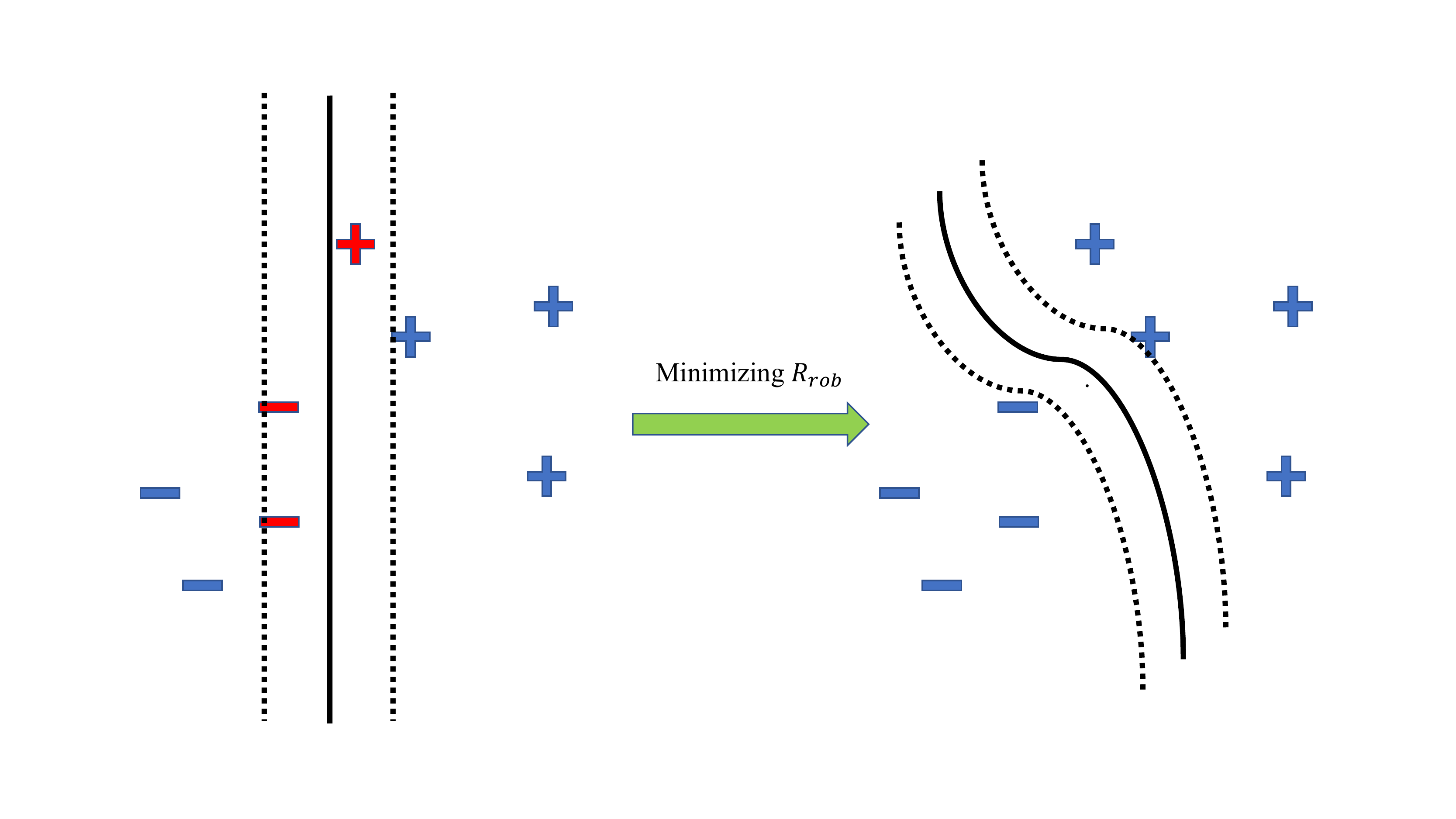}
    \caption{Solid line is the classifier. The area between dashed line is decision boundary of the classifier. 
    Minimizing robust risk $R_{\mathrm{rob}}$ is to find a classifier, where data is less likely located within decision boundary of the classifier.}
    \label{fig:theorem_show}
\end{figure}

\textbf{Lemma 1.}
For any classifier $f: \cX \to \cY$, any probability distribution $\mathcal{D}$ on $\cX \times \cY$, we have 
\begin{align*}
\mathcal{R}_{\mathrm{rob}}(f)  =  \mathcal{R}_{\mathrm{nat}}(f)  + 
\mathbb{E}_{(X,Y) \sim \mathcal{D}}  \mathds{1} \{\exists X'  \in  \epsball[X]: f(X) \neq f(X')\} \cdot \mathds{1} \{ f(X)=Y\}. 
\end{align*}
\begin{proof}
By the equation  $\mathcal{R}_{\mathrm{rob}}(f) =  \mathcal{R}_{\mathrm{nat}}(f) + \mathcal{R}_{\mathrm{bdy}}(f)$, 
\begin{align*}
    \mathcal{R}_{\mathrm{rob}}(f) 
    &= \mathcal{R}_{\mathrm{nat}}(f) + \mathcal{R}_{\mathrm{bdy}}(f) \\
    &=\mathcal{R}_{\mathrm{nat}}(f)  + \mathbb{E}_{(X,Y) \sim \mathcal{D}} \mathds{1} \{ X' \in \epsball[DB(f)], f(X) = Y\} \\
    &= \mathcal{R}_{\mathrm{nat}}(f)  + {\rm Pr} [ X' \in \epsball[DB(f)], f(X) = Y] \\
    &= \mathcal{R}_{\mathrm{nat}}(f)  + {\rm Pr }[ f(X) \neq f(X'), f(X) = Y] \\
    &= \mathcal{R}_{\mathrm{nat}}(f)  + \mathbb{E}_{(X,Y) \sim \mathcal{D}}  \mathds{1}\{\exists X' \in \epsball[X]: f(X) \neq f(X'), f(X) = Y) \} \\
    &= \mathcal{R}_{\mathrm{nat}}(f)  + \mathbb{E}_{(X,Y) \sim \mathcal{D}}  \mathds{1}\{\exists X' \in \epsball[X]: f(X) \neq f(X') \} \cdot \mathds{1}\{f(X) = Y\}
\end{align*}

The fourth equality comes from the definition of decision boundary of $f$, i.e., 
$\epsball[DB(f)] = \{\bx \in \cX:  \exists \bxprime \in \epsball[\bx] \quad  \mathrm{s.t.} \quad f(\bx)\neq f(\bxprime)  \}$.
\end{proof}

Figure~\ref{fig:theorem_show} illustrates the message of Lemma 1. Minimizing robust risk $R_{\mathrm{rob}}$ encourages the learning algorithm to find a classifier whose decision boundary contains less training data. Meanwhile, the classifier should correctly separate data from different classes.
Finding such a classifier is hard. As we can see in Figure~\ref{fig:theorem_show}, the hypothesis set of hyperplanes is enough for minimizing the natural risk (the left figure). However, a robust classifier (the right figure) has much curvatures, which is more complicated. \citet{nakkiran2019adversarial} states that robust classification needs more complex classifiers (exponentially more complex, in some examples). 
This implies that in order to learn a robust classifier, our learning algorithm needs (a) setting a large hypothesis set and (b) fine-tuning the decision boundary.  

\textbf{Theorem~\ref{lemma} (restated).} For any classifier $f$, any non-negative surrogate loss function $\ell$ which upper bounds $0/1$ loss, and any probability distribution $\mathcal{D}$, we have
\begin{align*}
   \mathcal{R}_{\mathrm{rob}}(f)  &\leq  \underbrace{ \mathbb{E}_{(X,Y) \sim \mathcal{D}} \ell (f(X), Y) }_{\text{For standard test accuracy}} \\ 
   &+ \underbrace{ \mathbb{E}_{(X,Y) \sim \mathcal{D}, X' \in \epsball[X, \epsilon]} \ell^{*} (f(X'), Y) }_{\text{For robust test accuracy}},
\end{align*}
where \[
    \ell^{*} = 
\begin{cases}
    \min \ell(f(X'), Y) + \rho,& \text{if } f(X') \neq Y; \\
    \max \ell(f(X'), Y),& \text{if } f(X') = Y.
\end{cases}
\]
$\rho$ is the small constant.
\begin{proof}
\begin{align*}
    \mathcal{R}_{\mathrm{rob}}(f) 
    &= \mathcal{R}_{\mathrm{nat}}(f) + \mathcal{R}_{\mathrm{bdy}}(f) \\
    &\leq \mathbb{E}_{(X,Y) \sim \mathcal{D}} \ell (f(X), Y)  + \mathcal{R}_{\mathrm{bdy}}(f) \\
    &= \mathbb{E}_{(X,Y) \sim \mathcal{D}} \ell (f(X), Y)  + \mathbb{E}_{(X,Y) \sim \mathcal{D}} \mathds{1} \{ X \in \epsball[DB(f)], f(X) = Y\} \\
    &= \mathbb{E}_{(X,Y) \sim \mathcal{D}} \ell (f(X), Y)  +  {\rm Pr} [ X \in \epsball[DB(f)], f(X) = Y] \\ 
    &= \mathbb{E}_{(X,Y) \sim \mathcal{D}} \ell (f(X), Y)   +  {\rm Pr }[ f(X) \neq f(X'), f(X) = Y]  \\
    &\leq \mathbb{E}_{(X,Y) \sim \mathcal{D}} \ell (f(X), Y)  + {\rm Pr }[ f(X') \neq Y] \\
    &= \mathbb{E}_{(X,Y) \sim \mathcal{D}} \ell (f(X), Y)  + \mathbb{E}_{(X,Y) \sim \mathcal{D}}\mathds{1} \{ \exists X' \in \epsball[X] : f(X') \neq Y\} \\ 
    &\leq \mathbb{E}_{(X,Y) \sim \mathcal{D}} \ell (f(X), Y)  + \mathbb{E}_{(X,Y) \sim \mathcal{D}, X' \in \epsball[X, \epsilon]} \ell^{*} (f(X'), Y)
\end{align*}
The first inequality comes from the assumption that surrogate loss function $\ell$ upper bound $0/1$ loss function. The second inequality comes from the fact that there exists misclassified natural data within the decision boundary set. Therefore, $ {\rm Pr }[ f(X) \neq f(X'), f(X) = Y] \cup  {\rm Pr }[ f(X) \neq f(X'), f(X) \neq Y] =  {\rm Pr }[ f(X') \neq Y ]$. The third inequality comes from the assumption that surrogate loss function $\ell$ upper bound $0/1$ loss function, i.e., in Figure~\ref{fig:lemma_show}, the adversarial data $X'$ (purple triangle) is on line of logistic loss (blue line), which is always above the 0/1 loss (yellow line). 
\end{proof}
Our Theorem~\ref{lemma} informs our strategy to fine tune the decision boundary. To fine-tune the decision boundary, the data ``near'' the classifier plays an important role. Those data are easily wrongly predicted with  small perturbations. As we show in the Figure~\ref{fig:lemma_show}, when adversarial data are wrongly predicted, our adversarial data (purple triangle) increases to minimize the loss by a violation of a small constant $\rho$.
Thus, our adversarial data can help fine-tune the decision boundary ``bit by bit'' over the training. 

\section{Alternative Adversarial Data Searching Algorithm}
\label{appendix:pgd_no_projection}
In this section, we give an alternative adversarial data searching algorithm to generate friendly adversarial data via modifying the method to update $\xadv$ in Algorithm~\ref{alg:PGD-k-t}. 
We remove the constraint of $\epsilon$-ball projection in Eq.~\eqref{PGD-k}, i.e., \begin{equation}
    \label{GD-k}
    {x}^{(t+1)} = {x}^{(t)} +\alpha \sign (\nabla_{{x}^{(t)}} \ell(f_{\theta}({x}^{(t)}), y )  )   , \forall { t \geq 0}
\end{equation}
where ${x}^{0}$ is a natural data and $\alpha > 0$ is step size. 

\begin{figure}[!htb]
    \centering
    \includegraphics[scale=0.33]{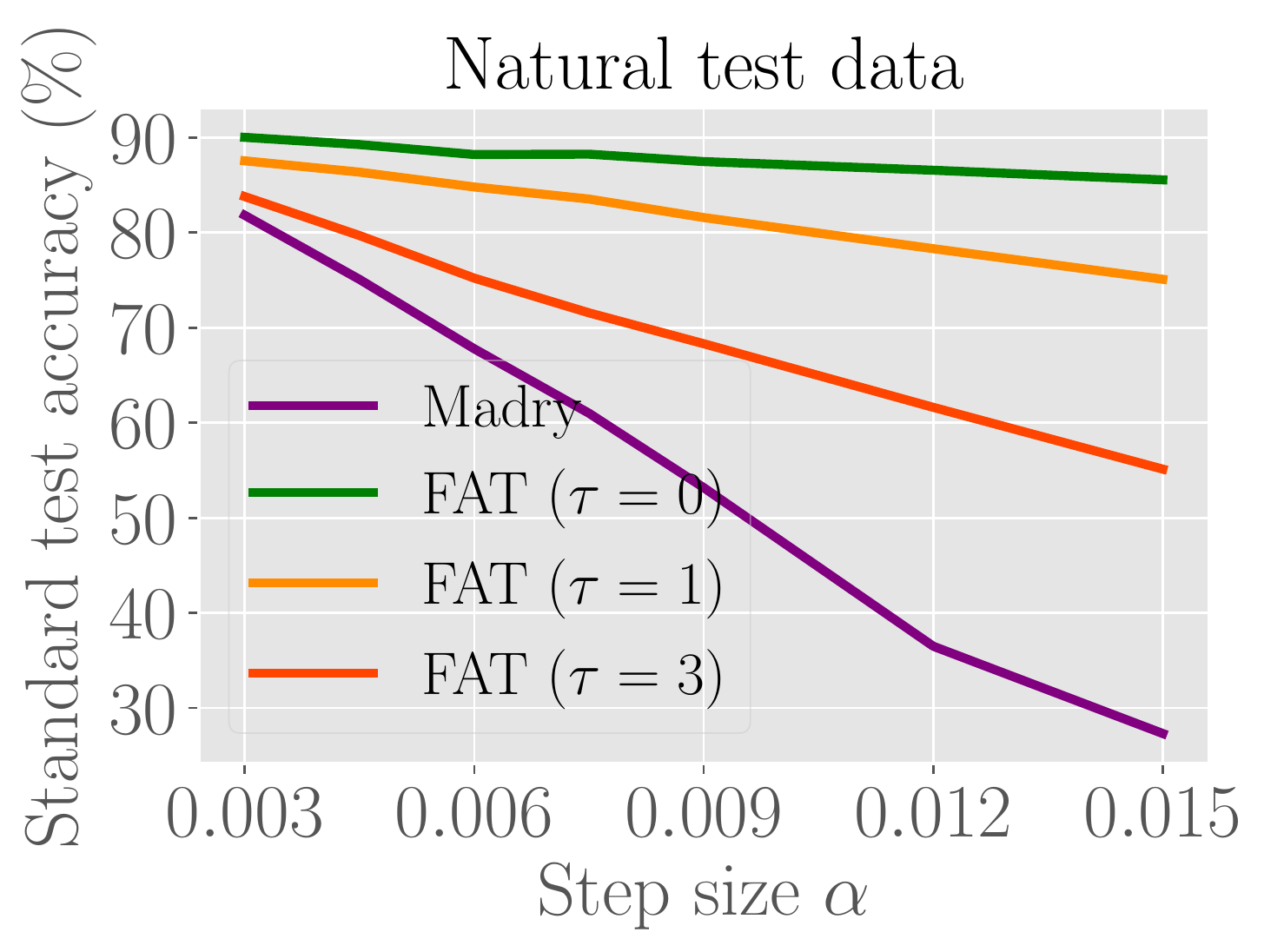}
    \includegraphics[scale=0.33]{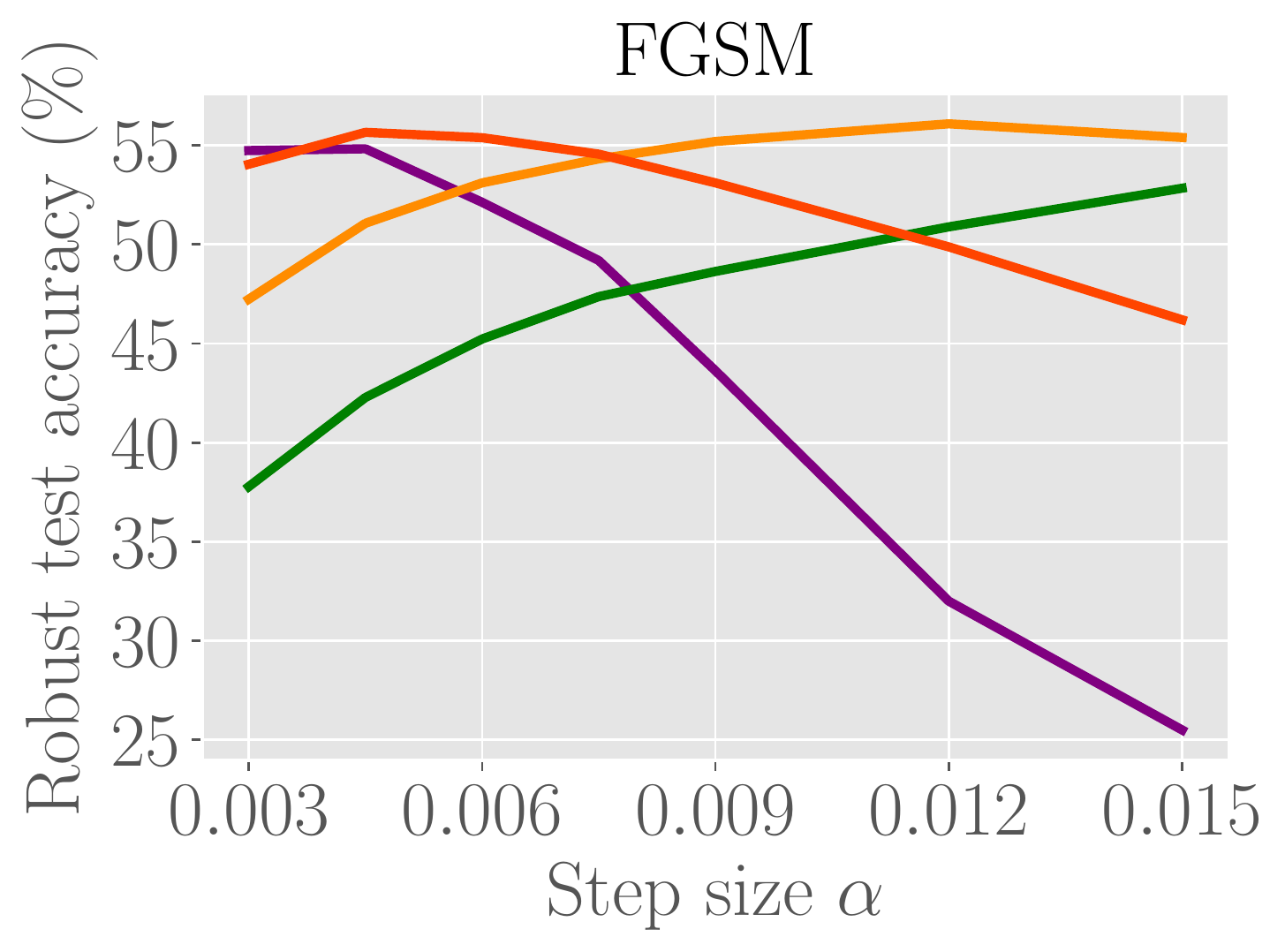}
    \includegraphics[scale=0.33]{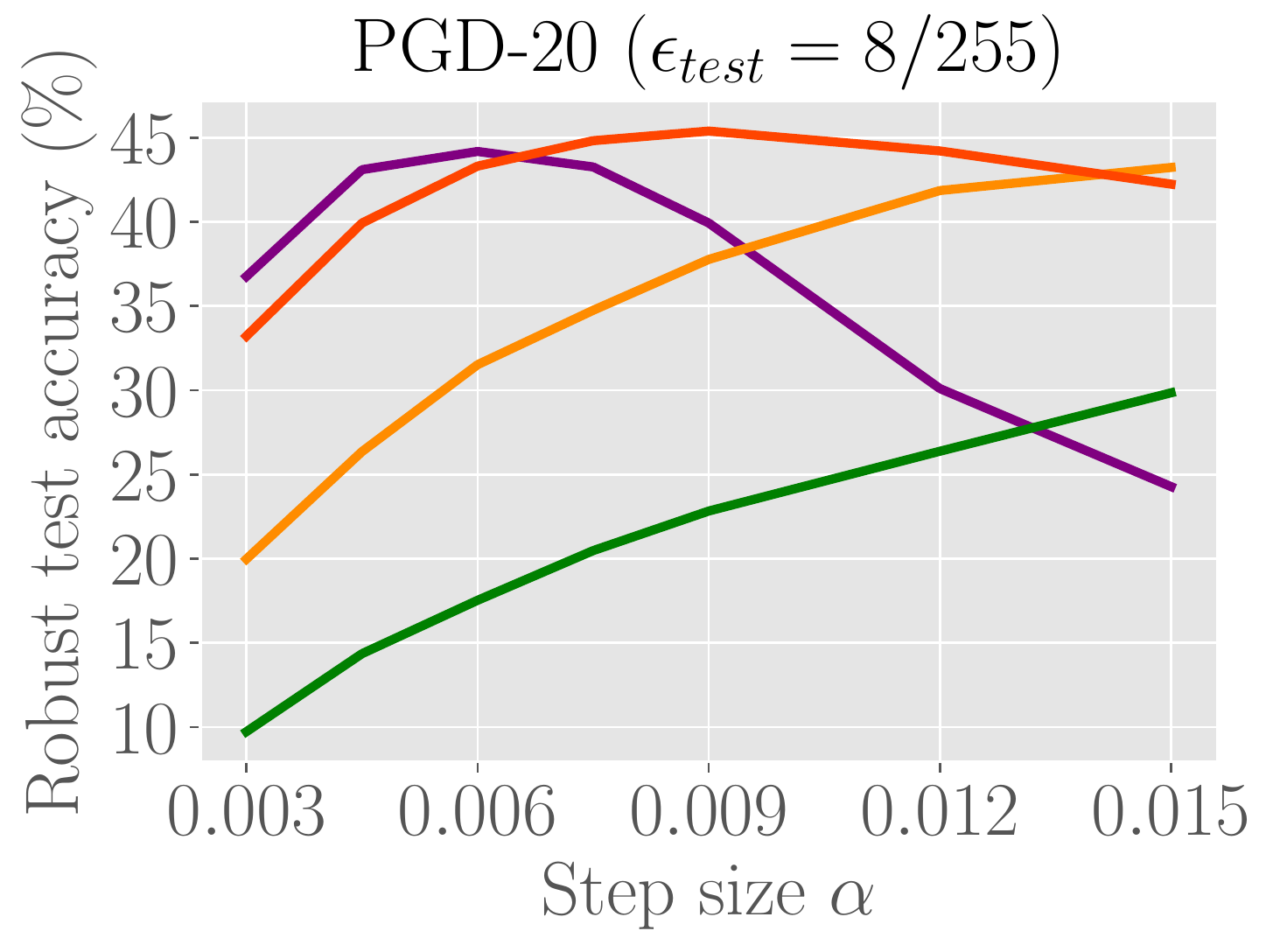}\\
    \includegraphics[scale=0.33]{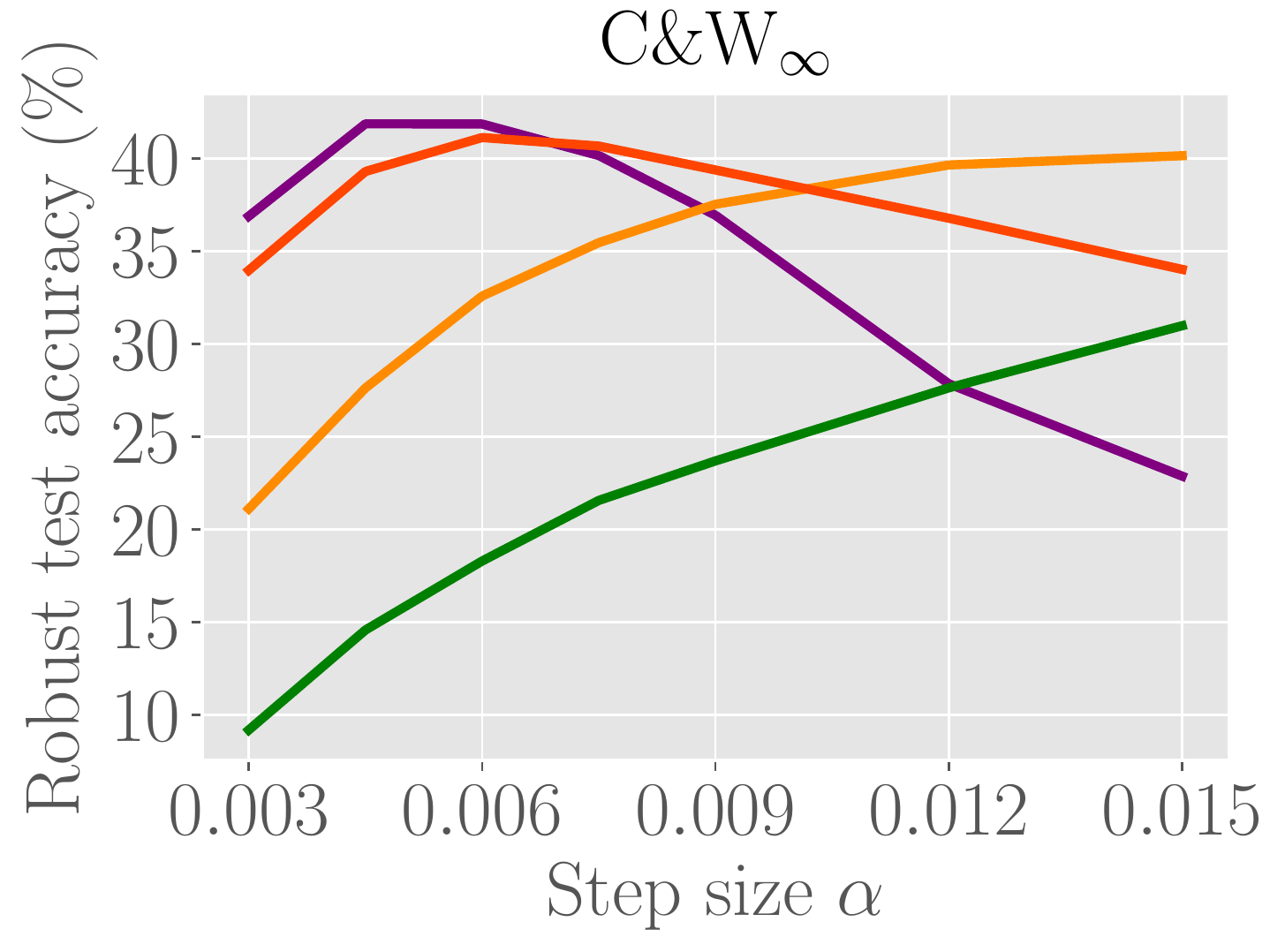}
    \includegraphics[scale=0.33]{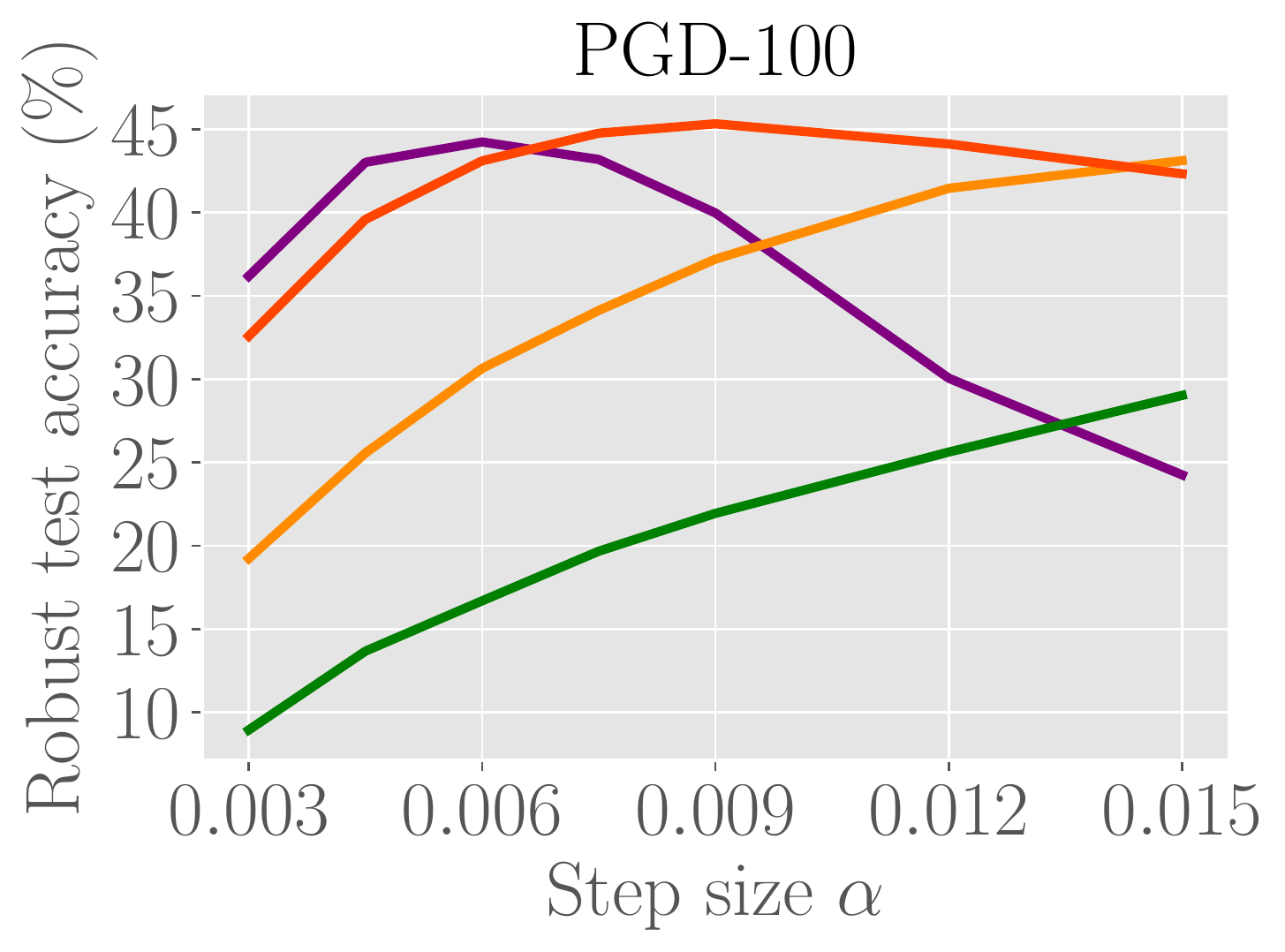}
    \includegraphics[scale=0.33]{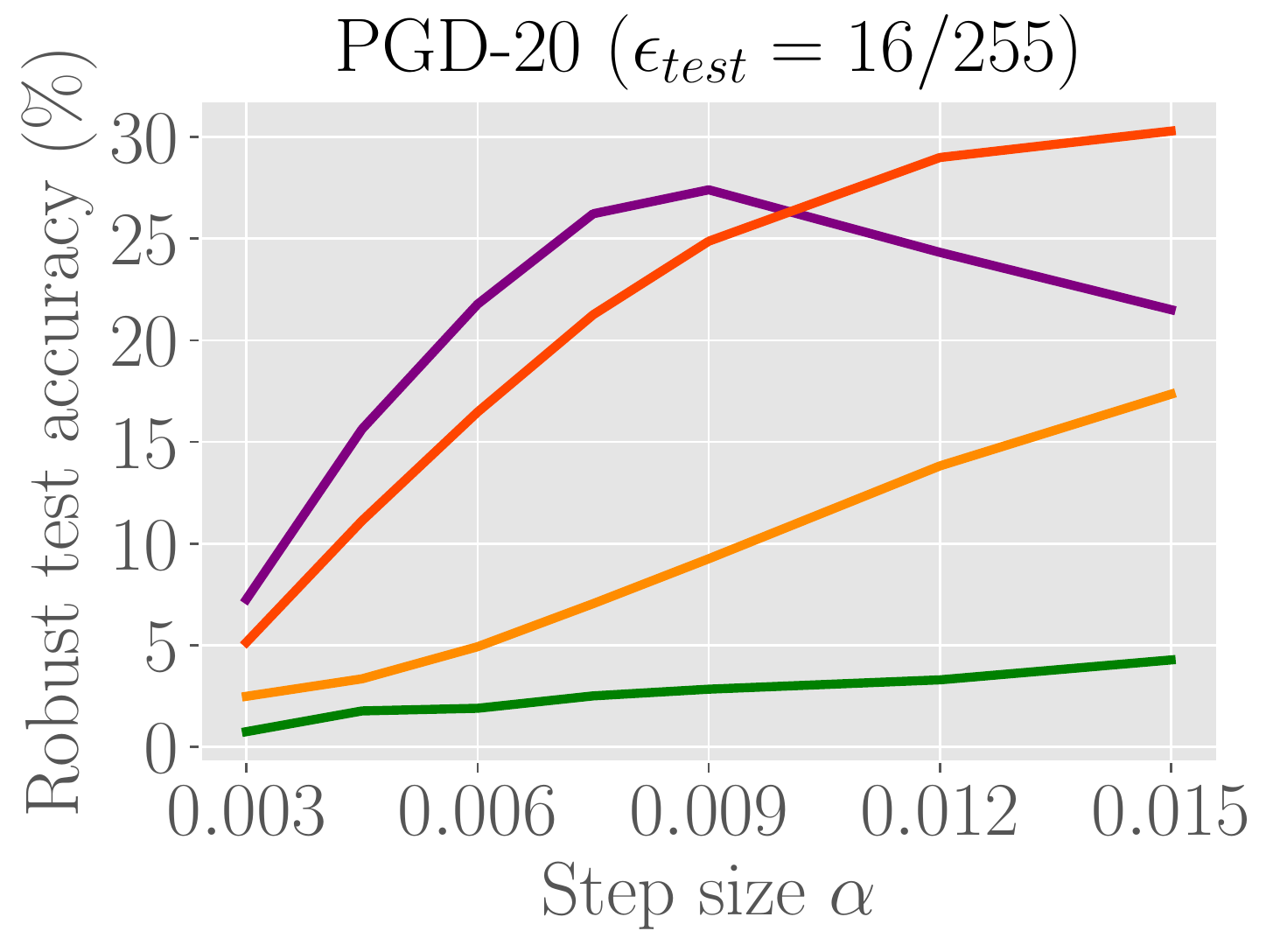}
    \caption{Test accuracy of Small CNN trained under different step size $\alpha$ on CIFAR-10}
    \label{fig:smallcnn_cifar10_dynamic_epsball_nopro}
\end{figure}

\begin{figure}[!htb]
    \centering
    \includegraphics[scale=0.33]{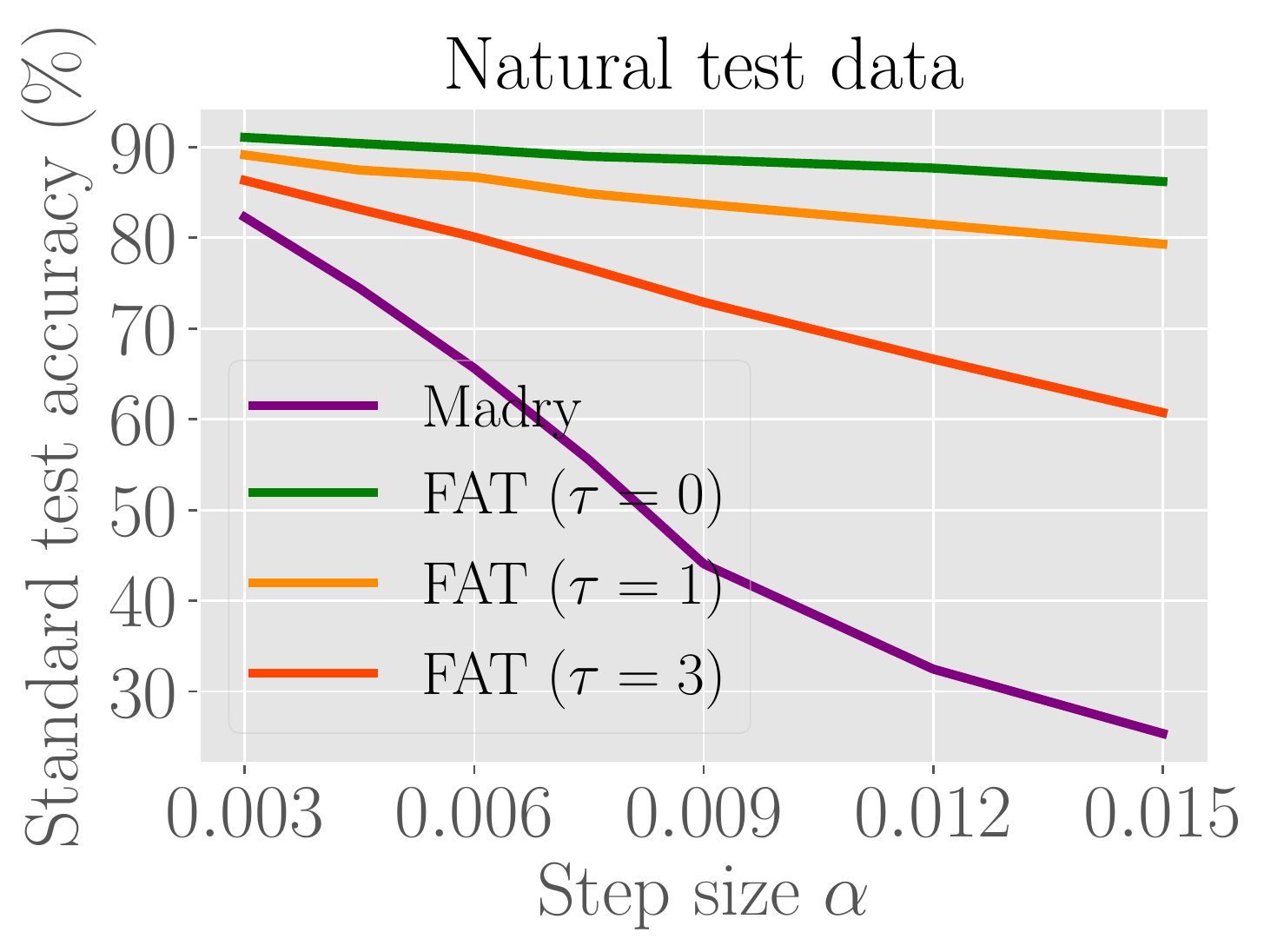}
    \includegraphics[scale=0.33]{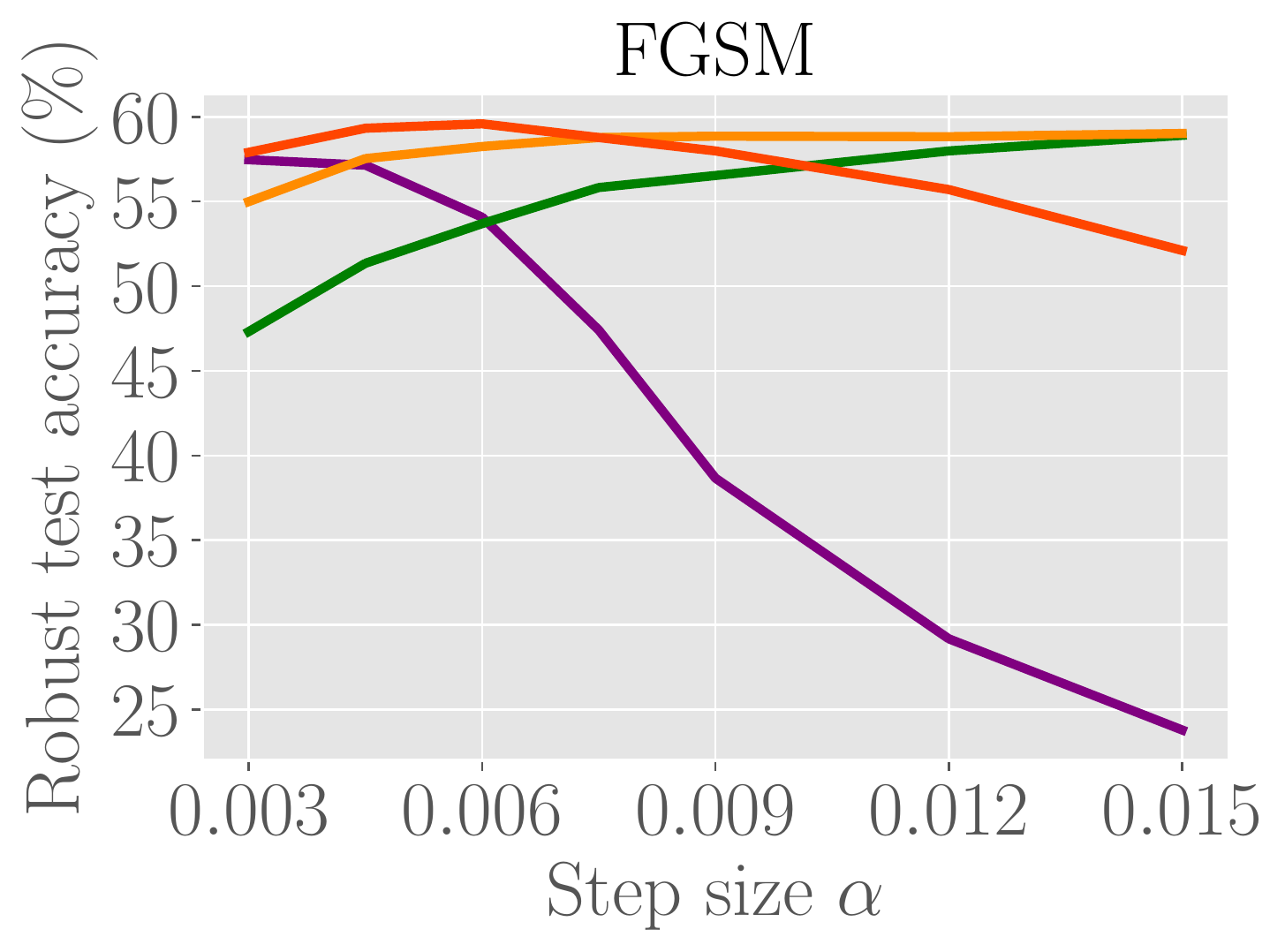}
    \includegraphics[scale=0.33]{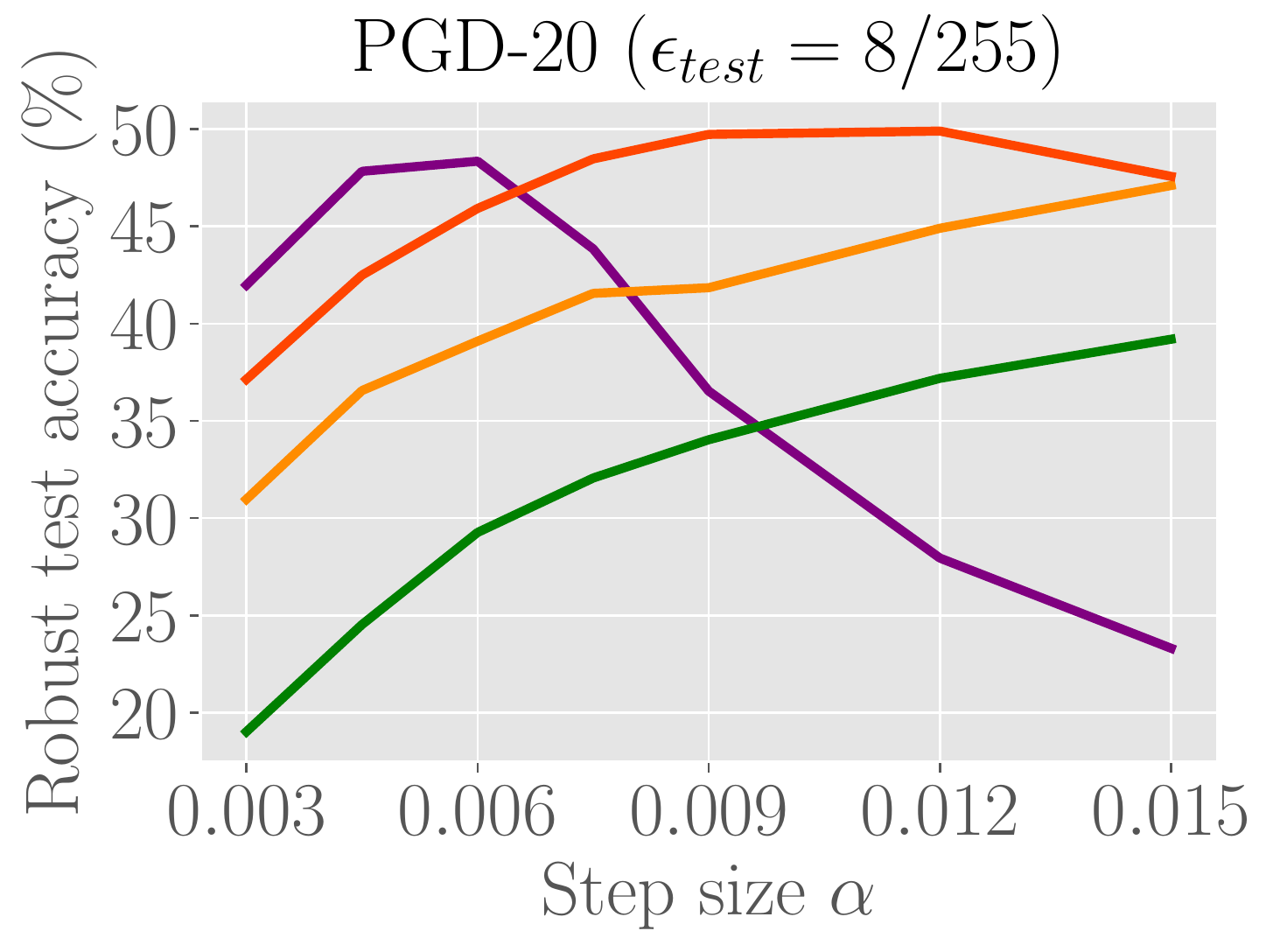}\\
    \includegraphics[scale=0.33]{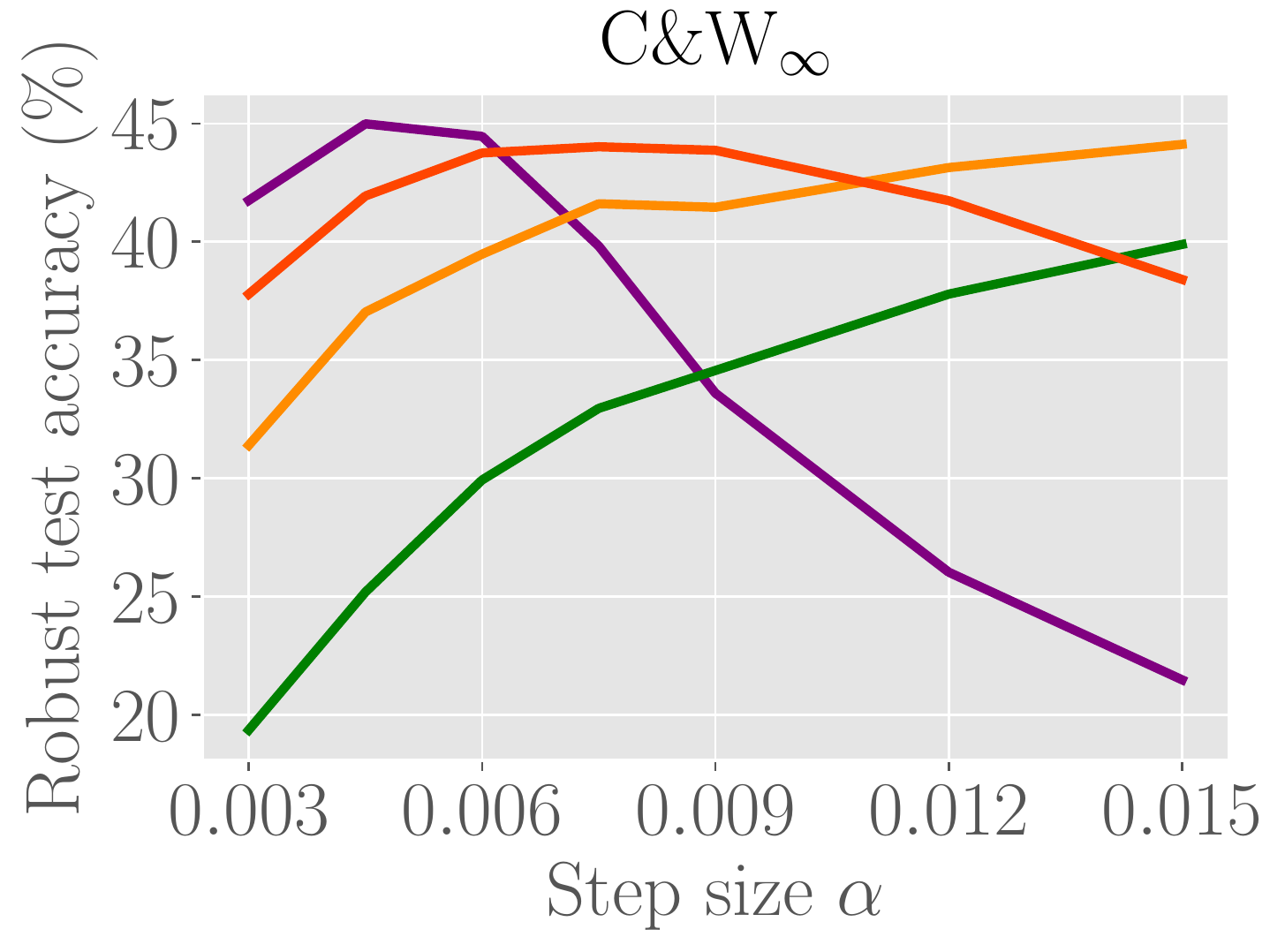}
    \includegraphics[scale=0.33]{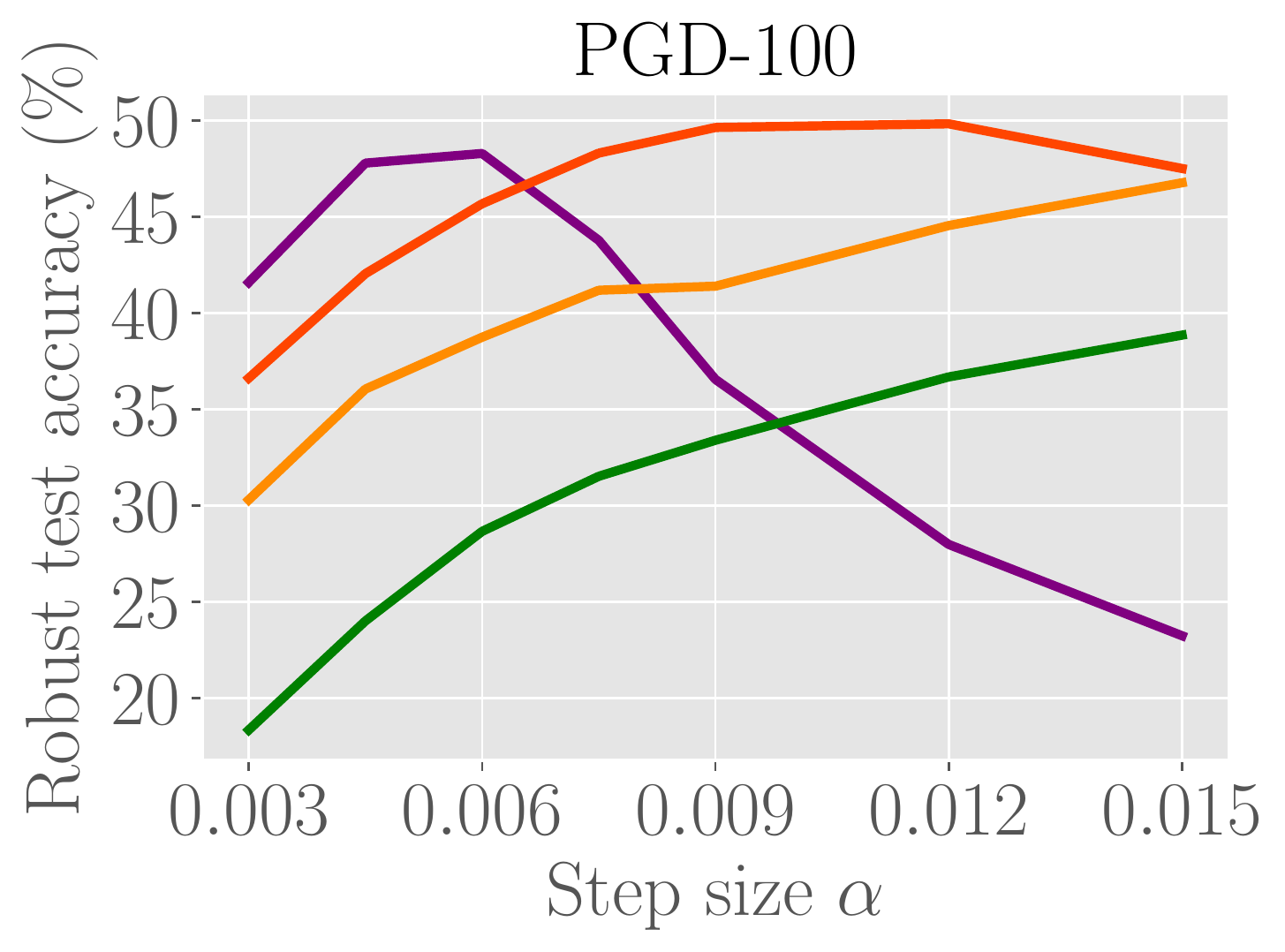}
    \includegraphics[scale=0.33]{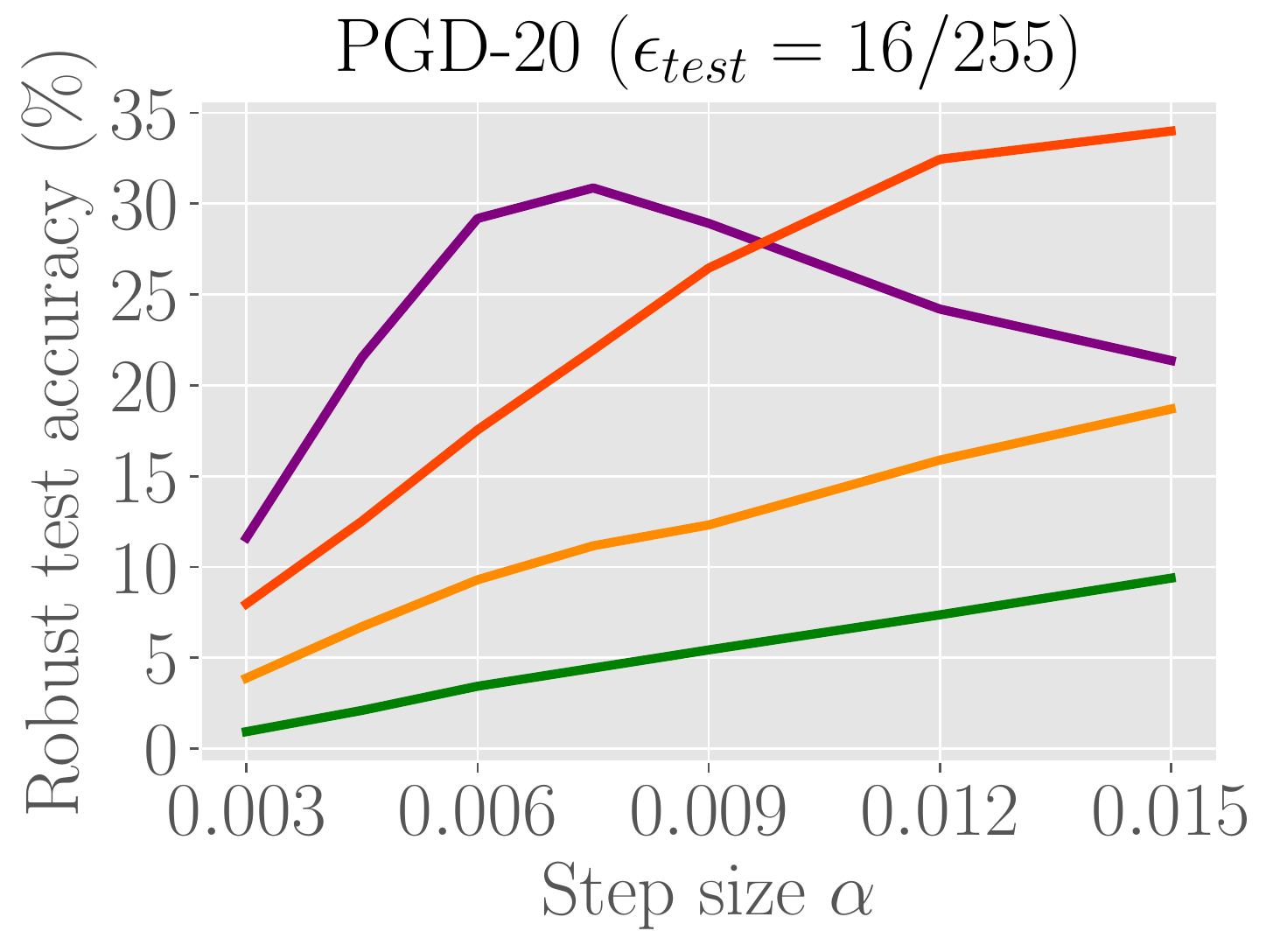}
    \caption{Test accuracy of ResNet-18 trained under different step size $\alpha$ on CIFAR-10}
    \label{fig:resnet18_cifar10_dynamic_epsball_nopro}
\end{figure}

We employ Small CNN and ResNet-18 to show the performance of deep models against FGSM, PGD-20, PGD-100 and C$\&$W$_{\infty}$ in Figure~\ref{fig:smallcnn_cifar10_dynamic_epsball_nopro} and Figure~\ref{fig:resnet18_cifar10_dynamic_epsball_nopro}. Deep models are trained using SGD with 0.9 momentum for 80 epochs with the initial learning rate 0.01 divided by 10 at 60 epoch. We compare FAT combined with the alternative adversarial data searching algorithm ($\tau = 0,1,3$) and standard adversarial training (Madry) with different step size $\alpha$, i.e., $\alpha \in [0.003,0.015]$. The maximum PGD step $K$ is fixed to 10. All the testing settings are the same as those are stated in Section~\ref{section:fat_enable_lager_epsilon}.  

\clearpage

\section{Mixture Alleviation}
\label{appendix:mixture_alleviation}
\subsection{Output Distributions of Small CNN's Intermediate Layers}
In Figure~\ref{fig:overshoot_cifar} in Section~\ref{Section:Mixture_alleviation}, we only visualize layer $\#7$’s output distribution by Small CNN (8-layer convolutional neural network with 6 convolutional layers and 2 fully connected layers). For completeness, we visualize the output distributions of layers $\#7$ and $\#8$.

\begin{figure}[h!]
	\subfigure[Output distributions visualized by PCA]{
		\begin{minipage}[b]{0.5\textwidth}
		    \centering
		    \includegraphics[scale=0.18]{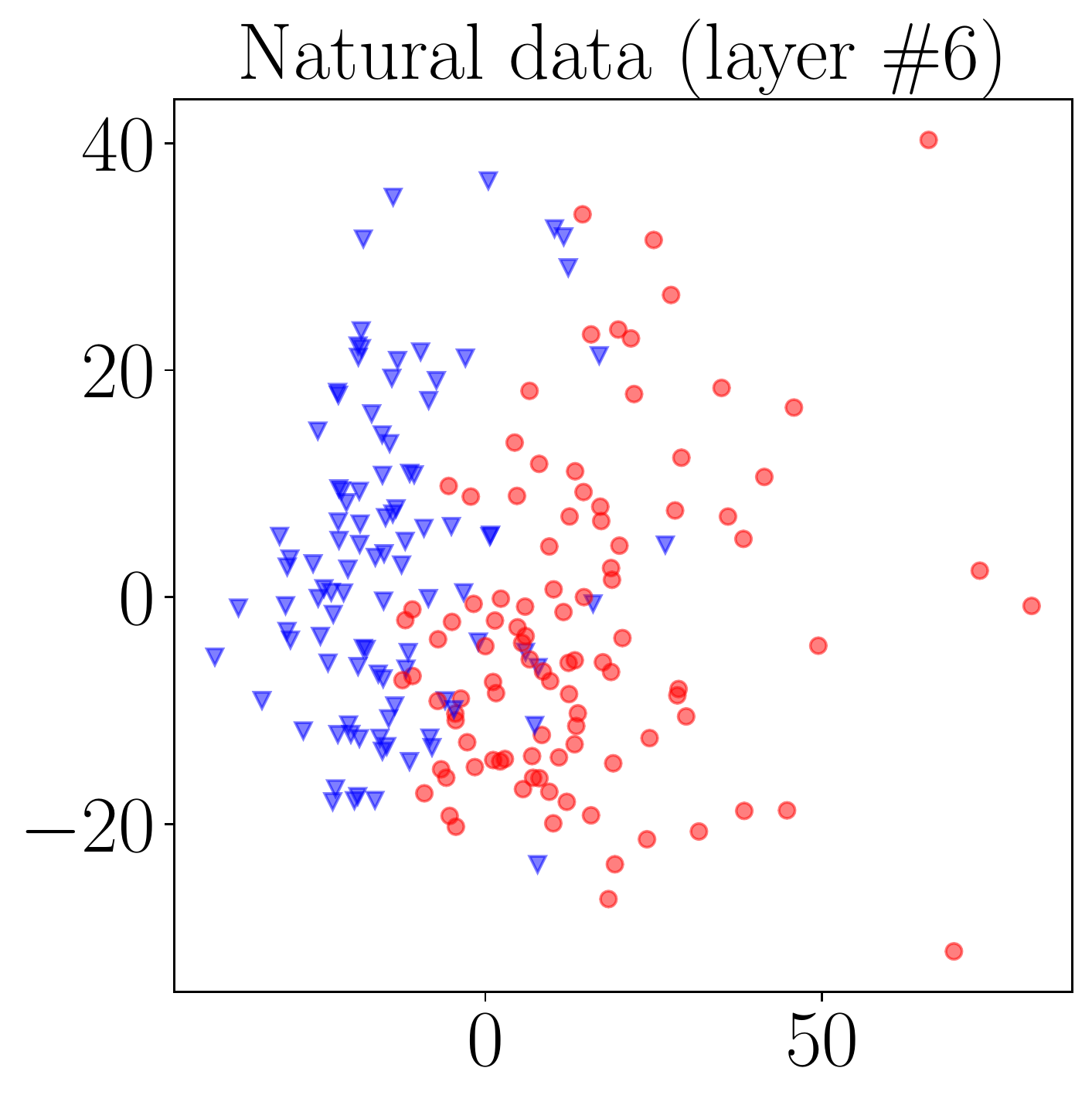} 
            \includegraphics[scale=0.18]{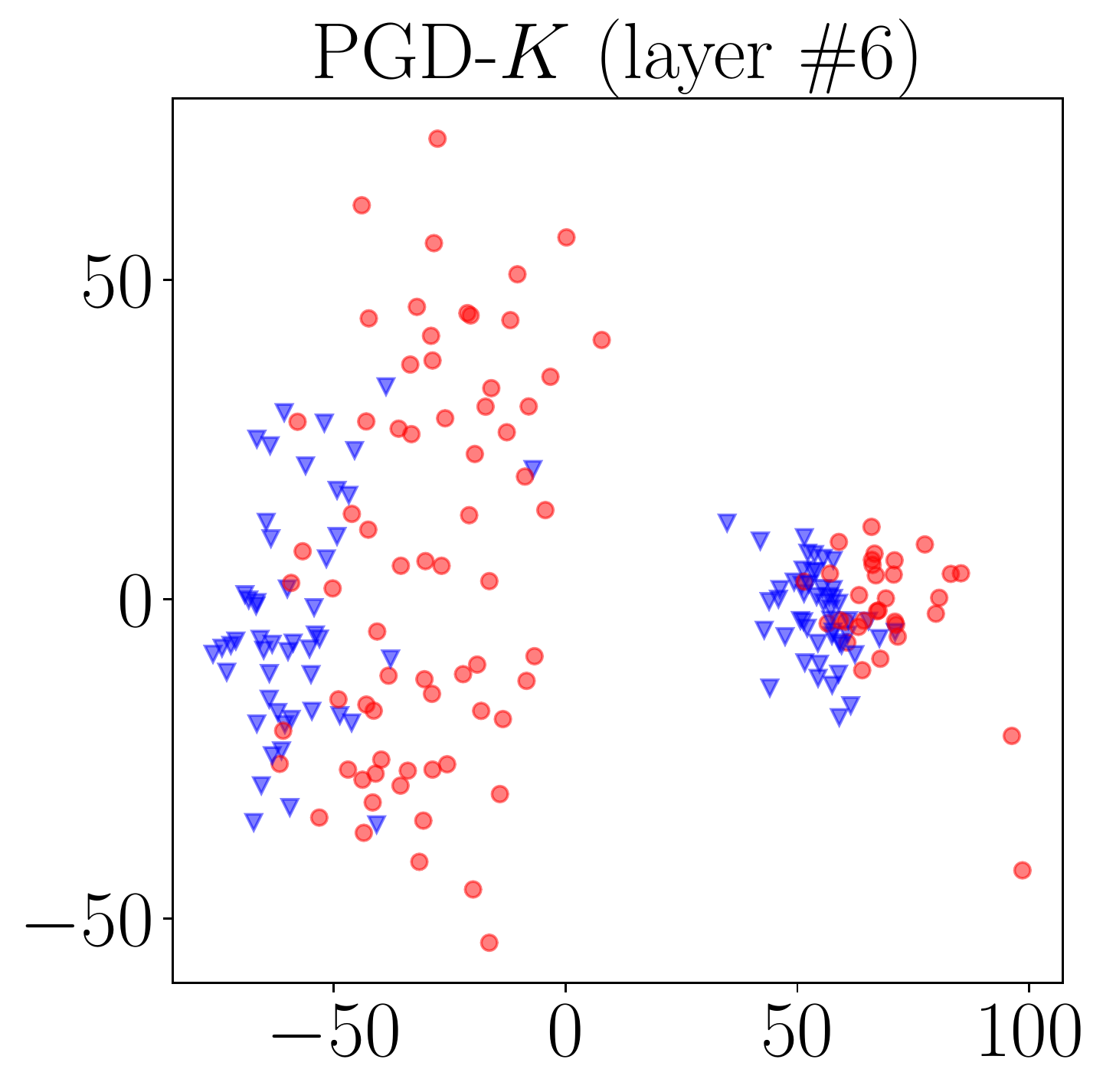}
            \includegraphics[scale=0.18]{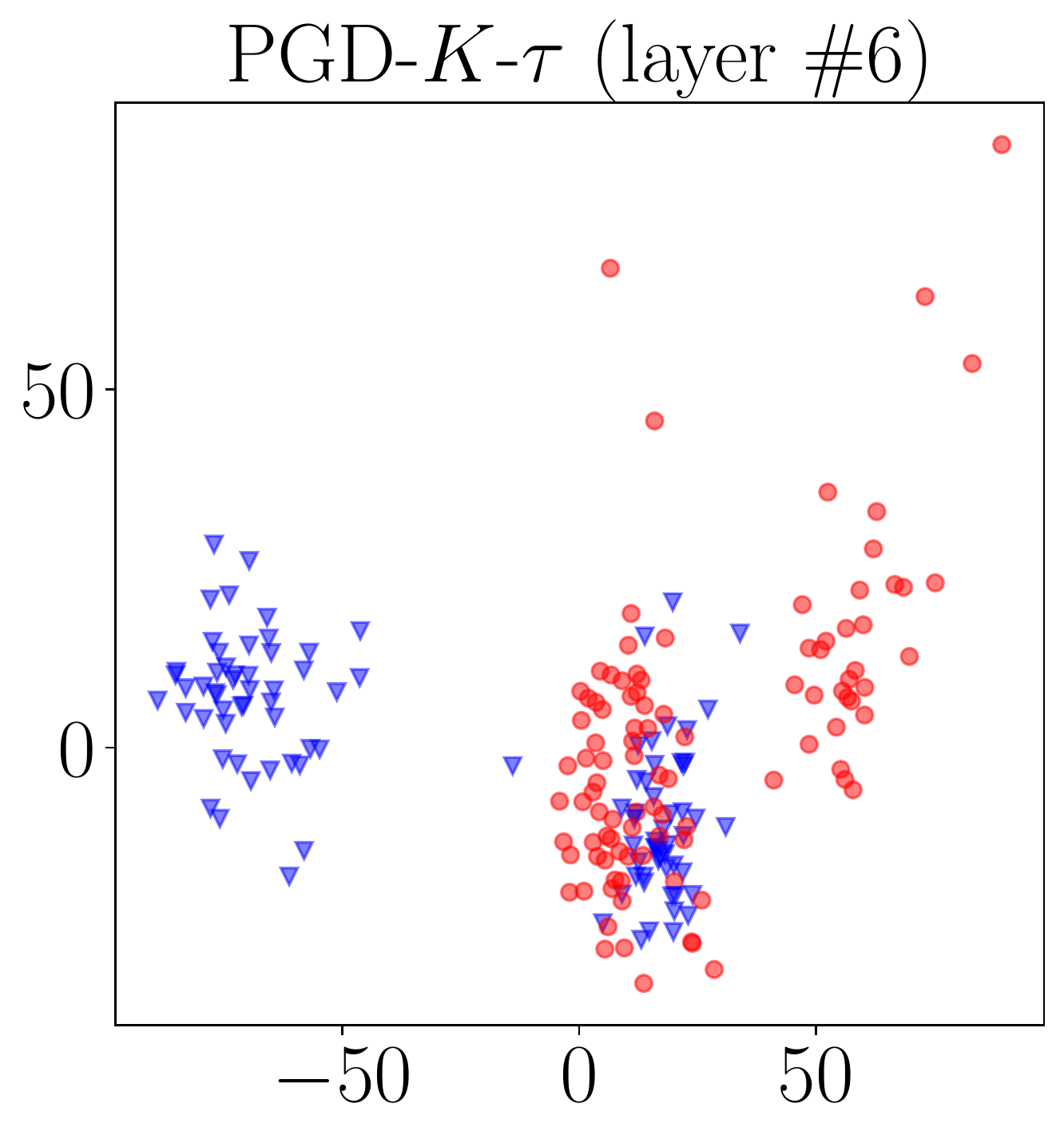}
            \quad
            \includegraphics[scale=0.18]{overshoot_issue_cifar/binary/24-l7/wo-clean--epoch80.pdf} 
            \includegraphics[scale=0.18]{{overshoot_issue_cifar/binary/24-l7/wo-pgd20-eps0.031-stepsize0.007-epoch80}.pdf}
            \includegraphics[scale=0.18]{{overshoot_issue_cifar/binary/24-l7/wo-pgd20-earlystop-eps0.031-stepsize0.007-epoch80}.pdf}
            \quad
            \includegraphics[scale=0.18]{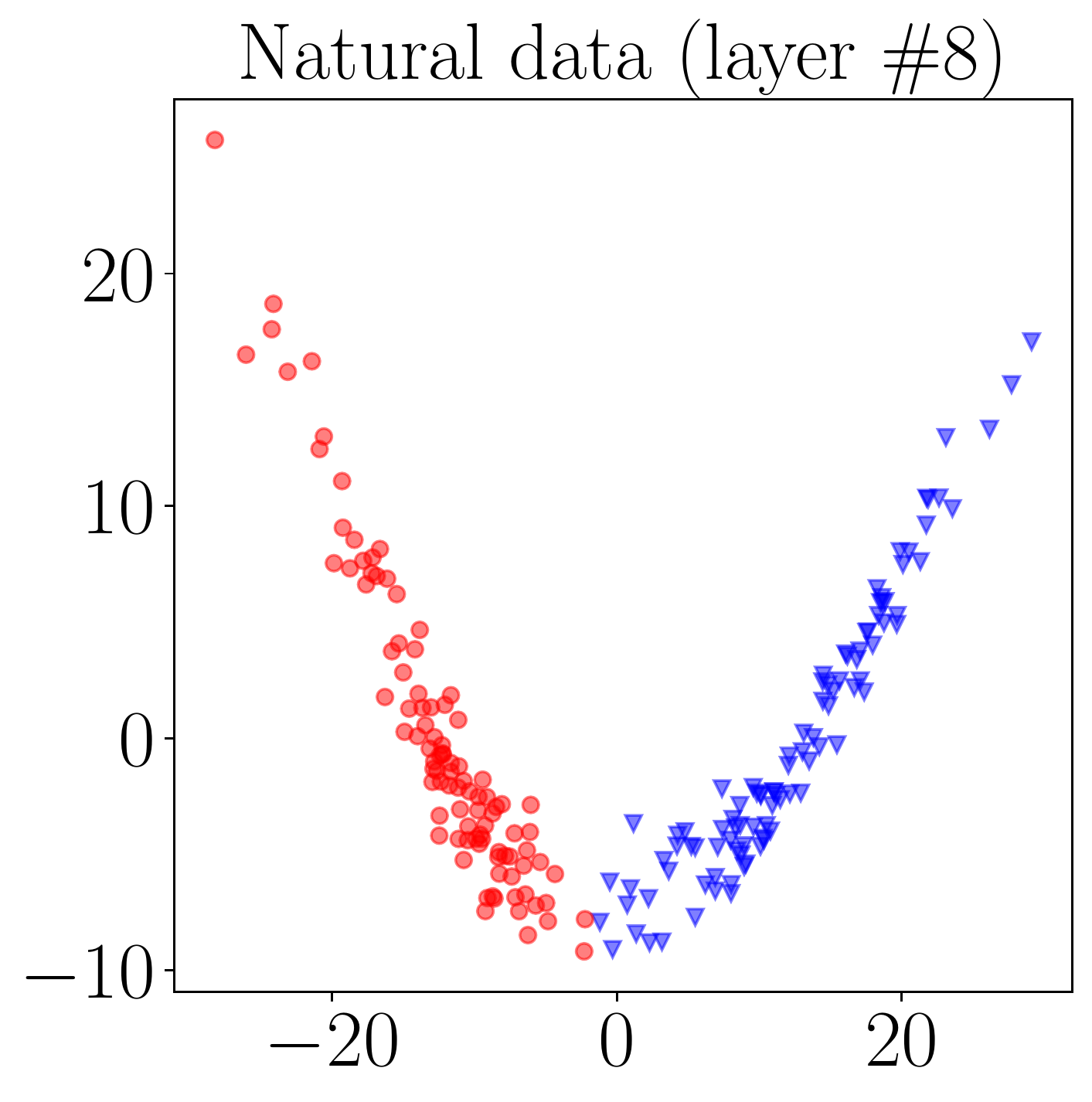} 
            \includegraphics[scale=0.18]{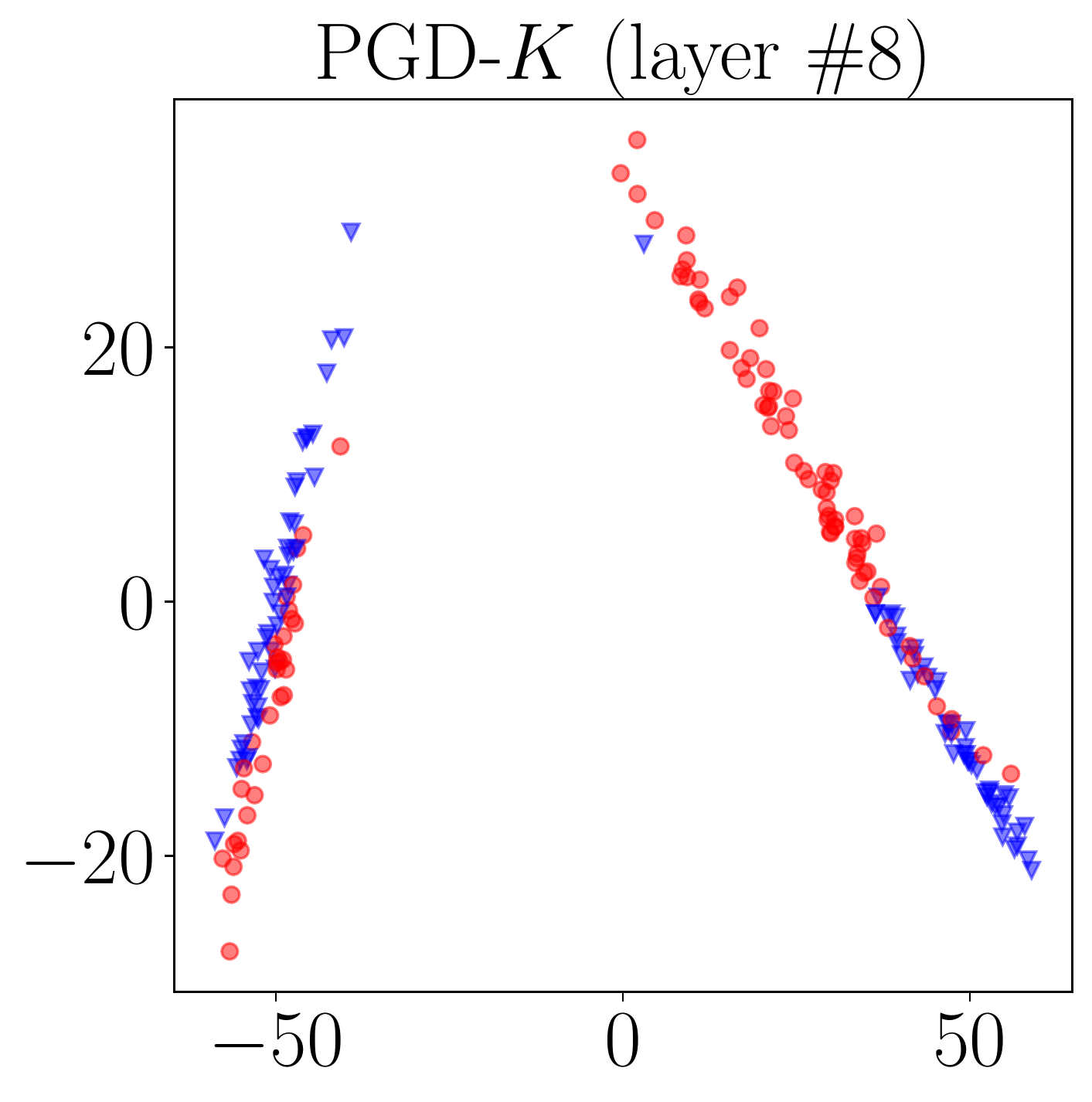}
            \includegraphics[scale=0.18]{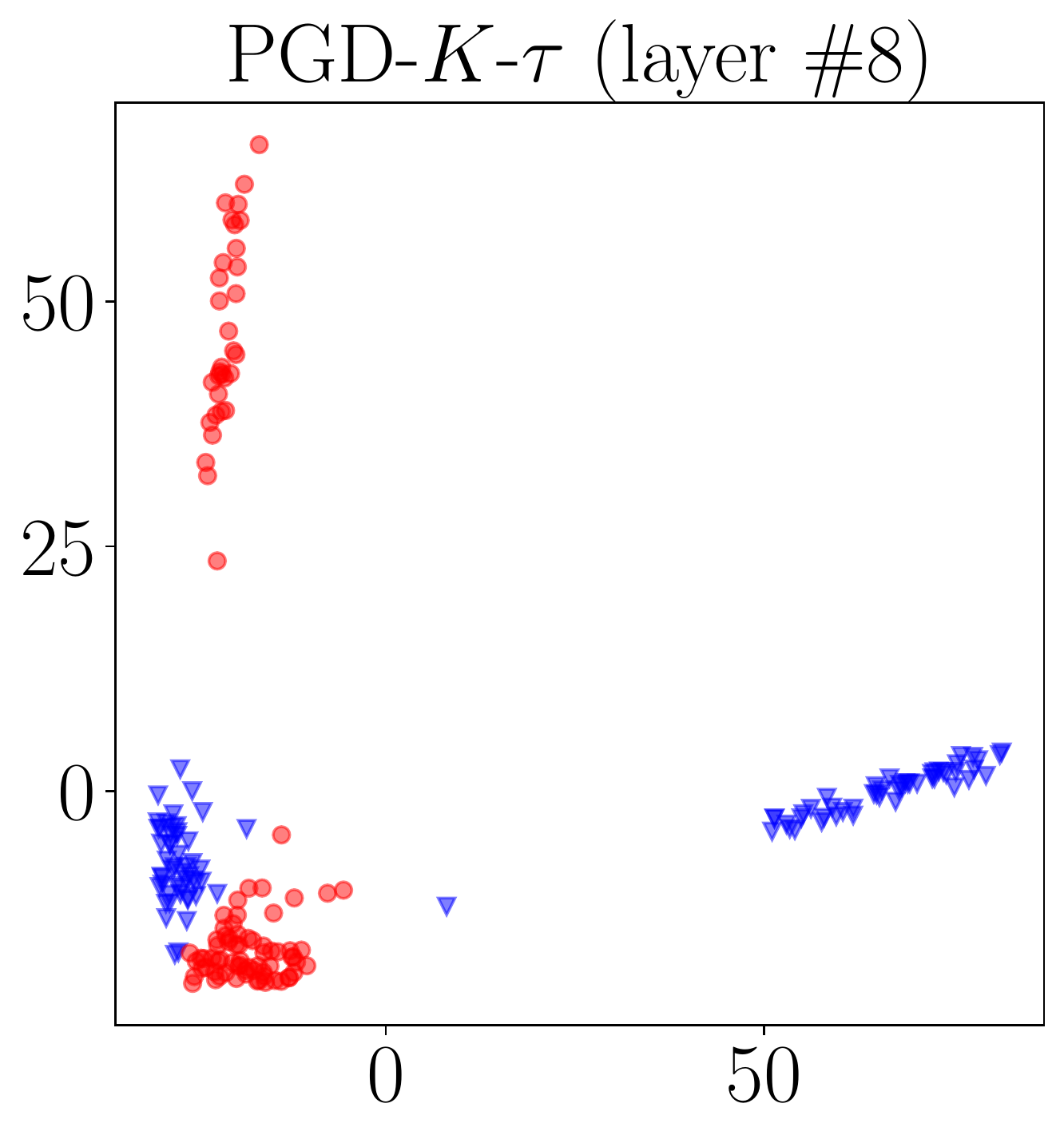}
		\end{minipage}
		\label{fig:appendix_smallcnn_overshoot_pca}
	}
	\subfigure[Output distributions visualized by t-SNE]{
		\begin{minipage}[b]{0.5\textwidth}
		    \centering
   	 	    \includegraphics[scale=0.18]{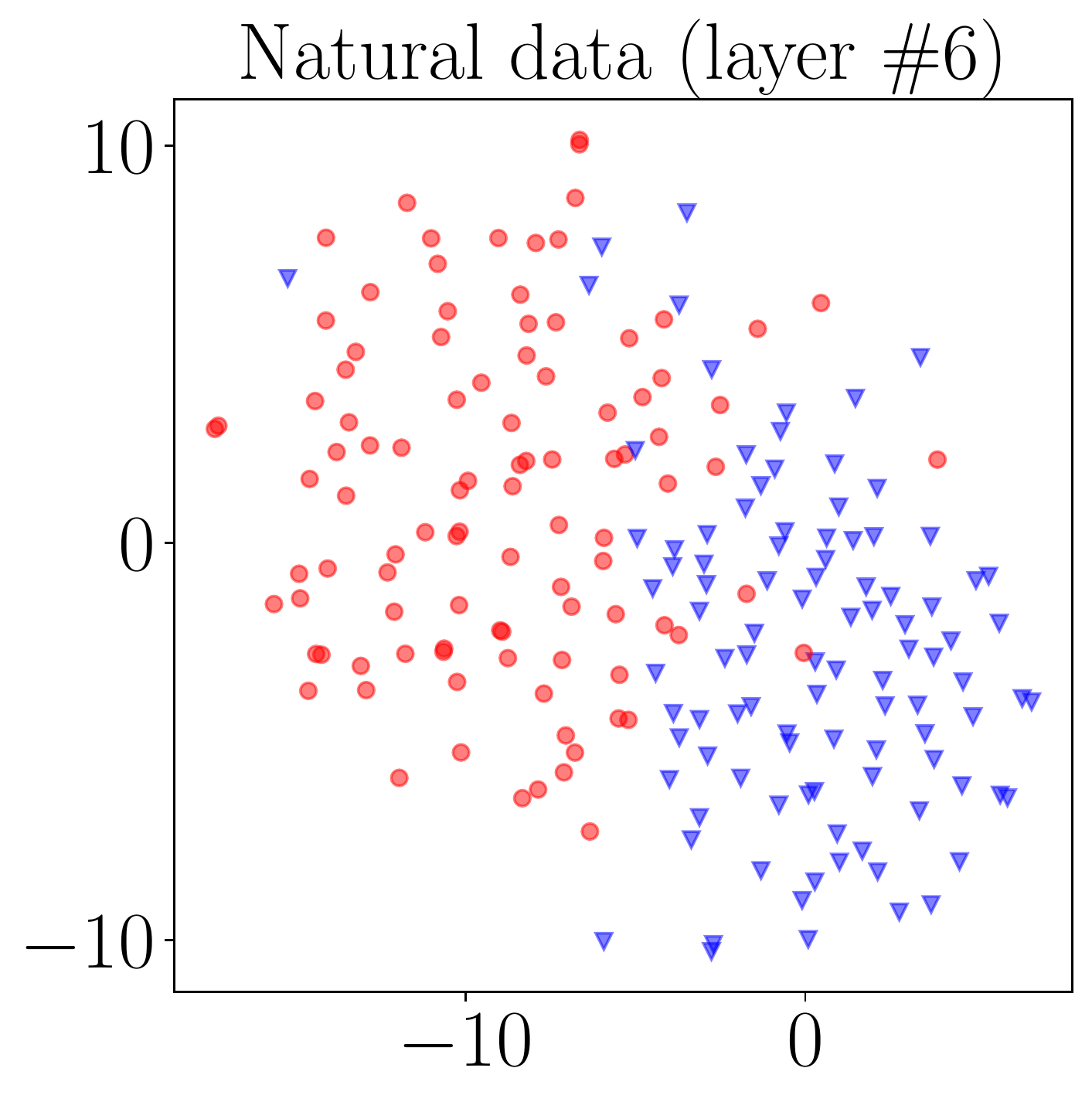} 
            \includegraphics[scale=0.18]{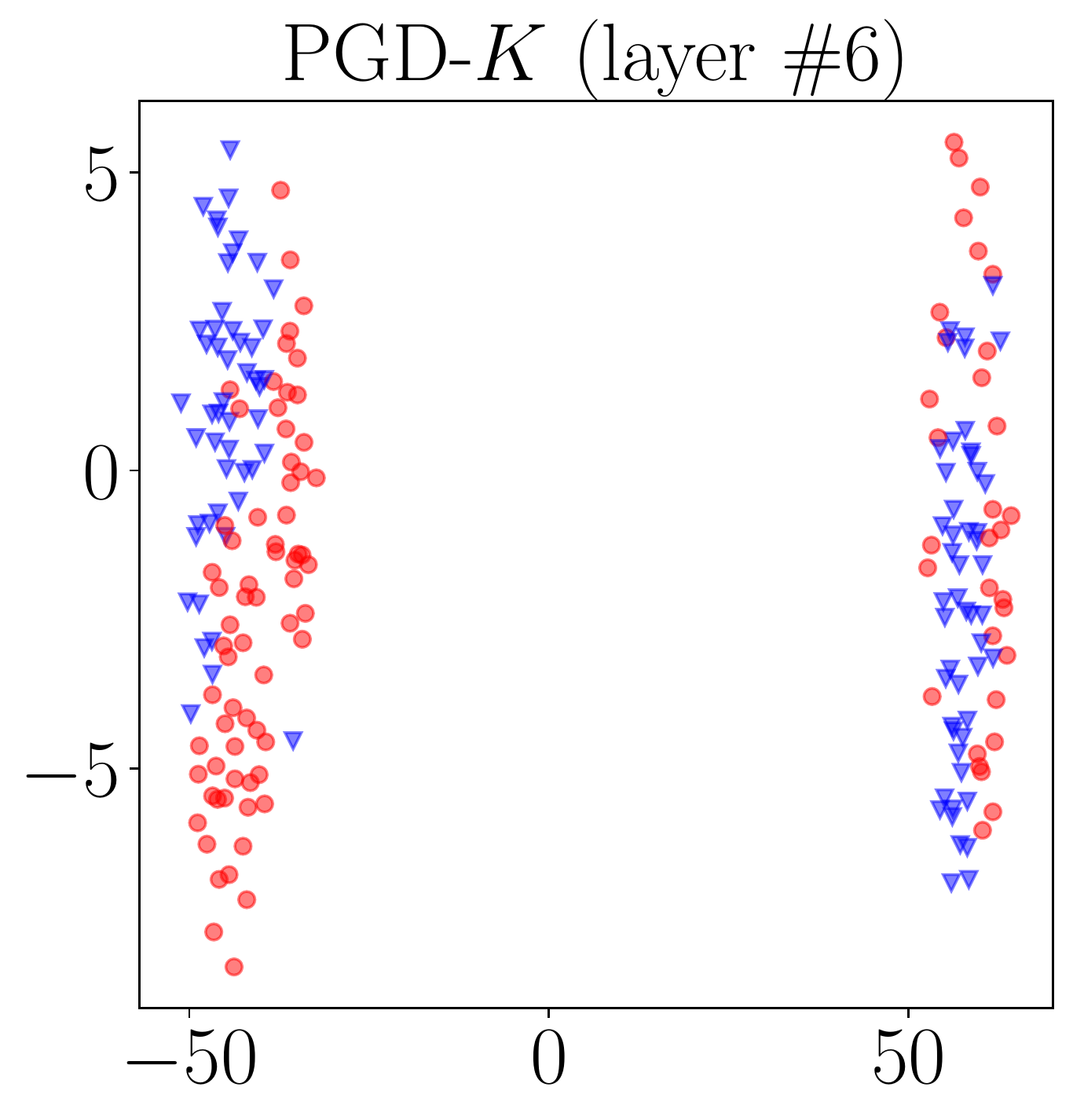}
            \includegraphics[scale=0.18]{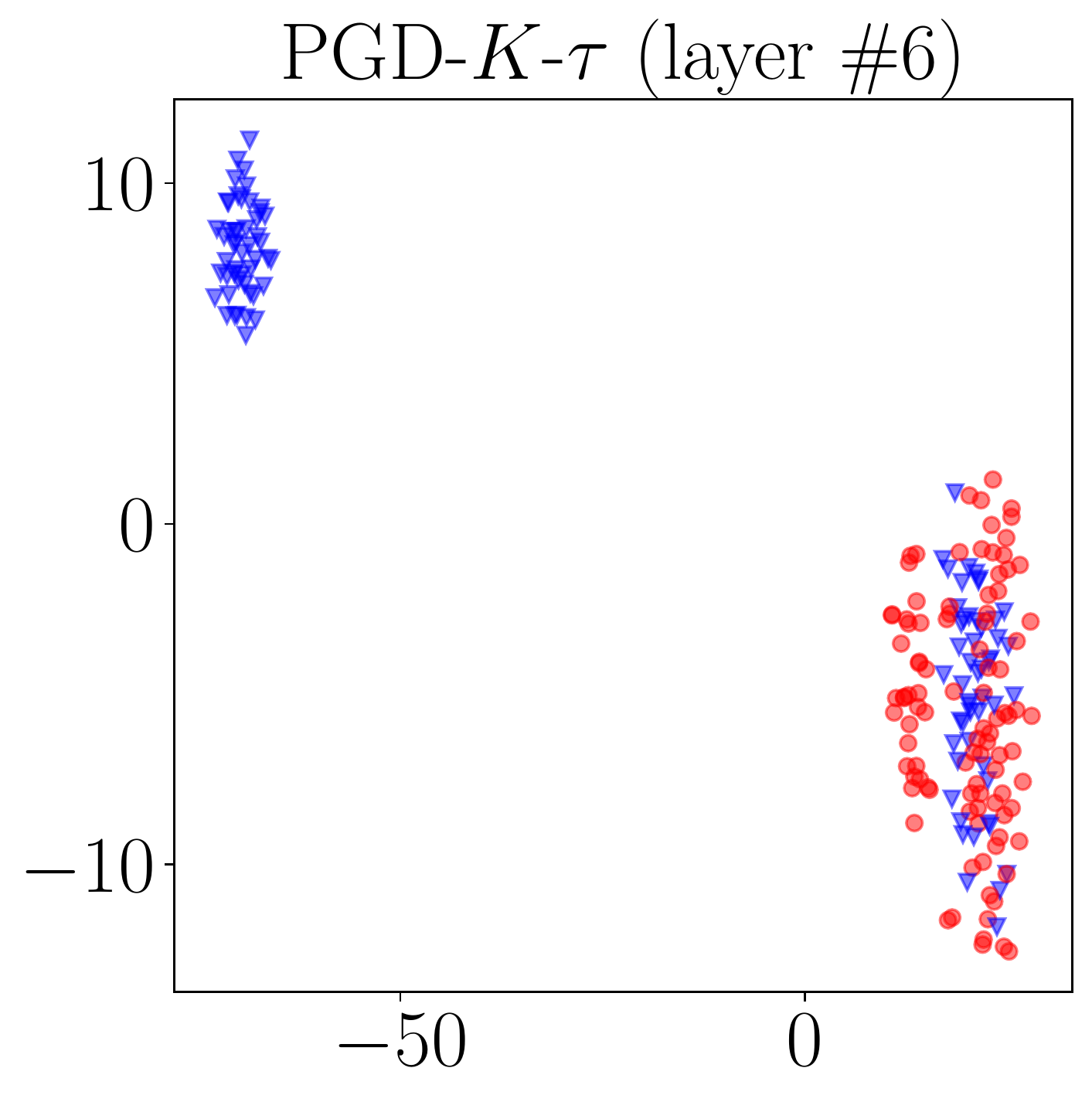}
            \quad
            \includegraphics[scale=0.18]{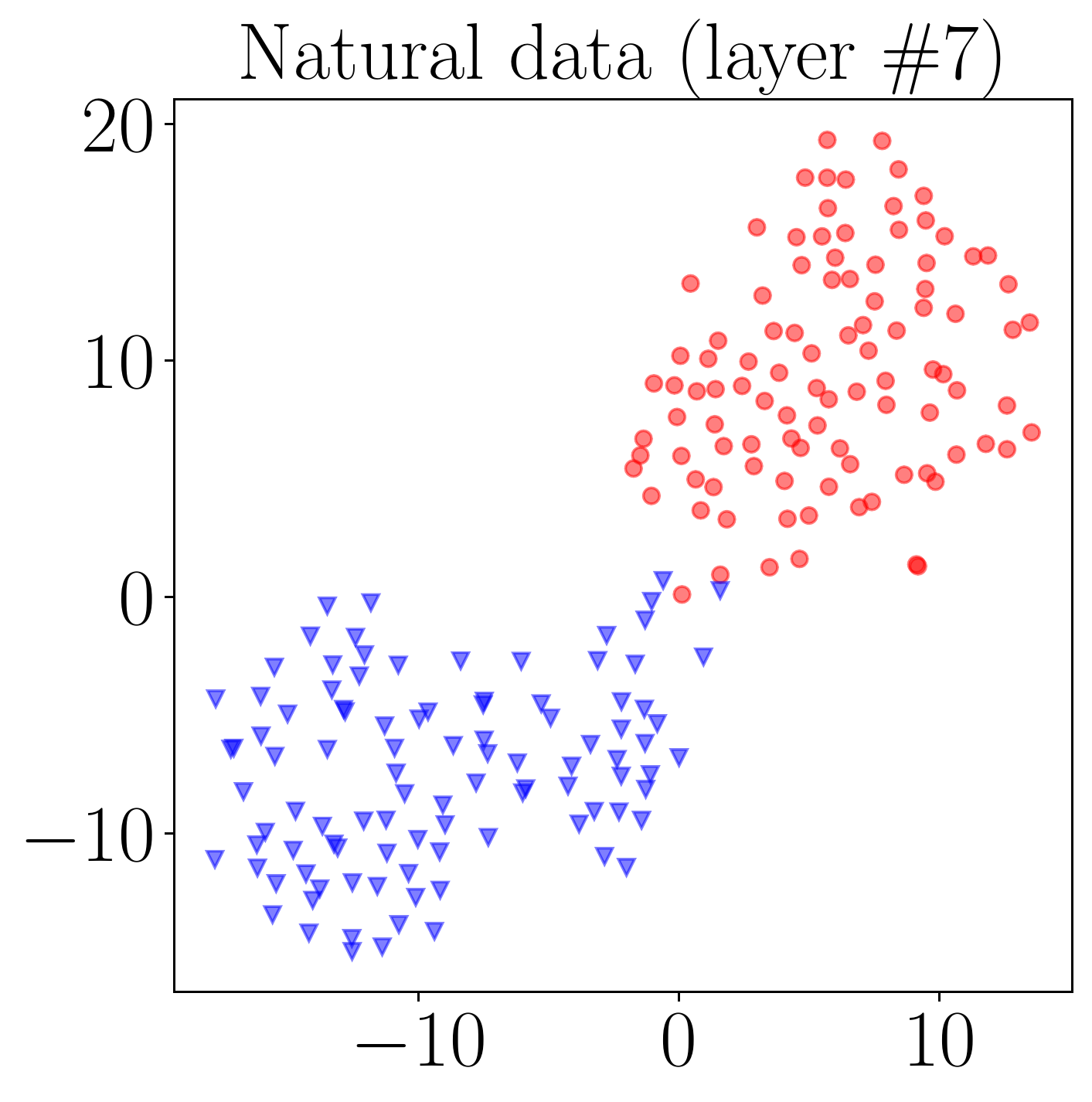} 
            \includegraphics[scale=0.18]{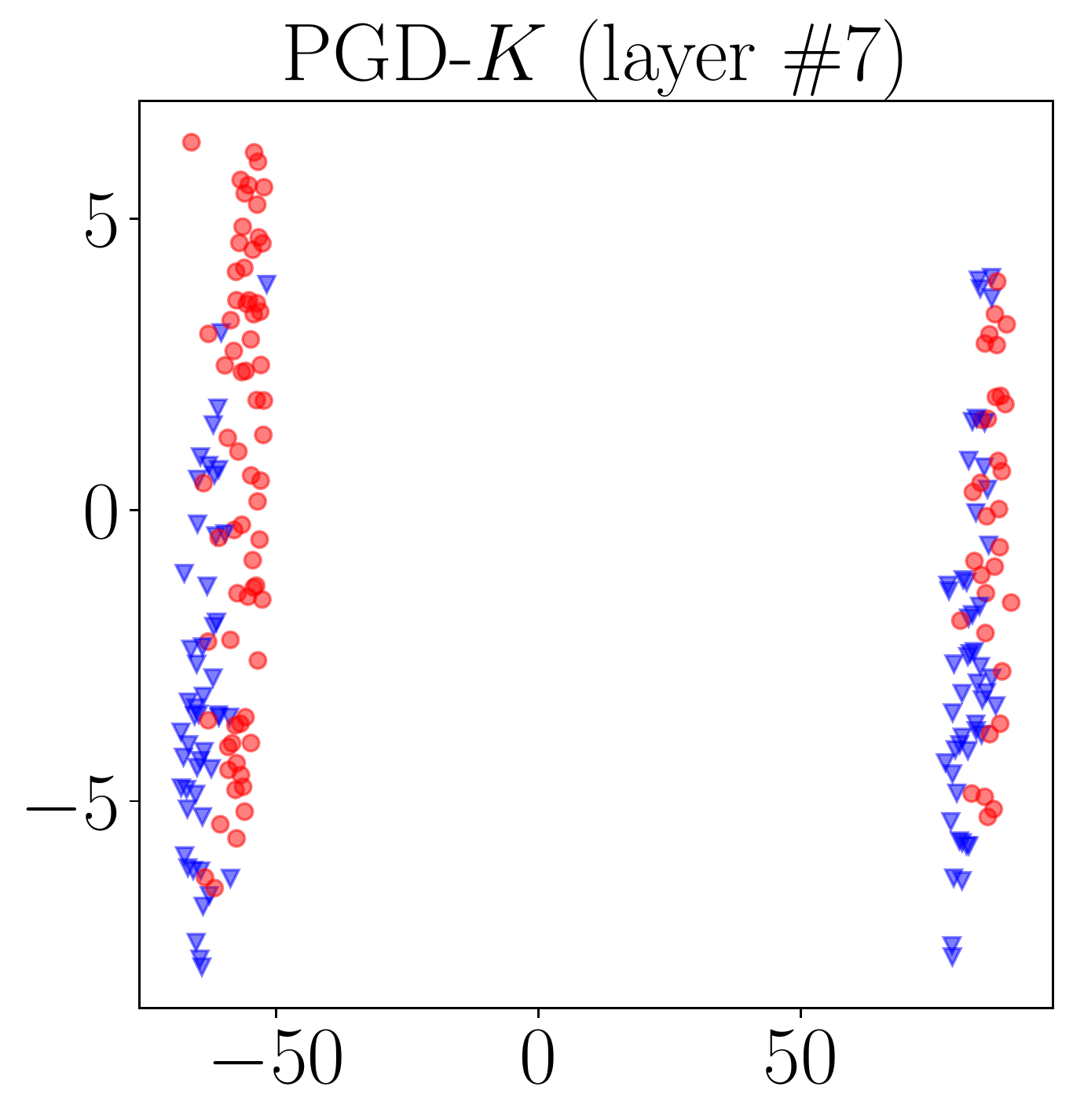}
            \includegraphics[scale=0.18]{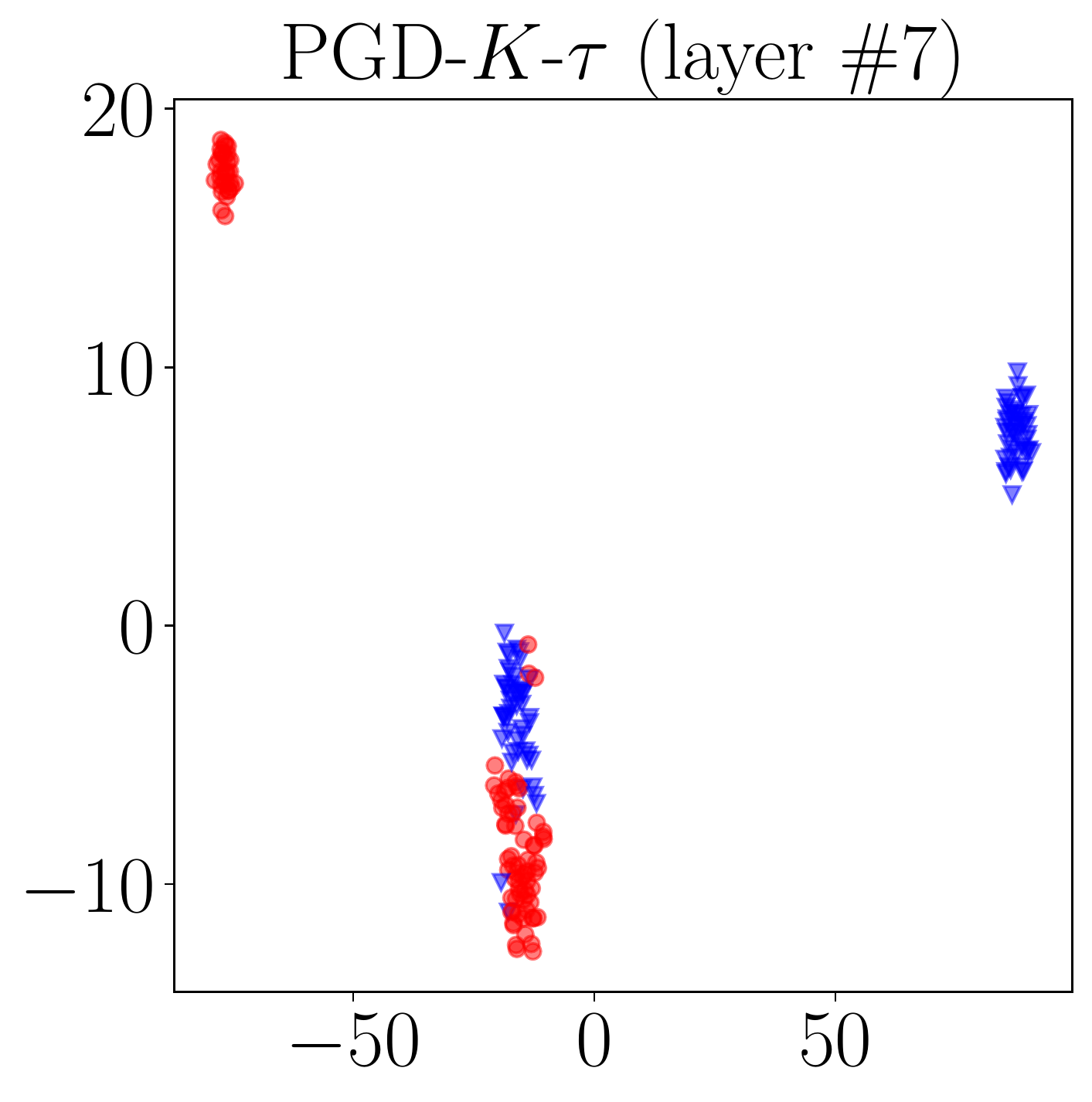}
            \quad
            \includegraphics[scale=0.18]{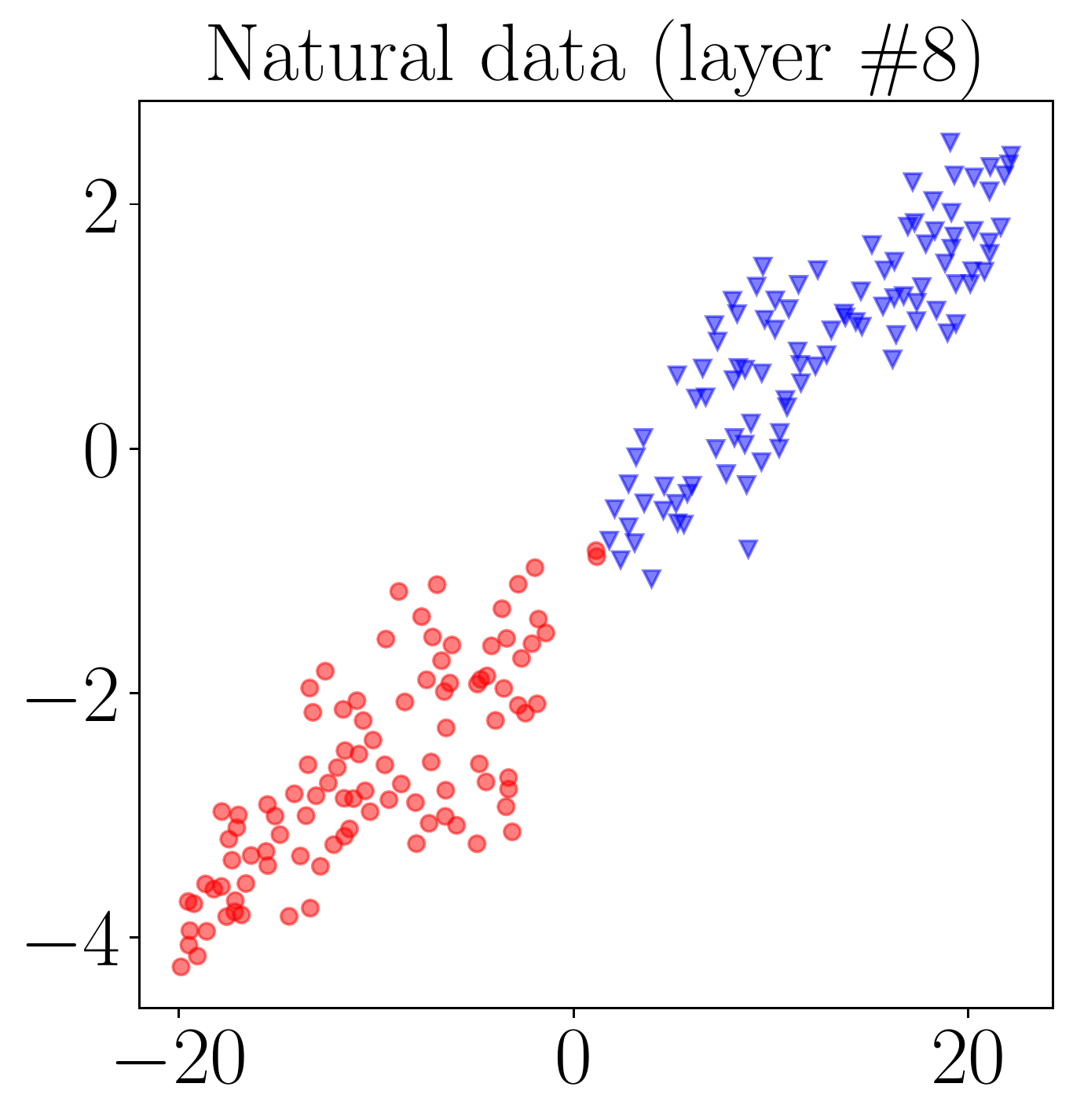} 
            \includegraphics[scale=0.18]{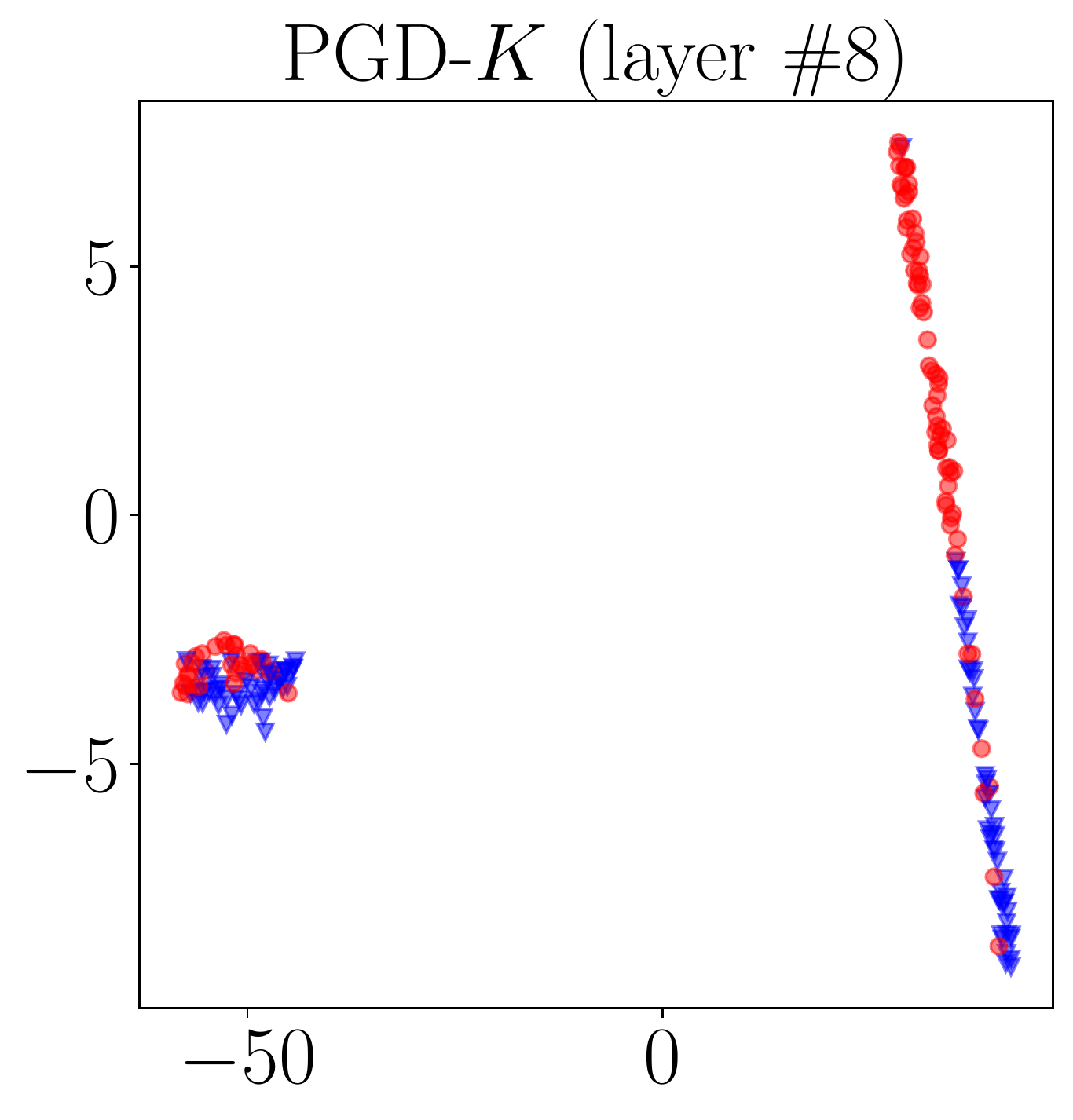}
            \includegraphics[scale=0.18]{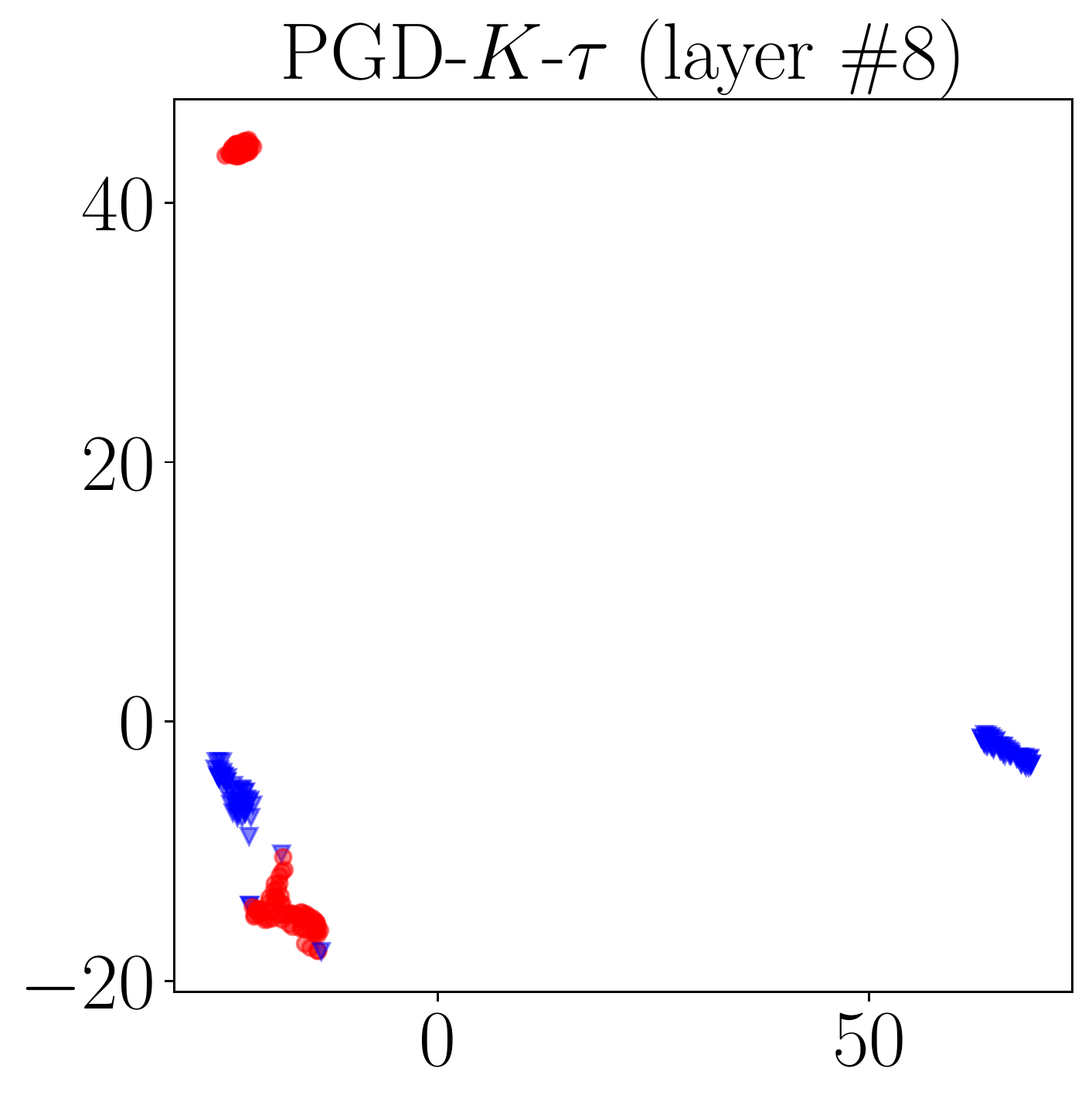}
		\end{minipage}
		\label{fig:appendix_smallcnn_overshoot_tsne}
	}
	\caption{Output distributions of Small CNN's intermediate layers. Left column: Intermediate layers' output distributions on natural data (not mixed). Middle column:  Intermediate layers' output distributions on adversarial data generated by PGD-20 (significantly mixed). Right column: Intermediate layers' output distributions on friendly adversarial data generated by PGD-20-0 (no significantly mixed).}
	\label{fig:appendix_overshoot_smallcnn}
\end{figure}

We conduct warm-up training using natural training data of two randomly selected classes (bird and deer) in CIFAR-10, then involve its adversarial variants generated by PGD-20 with step size $\alpha = 0.007$ and maximum perturbation $\epsilon = 0.031$. We show output distributions of layer $\#6$, $\#7$ and $\#8$ by PCA in Figure \ref{fig:appendix_smallcnn_overshoot_pca} and t-SNE in Figure \ref{fig:appendix_smallcnn_overshoot_tsne}.

\subsection{Output distributions of WRN-40-4's intermediate layers}
\begin{figure}[h!]
	\subfigure[Output distributions visualized by PCA]{
		\begin{minipage}[b]{0.5\textwidth}
		    \centering
		    \includegraphics[scale=0.18]{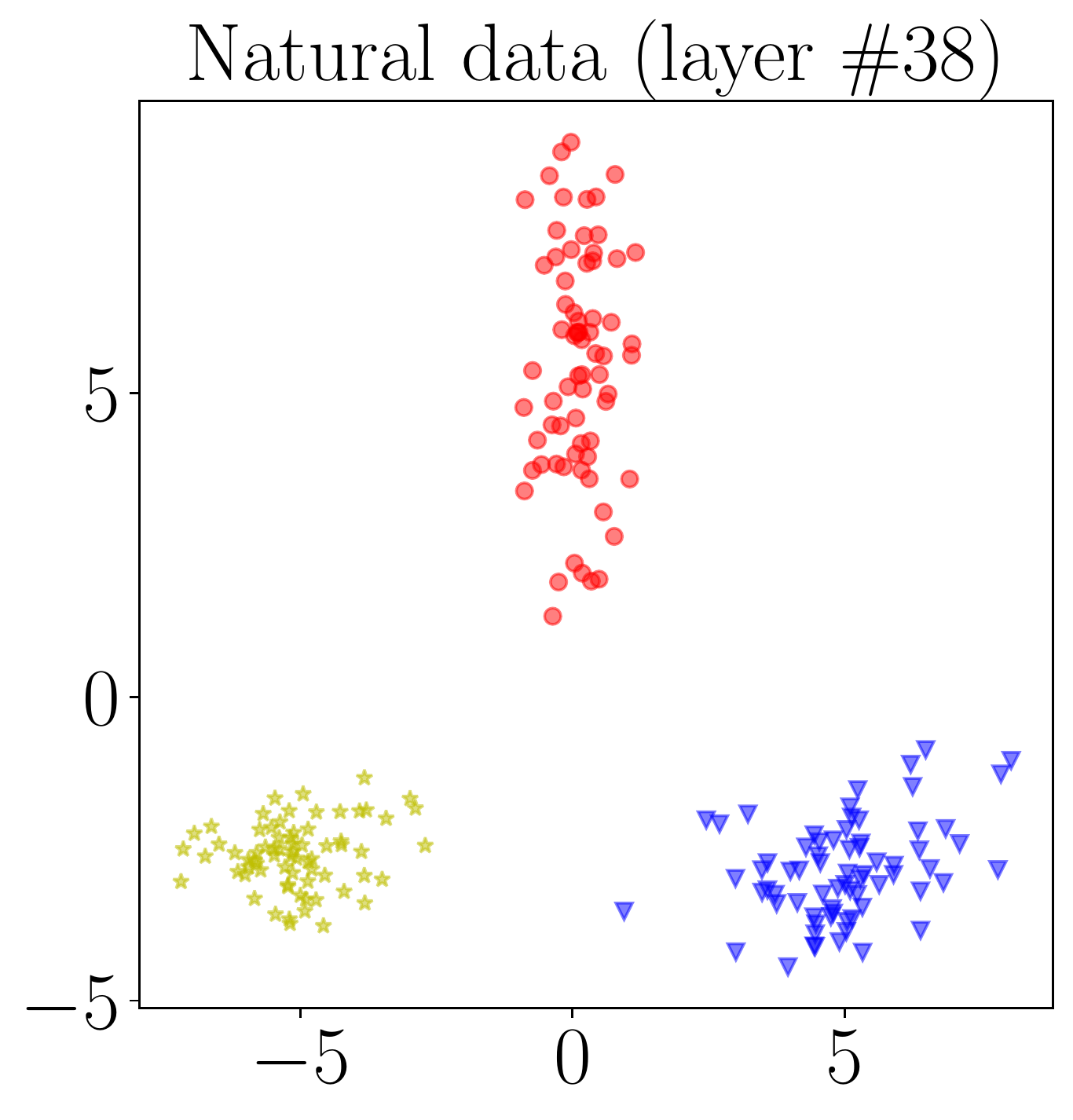} 
            \includegraphics[scale=0.18]{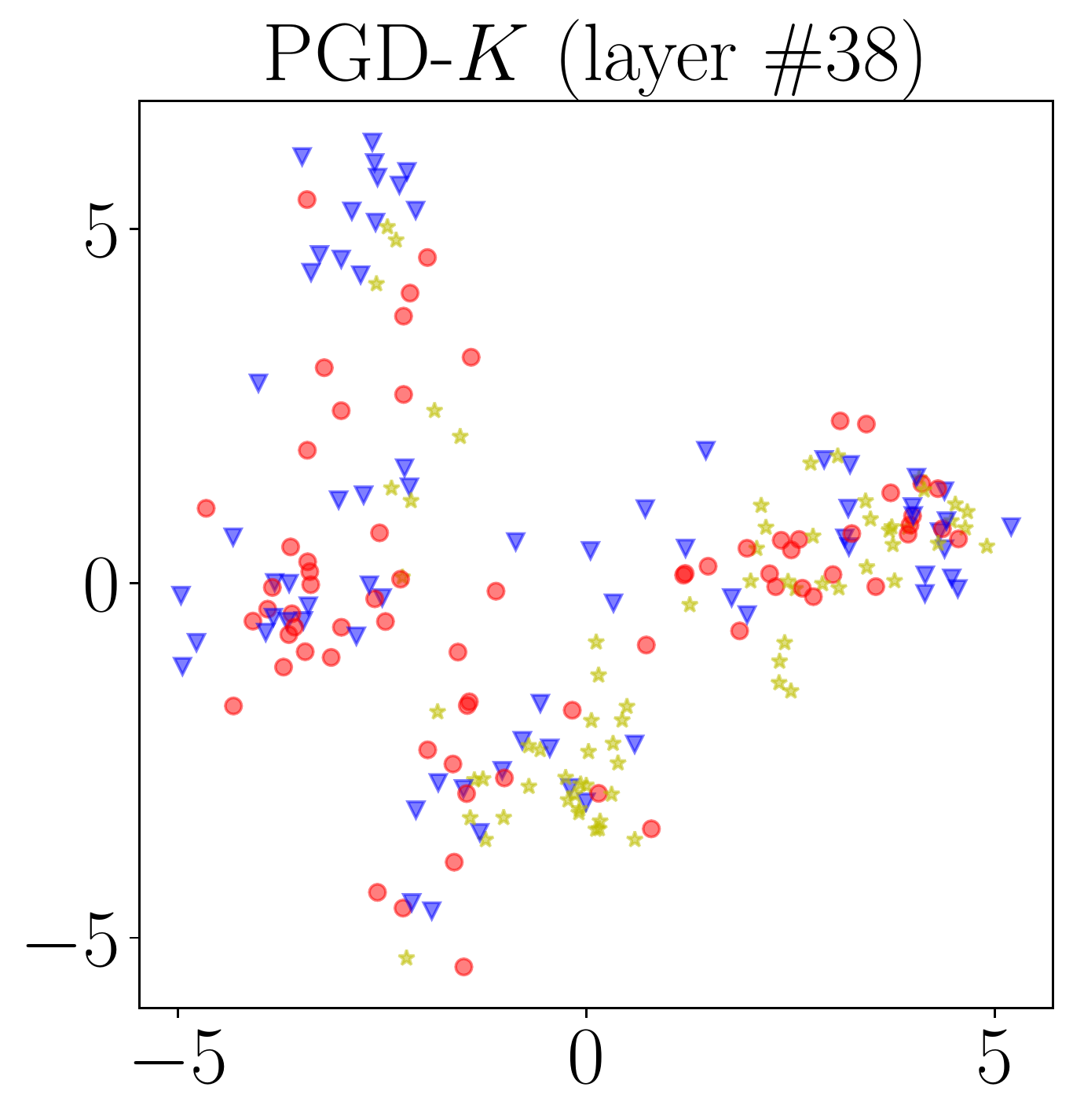}
            \includegraphics[scale=0.18]{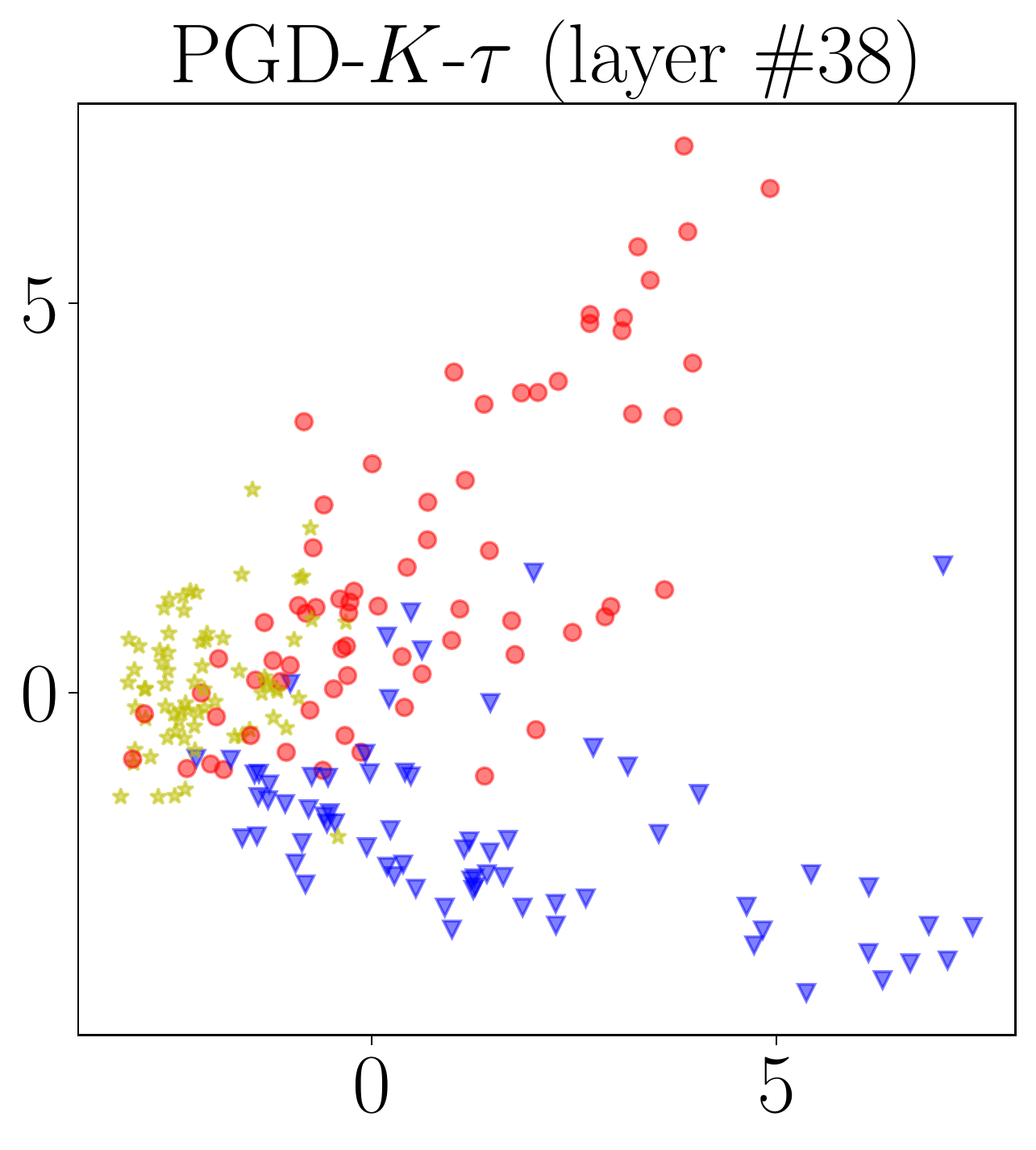}
            \quad
            \includegraphics[scale=0.18]{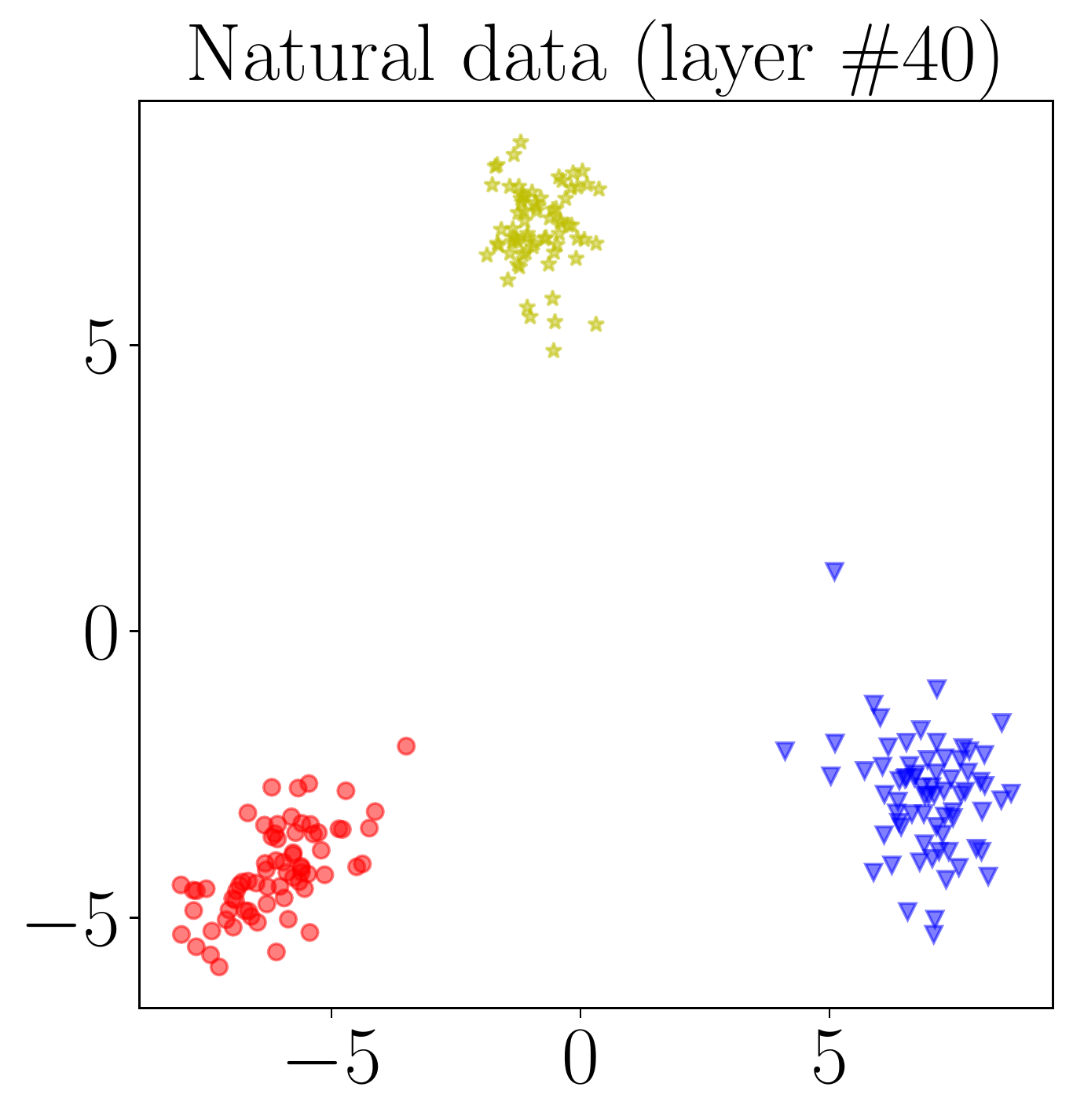} 
            \includegraphics[scale=0.18]{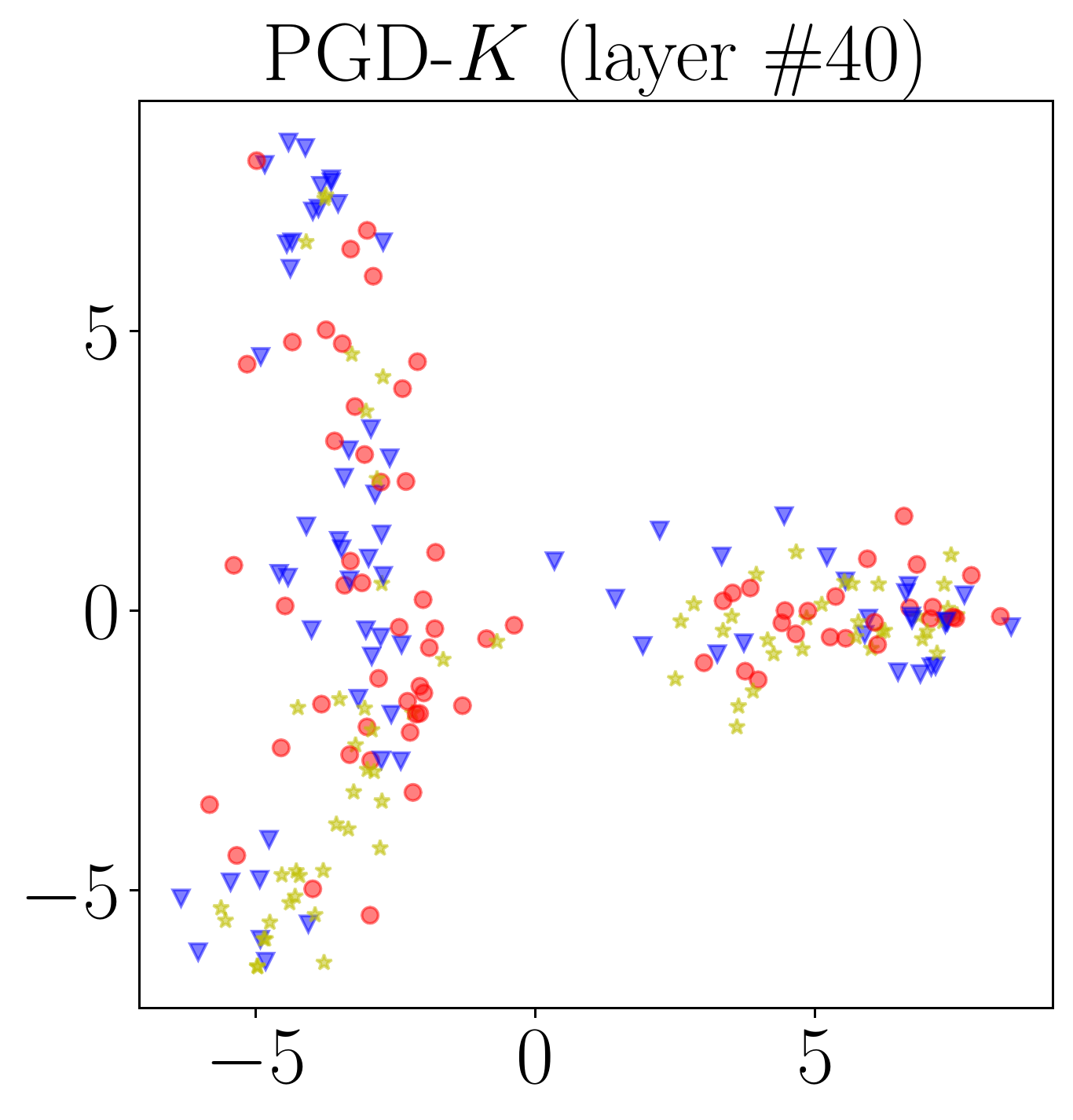}
            \includegraphics[scale=0.18]{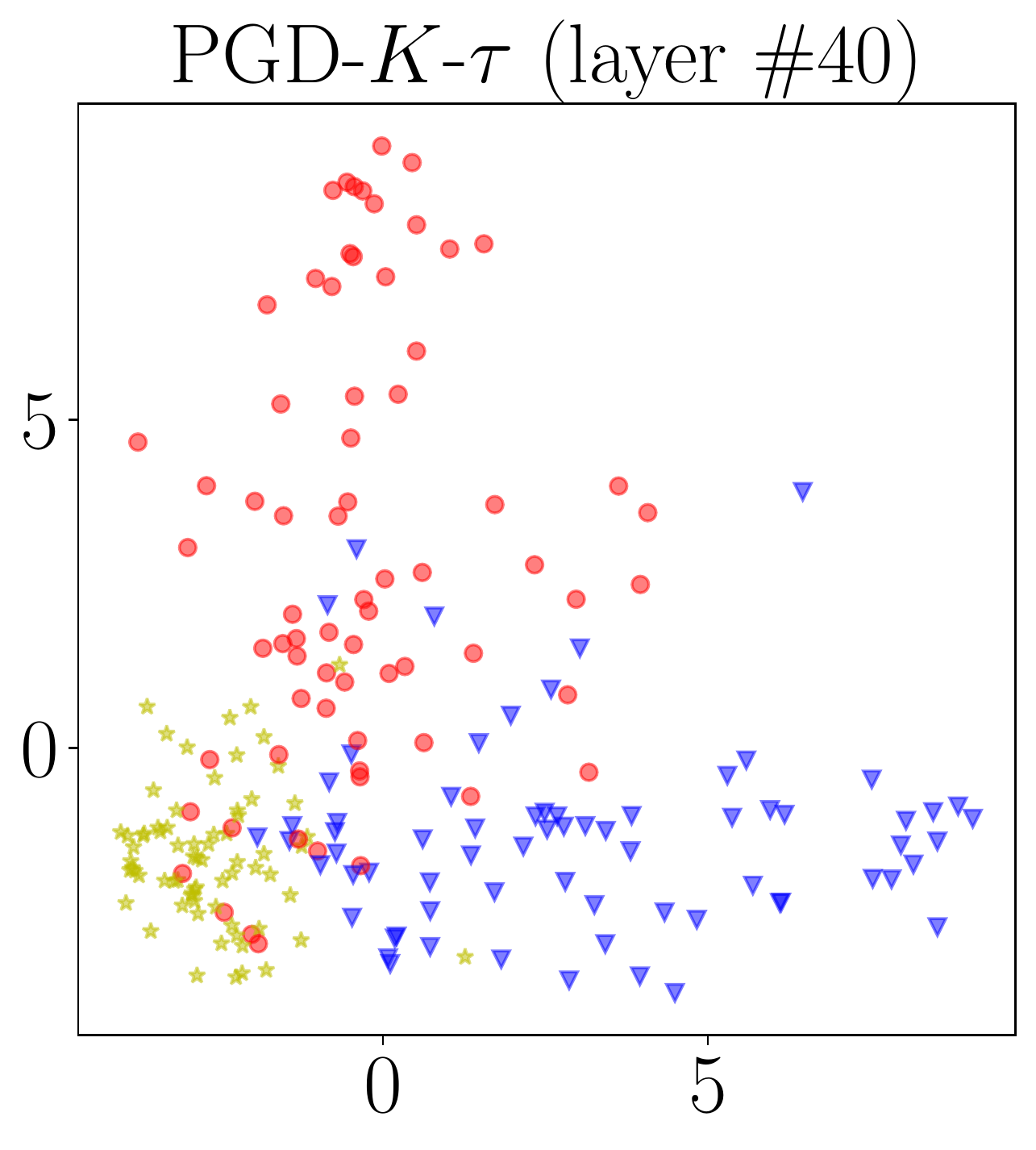}
            \quad
            \includegraphics[scale=0.18]{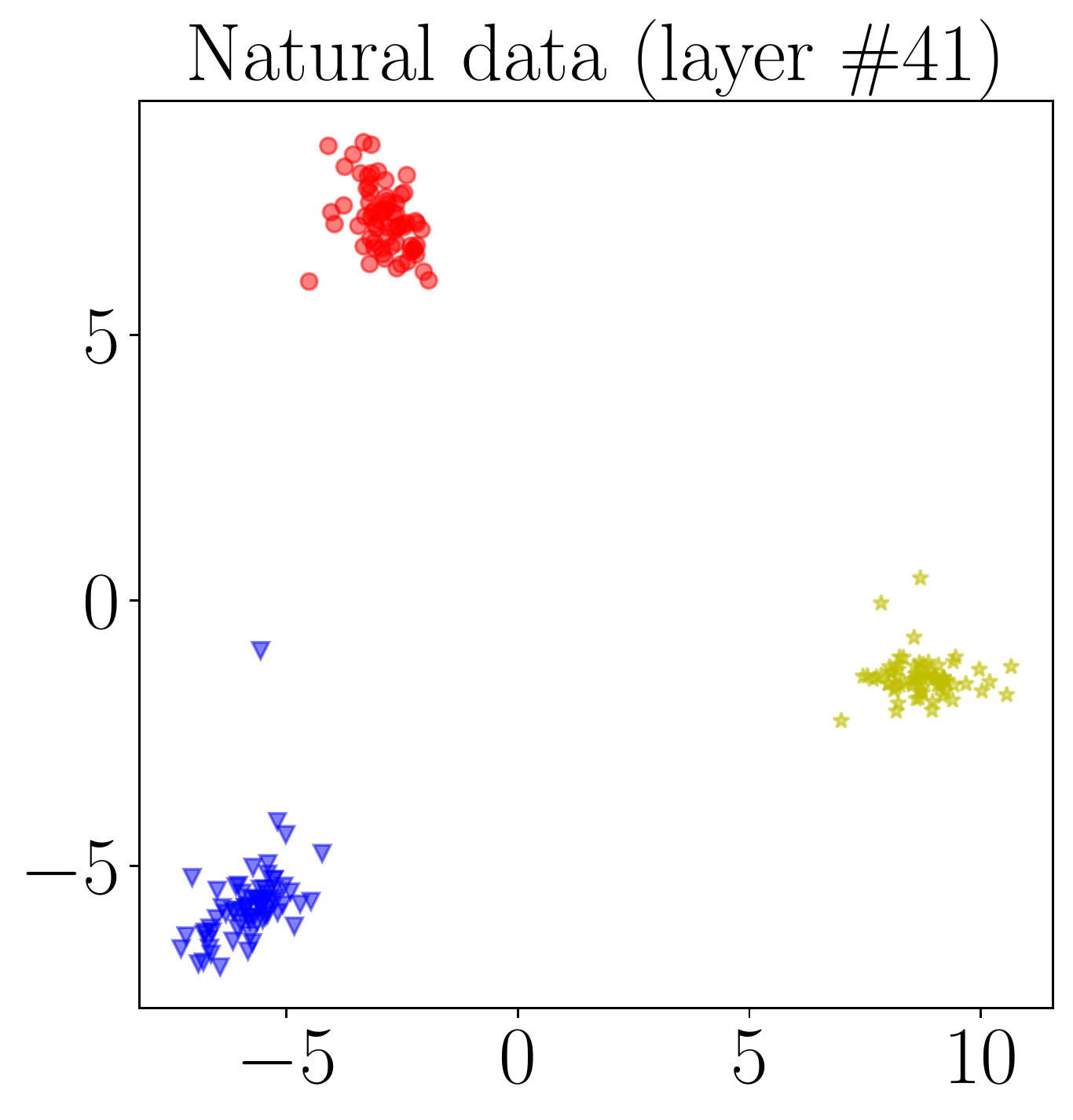} 
            \includegraphics[scale=0.18]{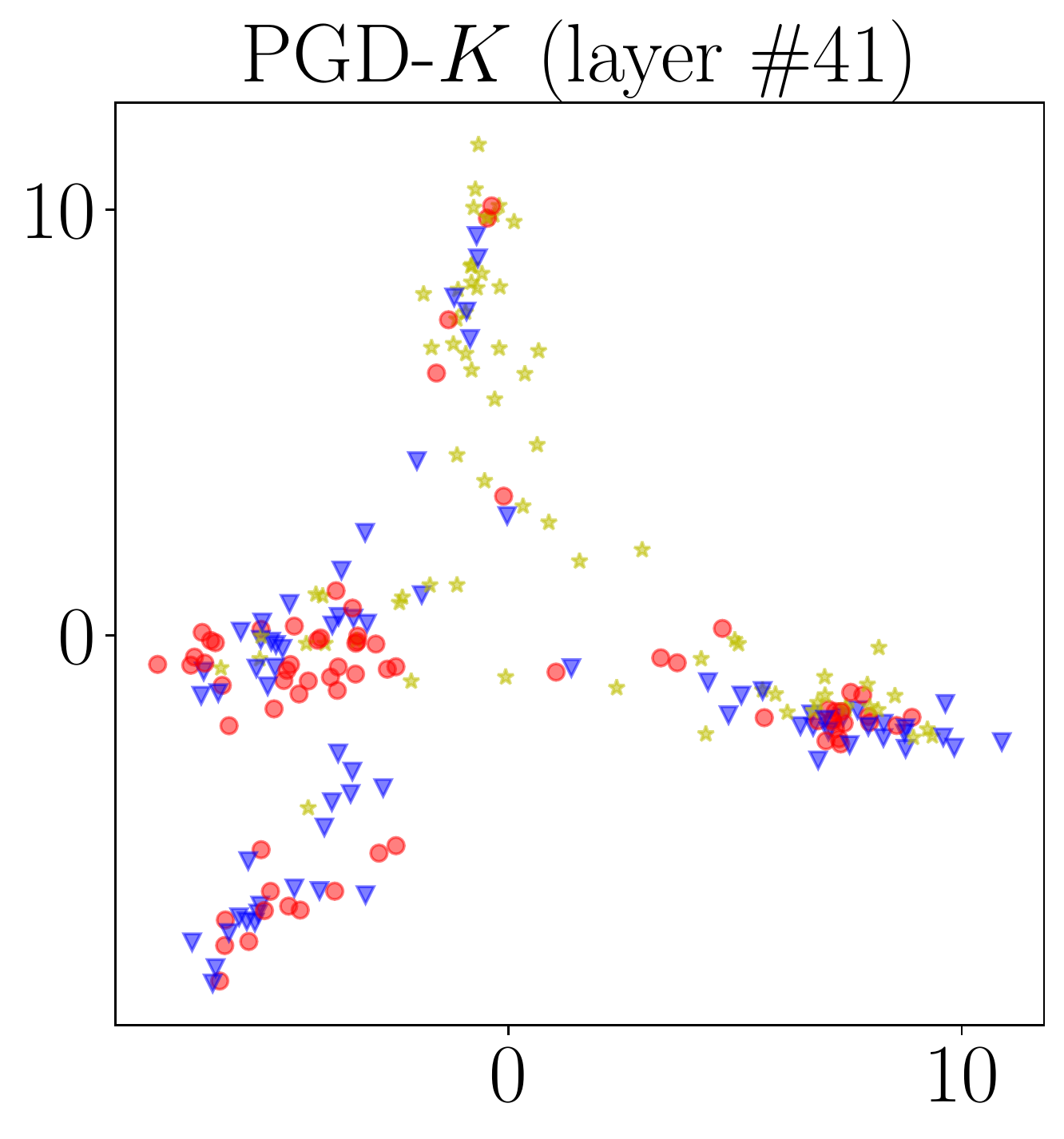}
            \includegraphics[scale=0.18]{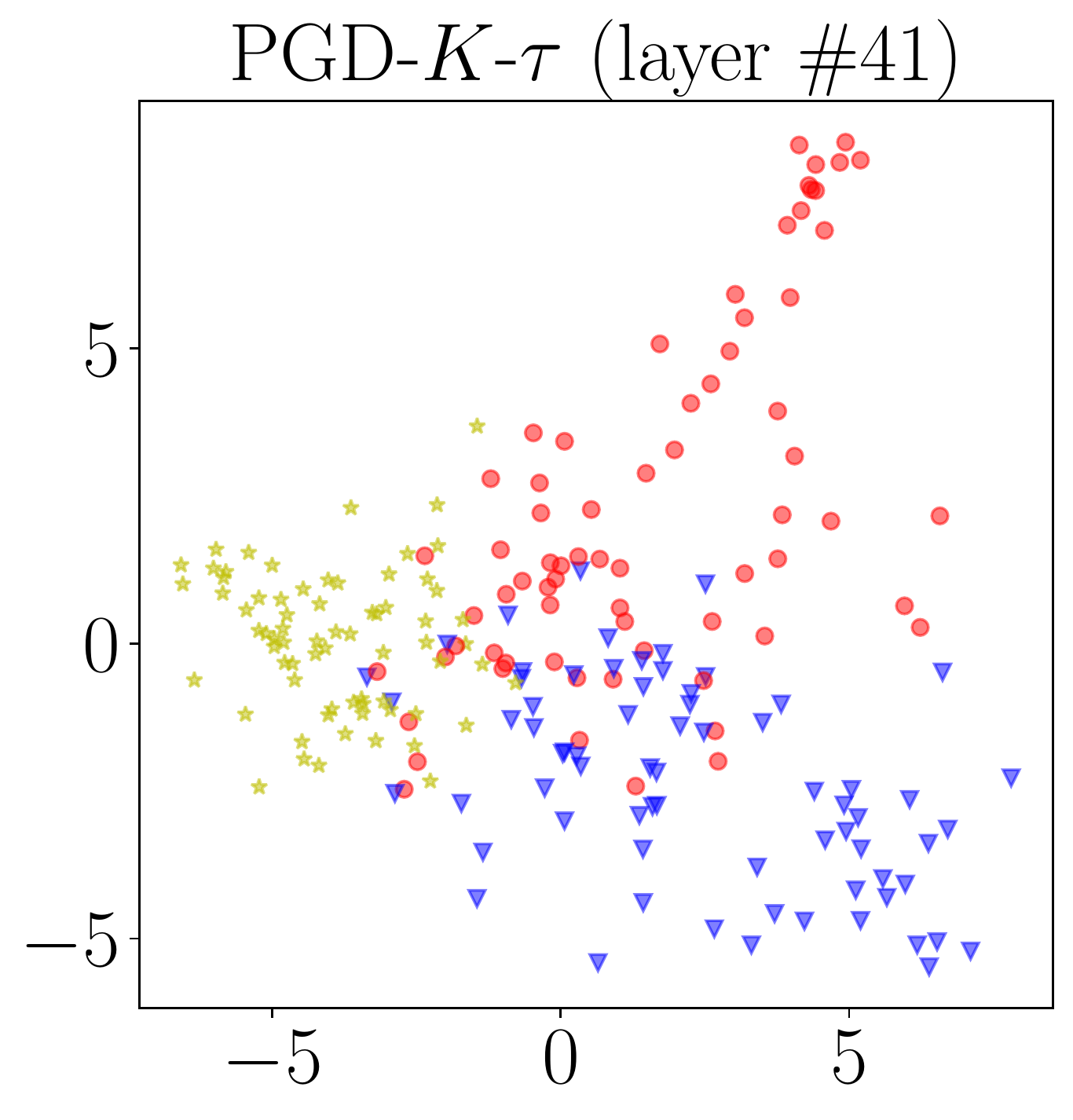}
		\end{minipage}
		\label{fig:appendix_overshoot_wrn_pca}
	}
	\subfigure[Output distributions visualized by t-SNE]{
		\begin{minipage}[b]{0.5\textwidth}
		    \centering
   	 	    \includegraphics[scale=0.18]{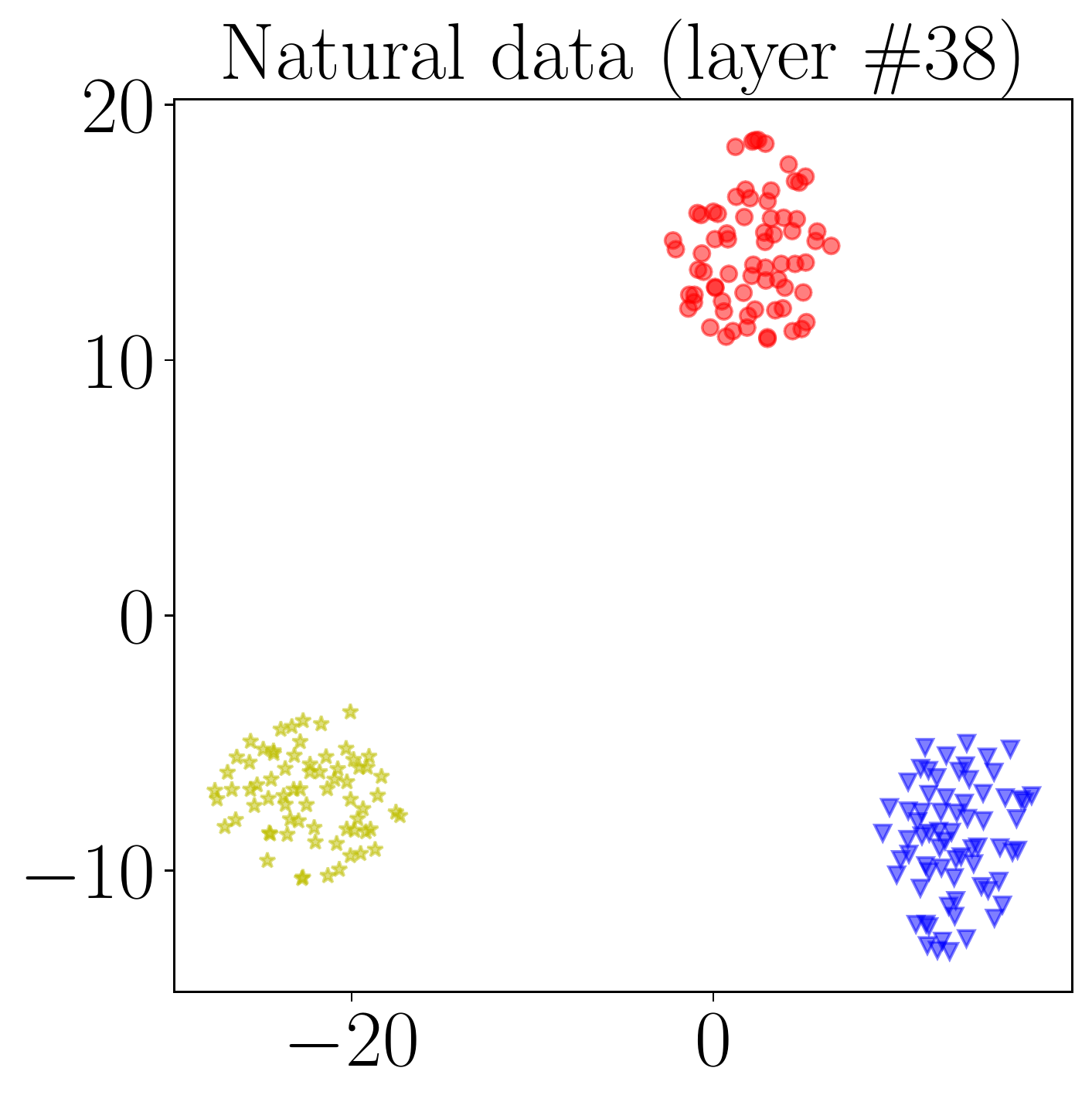} 
            \includegraphics[scale=0.18]{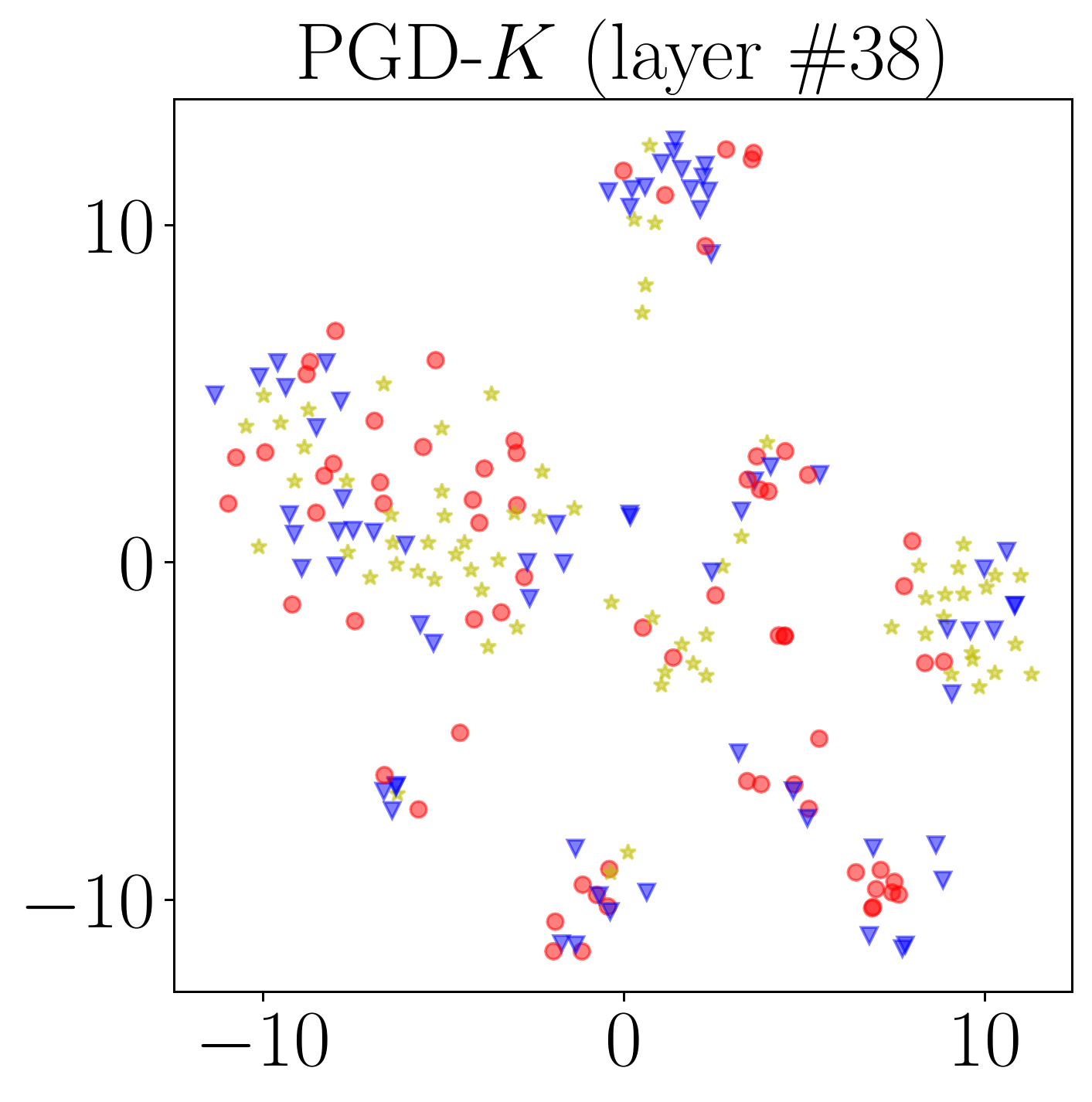}
            \includegraphics[scale=0.18]{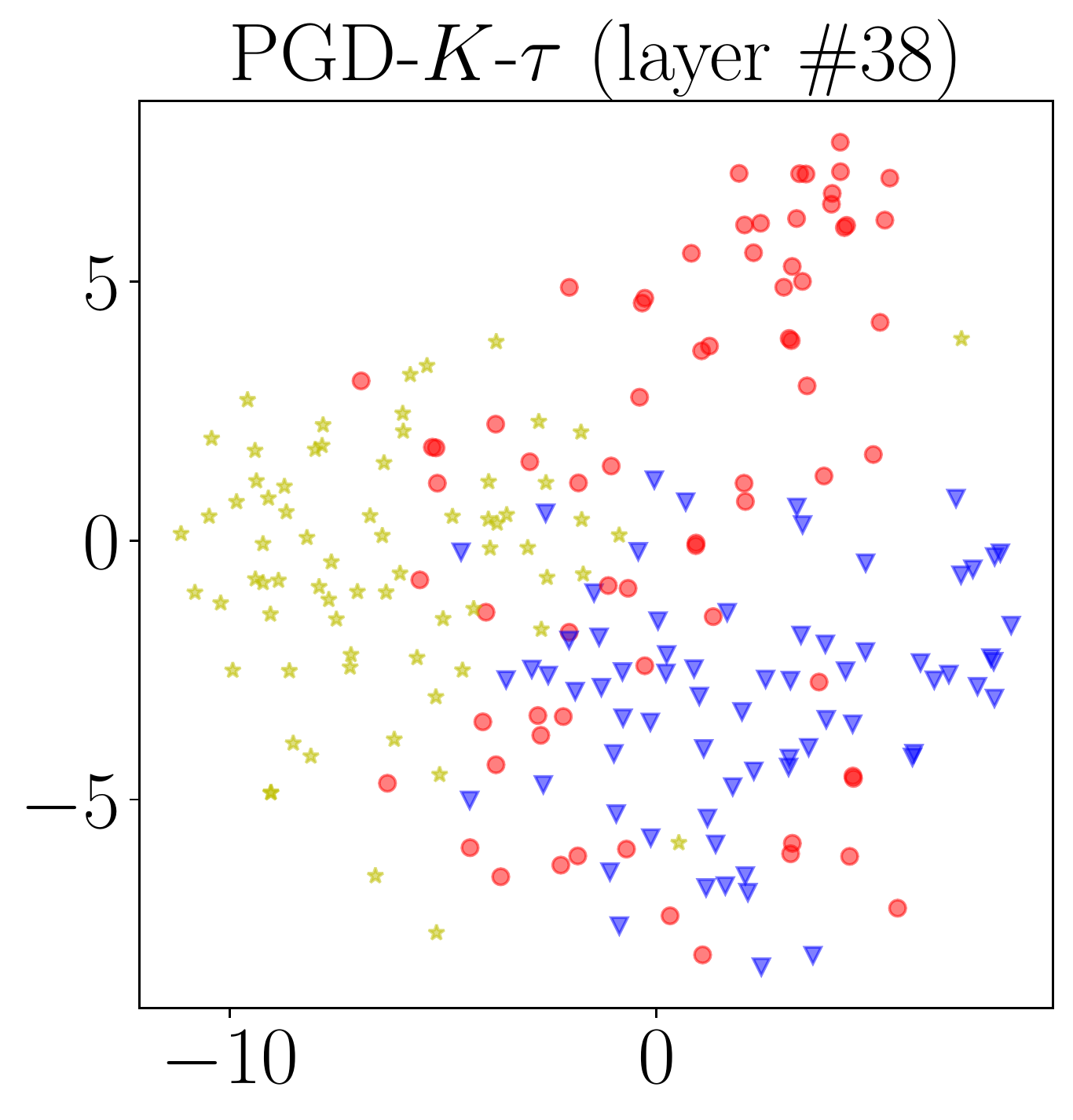}
            \quad
            \includegraphics[scale=0.18]{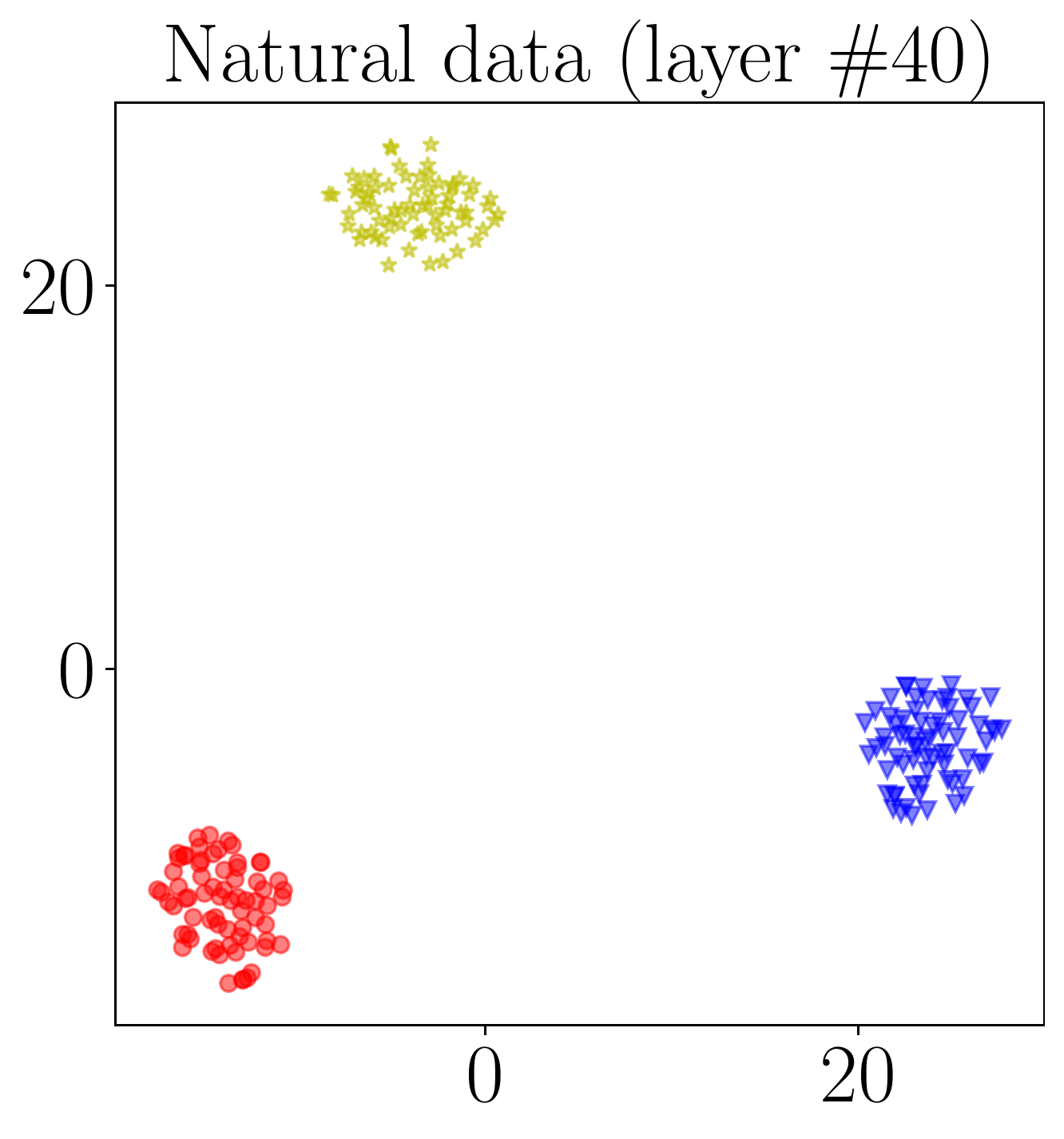} 
            \includegraphics[scale=0.18]{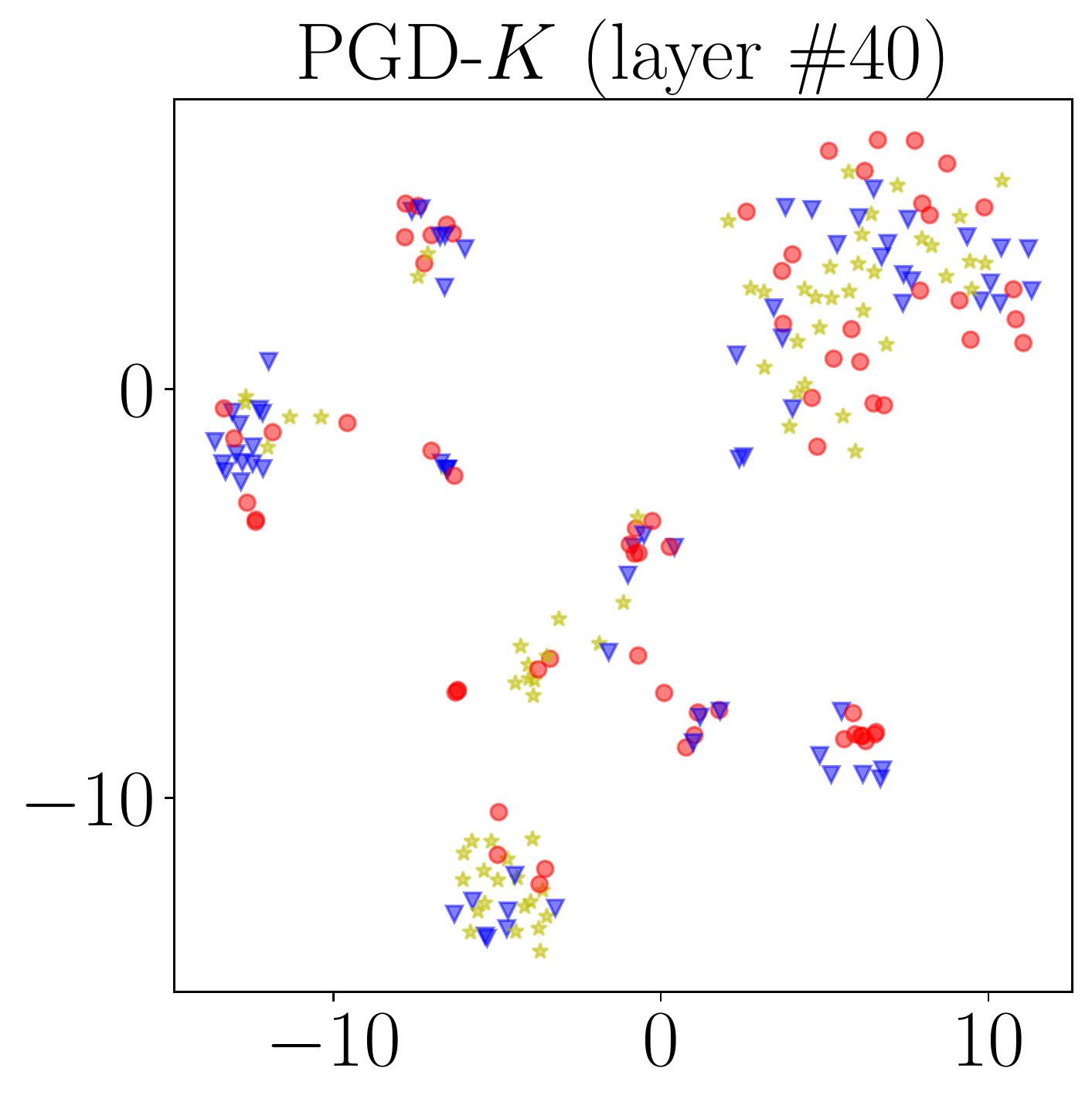}
            \includegraphics[scale=0.18]{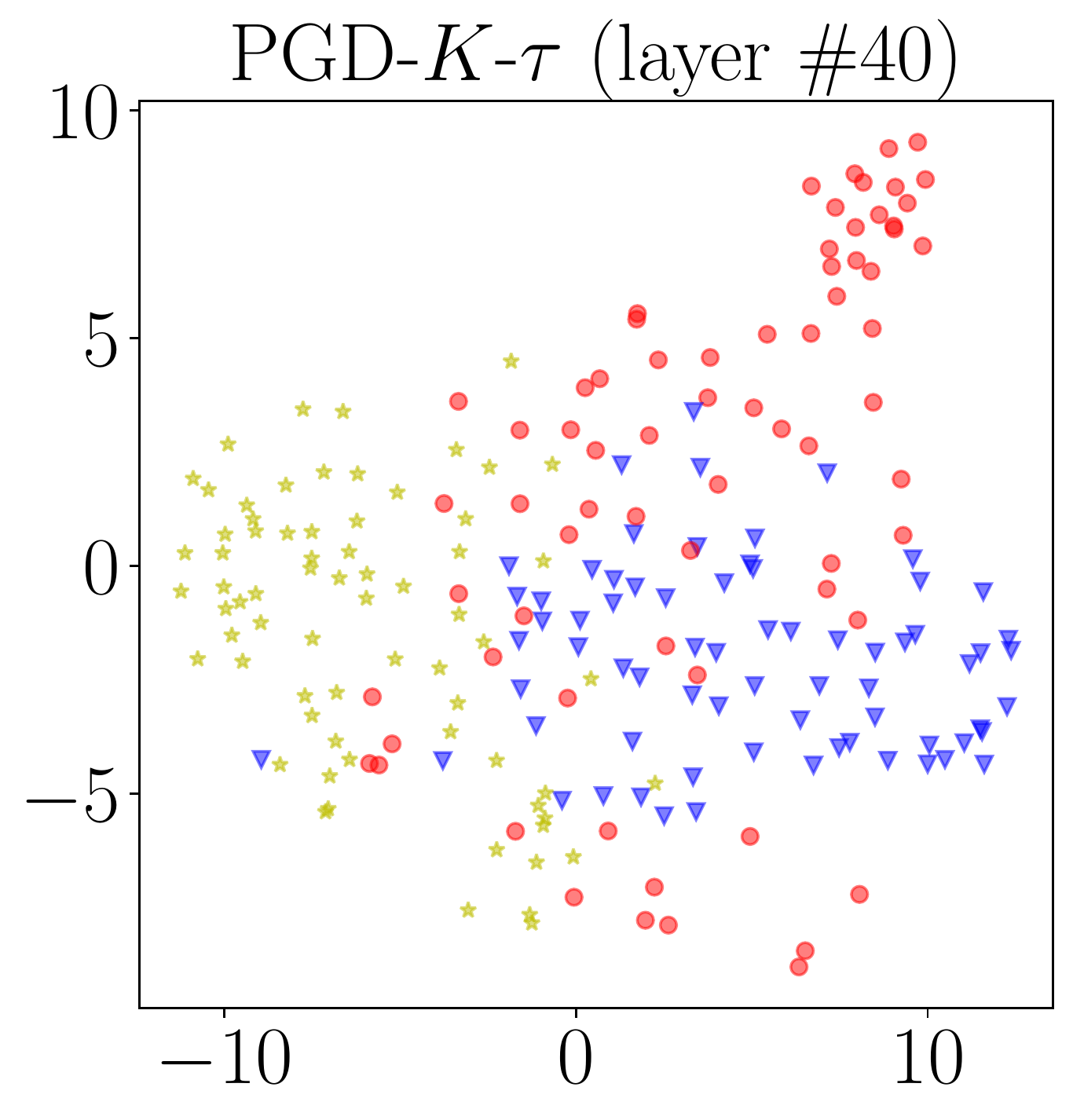}
            \quad
            \includegraphics[scale=0.18]{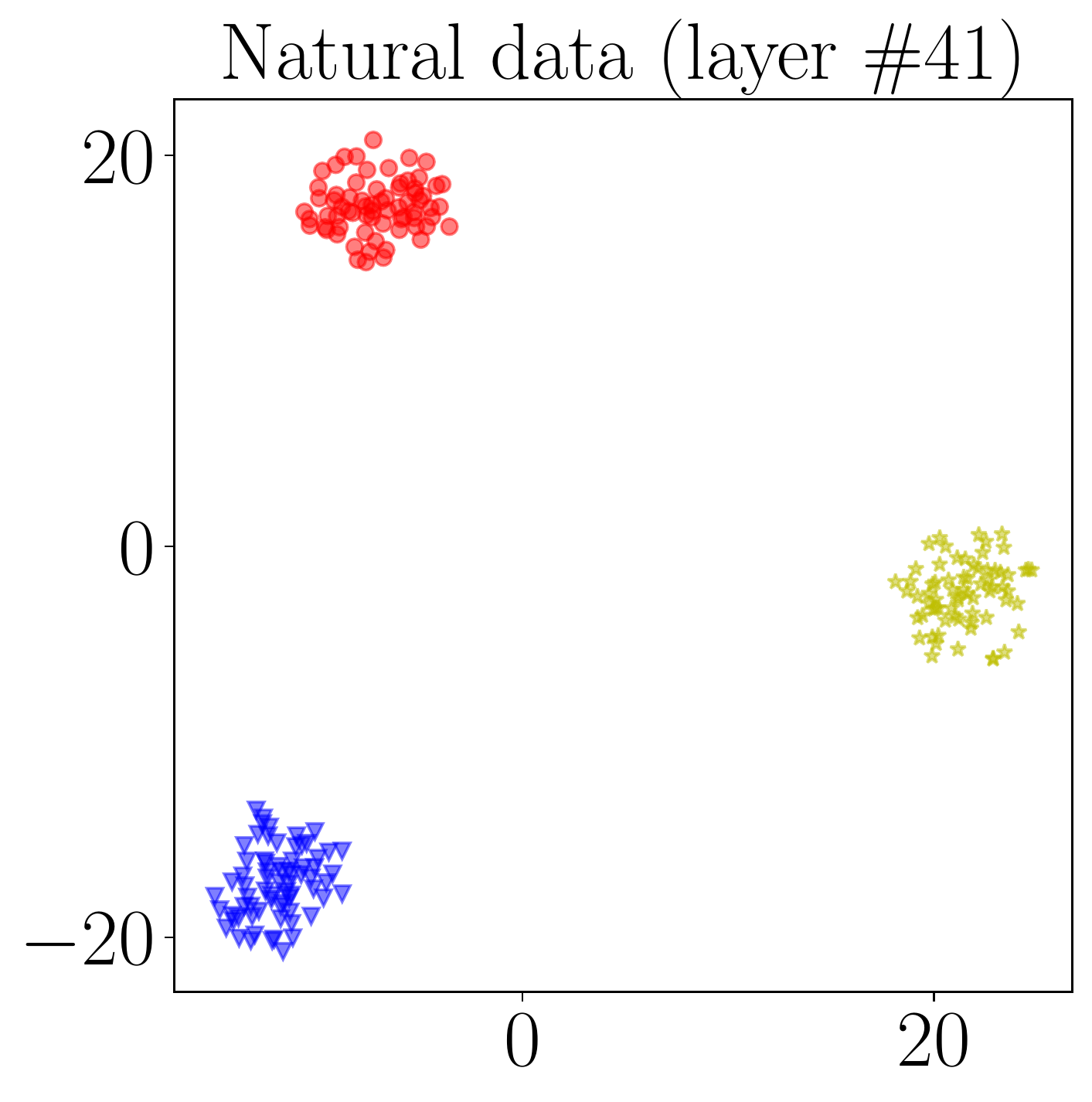} 
            \includegraphics[scale=0.18]{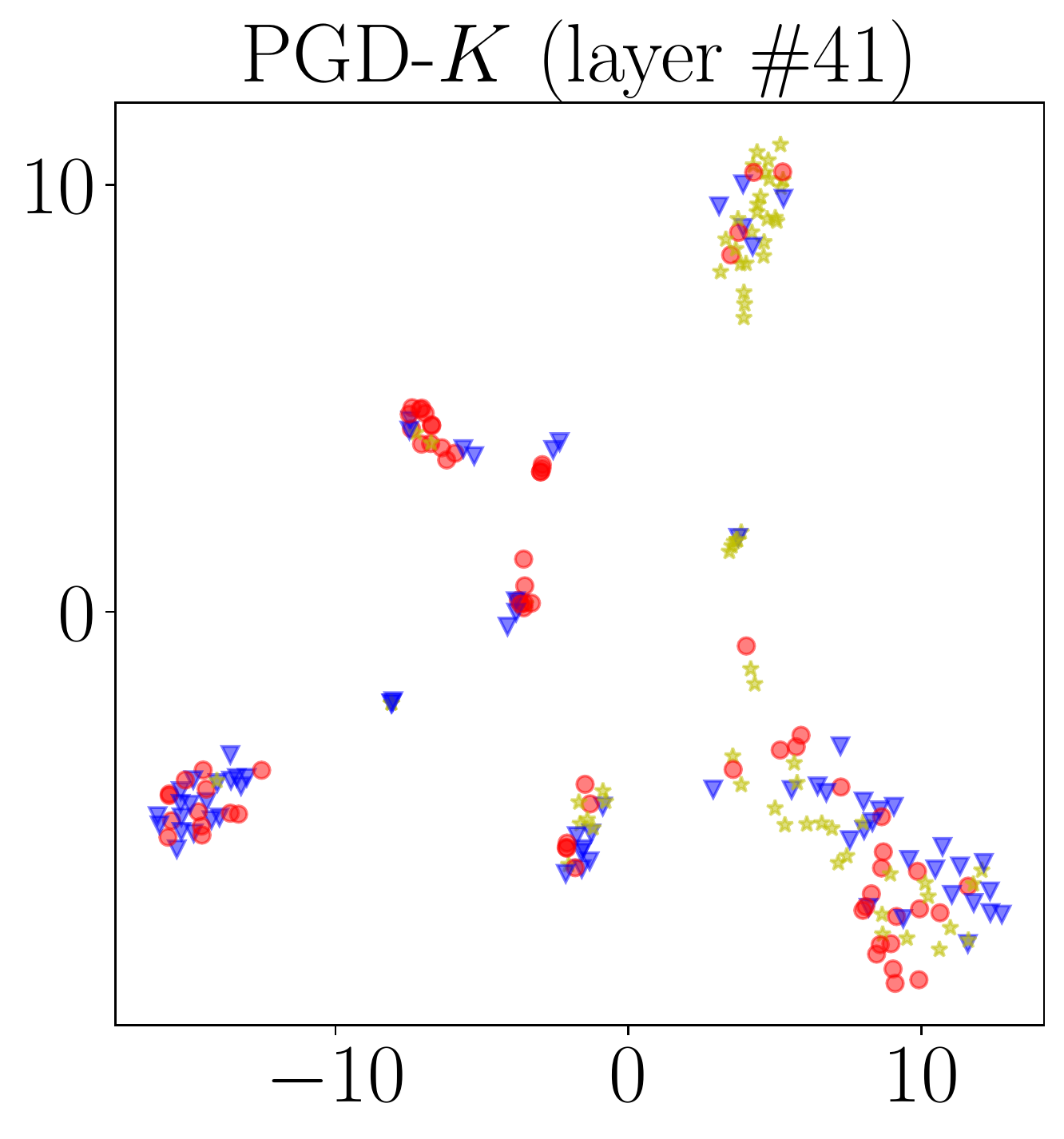}
            \includegraphics[scale=0.18]{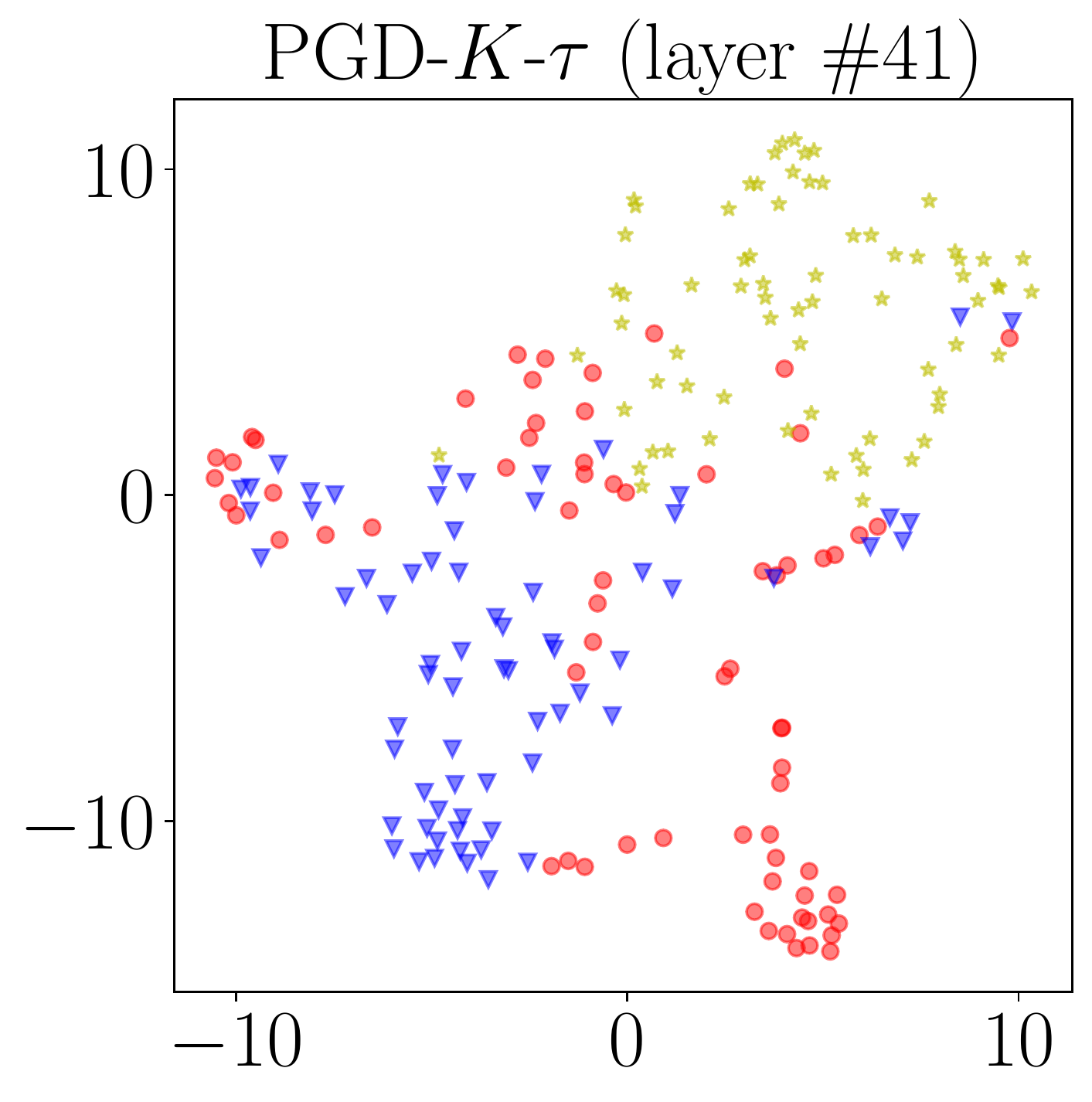}
		\end{minipage}
		\label{fig:appendix_overshoot_wrn_tsne}
	}
	\caption{Output distributions of WRN-40-4's intermediate layers. Left column: Intermediate layers' output distributions on natural data (not mixed). Middle column: Intermediate layers' output distributions on adversarial data generated by PGD-20 (significantly mixed). Right column: Intermediate layers' output distributions on friendly adversarial data generated by PGD-20-0 (no significantly mixed).}
	\label{fig:appendix_overshoot_WRN}
\end{figure}

We train a Wide ResNet (WRN-40-4, totally 41 layers) using natural data on 10 classes in CIFAR-10 and then include adversarial variants. We randomly select 3 classes (deer, horse and truck) for illustrating output distributions of WRN-40-4's intermediate layers. Adversarial data are generated by PGD-20 with step size $\alpha = 0.007$ and maximum perturbation $\epsilon = 0.031$ on WRN-40-4. We show output distributions by layer $\#38$, $\#40$ and $\#41$ by PCA in Figure \ref{fig:appendix_overshoot_wrn_pca} and t-SNE in Figure \ref{fig:appendix_overshoot_wrn_tsne}.

\clearpage

\section{FAT for TRADES}
\subsection{Learning Objective of TRADES}
\label{Appendix:Learning_obj_for_trades}
Besides the standard adversarial training, TRADES is another effective adversarial training method~\cite{Zhang_trades}, which trains on both natural data ${x}$ and adversarial data $\Tilde{{x}}$.

Similar to virtual adversarial training (VAT) adding a regularization term to the loss function~\cite{Miyato_VAT_ICLR}, which regularizes the output distribution by its local sensitivity of the output w.r.t. input, the objective of TRADES is
\begin{align}
\label{trades_adversarial_training_in_main_tex}
\min_{f\in\cF}\frac{1}{n}\sum_{i=1}^n 
   \bigg \{ \ell (  f({{x}}_i), y_i)+ \beta  \ell_{KL} (f(\Tilde{{x}}_i), f({x}_i))\bigg \},
\end{align}
where $\beta > 0$ is a regularization parameter, which controls the trade-off between standard accuracy and robustness accuracy, i.e., as $\beta$ increases, standard accuracy will decease while robustness accuracy will increase, and vice visa. Meanwhile, $\Tilde{{x}}_i$ in TRADES is dynamically generated by 
\begin{align}
\label{Eq:TRADES_inner_maximization}
\xadv_i = \argmax\nolimits_{\xadv\in\epsball[\bx_i]} \ell_{KL}(f(\xadv),f({x})),
\end{align}
and $\ell_{KL}$ is Kullback-Leibler loss that is calculated by
\begin{align*}
    \ell_{KL}(f(\Tilde{{x}}), f({x})) = \sum_{i=1}^{C} \ell_{\textrm{L}}^{i}( f({x}) ) \log \Bigg( \frac{\ell_{\textrm{L}}^{i} ( f({x}))} {\ell_{\textrm{L}}^{i} ( f(\Tilde{{x}}) )}\Bigg).
\end{align*}

\subsection{FAT for TRADES - Realization}
\label{Appendix:fat_for_trades}
\begin{algorithm}[h!]
   \caption{PGD-$K$-$\tau$ (Early Stopped PGD for TRADES)}
   \label{alg:PGD-k-t_for_trades}
\begin{algorithmic}
   \STATE {\bfseries Input:} data ${x}\in \cX$, label $y \in \cY$, model $f$, loss function $\ell_{KL}$, maximum PGD step $K$, step $\tau$, perturbation bound $\epsilon$, step size $\alpha$
   \STATE {\bfseries Output:} $\Tilde{{x}}$  
   \STATE $\Tilde{{x}} \gets {x} + \xi \mathcal{N}(\mathbf{0}, \mathbf{I}) $
    \WHILE{$K > 0 $}
    \IF{$\arg\max_{i} f (\Tilde{ {x} }) \neq y$ and $\tau = 0$}
    \BREAK
    \ELSIF{$\arg\max_{i} f (\Tilde{ {x} }) \neq y$}
     \STATE $\tau \gets \tau - 1$
   \ENDIF
   \STATE $\Tilde{{x}} \gets \Pi_{\mathcal{B}[{x},\epsilon]}\big( \alpha\sign(\nabla_{\Tilde{{x}}} \ell_{KL}( f(\Tilde{{x}}), f({x}))  +  \Tilde{{x}} \big) $ 
   \STATE $K \gets K-1$
    \ENDWHILE
\end{algorithmic}

\end{algorithm}
In Algorithm~\ref{alg:PGD-k-t_for_trades}, $\mathcal{N}(\mathbf{0}, \mathbf{I})$ generates a random unit vector of $d$ dimension. $\xi$ is a small constant. $\ell_{KL}$ is Kullback-Leibler loss.

Given a dataset $S = \{ ({x}_i, y_i)\}^n_{i=1}$, where ${x}_i \in \mathcal{R}^d$ and $y_i \in \{0, 1, ..., C-1\}$, adversarial training (TRADES) with early stopped PGD-$K$-$\tau$ returns a classifier $\theta^{*}$:
\begin{equation}
\label{trades_adversarial_training}
    \theta^{*} = \underset{\theta}{\arg \min}  \sum_{i = 1} ^{n}
    \Bigg \{ \ell_{CE} (  f_{\theta}({{x}}_i), y_i)+ \beta  \ell_{KL} (f_{\theta}(\Tilde{{x}}_i), f_{\theta}({x}_i)) \bigg \}
\end{equation}
where  $f_{\theta}: \mathcal{R}^d \rightarrow \mathcal{R}^C$ is DNN classification function, $f_{\theta}({\cdot})$ outputs predicted probability over $C$ classes,
The adversarial data $\Tilde{{x}}_i$ of ${x}_i$ is dynamically generated according to Algorithm~\ref{alg:PGD-k-t_for_trades}, 
$\beta > 0$ is a regularization parameter,
$\ell_{CE}$ is cross-entropy loss, 
$\ell_{KL}$ is Kullback-Leibler loss.

Based on our early stopped PGD-$K$-$\tau$ for TRADES in Algorithm~\ref{alg:PGD-k-t_for_trades}, our friendly adversarial training for TRADES (FAT for TRADES) is
\begin{algorithm}[h!]
   \caption{Friendly Adversarial Training for TRADES (FAT for TRADES)}
   \label{alg:FAT_for_TRADES}
\begin{algorithmic}
   \STATE {\bfseries Input:} network $f_{\mathbf{\theta}}$, training dataset $S = \{(\bx_i, y_i) \}^{n}_{i=1}$, learning rate $\eta$, number of epochs $T$, batch size $m$, number of batches $M$
   \STATE {\bfseries Output:} adversarially robust network $f_{\mathbf{\theta}}$  
  \FOR{epoch $= 1$, $\dots$, $T$}
    \FOR {mini-batch $=1$, $\dots$, $M$ }
    \STATE Sample a mini-batch $\{(\bx_i, y_i) \}^{m}_{i=1}$ from $S$
        \FOR{$i = 1$, $\dots$, $m$ (in parallel) }
         \STATE Obtain adversarial data $\bxtidle_i$of $\bx_i$ by Algorithm~\ref{alg:PGD-k-t_for_trades}
         \ENDFOR
    \STATE $\mathbf{\theta} \gets \mathbf{\theta} - \eta \frac{1}{m} \sum^{m}_{i - 1} \nabla_{\mathbf{\theta}} \big [ \ell_{CE}(f_{\mathbf{\theta}}(\bxtidle_i), y_i) + \beta  \ell_{KL} (f_{\theta}(\Tilde{{x}}_i), f_{\theta}({x}_i)) \big]   $
  \ENDFOR
 \ENDFOR
\end{algorithmic}
\end{algorithm}

\section{FAT for MART}
\label{Appendix:fat_for_mart}
\subsection{Learning Objective of MART}
\label{Appendix:Learning_obj_for_mart}
MART~\cite{wang2020improving_MART} emphasizes the importance of misclassified natural data on the adversarial robustness. \citet{wang2020improving_MART} propose a regularized adversarial learning objective which contains an explicit differentiation of misclassified data as the regularizer. The learning objective of MART is
\begin{align}
\label{mart_adversarial_training_in_main_tex}
\min_{f\in\cF}\frac{1}{n}\sum_{i=1}^n 
   \bigg \{ \ell_{BCE} (  f(\tilde{{x}}_i), y_i)+ \beta \cdot \ell_{KL} (f(\Tilde{{x}}_i), f({x}_i)) \cdot (1 - \ell_L^{y_i}(f(x_i))) \bigg \},
\end{align}
where $\beta > 0$ is a regularization parameter which balances the two parts of the final loss. $\ell_{KL}$ is Kullback-Leibler loss. $\ell_L^k$ stands for the k-th element of the soft-max output and $\Tilde{{x}}_i$ in MART is dynamically generated according to Eq.~\ref{Eq:madry_inner_maximization} that is realized by PGD-K.
The first part $\ell_{BCE}$ is the proposed BCE loss in MART that is calculated by
\begin{align*}
    \ell_{BCE}(f(\Tilde{{x}}_i), y_i) = - \log (\ell_L^{y_i}(f(\tilde{{x}}_i))) 
    - \log (1 - \max_{k \neq y_i} \ell_L^k(f(\tilde{{x}}_i)))
\end{align*}
where the first term $- \log (\ell_L^{y_i}(f(\tilde{{x}}_i)))$ is the cross-entropy loss and the second term $- \log (1 - \max_{k \neq y_i} \ell_L^k(f(\tilde{{x}}_i)))$ is a margin term used to increase the distance between $\ell_L^{y_i}(f(\tilde{{x}}_i))$ and $ \max_{k \neq y_i} \ell_L^k(f(\tilde{{x}}_i))$. This is a similar to C$\&$W~\cite{Carlini017_CW} attack that is to improve attack strength. 
For the second part, they combine Kullback-Leibler loss~\cite{Zhang_trades} and emphases on misclassified examples. This part of loss will be large for misclassified examples and small for correctly classified examples.
\subsection{FAT for MART - Realization}
\label{Appendix:fat_for_mart_realization}
Given a dataset $S = \{ ({x}_i, y_i)\}^n_{i=1}$, where ${x}_i \in \mathcal{R}^d$ and $y_i \in \{0, 1, ..., C-1\}$, FAT for MART returns a classifier $\theta^{*}$:
\begin{equation}
\label{mart_adversarial_training}
    \theta^{*} = \underset{\theta}{\arg \min}  \sum_{i = 1} ^{n}
    \Bigg \{ \ell_{BCE} (f_{\theta}(\tilde{{x}}_i), y_i)+ \beta \cdot \ell_{KL} (f_{\theta}(\Tilde{{x}}_i), f_{\theta}({x}_i)) \cdot (1 - \ell_L^{y_i}(f_{\theta}({x}_i))) \bigg \}
\end{equation}
where  $f_{\theta}: \mathcal{R}^d \rightarrow \mathcal{R}^C$ is DNN classification function, $f_{\theta}(\mathbf{\cdot})$ outputs predicted probability over $C$ classes.
The adversarial data $\Tilde{{x}}_i$ of ${x}_i$ is dynamically generated according to Algorithm~\ref{alg:PGD-k-t}, 
$\ell_L$ is the soft-max activation,
$\beta > 0$ is a regularization parameter,
$\ell_{BCE}$ is the proposed BCE loss in MART and $\ell_{KL}$ is Kullback-Leibler loss.
Based on our early stopped PGD-$K$-$\tau$ in Algorithm~\ref{alg:PGD-k-t}, FAT for MART Algorithm~\ref{alg:FAT_for_MART}.
\begin{algorithm}[h!]
   \caption{Friendly Adversarial Training for MART (FAT for MART)}
   \label{alg:FAT_for_MART}
\begin{algorithmic}
   \STATE {\bfseries Input:} network $f_{\mathbf{\theta}}$, training dataset $S = \{(\bx_i, y_i) \}^{n}_{i=1}$, learning rate $\eta$, number of epochs $T$, batch size $m$, number of batches $M$
   \STATE {\bfseries Output:} adversarially robust network $f_{\mathbf{\theta}}$  
  \FOR{epoch $= 1$, $\dots$, $T$}
    \FOR {mini-batch $=1$, $\dots$, $M$ }
    \STATE Sample a mini-batch $\{(\bx_i, y_i) \}^{m}_{i=1}$ from $S$
        \FOR{$i = 1$, $\dots$, $m$ (in parallel) }
         \STATE Obtain adversarial data $\bxtidle_i$of $\bx_i$ by Algorithm~\ref{alg:PGD-k-t}
         \ENDFOR
    \STATE $\mathbf{\theta} \gets \mathbf{\theta} - \eta \frac{1}{m} \sum^{m}_{i - 1} \nabla_{\mathbf{\theta}} \big [ \ell_{BCE} (f_{\theta}(\tilde{{x}}_i), y_i)+ \beta \cdot \ell_{KL} (f_{\theta}(\Tilde{{x}}_i), f_{\theta}({x}_i)) \cdot (1 - \ell_L^{y_i}(f_{\theta}(x_i))) \big] $
  \ENDFOR
 \ENDFOR
\end{algorithmic}
\end{algorithm}

\section{Experimental Setup}
\subsection{Selection of Step $\tau$}
\label{APPENDIX:selection_of_slippery_step}
Figure~\ref{fig:tau_effect} presents empirical results on CIFAR-10 via our FAT algorithm,where we train 8-layer convolutional neural network (Small CNN, blue line) and 18-layer residual neural network (ResNet-18, red line)~\cite{he2016deep}. The maximum step $K = 10$, $\epsilon_{train} = 8/255$, step size $\alpha = 0.007$, and step $\tau \in \{0,1, \dots, 10\}$. We train deep networks for 80 epochs using SGD with 0.9 momentum, where learning rate starts at 0.1 and divided by 10 at 60 epoch.

For each $\tau$, we take five trials, where each trial will obtain standard test accuracy evaluated on natural test data and robust test accuracy evaluated on adversarial test data that are generated by attacks FGSM~\cite{Goodfellow14_Adversarial_examples}, PGD-10 and PGD-20, PGD-100~\cite{Madry_adversarial_training} and C$\&$W attack~\cite{Carlini017_CW} respectively. 
All those attacks are white box attacks, which are constrained by the same perturbation bound $\epsilon_{test} = 8/255$. Following \citet{Zhang_trades}, all attacks have the random start, and the step size $\alpha$ in PGD-10, PGD-20, PGD-100 and C$\&$W is fixed to $0.003$.

\section{Supplementary Experiments - FAT Enabling Larger $\epsilon_{train}$ }
\label{APPENDIX:exp_larger_eps_ball}
In this section, we provide extensive experimental results. 
The test settings are the same as those are stated in Section~\ref{section:fat_enable_lager_epsilon}.
In Section~\ref{appendix:exp_smallcnn}, instead of using ResNet-18, we conduct adversarial training on the deep model of Small CNN. 
In Section~\ref{appendix:exp_fat_for_trades}, instead of applying FAT, we compare our FAT for TRADES and TRADE~\cite{Zhang_trades} under different values of perturbation bound $\epsilon_{train}$ on the deep models ResNet-18 and Small CNN.
In Section~\ref{appendix:exp_pgd20}, we set maximum PGD steps $K = 20$ and report results of FAT and FAT for TRADES over existing methods with larger perturbation bound $\epsilon_{train}$.
To sum up, all those extensive results verify that FAT and FAT for TRADES can enable deep models trained under larger values of perturbation bound $\epsilon_{train}$.

\subsection{A Different Deep Model - Small CNN}
\label{appendix:exp_smallcnn}
\begin{figure}[!htb]
    \centering
    \includegraphics[scale=0.33]{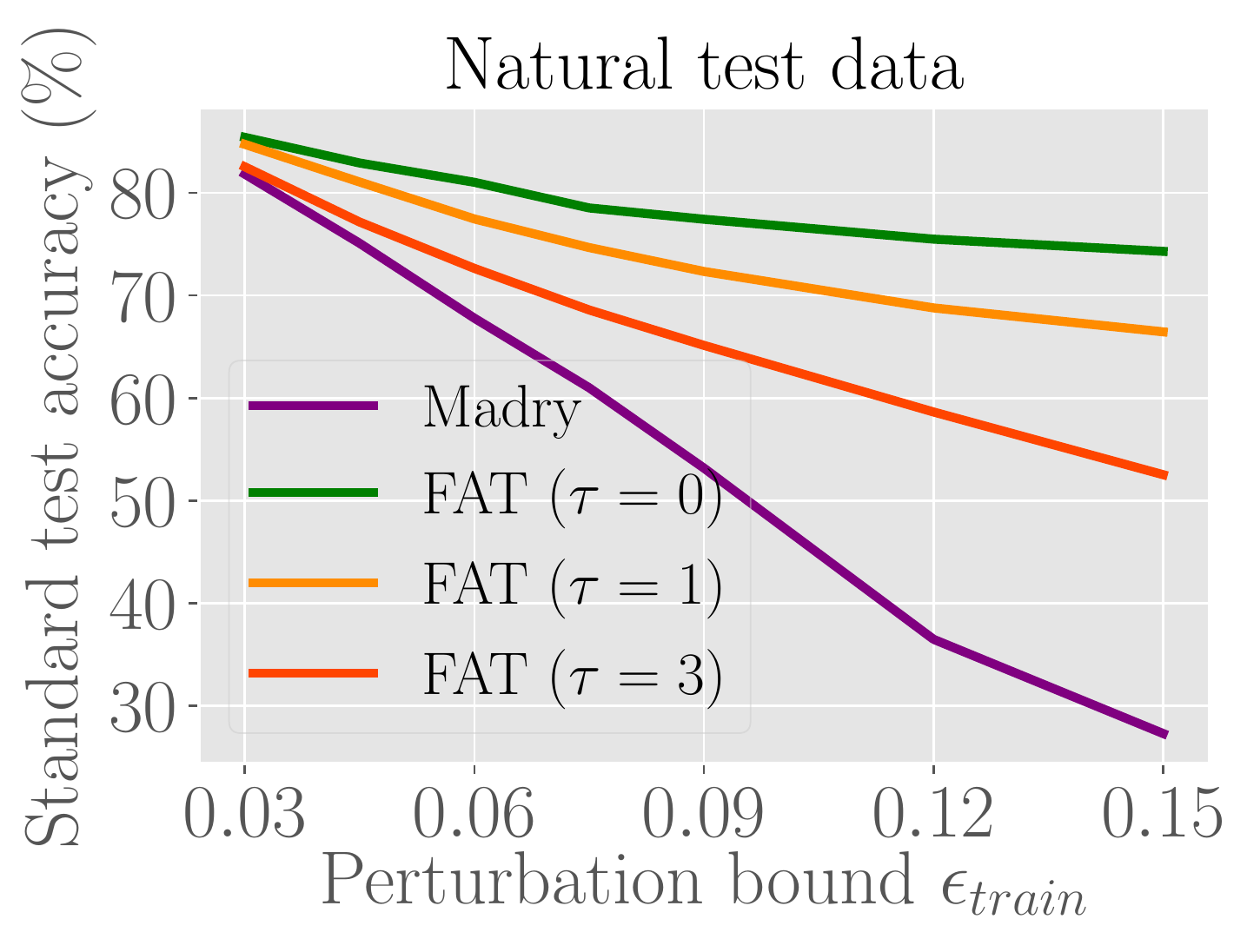}
    \includegraphics[scale=0.33]{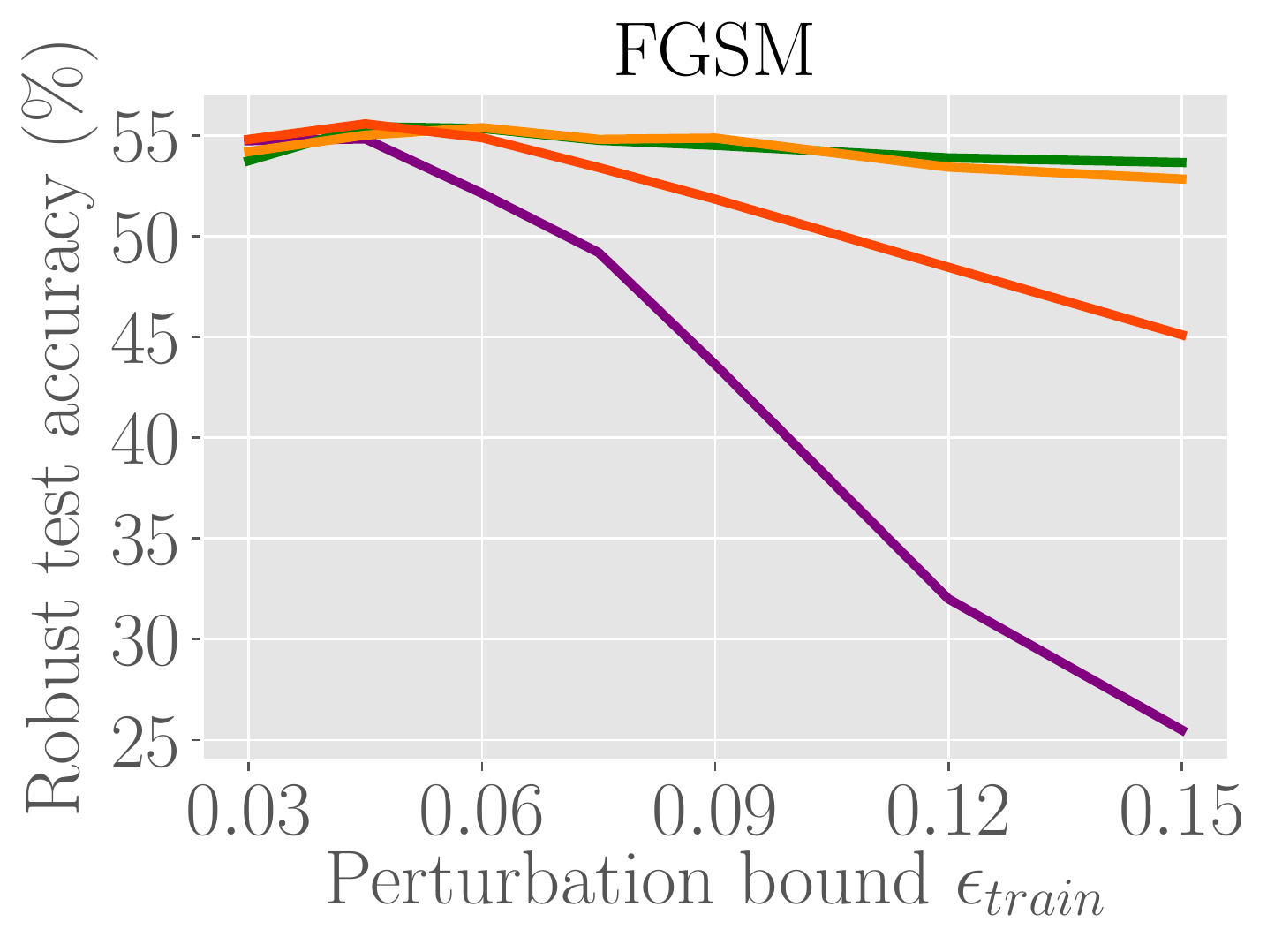}
    \includegraphics[scale=0.33]{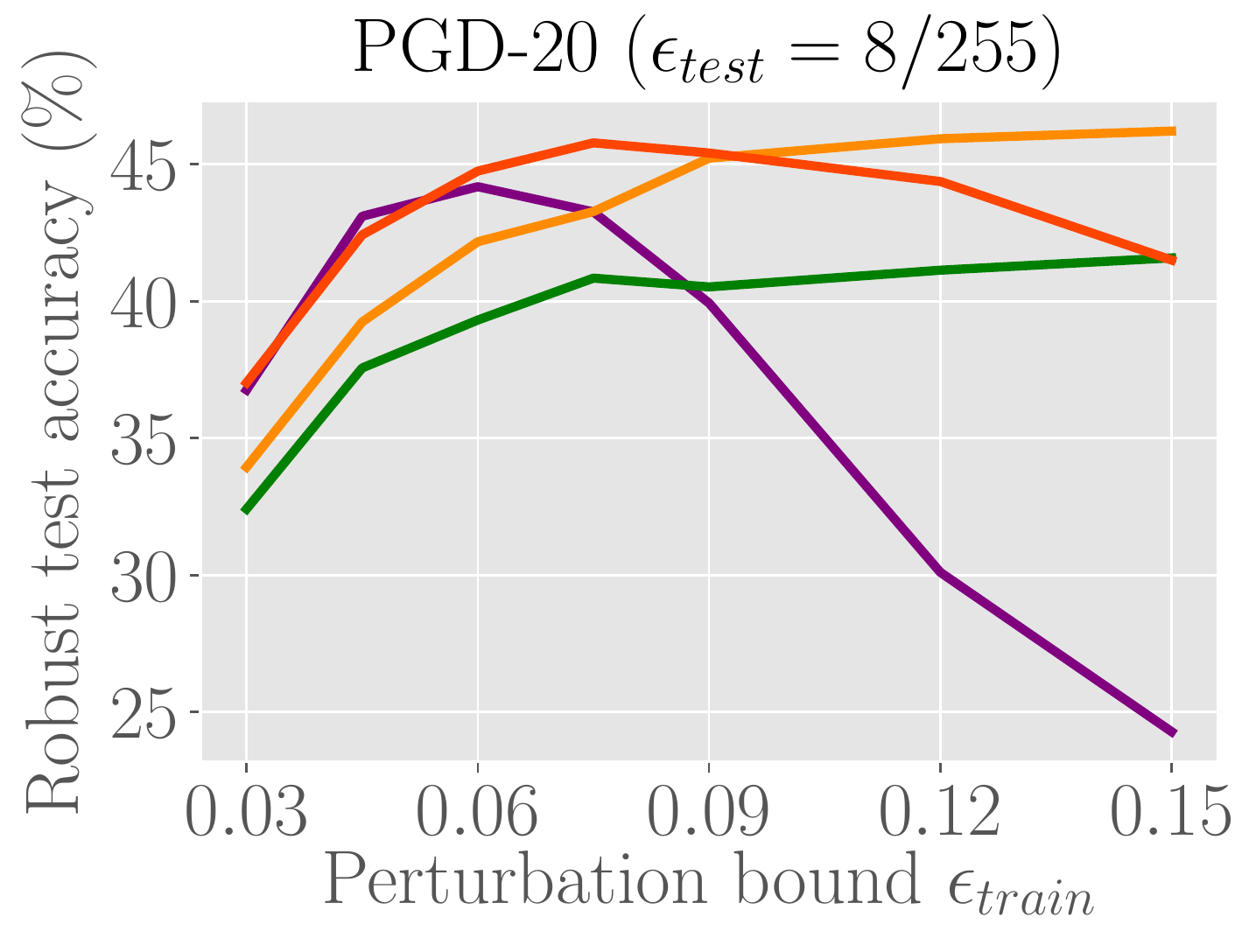}\\
    \includegraphics[scale=0.33]{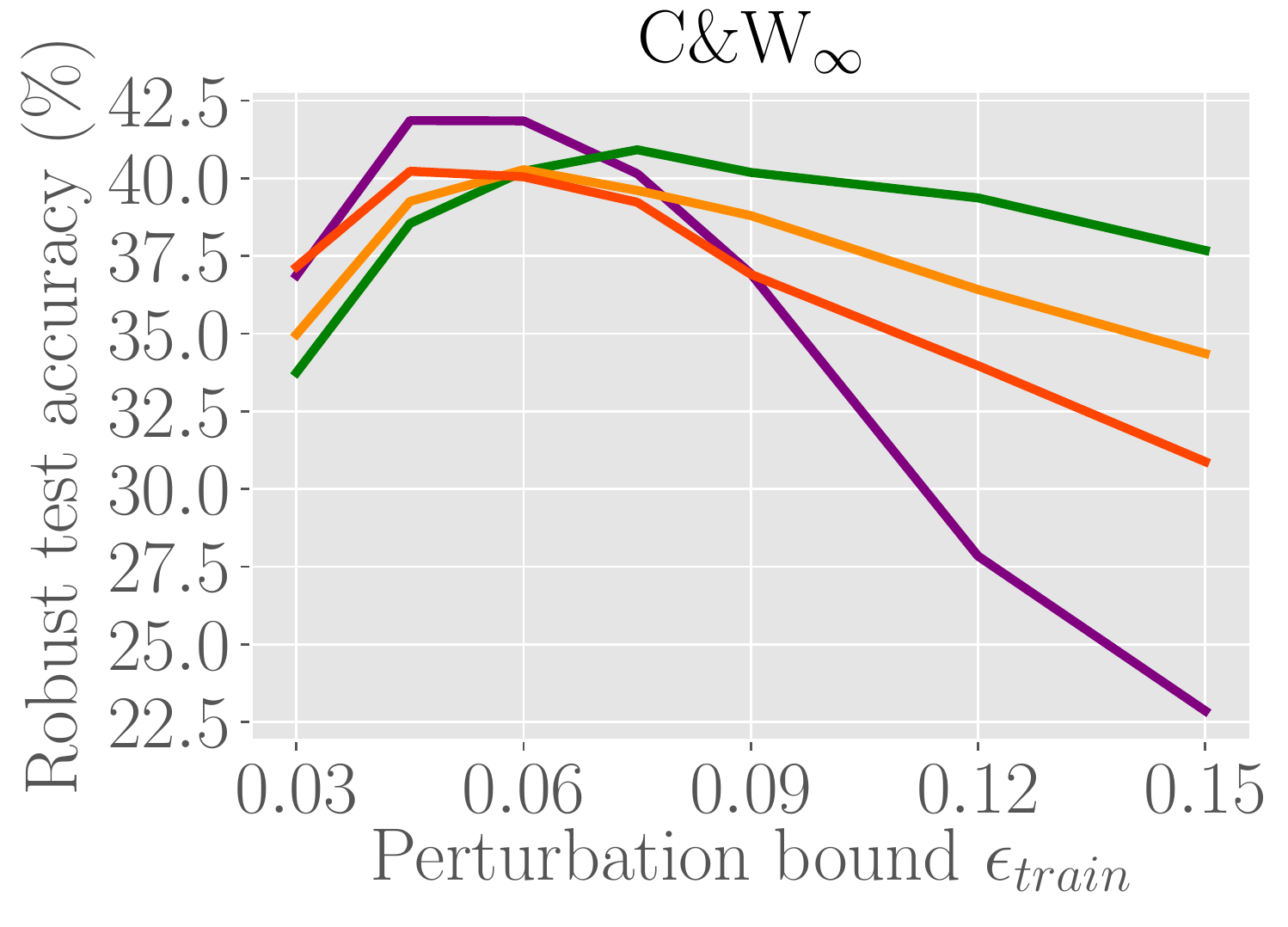}
    \includegraphics[scale=0.33]{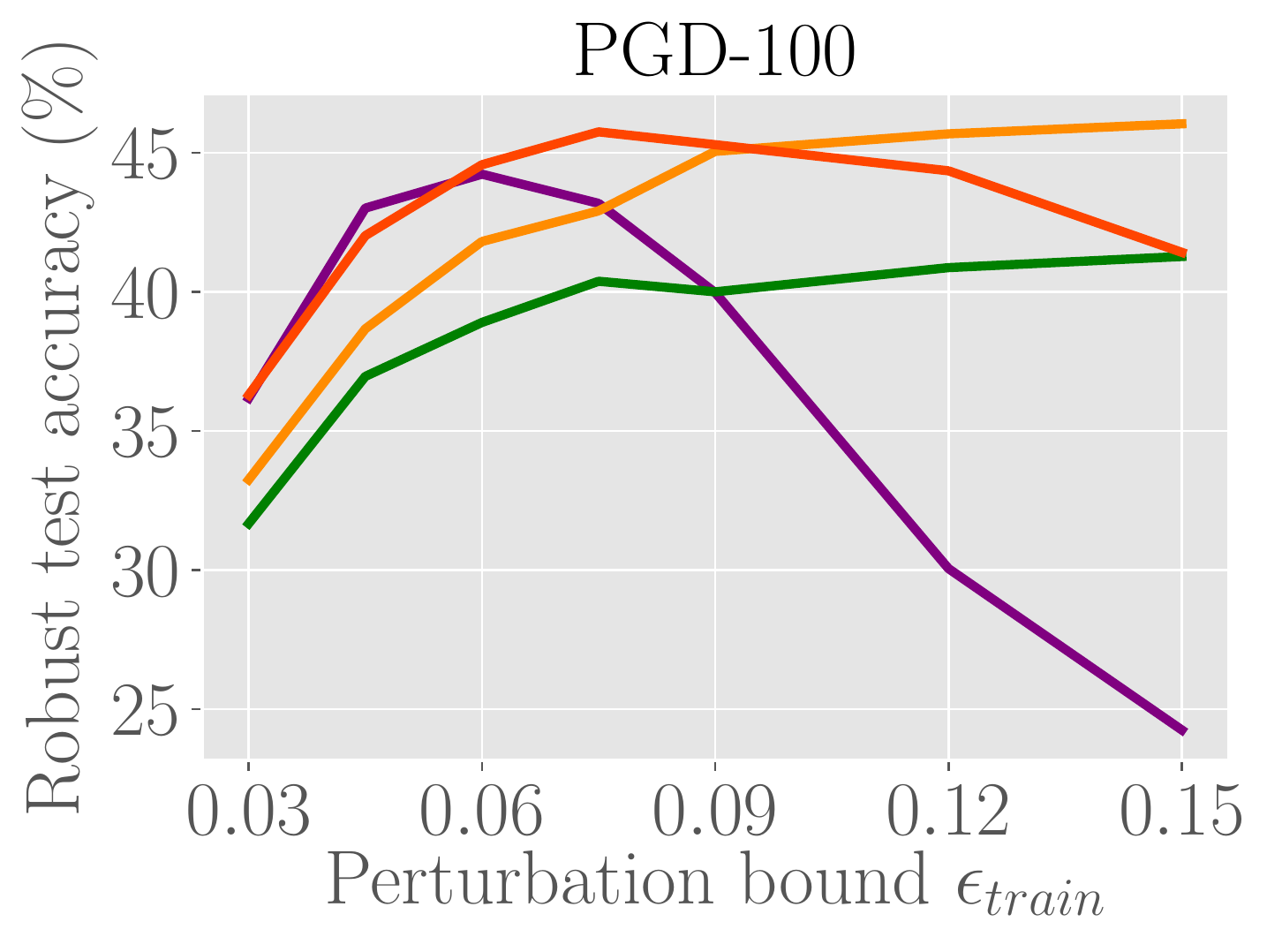}
    \includegraphics[scale=0.33]{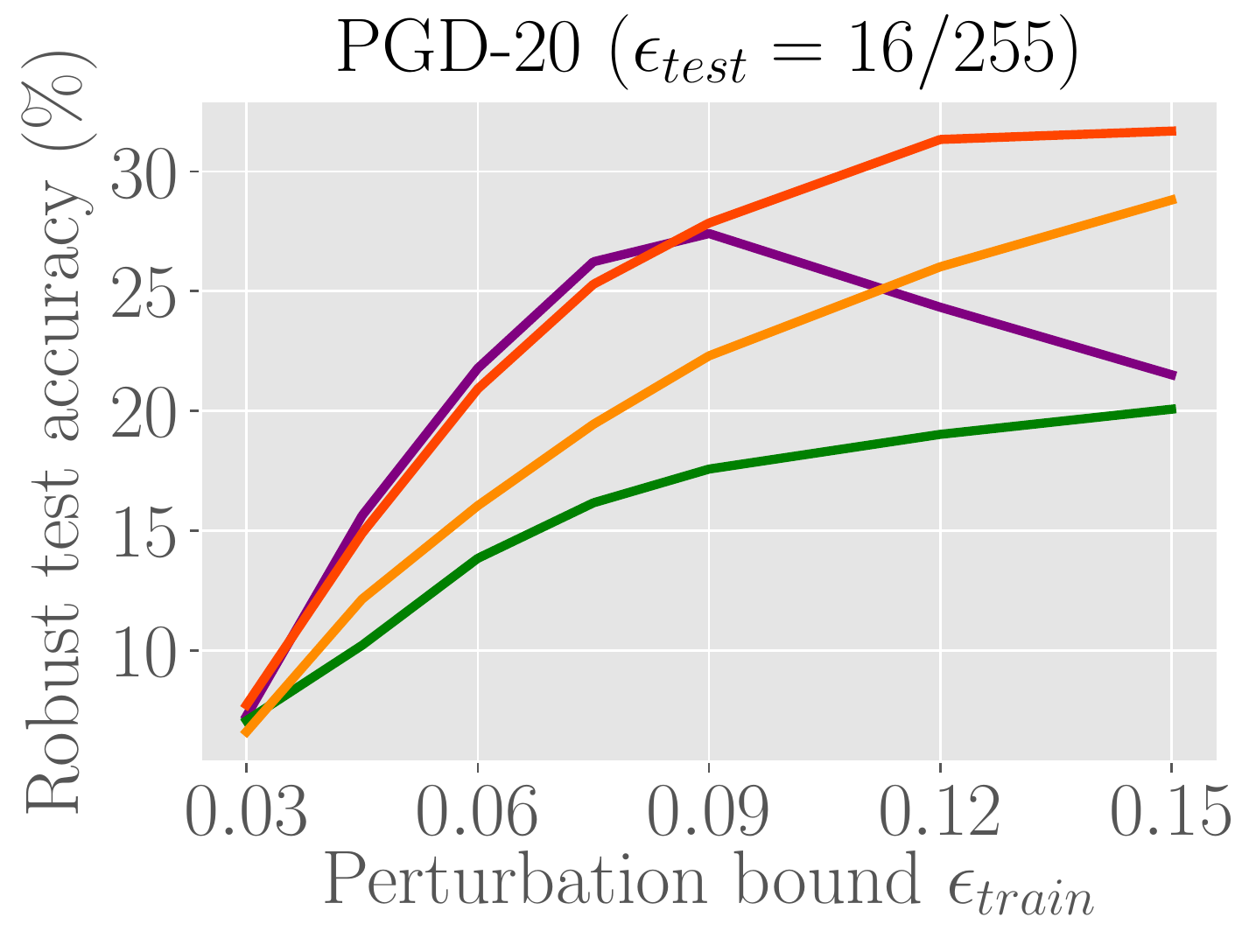}
    \caption{Test accuracy of Small CNN trained under different values of $\epsilon_{train}$ on CIFAR-10 dataset.}
    \label{fig:smallcnn_cifar10_dynamic_epsball}
\end{figure}

\begin{figure}[!htb]
    \centering
    \includegraphics[scale=0.33]{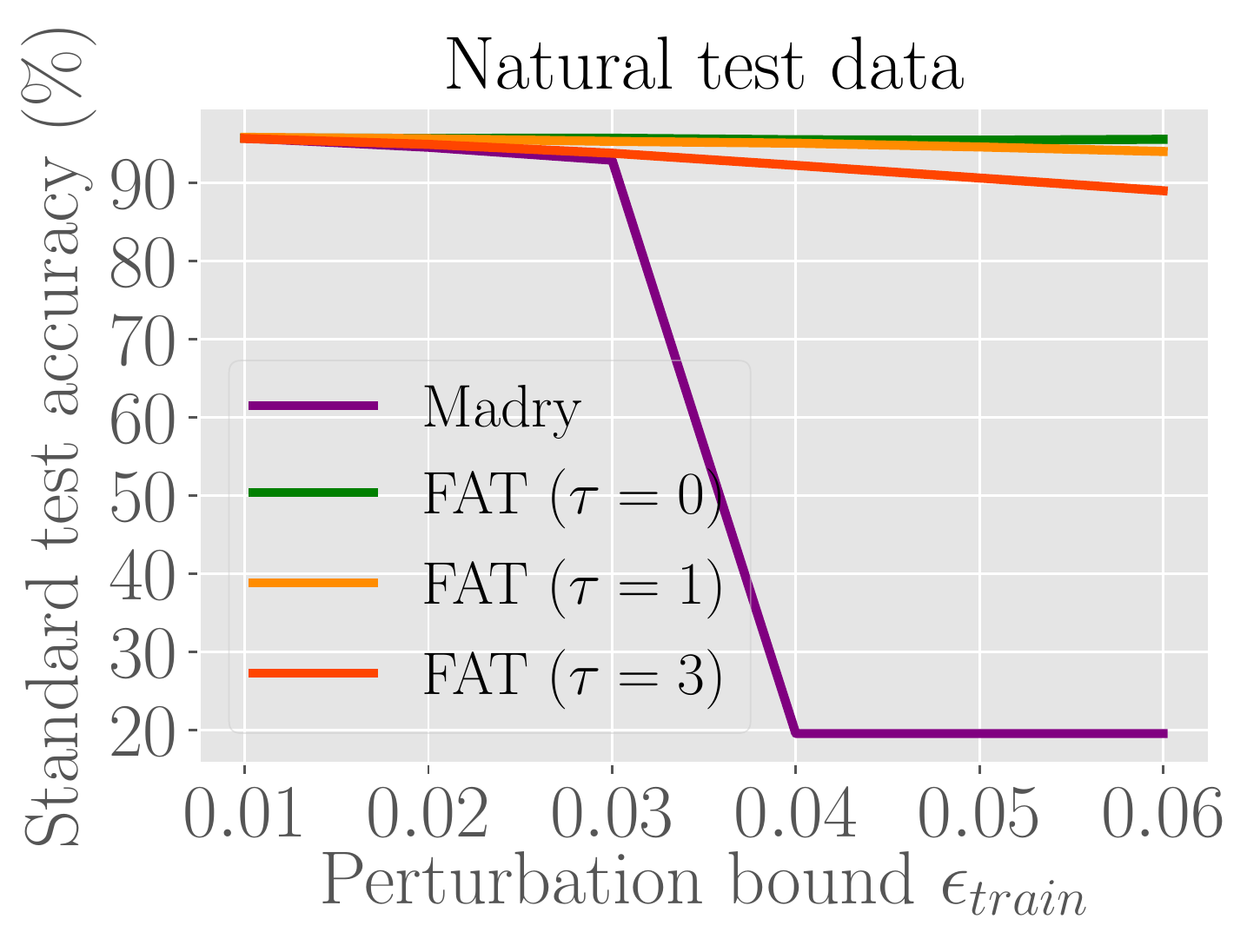}
    \includegraphics[scale=0.33]{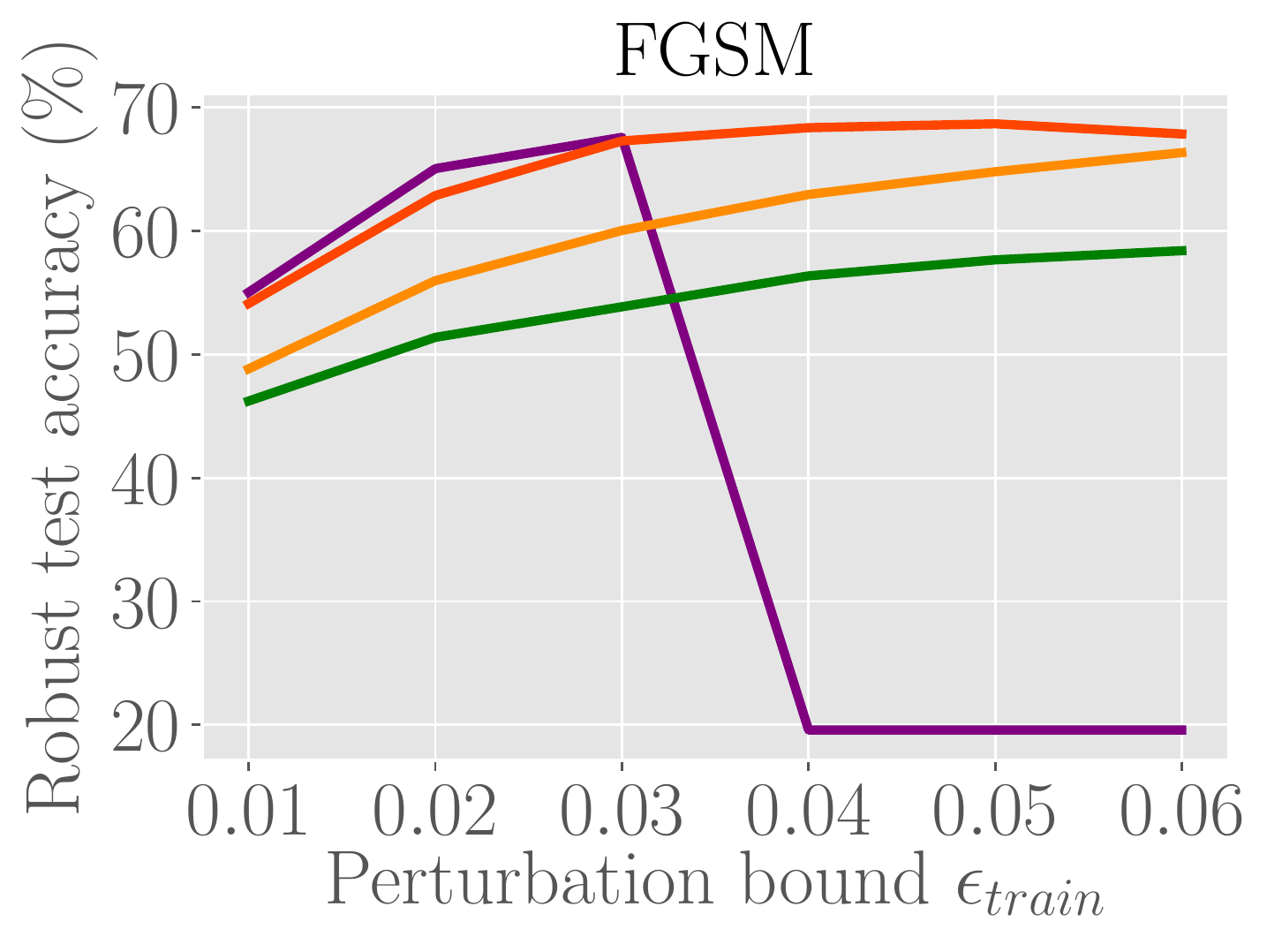}
    \includegraphics[scale=0.33]{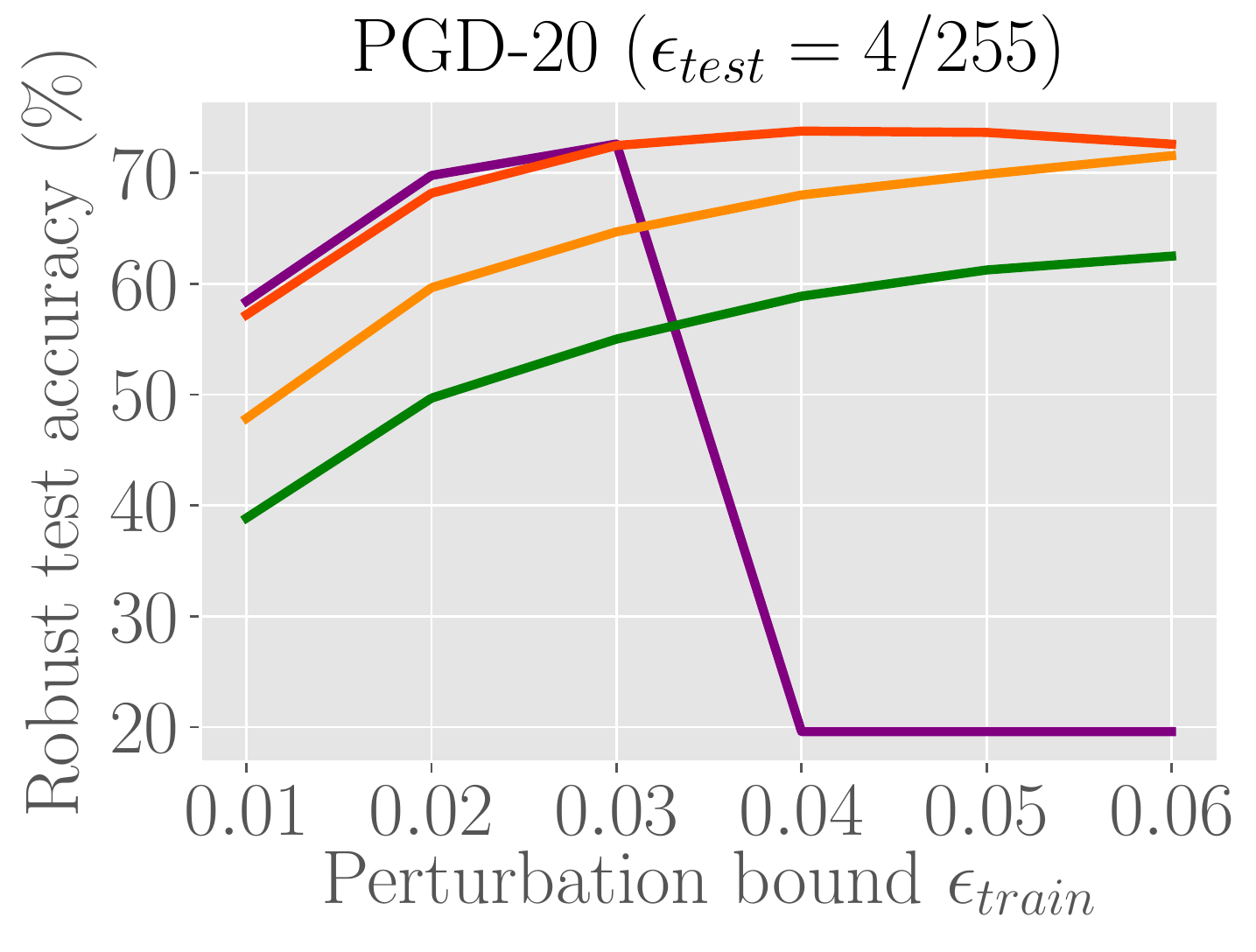}\\
    \includegraphics[scale=0.33]{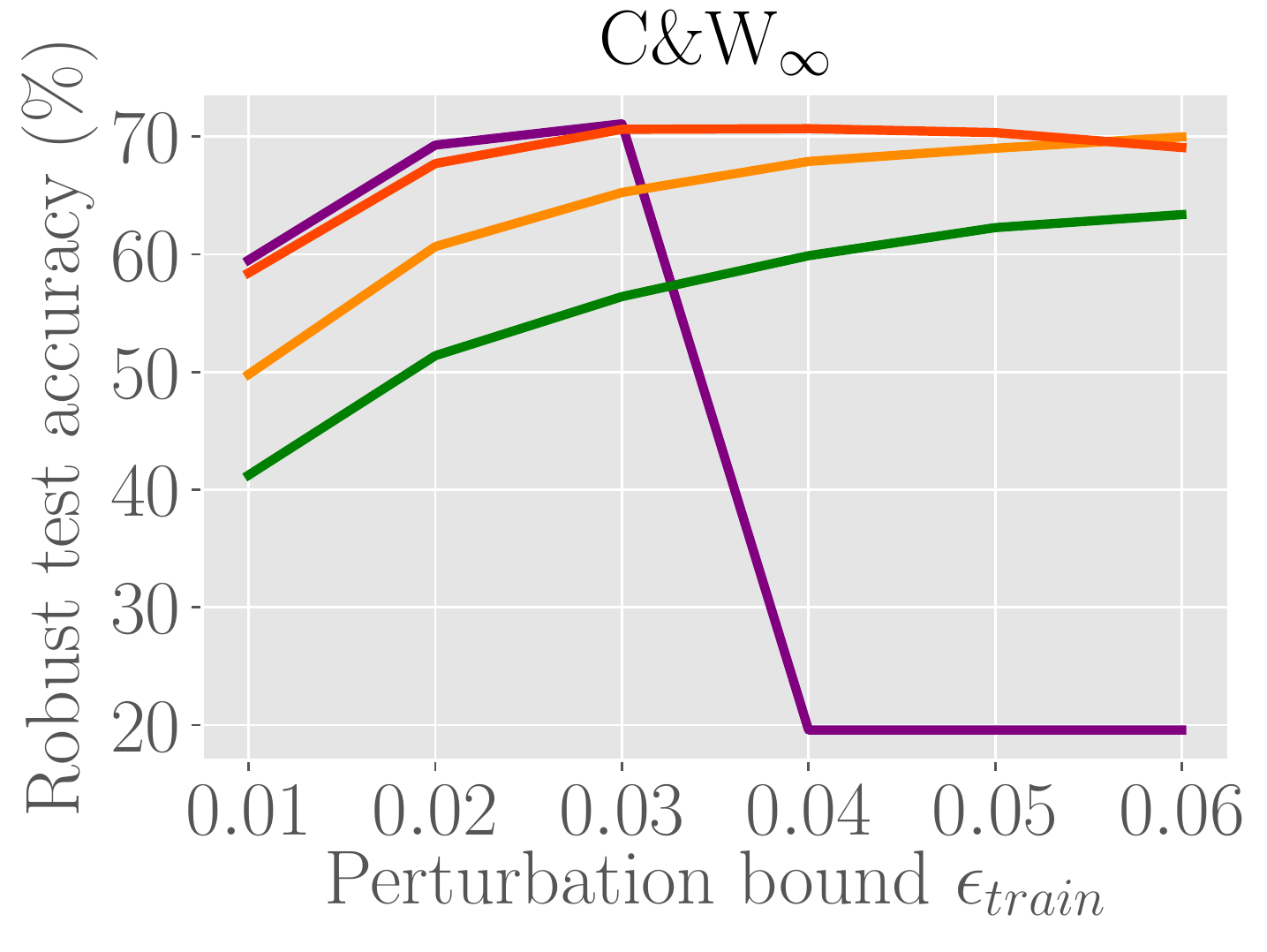}
    \includegraphics[scale=0.33]{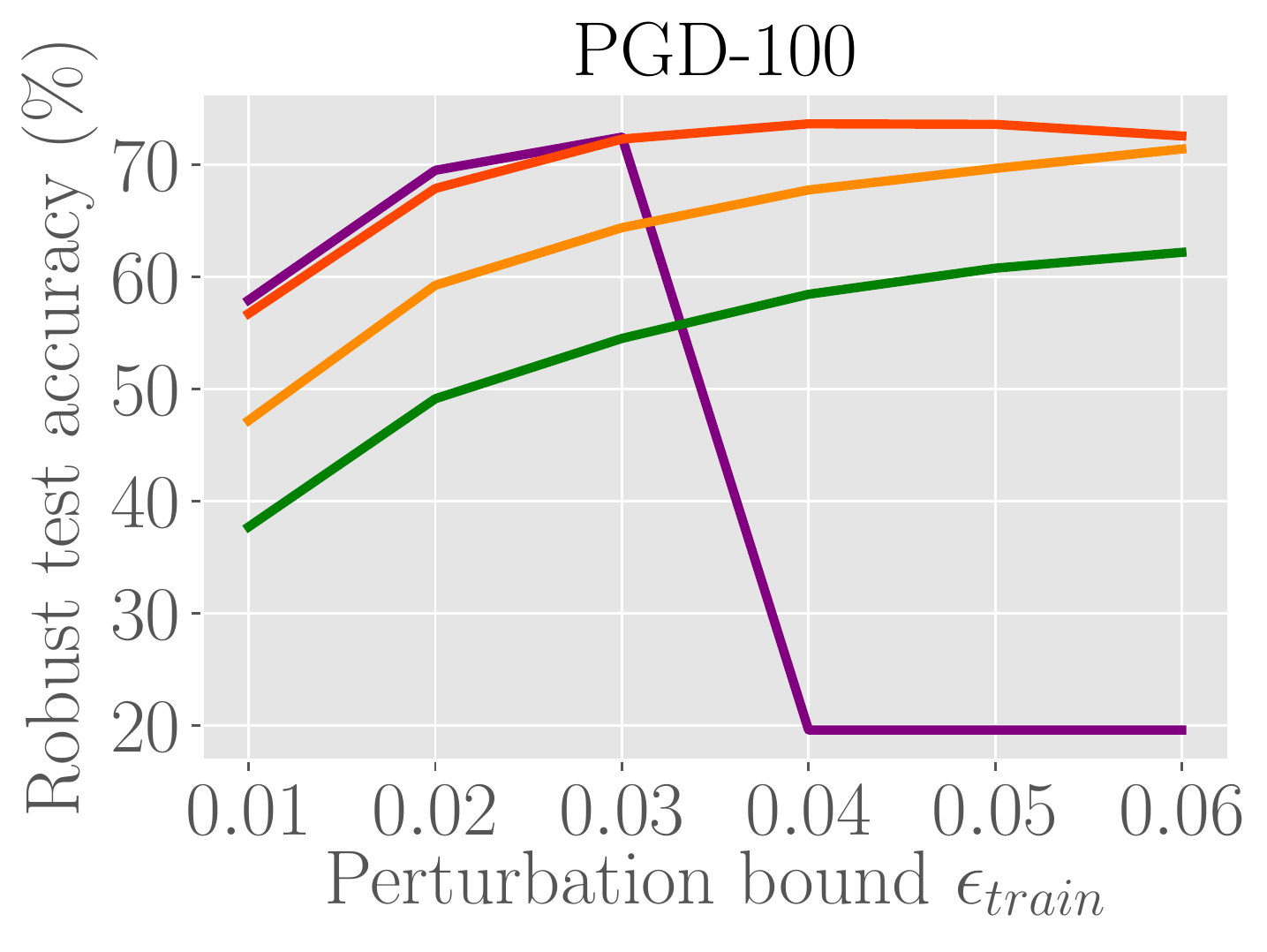}
    \includegraphics[scale=0.33]{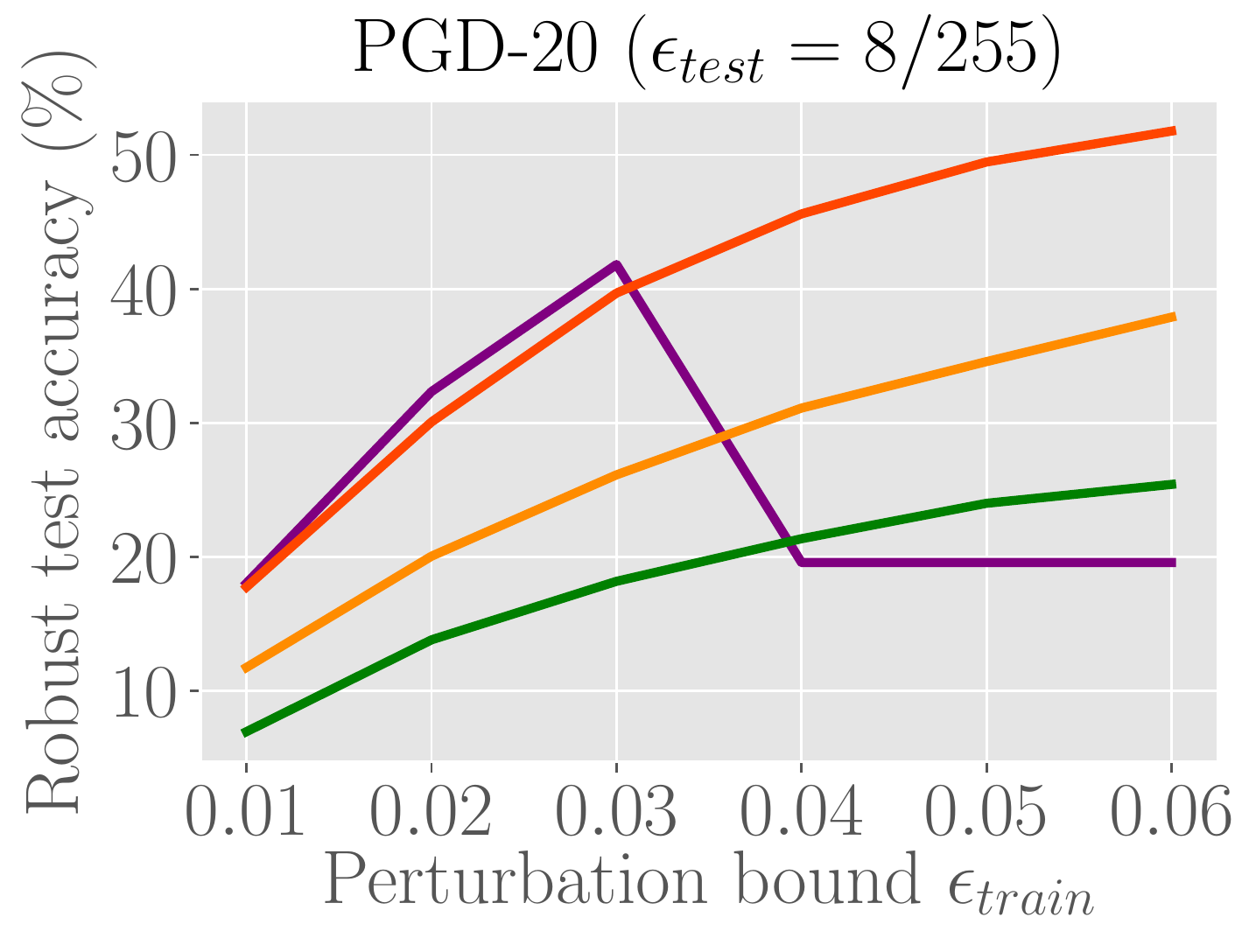}
    \caption{Test accuracy of Small CNN trained under different values of $\epsilon_{train}$ on SVHN dataset.}
    \label{fig:smallcnn_svhn_dynamic_epsball}
\end{figure}

We train Small CNN on CIFAR-10 and SVHN using the same settings as those stated in Section~\ref{section:fat_enable_lager_epsilon}. We show standard and robust test accuracy of deep model (Small CNN) on CIFAR-10 dataset (Figure~\ref{fig:smallcnn_cifar10_dynamic_epsball}) and SVHN dataset (Figure~\ref{fig:smallcnn_svhn_dynamic_epsball}).

\subsection{FAT for TRADES}
\label{appendix:exp_fat_for_trades}
\begin{figure}[!htb]
    \centering
    \includegraphics[scale=0.33]{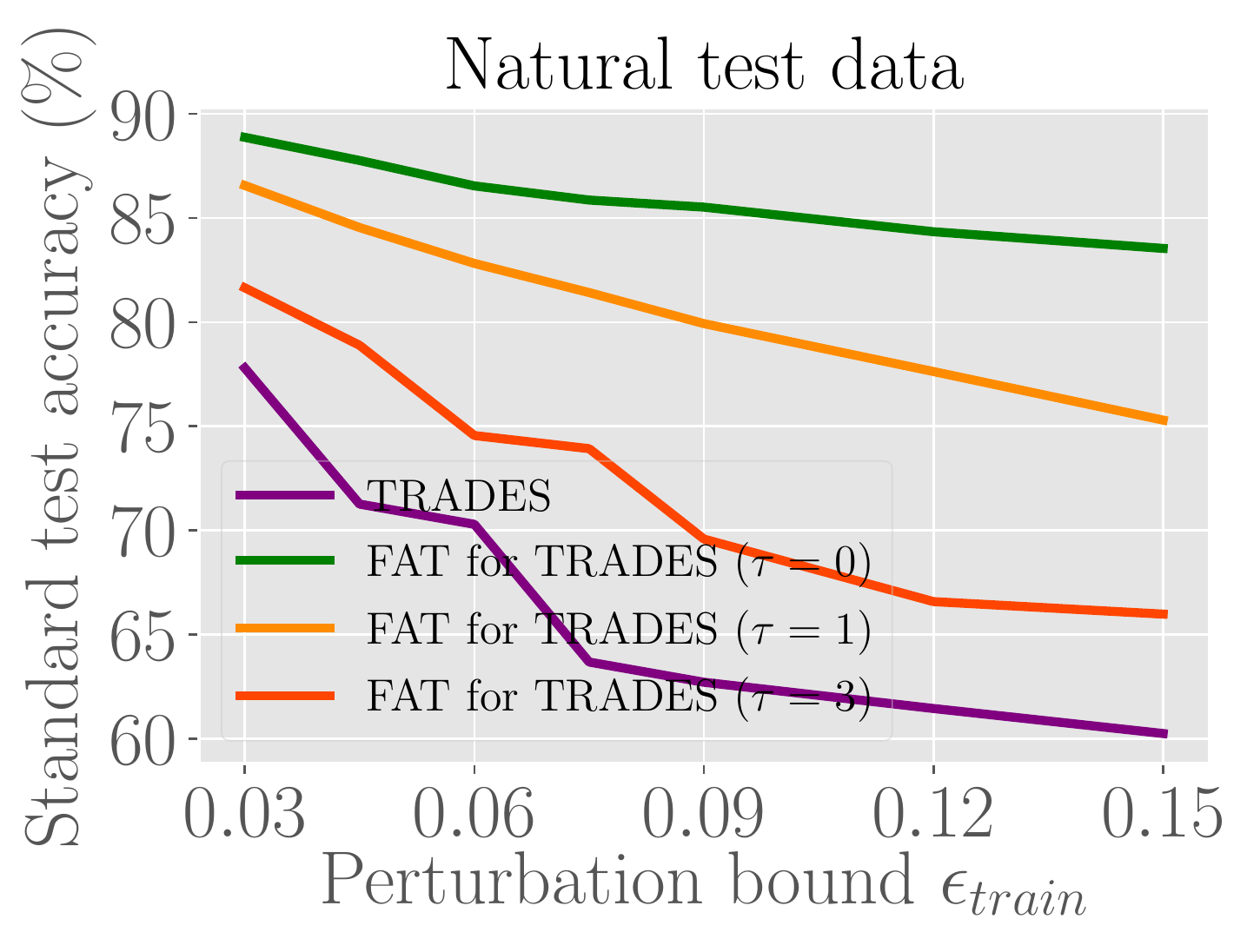}
    \includegraphics[scale=0.33]{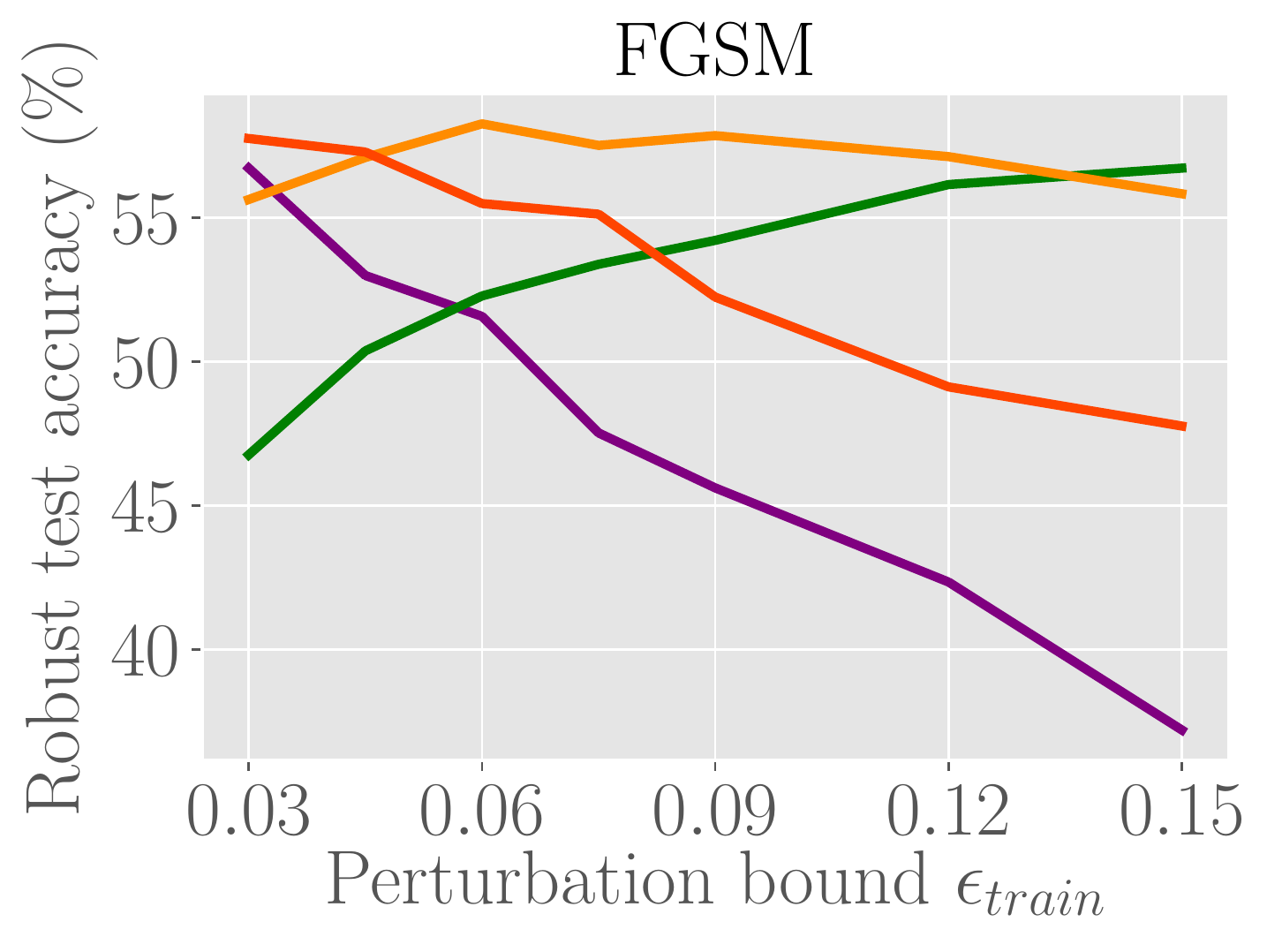}
    \includegraphics[scale=0.33]{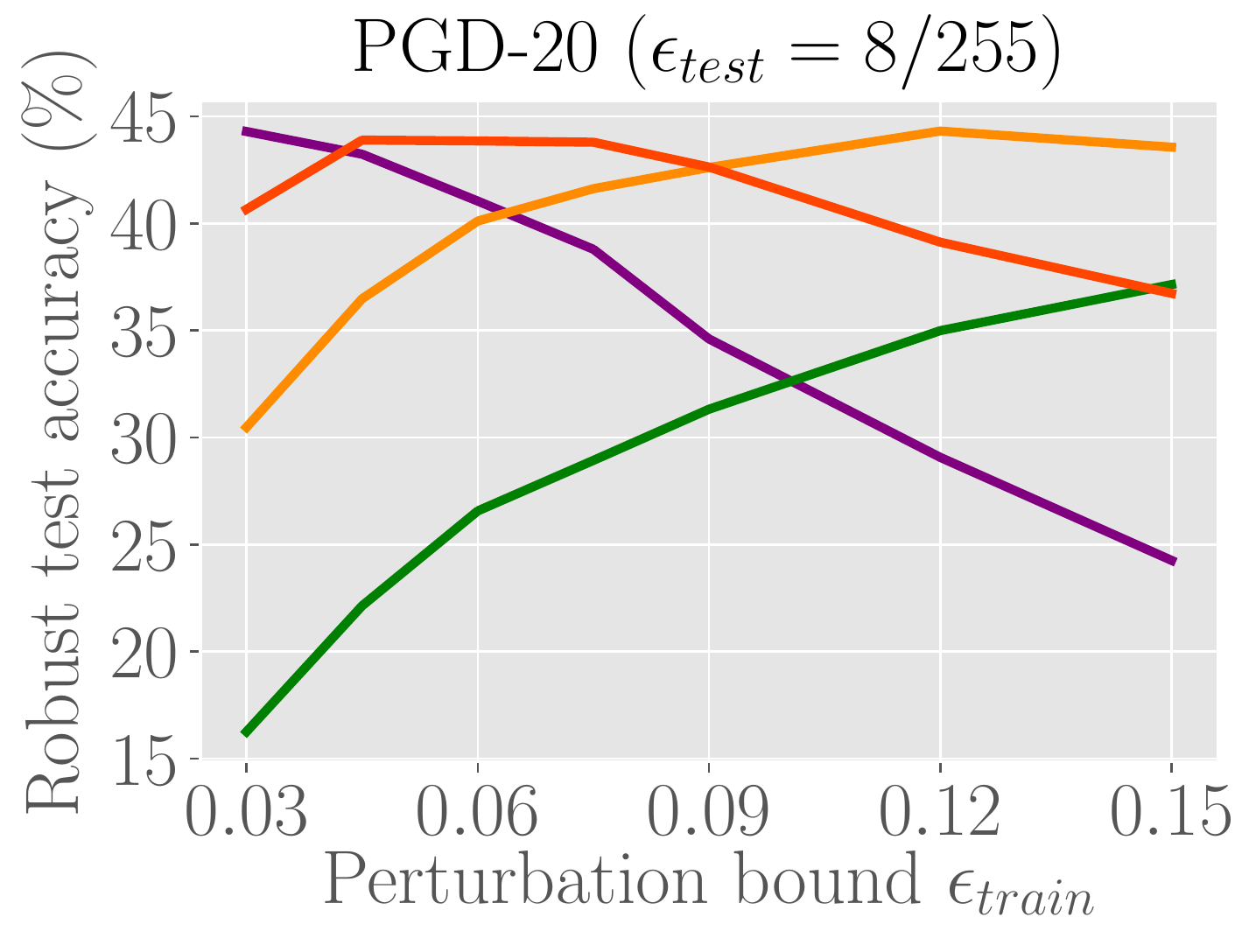}\\
    \includegraphics[scale=0.33]{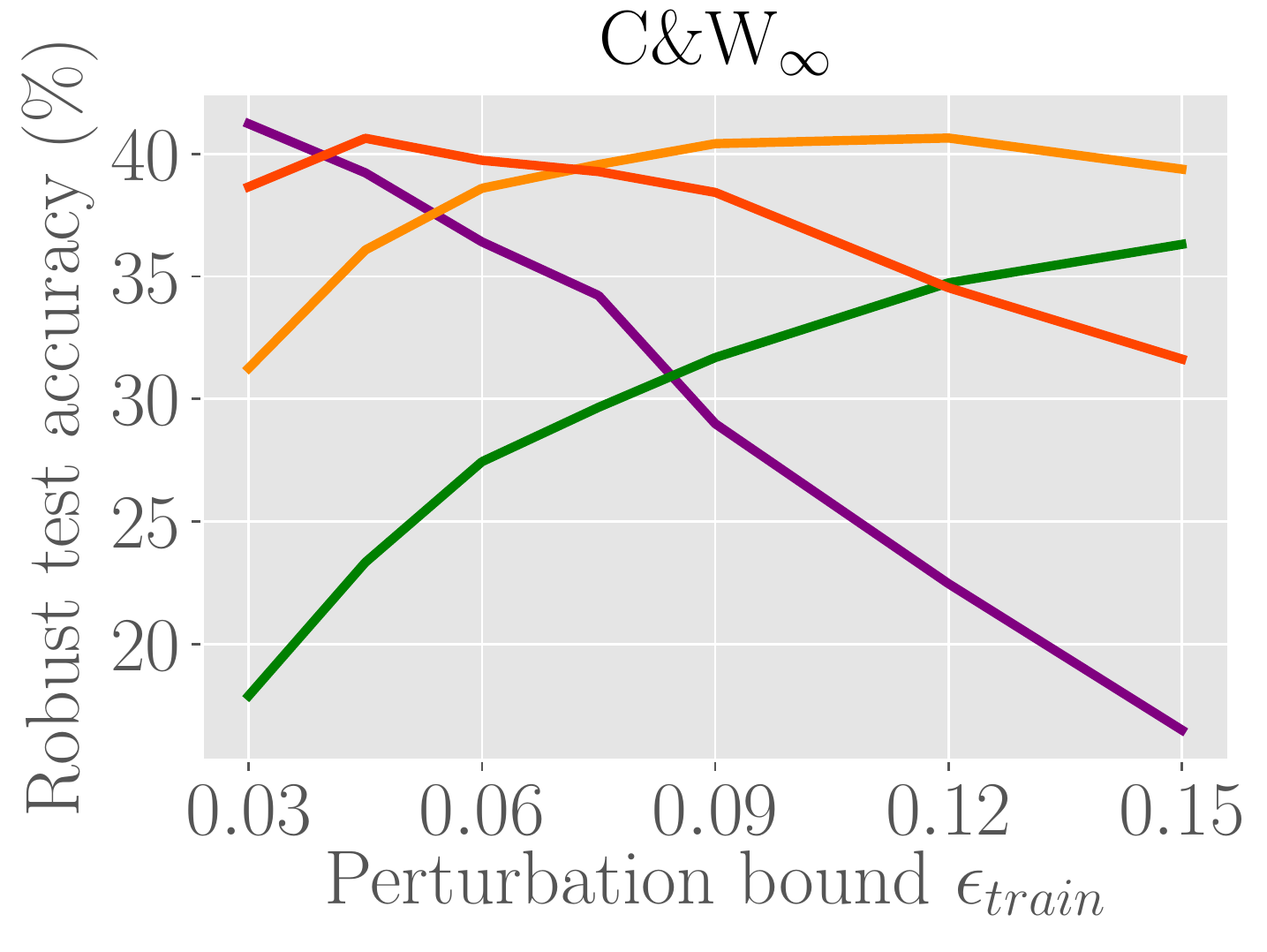}
    \includegraphics[scale=0.33]{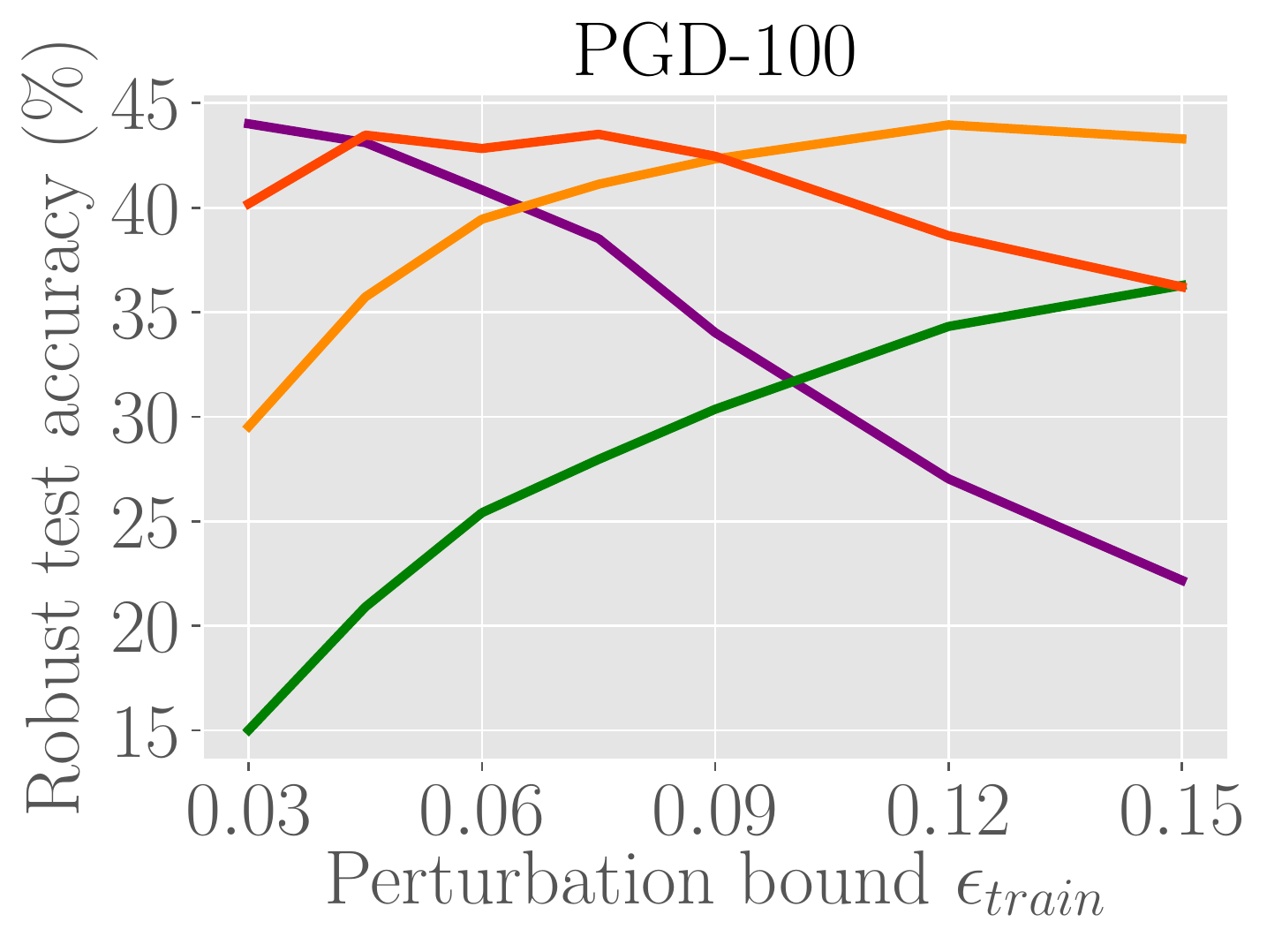}
    \includegraphics[scale=0.33]{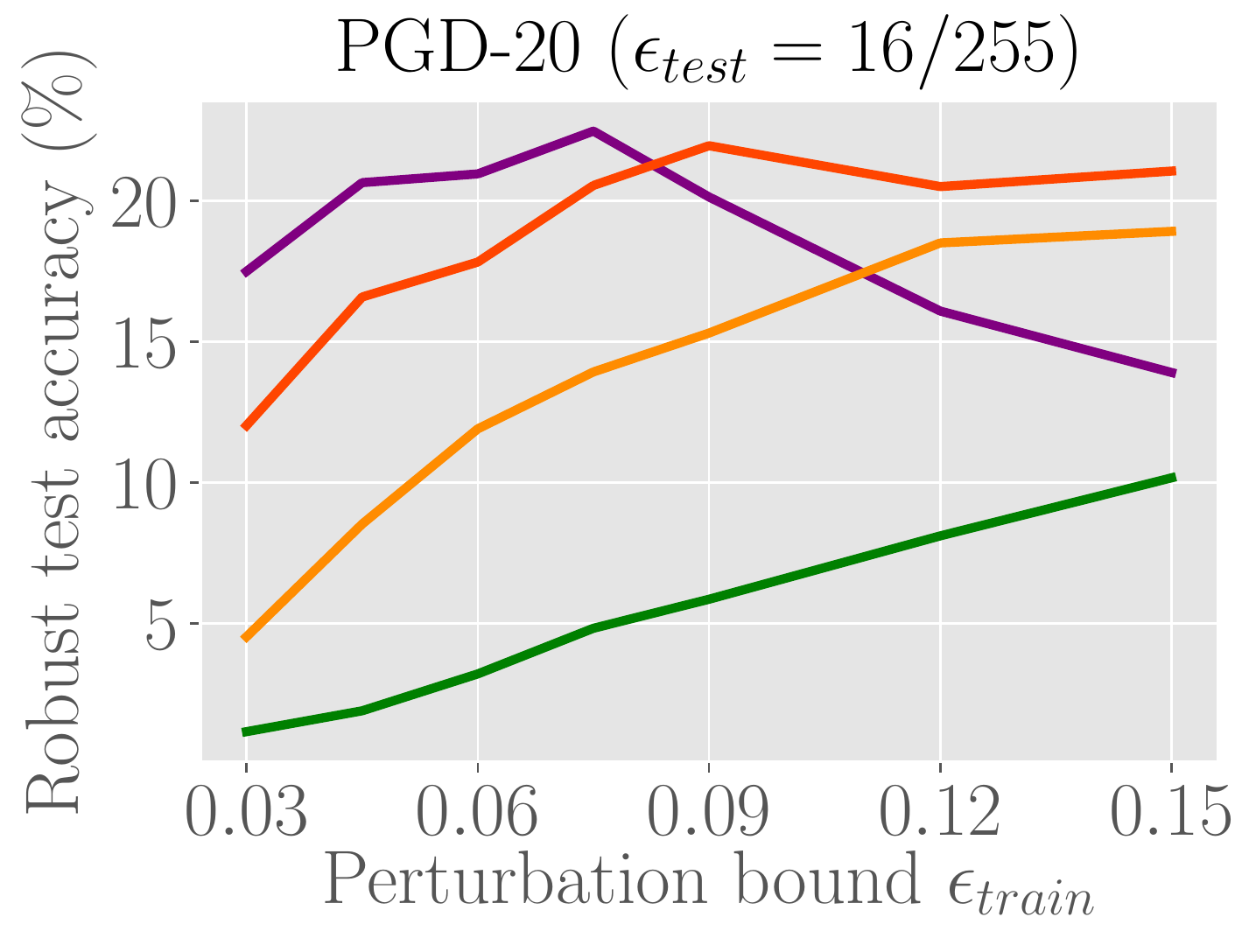}
    \caption{Test accuracy of Small CNN trained by FAT for TRADES ($\tau=0,1,3$) and TRADES under different values of  $\epsilon_{train}$ on CIFAR-10 dataset.}
    \label{fig:smallcnn_cifar10_fat_trades_dynamic_epsball}
\end{figure}

\begin{figure}[!htb]
    \centering
    \includegraphics[scale=0.33]{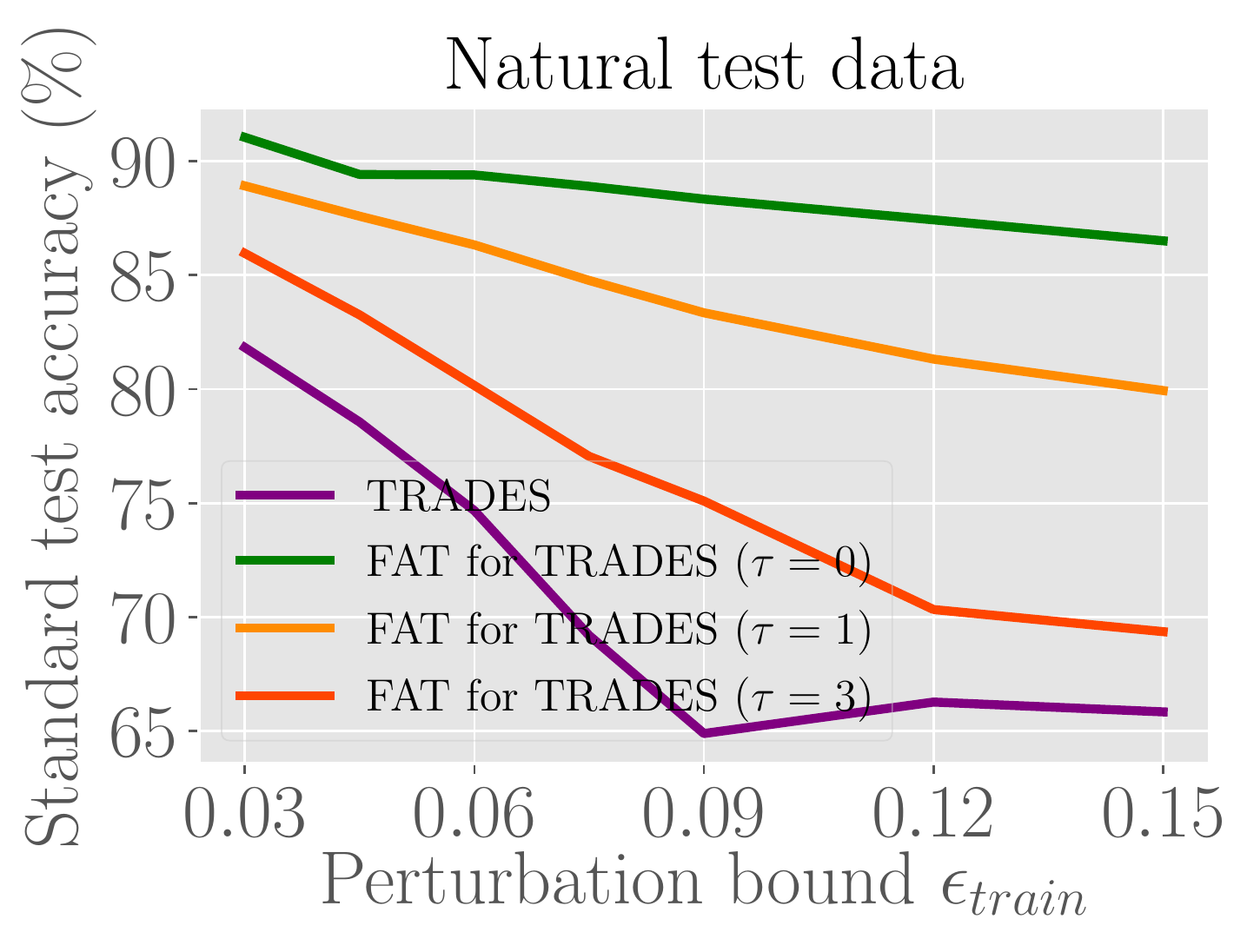}
    \includegraphics[scale=0.33]{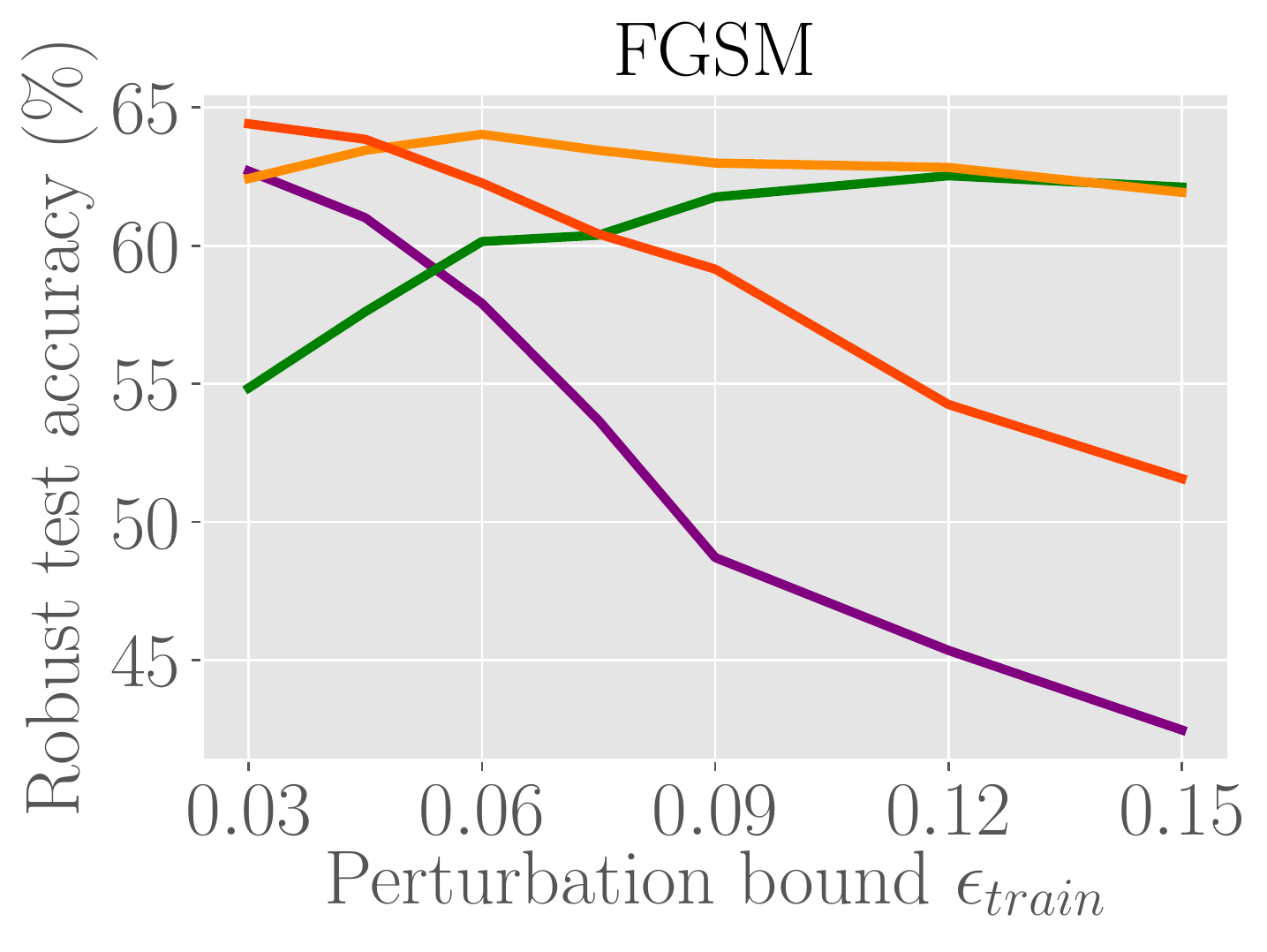}
    \includegraphics[scale=0.33]{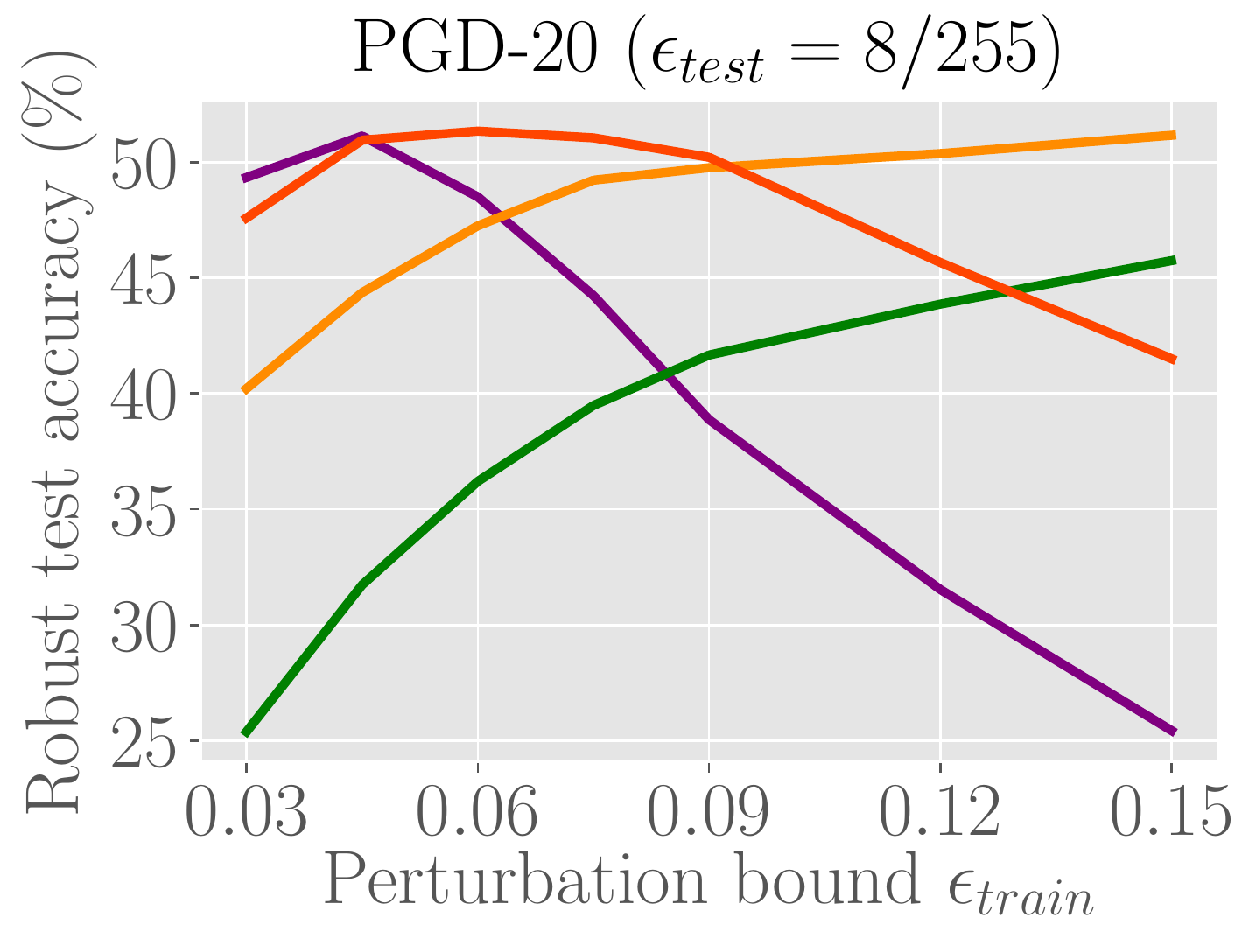}\\
    \includegraphics[scale=0.33]{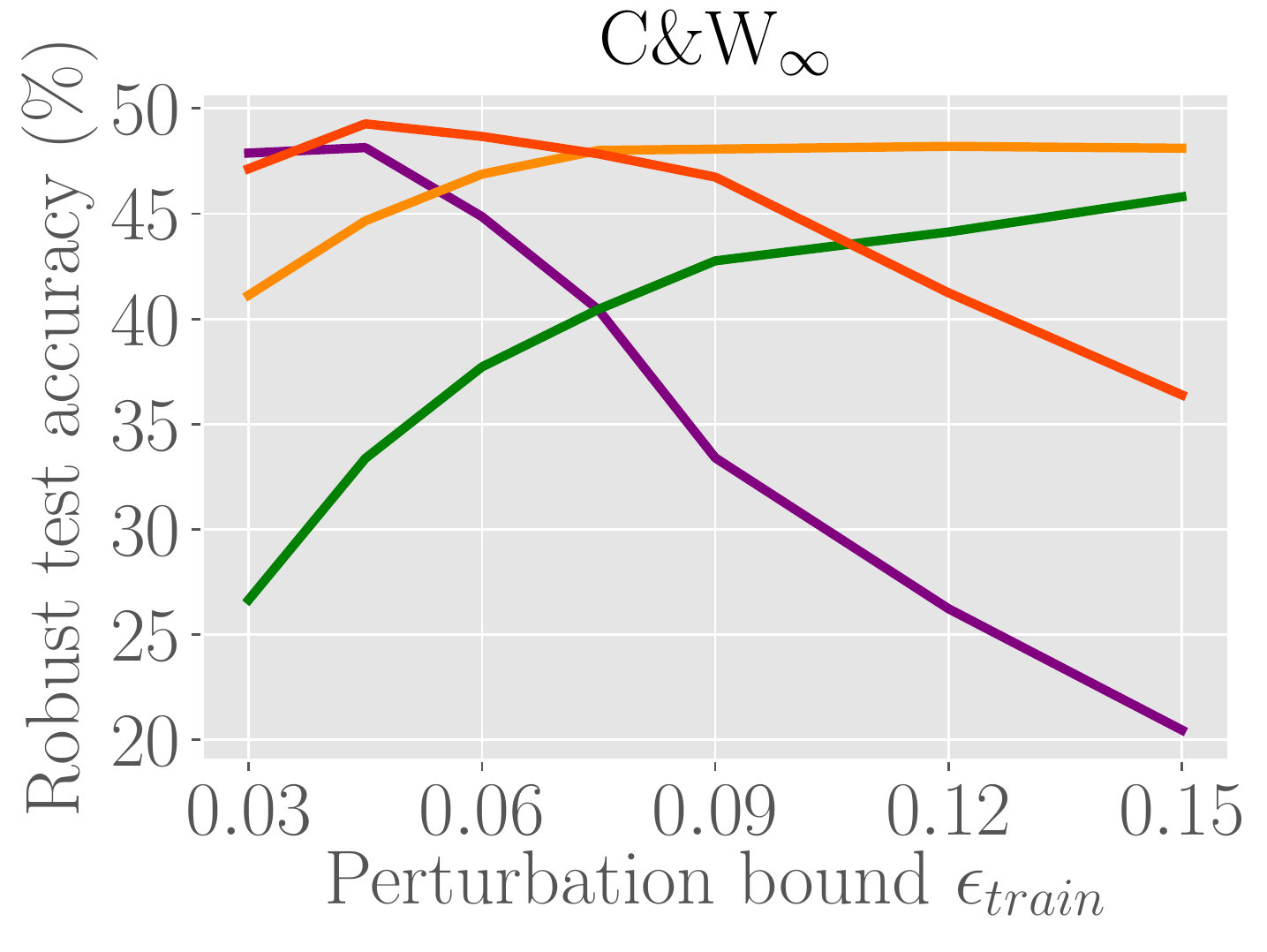}
    \includegraphics[scale=0.33]{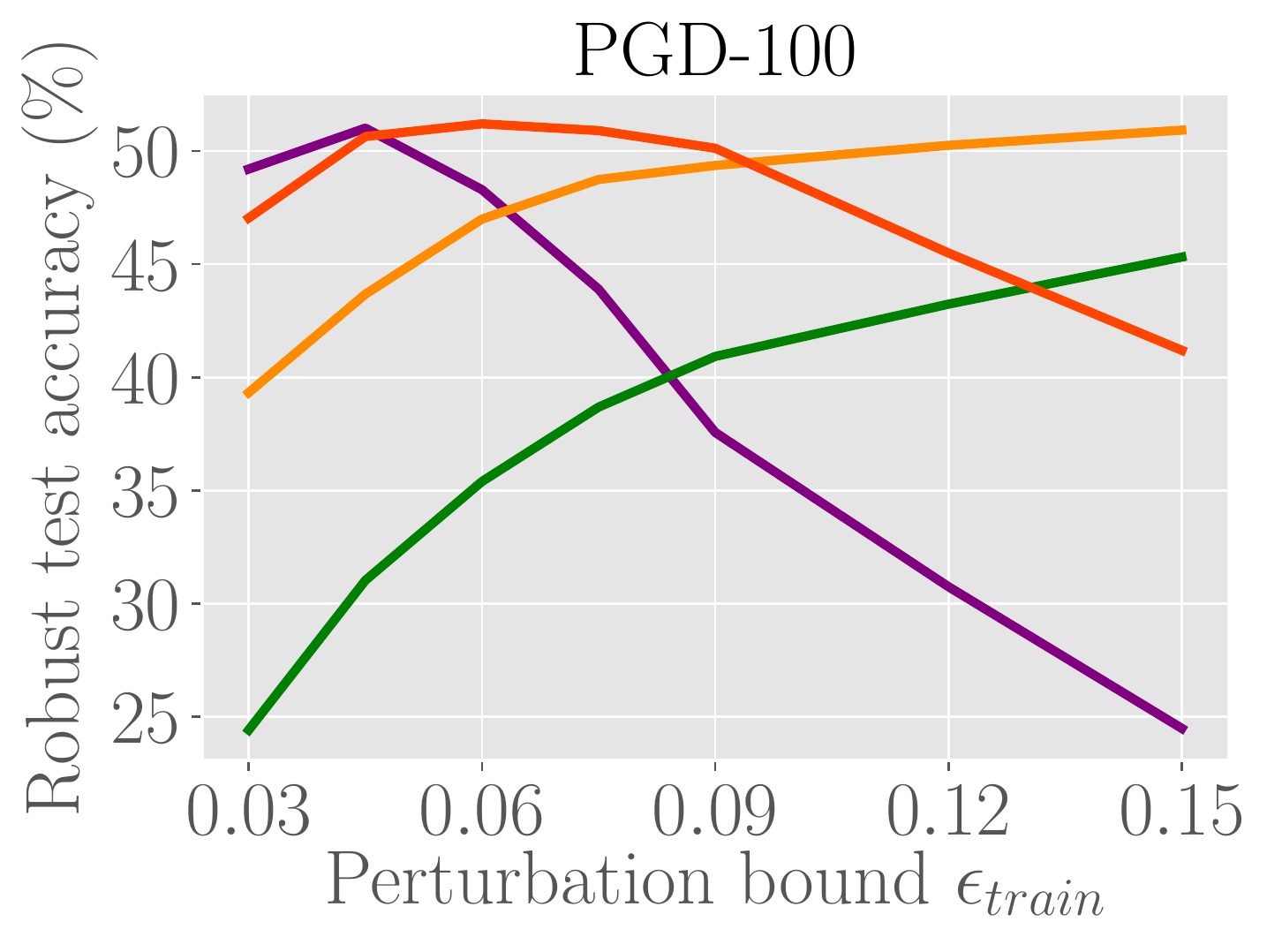}
    \includegraphics[scale=0.33]{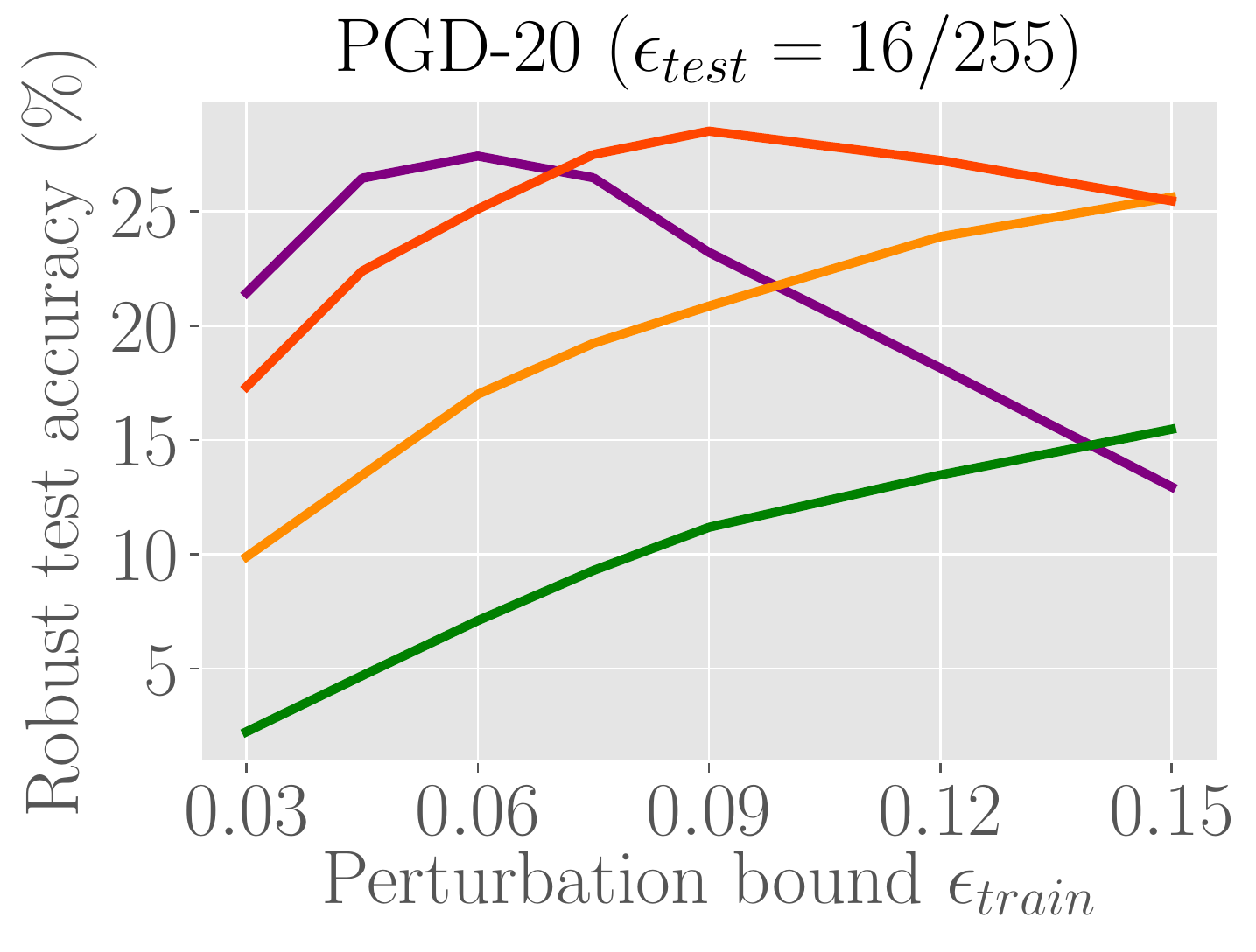}
    \caption{Test accuracy of ResNet-18 trained by FAT for TRADES ($\tau=0,1,3$) and TRADES under different values of $\epsilon_{train}$ on CIFAR-10 dataset.}
    \label{fig:resnet18_cifar10_fat_trades_dynamic_epsball}
\end{figure}

We apply FAT for TRADES(Algorithm~\ref{alg:FAT_for_TRADES}) to Small CNN and ResNet-18 on CIFAR-10 dataset. All training settings are the same as those are stated in Section~\ref{section:fat_enable_lager_epsilon}. Regularization parameter $\beta = 6$.  We present standard and robust test results of Small CNN (Figure~\ref{fig:smallcnn_cifar10_fat_trades_dynamic_epsball}) and ResNet-18 (Figure~\ref{fig:resnet18_cifar10_fat_trades_dynamic_epsball}).

\subsection{Maximum PGD Step $K = 20$}
\label{appendix:exp_pgd20}
\begin{figure}[!htb]
    \centering
    \includegraphics[scale=0.33]{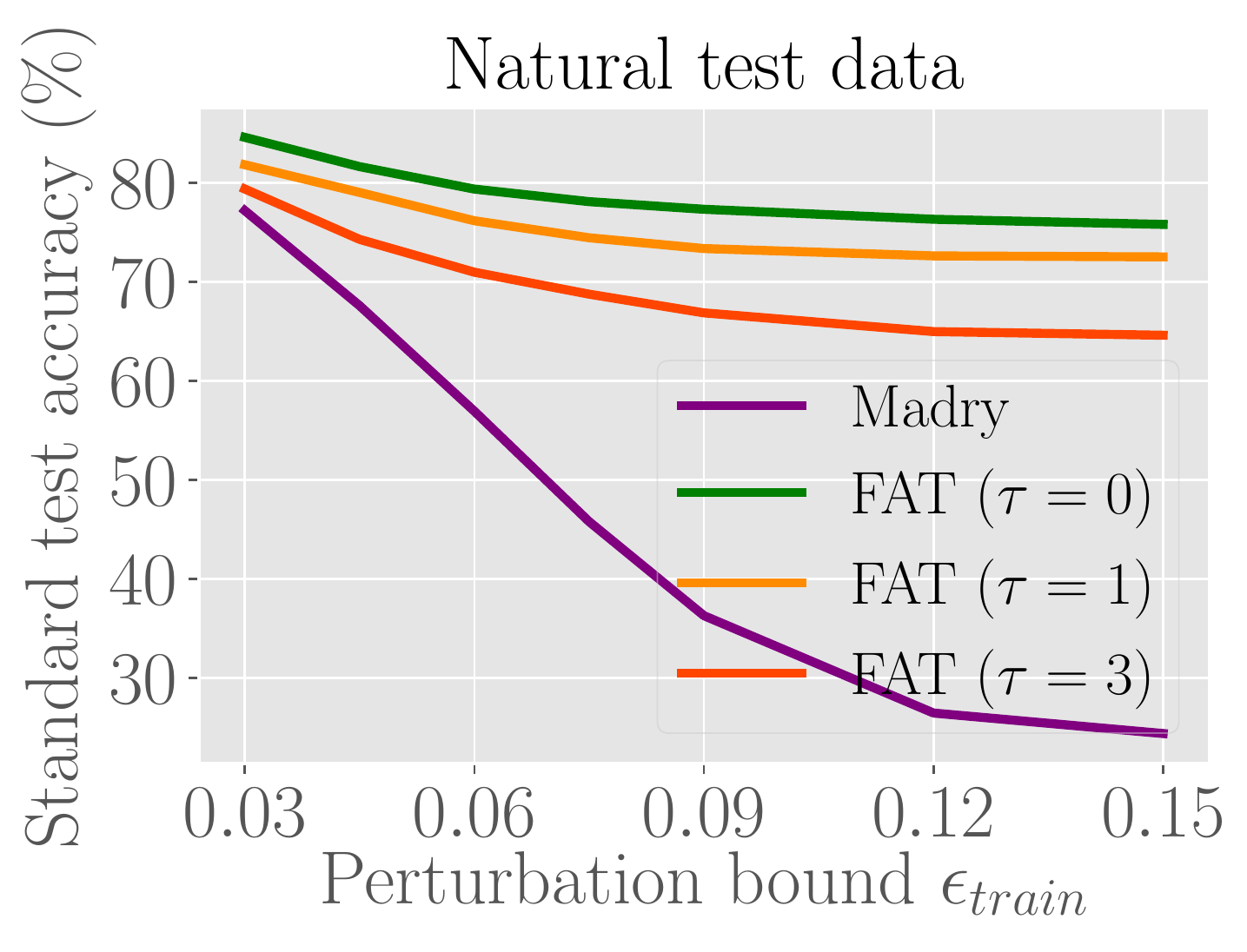}
    \includegraphics[scale=0.33]{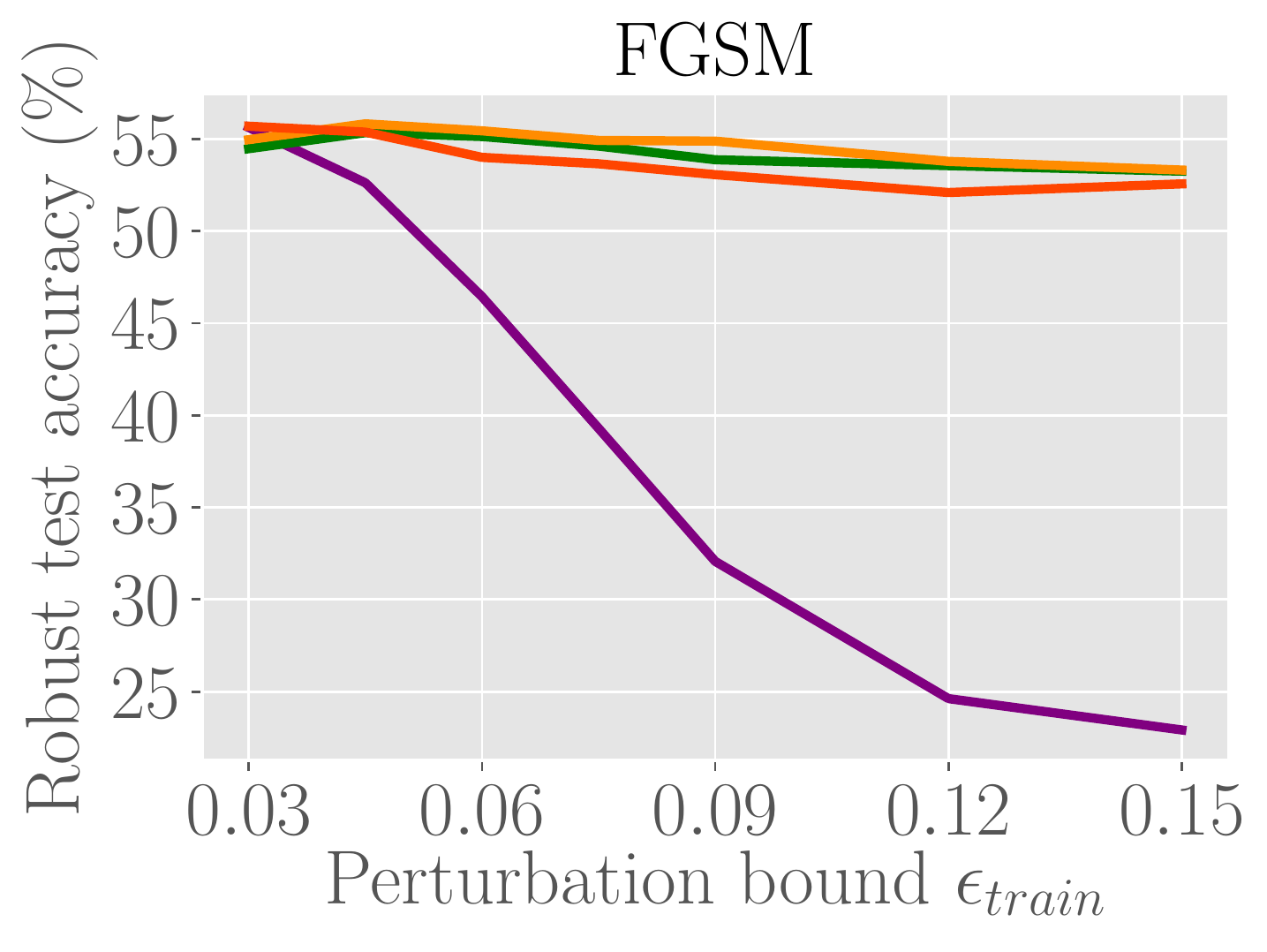}
    \includegraphics[scale=0.33]{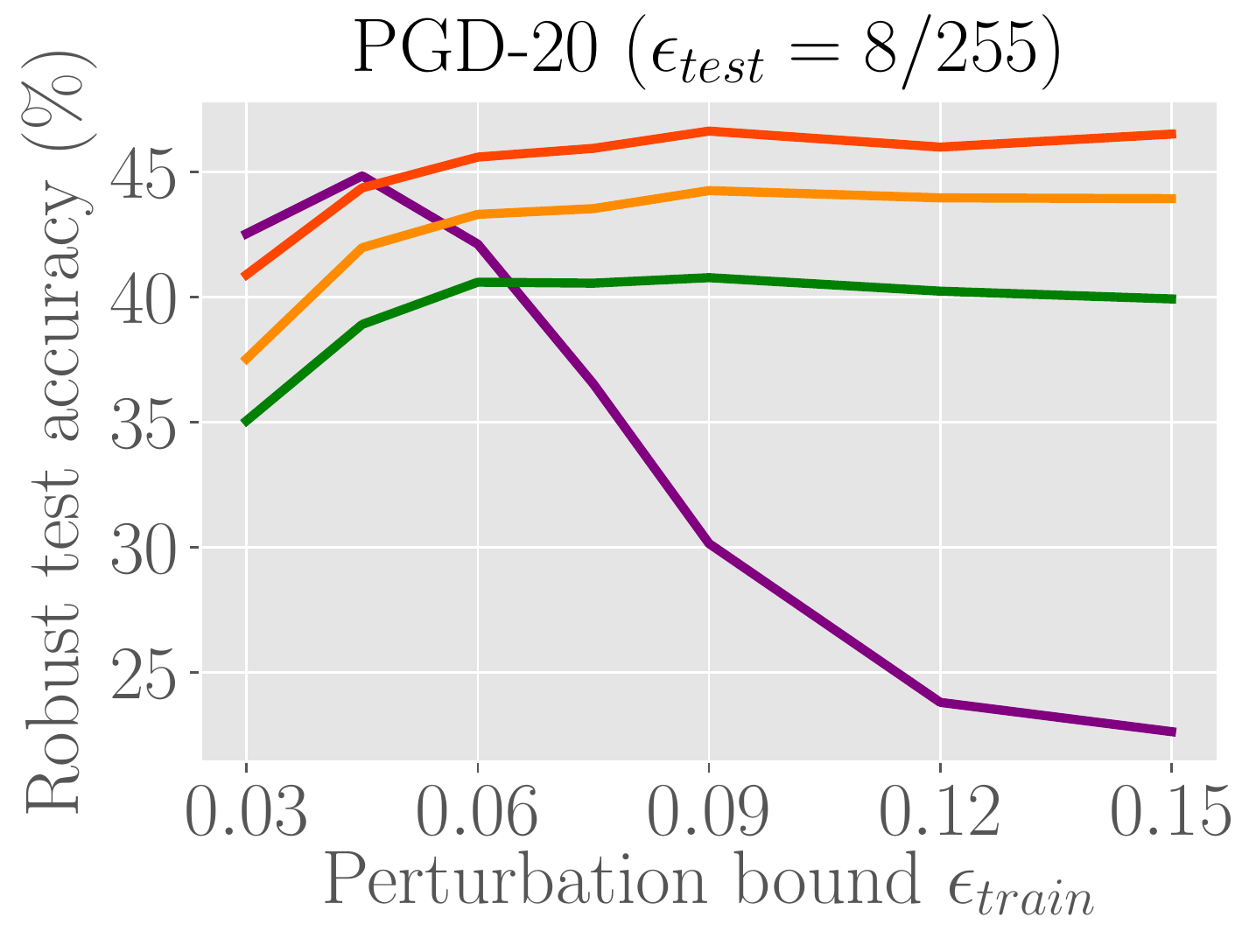}\\
    \includegraphics[scale=0.33]{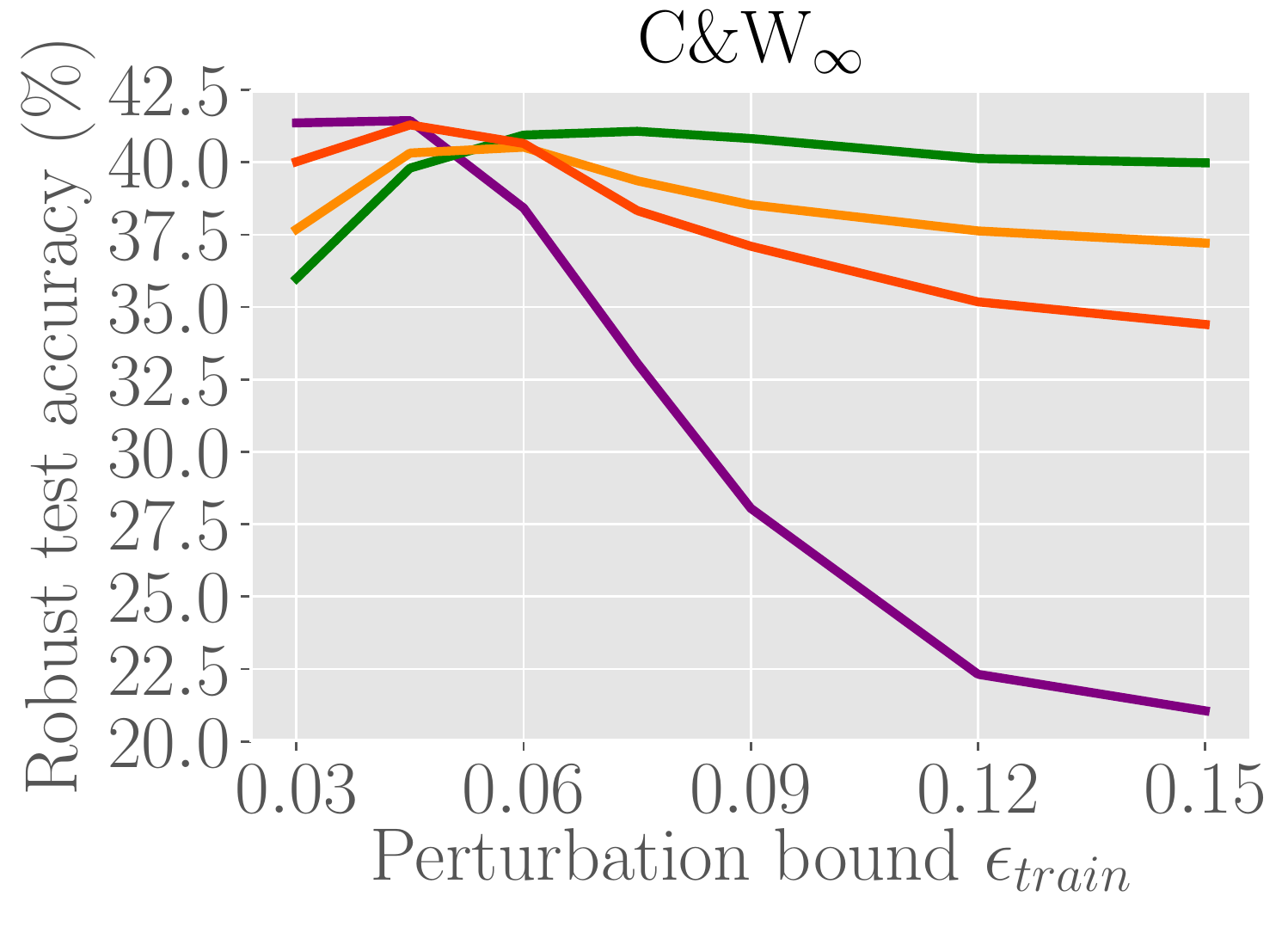}
    \includegraphics[scale=0.33]{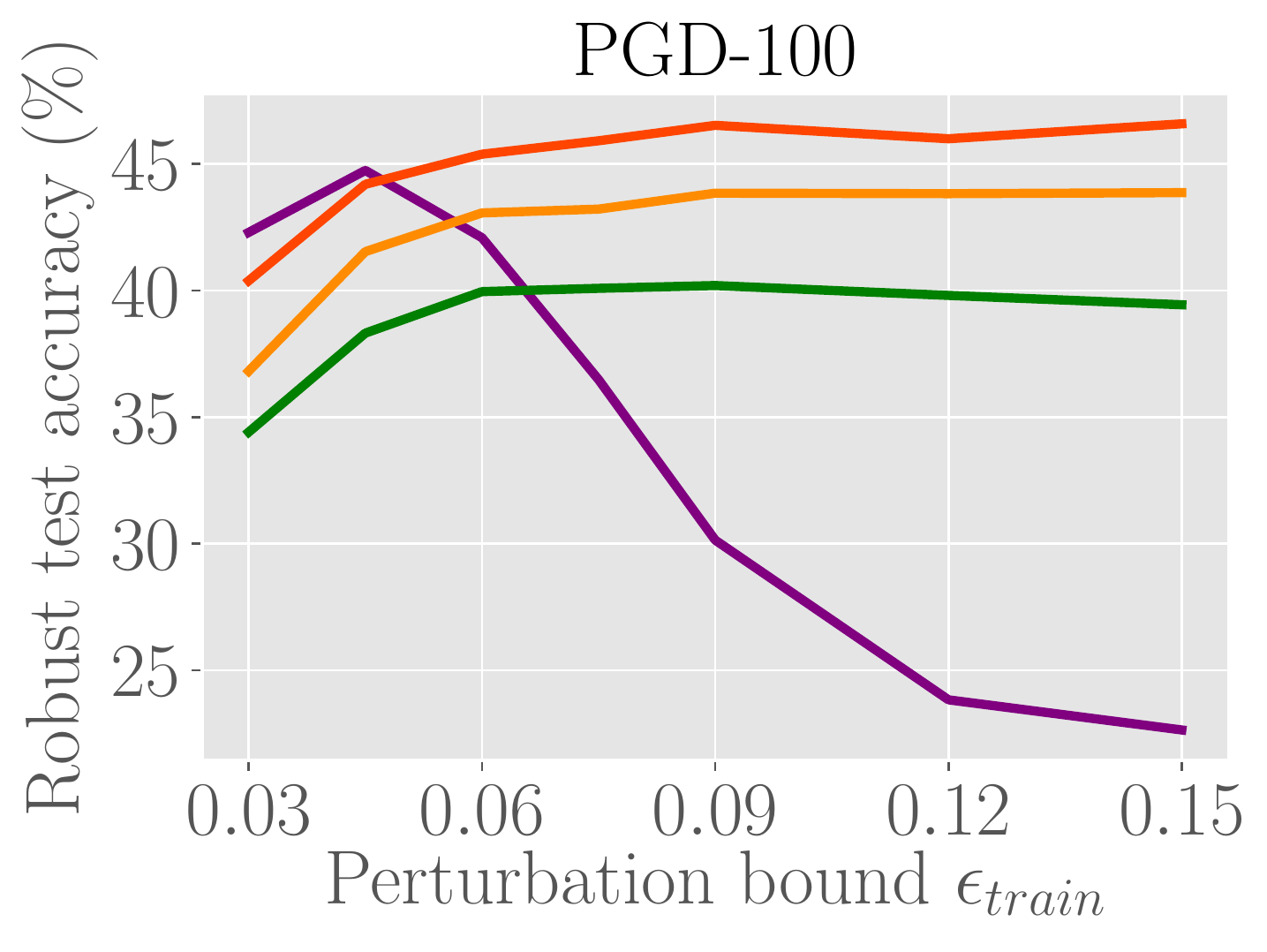}
    \includegraphics[scale=0.33]{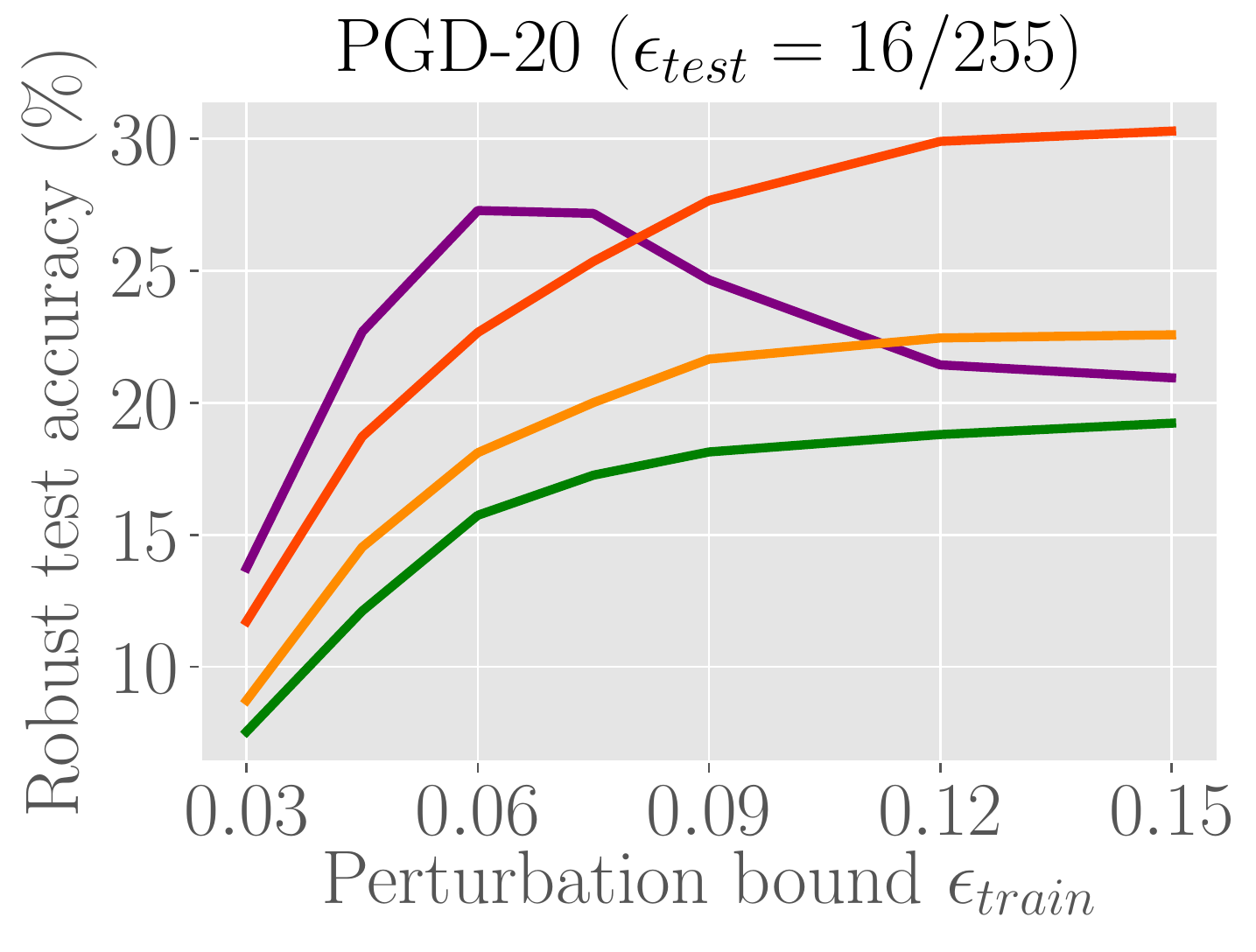}
    \caption{Test accuracy of Small CNN trained by FAT and standard adversarial training (Madry) with maximum PGD step $K = 20$ under different values of $\epsilon_{train}$ on CIFAR-10 dataset.}
    \label{fig:smallcnn_cifar10_pgd20_dynamic_epsball}
\end{figure}

\begin{figure}[!htb]
    \centering
    \includegraphics[scale=0.33]{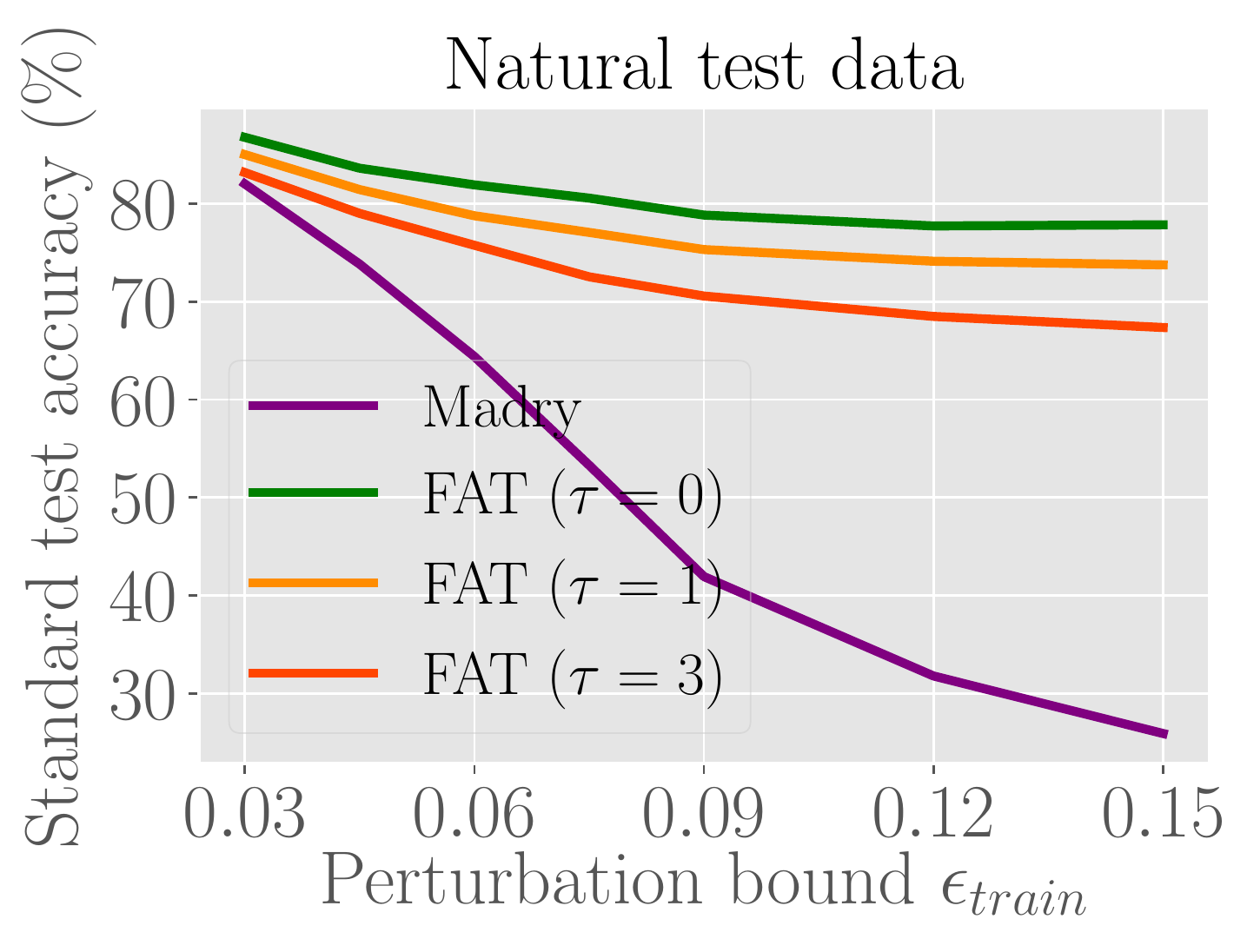}
    \includegraphics[scale=0.33]{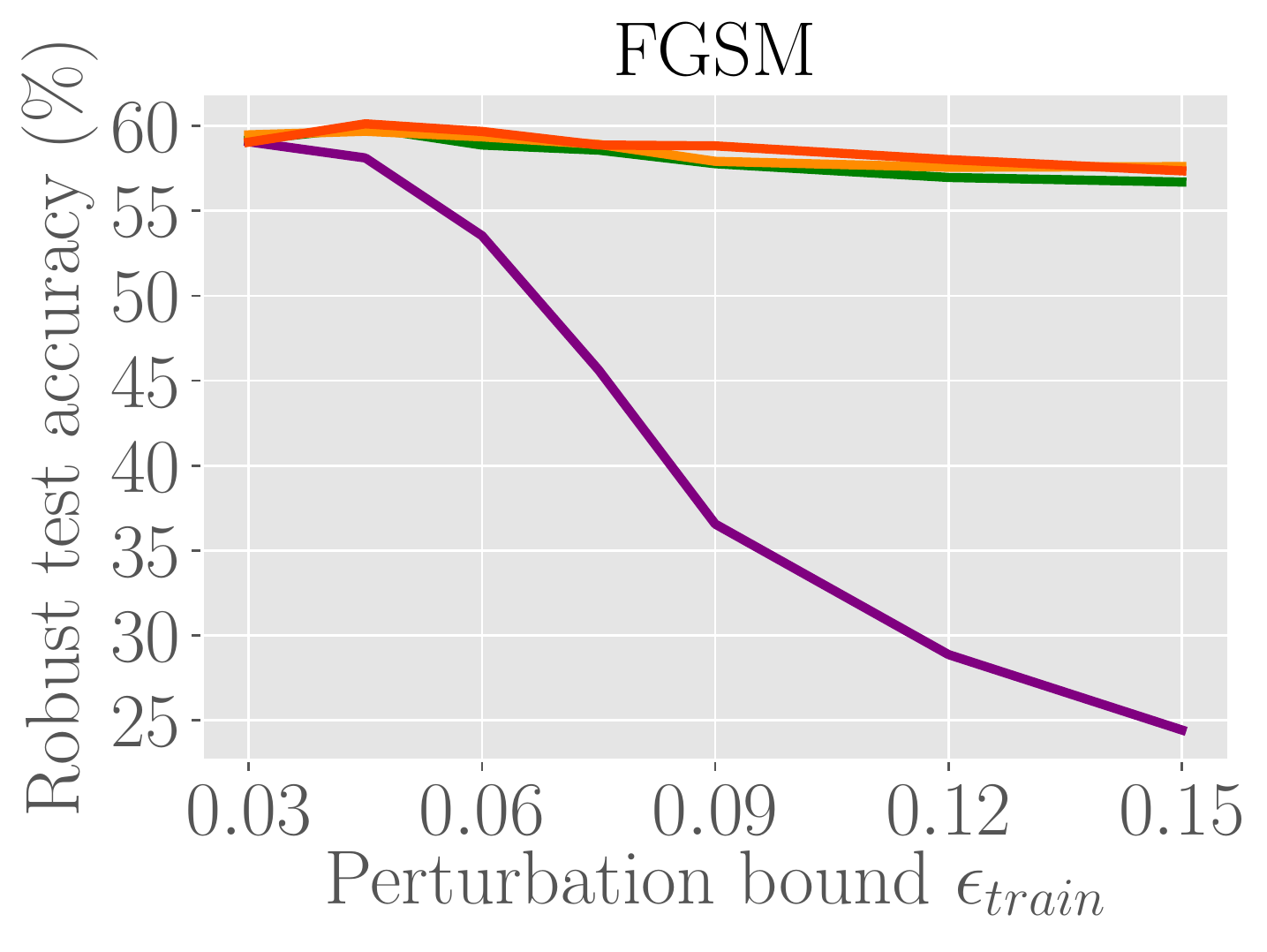}
    \includegraphics[scale=0.33]{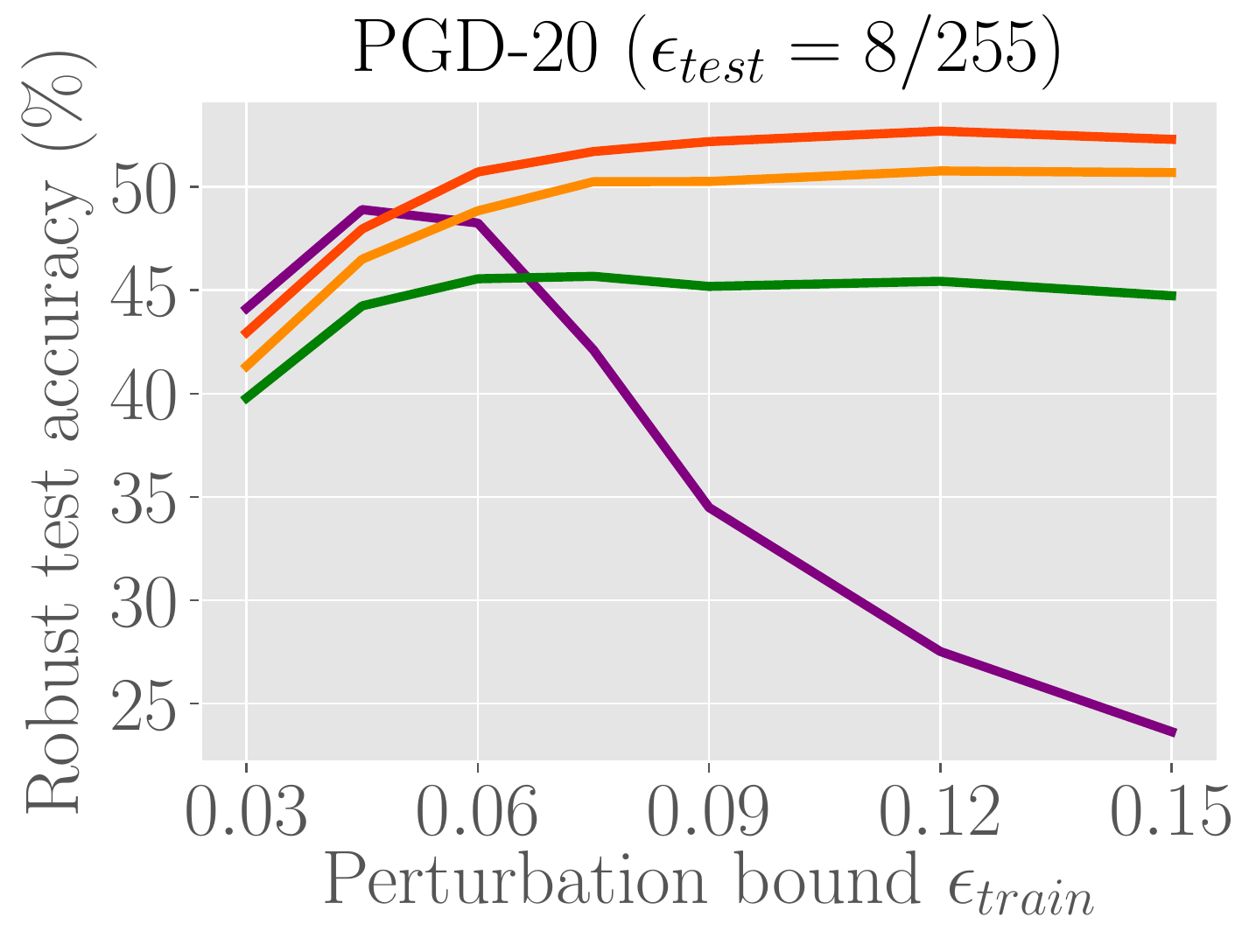}\\
    \includegraphics[scale=0.33]{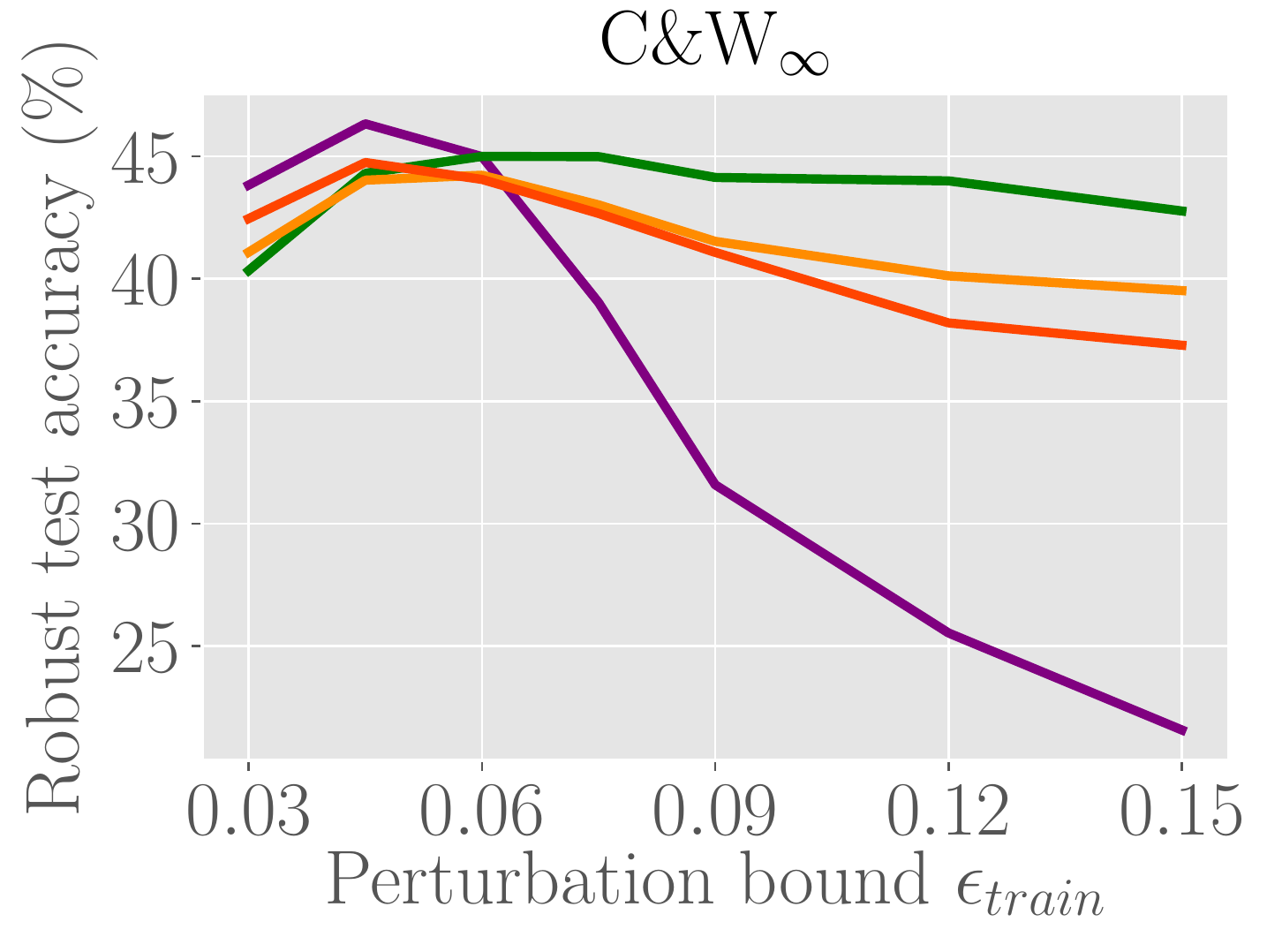}
    \includegraphics[scale=0.33]{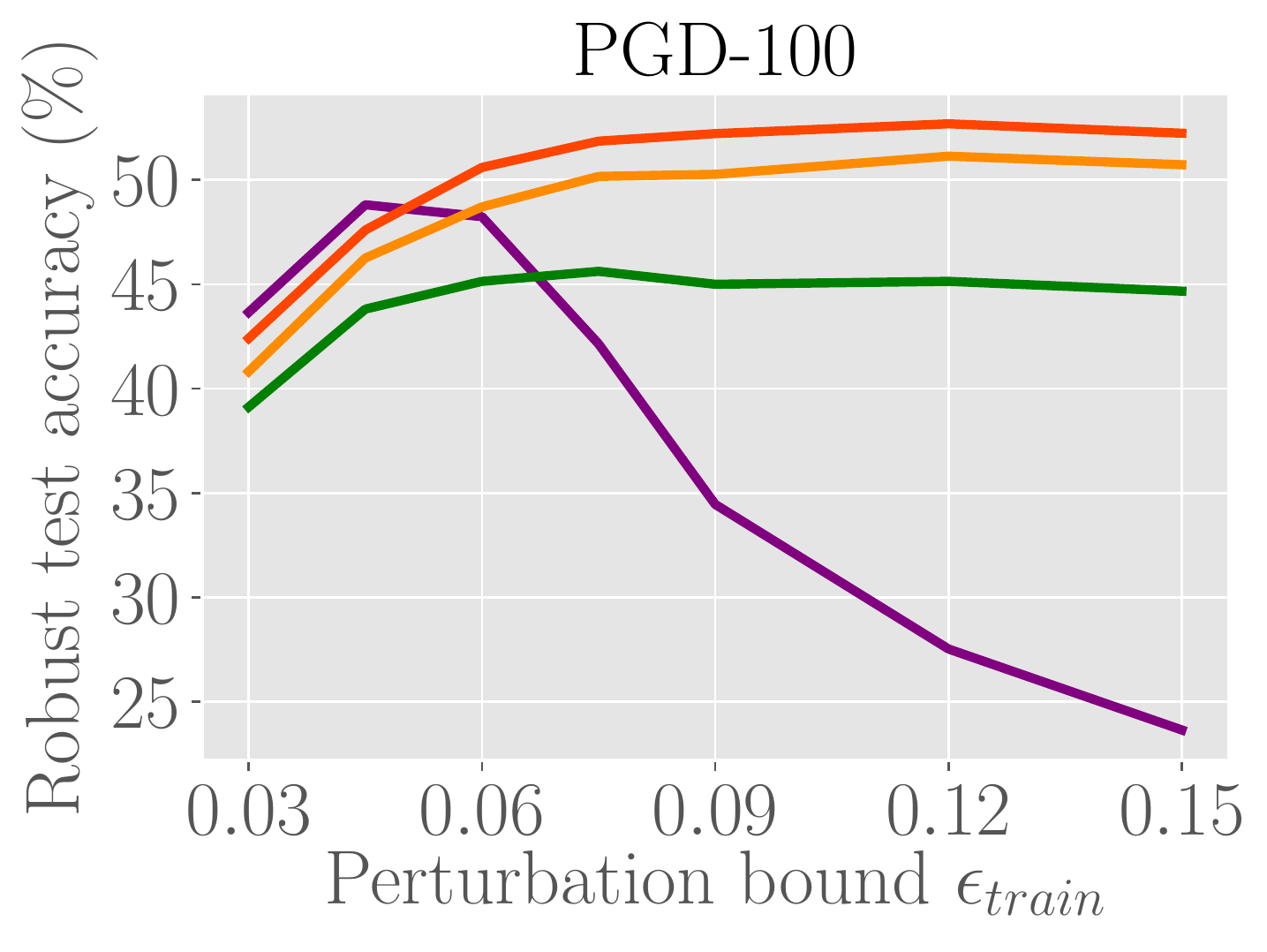}
    \includegraphics[scale=0.33]{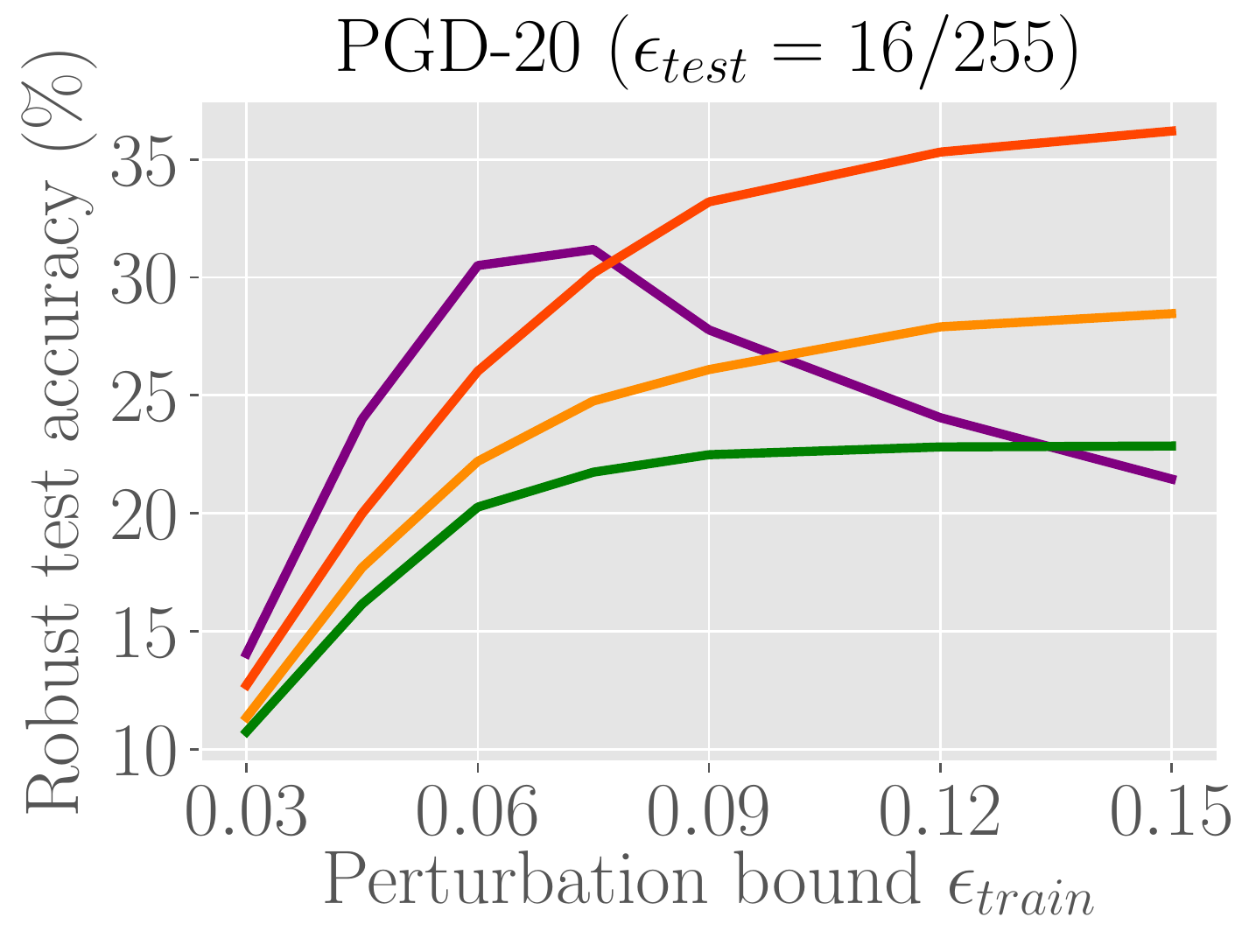}
    \caption{Test accuracy of ResNet-18 trained by FAT standard adversarial training (Madry) with maximum PGD step $K = 20$ under different values of $\epsilon_{train}$ on CIFAR-10 dataset.}
    \label{fig:resnet18_cifar10_pgd20_dynamic_epsball}
\end{figure}

\begin{figure}[!htb]
    \centering
    \includegraphics[scale=0.33]{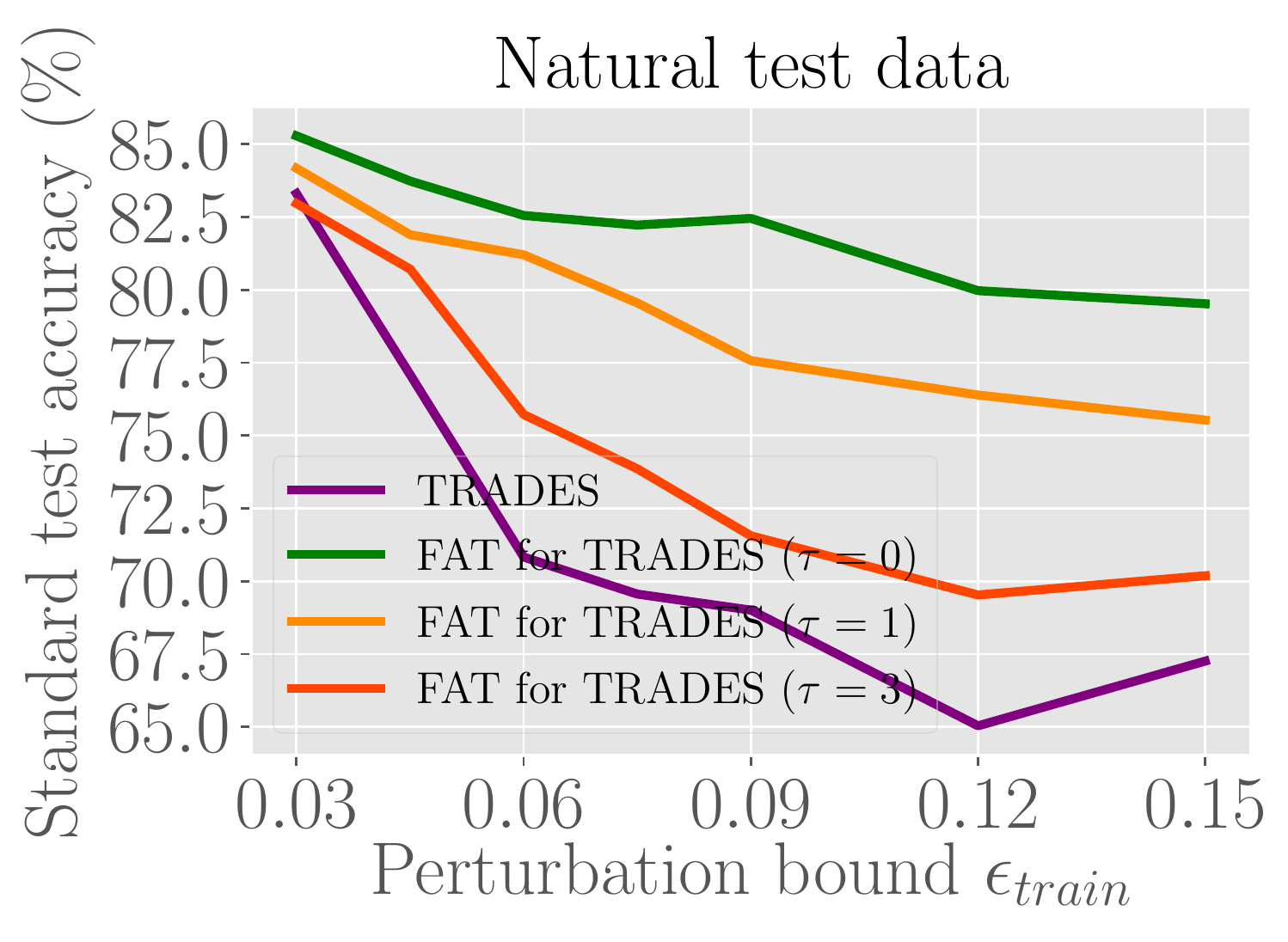}
    \includegraphics[scale=0.33]{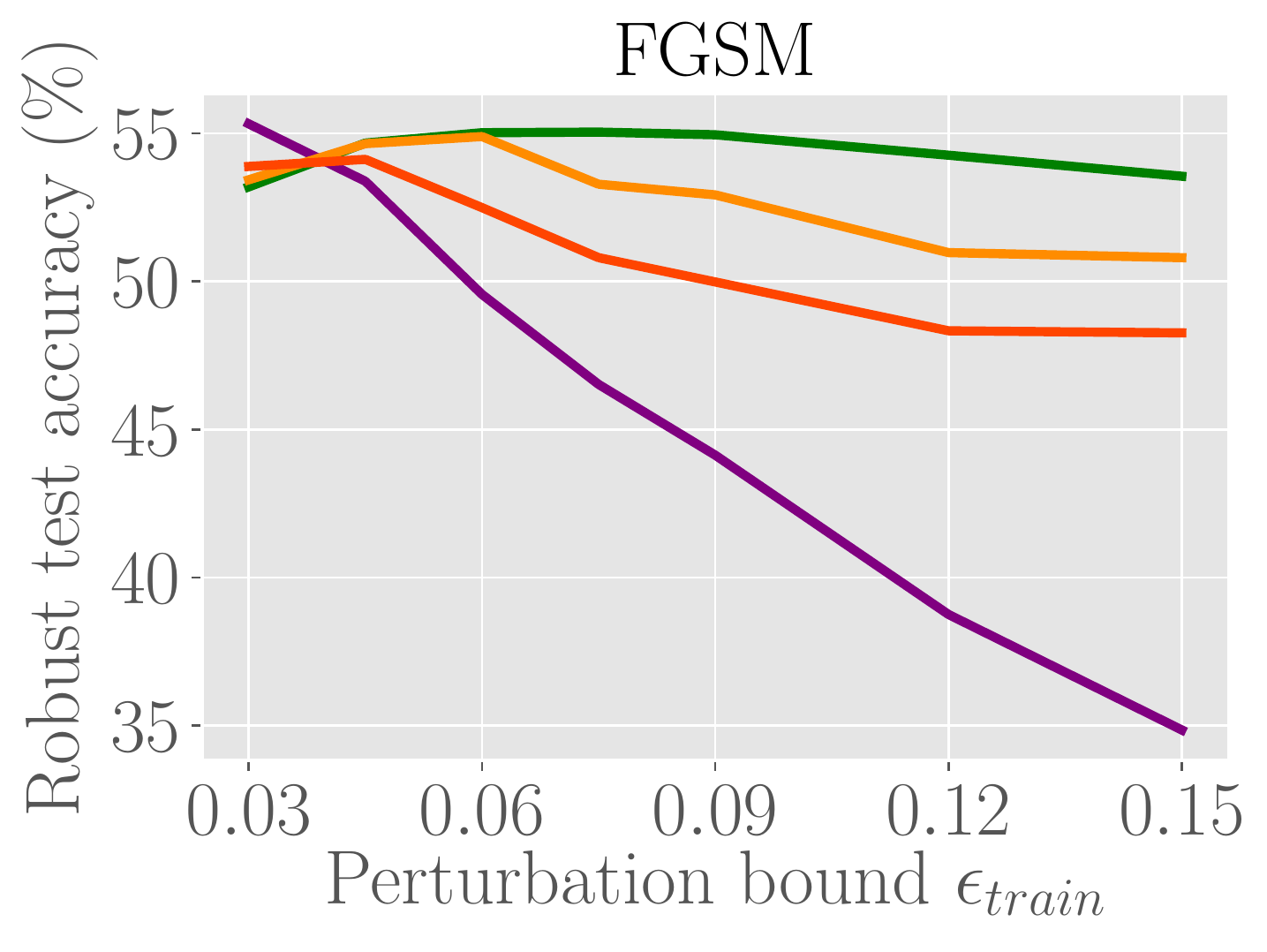}
    \includegraphics[scale=0.33]{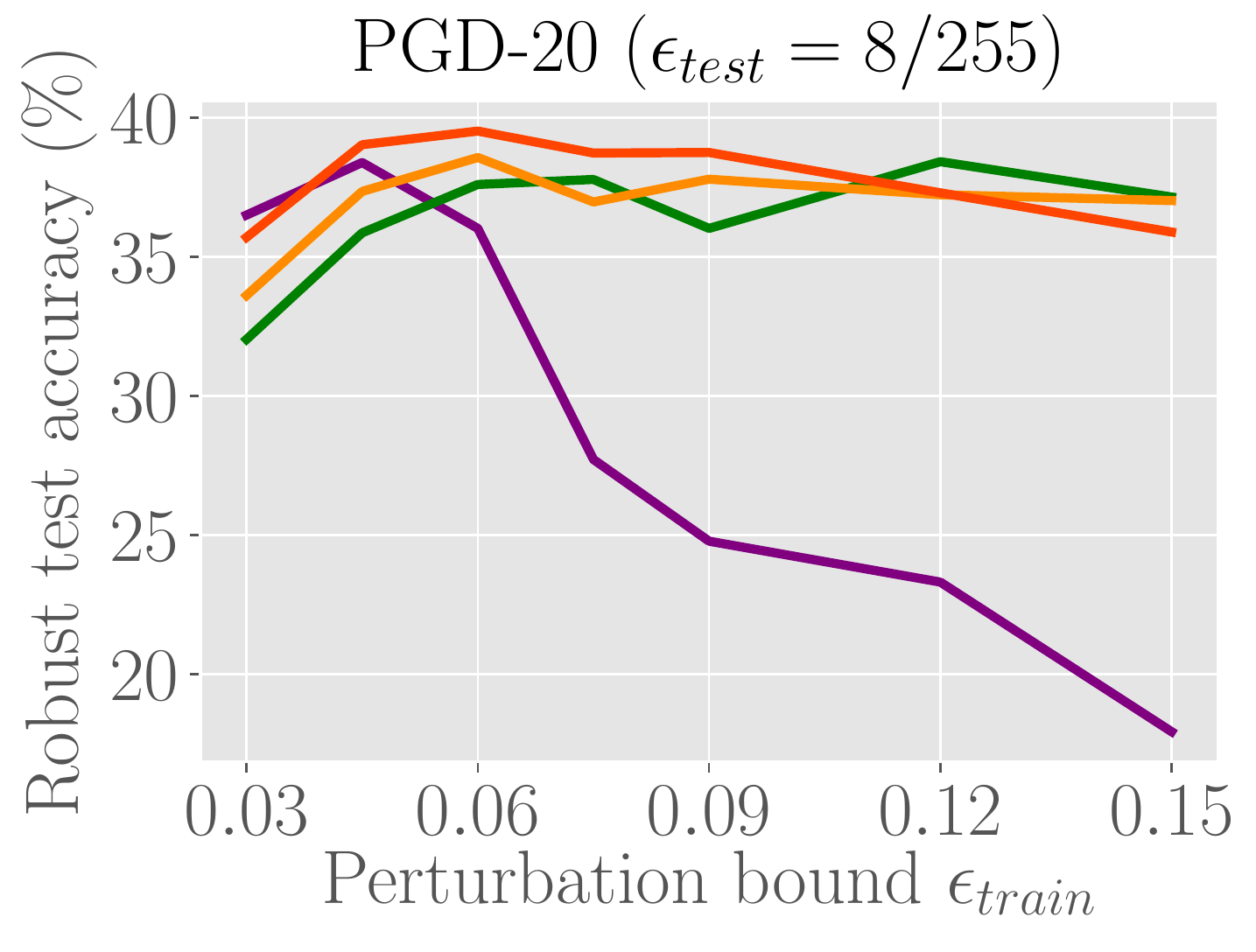}\\
    \includegraphics[scale=0.33]{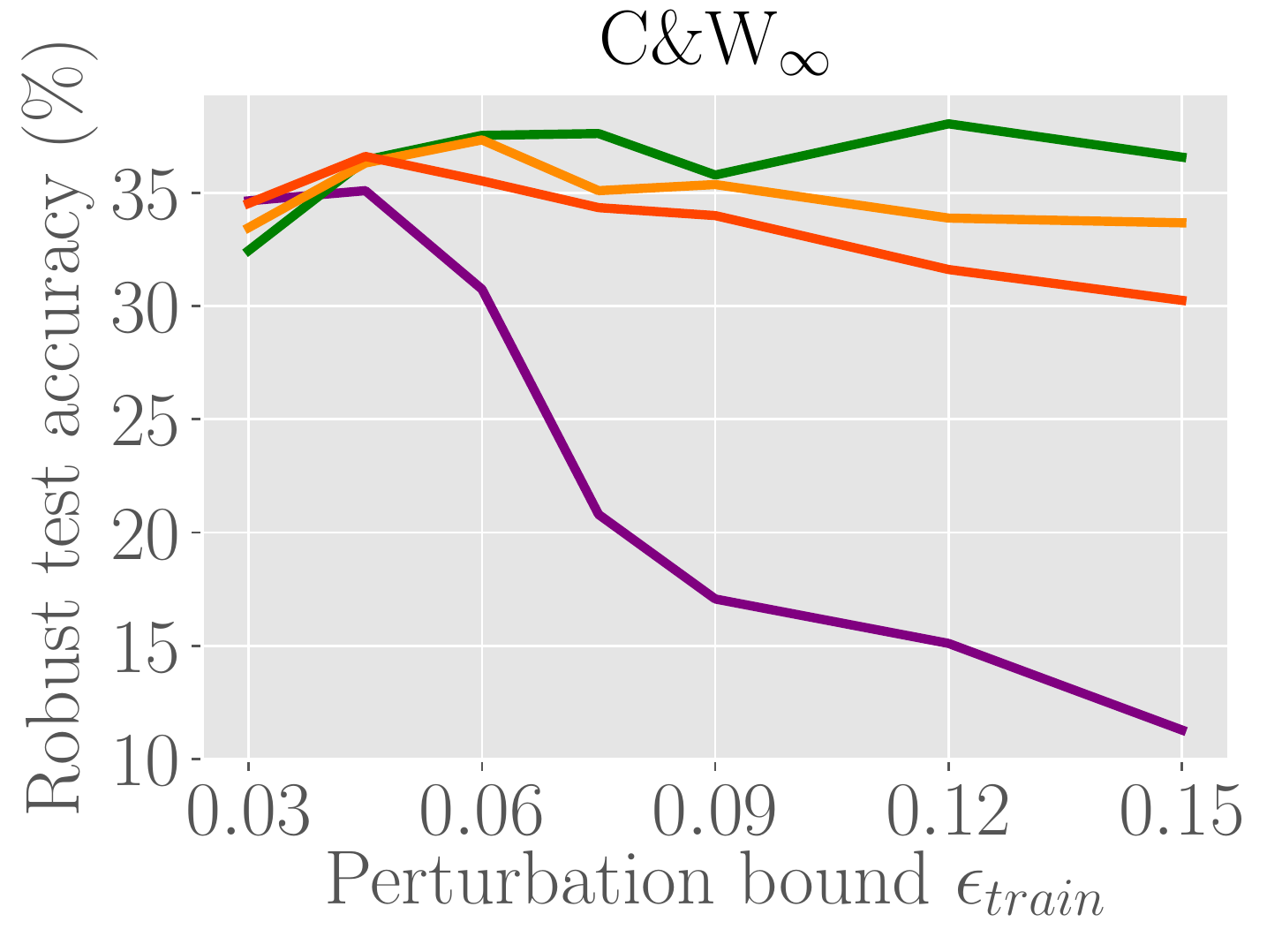}
    \includegraphics[scale=0.33]{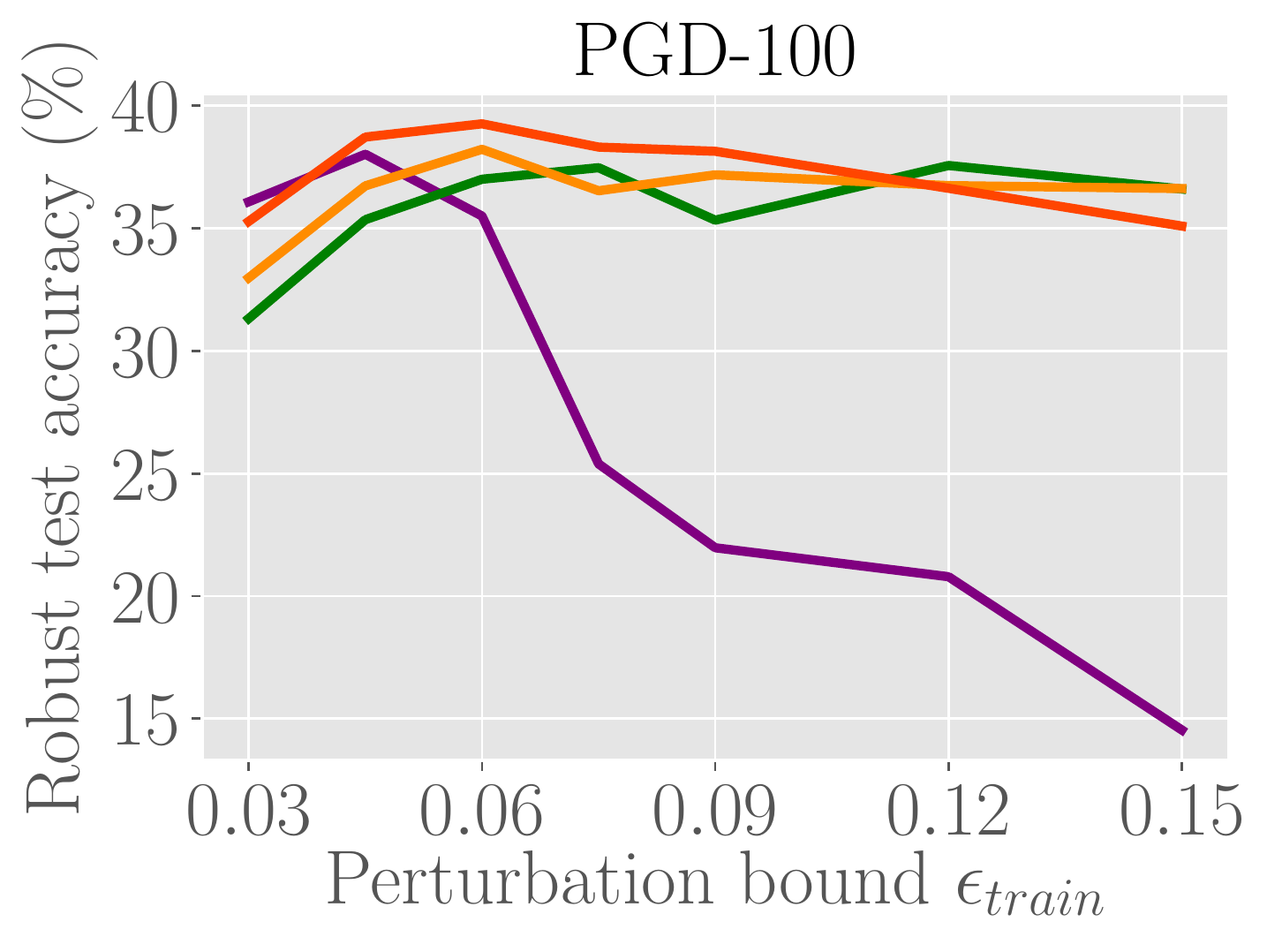}
    \includegraphics[scale=0.33]{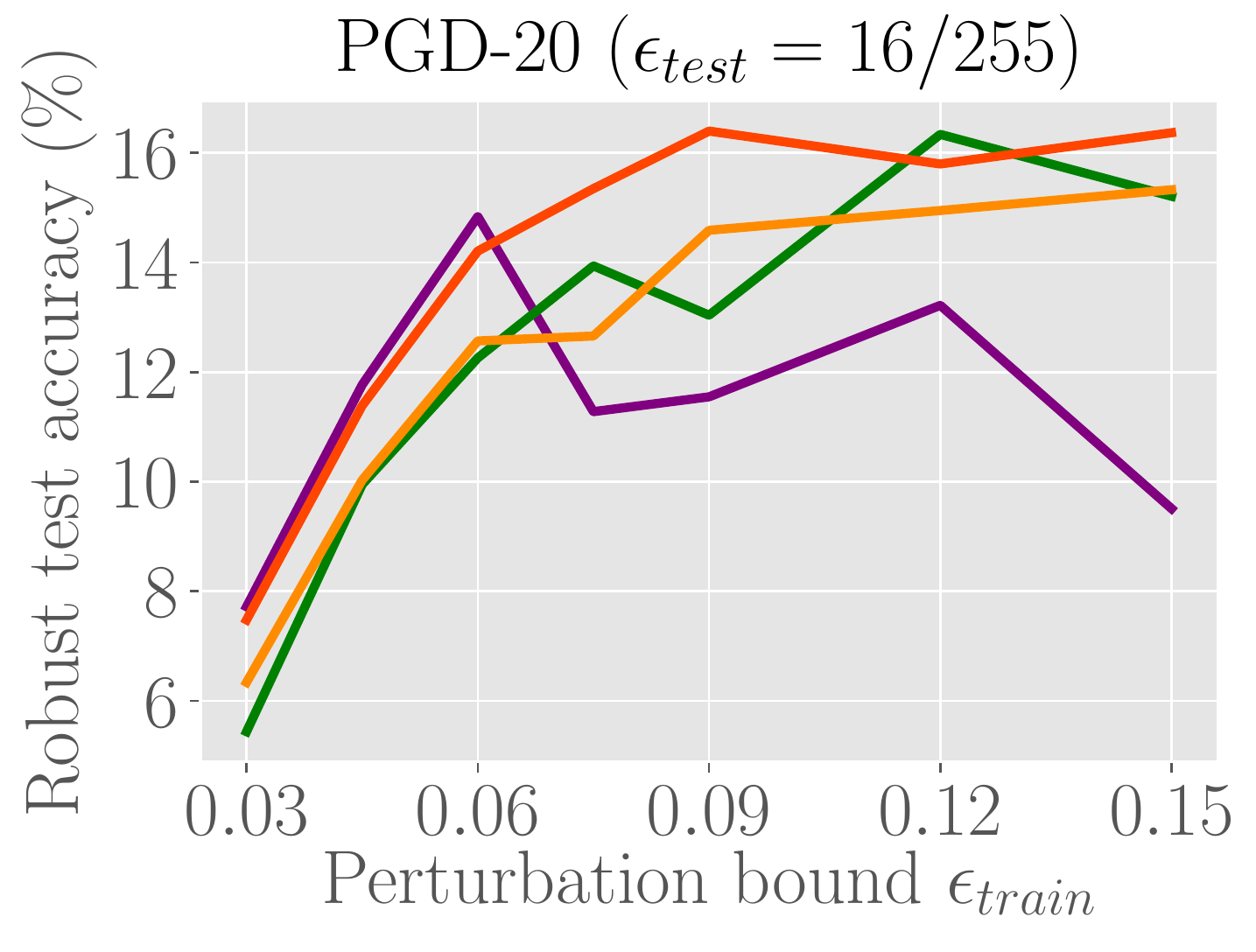}
    \caption{Test accuracy of Small CNN trained by FAT for TRADES and TRADES with maximum PGD step $K = 20$ under different values of $\epsilon_{train}$ on CIFAR-10 dataset.}
    \label{fig:smallcnn_fat_trades_cifar10_pgd20_dynamic_epsball}
\end{figure}

\begin{figure}[!htb]
    \centering
    \includegraphics[scale=0.33]{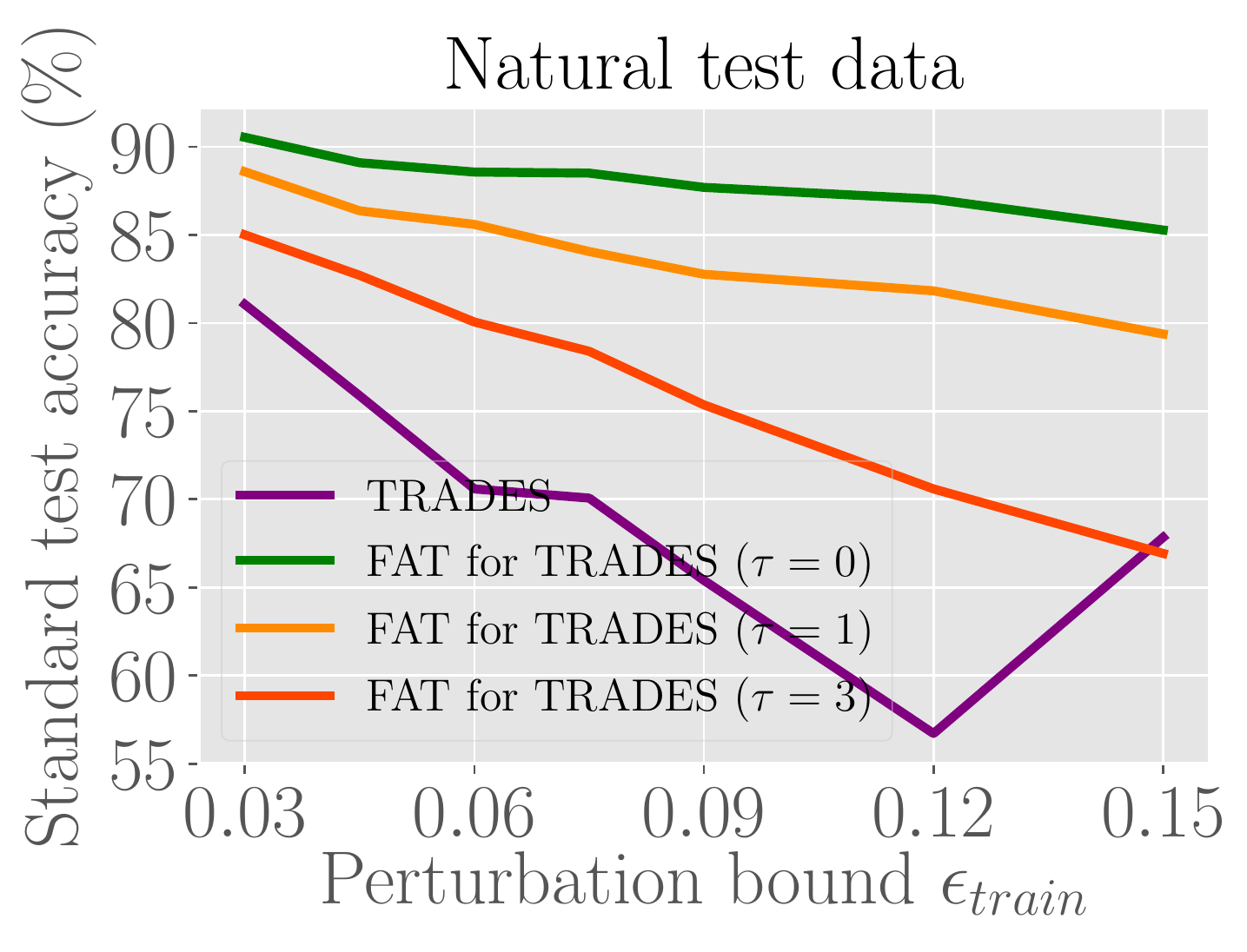}
    \includegraphics[scale=0.33]{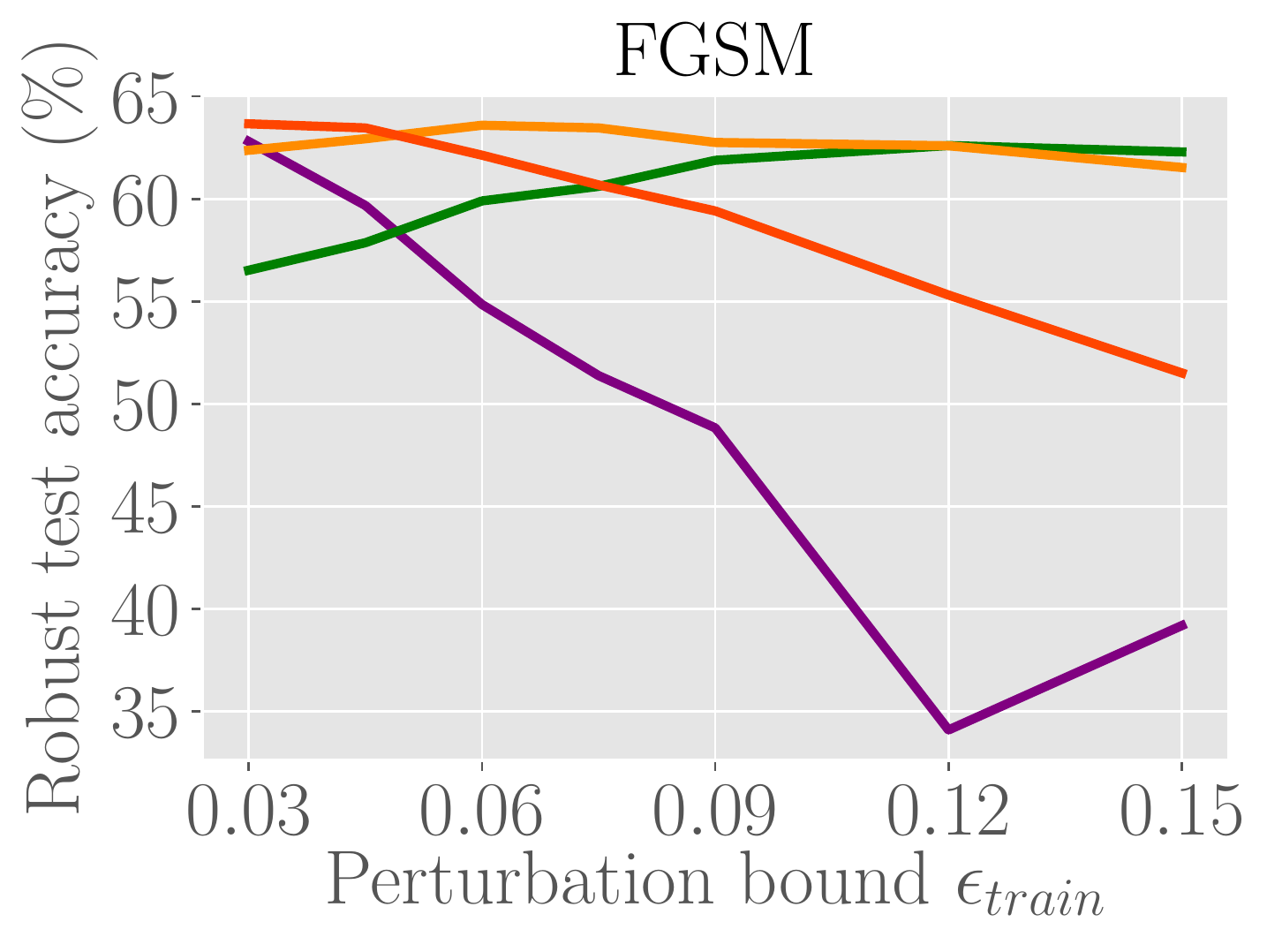}
    \includegraphics[scale=0.33]{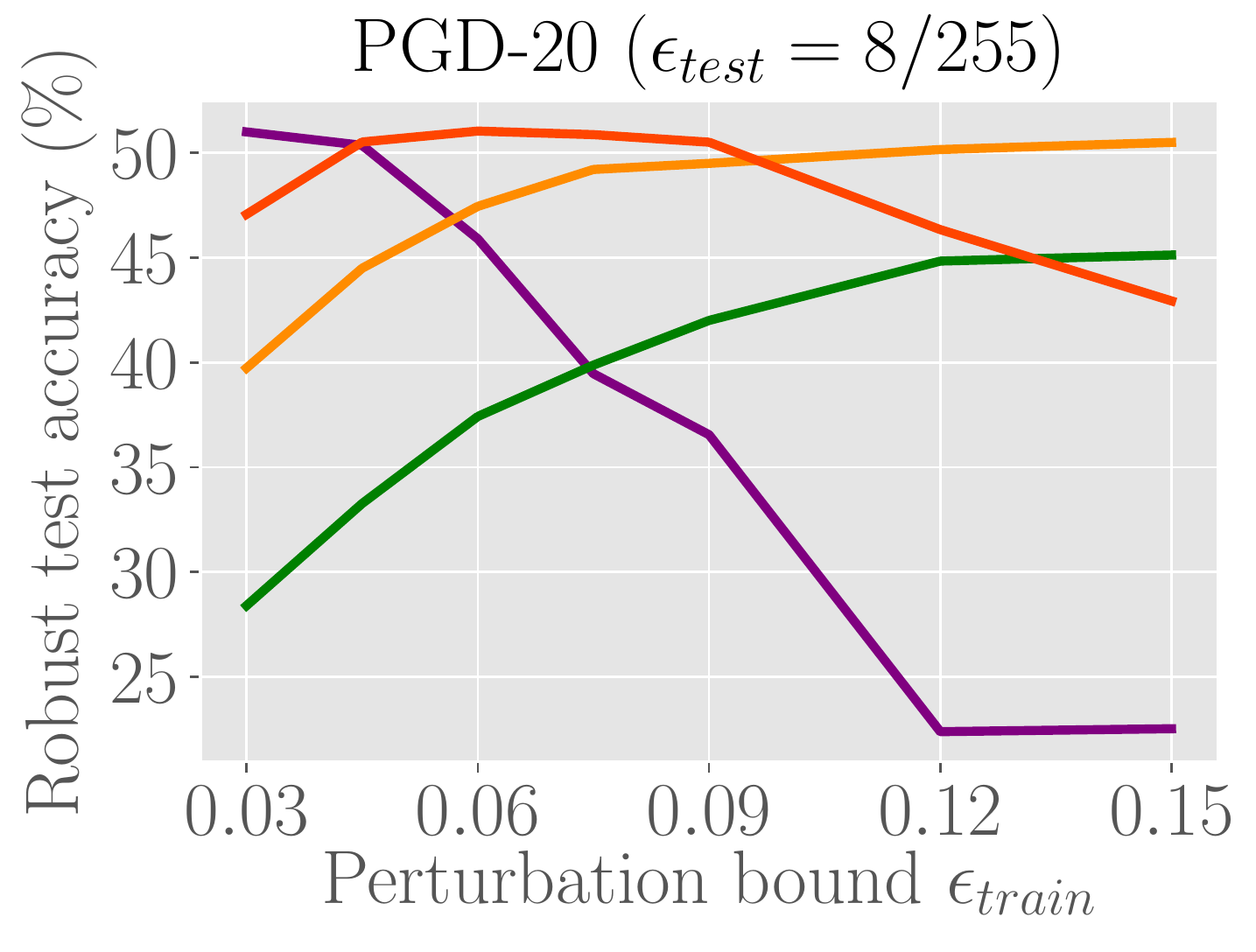}\\
    \includegraphics[scale=0.33]{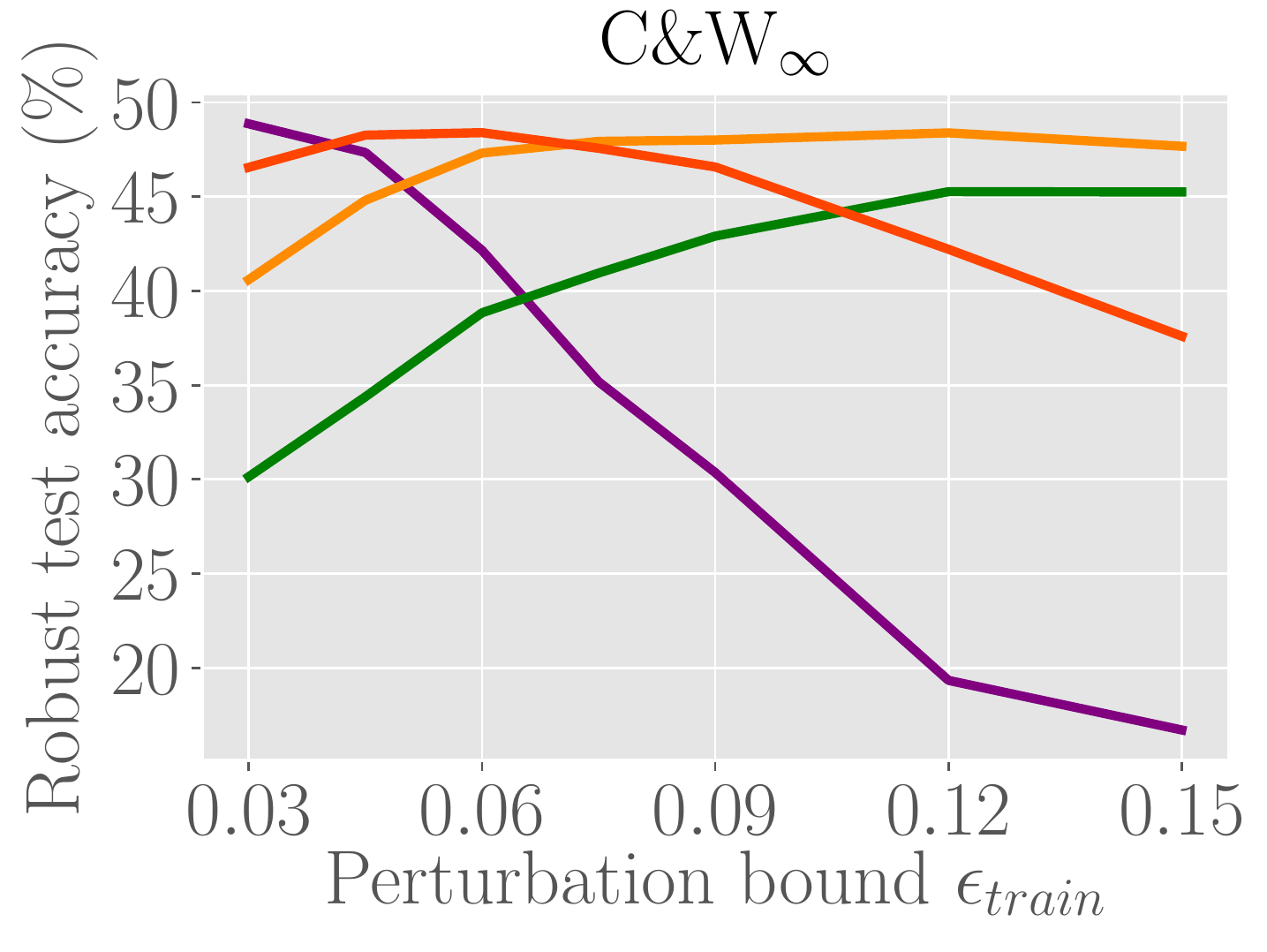}
    \includegraphics[scale=0.33]{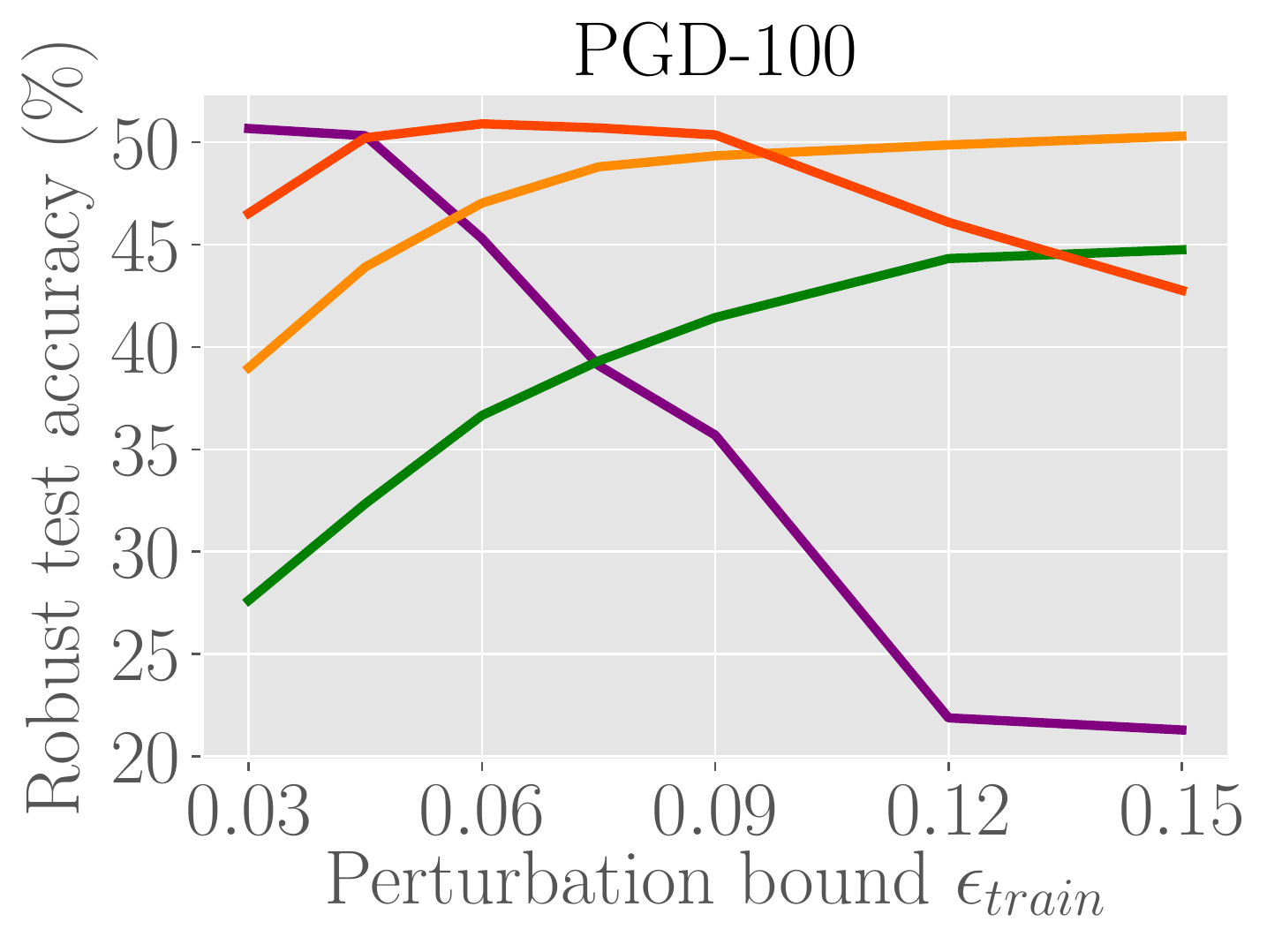}
    \includegraphics[scale=0.33]{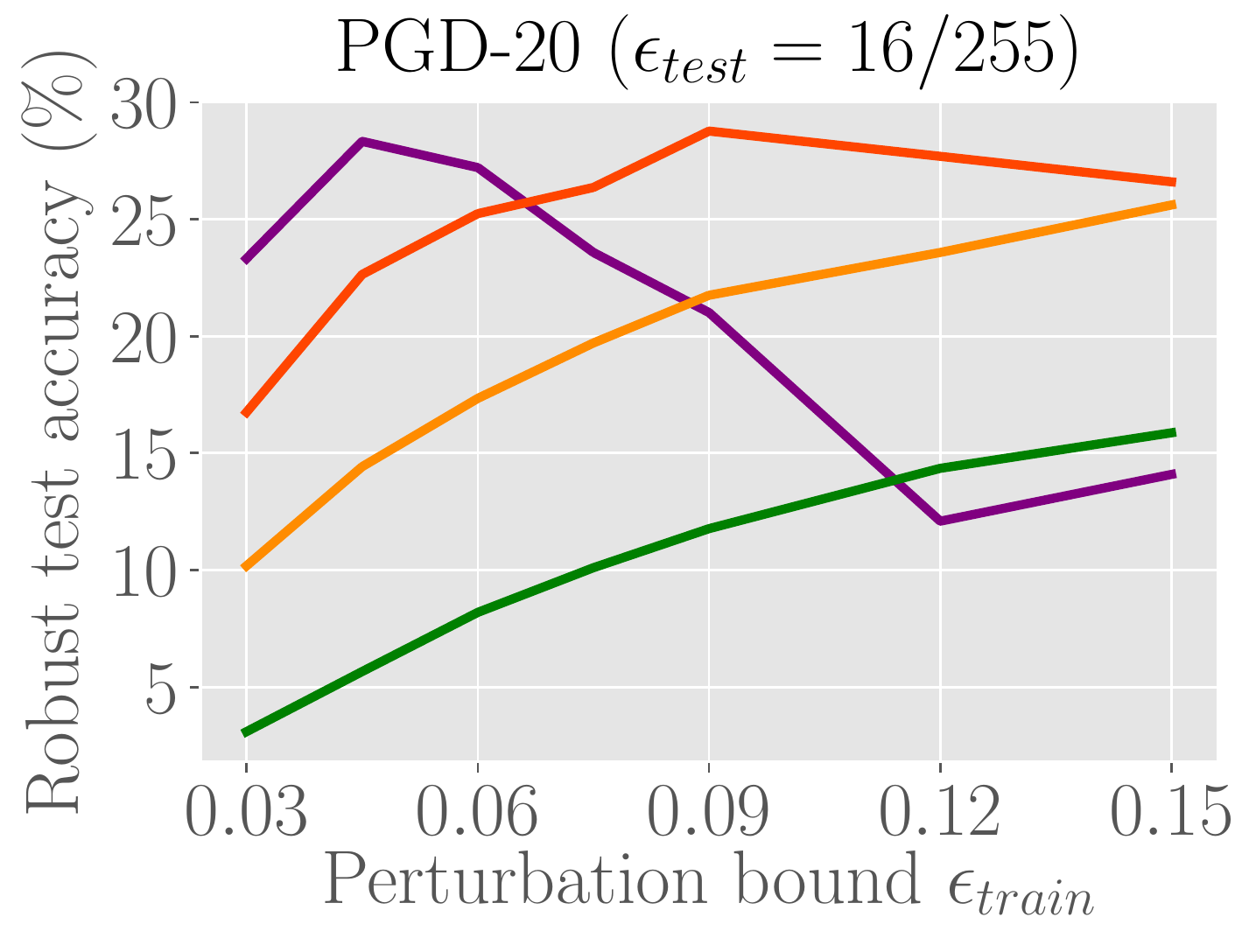}
    \caption{Test accuracy of ResNet-18 trained by FAT for TRADES and TRADES with maximum PGD step $K = 20$ under different values of $\epsilon_{train}$ on CIFAR-10 dataset.}
    \label{fig:resnet18_fat_trades_cifar10_pgd20_dynamic_epsball}
\end{figure}

By setting maximum PGD step $K = 20$, we conduct more experiments on Small CNN and ResNet-18 using FAT and FAT for TRADES. Except maximum PGD steps $K=20$, training settings are the same as those are stated in Section~\ref{section:fat_enable_lager_epsilon}.  Test results of robust deep models are shown in Figures~\ref{fig:smallcnn_cifar10_pgd20_dynamic_epsball},~\ref{fig:resnet18_cifar10_pgd20_dynamic_epsball},~\ref{fig:smallcnn_fat_trades_cifar10_pgd20_dynamic_epsball} and~\ref{fig:resnet18_fat_trades_cifar10_pgd20_dynamic_epsball}.

\clearpage
\subsection{C$\&$W Attack Analysis}
\label{Appendix:reasons-cw-performance}
As is shown in Figure~\ref{fig:resnet18_cifar10_dynamic_epsball} along with Figure~\ref{fig:smallcnn_cifar10_dynamic_epsball} in Section~\ref{appendix:exp_smallcnn}, both standard adversarial training and friendly adversarial training do not perform well under C$\&$W~\cite{Carlini017_CW} attack with larger $\epsilon_{train}$ (e.g., $\epsilon_{train} > 0.075$). We discuss the reasons for these phenomena.  

\paragraph{C$\&$W attack.} Given $x$, we choose a target class $t$ and then search for adversarial data $\Tilde{x}$ under C$\&$W attack in the $L_p$ metric by solving 
\begin{align*}
    \mathrm{minimize} \quad  \| \Tilde{x} - x \|_p + c \cdot h(\Tilde{x})
\end{align*}
with $h$ defined as 
\begin{align*}
   h(\Tilde{x}) = \max(\max_{i \neq t}f(\Tilde{x})_i-f(\Tilde{x})_t,-\kappa).
\end{align*}
The parameter $c>0$ balances two parts of loss. $\kappa > 0$ encourages the solver to find adversarial data $\Tilde{x}$ that will be classified as class $t$ with high confidence. Note that this paper follows the implementation of C$\&$W$_{\infty}$ attack in \cite{Cai_CAT}\footnote{\href{https://github.com/sunblaze-ucb/curriculum-adversarial-training-CAT}{curriculum adversarial training GitHub}} and \cite{Wang_Xingjun_MA_FOSC_DAT}\footnote{\href{https://github.com/YisenWang/dynamic_adv_training}{dynamic adversarial training GitHub}} where they replace the cross-entropy loss with $h(\Tilde{x})$ in PGD, i.e.,
\begin{equation}
\label{Eq:cw_inner_maximization}
\xadv_i = \argmax \nolimits_{\xadv\in\epsball[\bx_i]} (\max_{i \neq y_i}f(\Tilde{x})_i - f(\Tilde{x})_{y_i} - \kappa).
\end{equation}

\paragraph{Analysis.} 
In Figure~\ref{fig:resnet18_cifar10_dynamic_epsball}, with larger $\epsilon_{train}$, the performance evaluated by PGD attacks increases, while performance evaluated by C$\&$W attack decreases. The reason is that
C$\&$W and PGD have different ways of generating adversarial data according to Eq.~\eqref{Eq:cw_inner_maximization} and Eq.~\eqref{Eq:madry_inner_maximization} respectively. 
The two interactive methods search adversarial data in different directions due to gradients w.r.t. different loss.
Therefore, the distributions of C$\&$W and PGD adversarial data are inconsistent. 
As perturbation bound $\epsilon_{train}$ increases, there are more PGD adversarial data generated within $\epsilon_{train}$-ball.
A DNN learned from more PGD adversarial data becomes more defensive to PGD attacks, but this deep model may not effectively defend C$\&$W adversarial data.



\clearpage
\section{Extensive State-of-the-art Results on Wide ResNet}
\label{appendix:sota-WRN}
\subsection{Training Details of FAT on WRN-32-10}
\label{APPENDIX:training_details_sota_FAT}
In Table~\ref{table:sota_result_madry}, we compare our FAT with standard adversarial training (Madry), CAT~\cite{Cai_CAT} and DAT~\cite{Wang_Xingjun_MA_FOSC_DAT}. 

We use FAT ($\epsilon_{train} = 8/255$ and $16/255$ respectively) to train WRN-32-10 for 120 epochs using SGD with 0.9 momentum, and weight decay is 0.0002. Maximum PGD step is 10 and step size is fixed to 0.007. The initial learning rate is 0.1 reduced to 0.01, 0.001 and 0.0005 at epoch 60, 90 and 110. We set step $\tau = 0$ initially and increase $\tau$ by one at epoch 50 and 90 respectively. The maximum step $K = 10$. 
We report performance of the deep model at the last epoch. 
For fair comparison, in Table~\ref{table:sota_result_madry} we use the same test settings as those in DAT~\cite{Wang_Xingjun_MA_FOSC_DAT}. 
Performance of robust deep model is evaluated standard test accuracy for natural data and robust test accuracy for adversarial data, that are generated by FGSM, PGD-20 (20-steps PGD with random start), PGD-100 and C$\&$W$_{\infty}$(L$_\infty$ version of C$\&$W optimized by PGD-30).

All attacks have the same perturbation bound $\epsilon_{test} = 0.031$ and step size in PGD is $\alpha = \epsilon_{test}/4$. The same as DAT~\cite{Wang_Xingjun_MA_FOSC_DAT}, there is random start in PGD attack, i.e., uniformly random perturbations ($[-\epsilon_{test}, +\epsilon_{test}]$) added to natural data before PGD perturbations. We report the median test accuracy and its standard deviation over 5 repeated trails of adversarial training in Table~\ref{table:sota_result_madry}. 

\subsection{Training details of FAT for TRADES on Wide ResNet}
\label{APPENDIX:training_details_sota_TRADES}
In Table~\ref{table:sota_result_trades}, we use FAT for TRADES ($\epsilon_{train} = 8/255$ and $16/255$ respectively) train WRN-34-10 by FAT for TRADES for 85 epochs using SGD with 0.9 momentum and 0.0002 weight decay. Maximum PGD step $K = 10$ and step size $\alpha = 0.007$. The initial learning rate is 0.1 and divided 10 at epoch 75. We set step $\tau = 0$ initially and increased by one at epoch 30, 50 and 70. 
Since TRADES has a trade-off parameter $\beta$, for fair comparison, our FAT for TRADES use the same $\beta$. In Table~\ref{table:sota_result_trades}, we set $\beta = 1$ and $6$ separately, which are endorsed by~\cite{Zhang_trades}.

For fair comparison, we use the same test settings as those are stated in TRADES~\cite{Zhang_trades}. 
All attacks have the same perturbation bound $\epsilon_{test} = 0.031$ ( without random start), and step size $\alpha = 0.003$, which is the same as stated in the paper~\cite{Zhang_trades}. 
Performance of robust deep model is evaluated standard test accuracy for natural data and robust test accuracy for adversarial data, that are generated by FGSM, PGD-20, PGD-100 and C$\&$W$_{\infty}$(L$_\infty$ version of C$\&$W optimized by PGD-30). We report the median test accuracy and its standard deviation of the deep model at the last epoch over 3 repeated trials of adversarial training in Table~\ref{table:sota_result_trades}.

\paragraph{Fair comparison based on TRADES's experimental setting.} However, in TRADES's experimental testing\footnote{\href{https://github.com/yaodongyu/TRADES}{TRADES GitHub}}, they use random start before PGD perturbation that is deviated from the statements in the paper~\cite{Zhang_trades}. 
For fair comparison, we also retest the robust deep models under PGD attacks with random start. We evaluate their publicly released robust deep model\footnote{\href{https://drive.google.com/file/d/10sHvaXhTNZGz618QmD5gSOAjO3rMzV33/view}{TRADES's pre-trained model}} WRN-34-10 and compare it with ours trained by FAT for TRADES. The test results are reported in Table~\ref{table:sota_result_trades—randominit}. 

\paragraph{FAT for TRADES on larger WRN-58-10.} We employ Wide ResNet with larger capacity, i.e., WRN-58-10 to show our superior performance achieved by FAT for TRADES in Table~\ref{table:sota_result_trades—randominit}.
All the training settings are the same as details on WRN-34-10 in this section. The regularization parameter $\beta$ is fixed to 6.0. 
All attacks have the same perturbation bound $\epsilon_{test} = 0.031$ and step size $\alpha = 0.003$, which is the same as TRADES's experimental setting. Robustness against FGSM, PGD-20(20-steps PGD with random start) and C$\&$W$_{\infty}$ is reported in Table~\ref{table:sota_result_trades—randominit}.

\begin{table}[!htb]
	\centering
	\caption{Robustness (test accuracy) of deep models on CIFAR-10 dataset (evaluated with random start)}
	\label{table:sota_result_trades—randominit}
	\begin{tabular}{c|c|cccc}
		\hline
		Model & Defense & Natural & FGSM & PGD-20 & C$\&$W$_{\infty}$ \\ \hline
		{WRN-34-10} & TRADES ($\beta = 6.0$) & 84.92 & 67.00 & 57.18 & 54.72  \\ 
		& FAT for TRADES ($\epsilon_{train} = 8/255$) & 86.38 $\pm$ 0.548 & 67.64 $\pm$ 0.572 & 56.65 $\pm$ 0.262 & 54.51 $\pm$ 0.299  \\ 
		& FAT for TRADES ($\epsilon_{train} = 16/255$) & 84.39 $\pm$ 0.030 & 67.38 $\pm$ 0.370 & 57.67 $\pm$ 0.198 & 54.62 $\pm$ 0.140  \\ \hline
		{WRN-58-10} & FAT for TRADES ($\epsilon_{train} = 8/255$) & \textbf{87.09} & \textbf{68.7} & 57.17 & 55.43  \\ 
		& FAT for TRADES ($\epsilon_{train} = 16/255$) & 85.28 & 68.08 & \textbf{58.39} & \textbf{55.89}  \\ \hline
	\end{tabular}
\end{table}

\subsection{FAT for MART on Wide ResNet}
\label{APPENDIX:training_details_sota_MART}
We train WRN-34-10 by FAT for MART ($\epsilon_{train} = 8/255$ and $16/255$ respectively) using SGD with 0.9 momentum and 0.0002 weight decay. Maximum PGD step $K = 10$ and step size $\alpha = 0.007$. The initial learning rate is 0.1 and divided 10 at epoch 60 and 90 respectively. We set step $\tau = 0$ initially and increase $\tau$ by one at epoch 20, 40, 60 and 80. The regularization parameter $\beta$ is fixed to 6.0. The maximum step size $K = 10$.


\paragraph{Fair comparison based on MART's experimental setting.} For fair comparison, all attacks have the same perturbation bound $\epsilon_{test} = 8/255$ and step size $\alpha = \epsilon_{test}/10$, which is the same setting in MART~\cite{wang2020improving_MART}. White-box robustness of the deep model against attacks such as FGSM, PGD-20 (20-steps PGD with random start) and C$\&$W$_{\infty}$ (L$_\infty$ version of C$\&$W optimized by PGD-30) is reported. We evaluate \citet{wang2020improving_MART} publicly released robust deep model\footnote{\href{https://drive.google.com/file/d/1QjEwSskuq7yq86kRKNv6tkn9I16cEBjc/view}{MART's pre-trained model}} WRN-34-10 and compare it with ours trained by FAT for MART. In Table~\ref{table:sota_result_mart}, we report the median test accuracy and its standard deviation over 3 repeated trails of FAT for MART on WRN-34-10.

\paragraph{FAT for MART on larger WRN-58-10.} In Table~\ref{table:sota_result_mart}, we also employ WRN-58-10 to show the performance achieved by FAT for MART. All the training and testing settings are the same as those on WRN-34-10.

\begin{table}[!htb]
	\centering
	\caption{Robustness (test accuracy) of deep models on CIFAR-10 dataset}
    \label{table:sota_result_mart}
	\begin{tabular}{c|c|cccc}
		\hline
		Model & Defense & Natural & FGSM & PGD-20 & C$\&$W$_{\infty}$ \\ \hline
		{WRN-34-10} & MART ($\beta = 6.0$) & 83.62 & 67.38 & 58.24 & \textbf{53.67}  \\  
		& FAT for MART ($\epsilon_{train} = 8/255$) & 86.40 $\pm$ 0.071 & 68.94 $\pm$ 0.195 & 57.89 $\pm$ 0.144 & 52.28 $\pm$ 0.110  \\ 
		& FAT for MART ($\epsilon_{train} = 16/255$) & 84.39 $\pm$ 0.390 & 68.52 $\pm$ 0.297 & 59.13 $\pm$ 0.180 & 52.85 $\pm$ 0.459  \\ \hline
		{WRN-58-10} & FAT for MART ($\epsilon_{train} = 8/255$) & \textbf{87.10} & \textbf{69.52} & 58.57 & 52.73 \\ 
		& FAT for MART ($\epsilon_{train} = 16/255$) & 85.19 & 69.00 & \textbf{59.82} & 53.01  \\ \hline
	\end{tabular}
\end{table}


\end{document}